\documentclass{article} 
\usepackage{iclr2025_conference,times}

\usepackage{amsmath,amsfonts,bm}

\def\eqref#1{equation~\ref{#1}}

\def\1{\bm{1}}

\def\ve{{\bm{e}}}

\def\mA{{\bm{A}}}
\def\mB{{\bm{B}}}

\def\mW{{\bm{W}}}
\def\mX{{\bm{X}}}

\DeclareMathAlphabet{\mathsfit}{\encodingdefault}{\sfdefault}{m}{sl}
\SetMathAlphabet{\mathsfit}{bold}{\encodingdefault}{\sfdefault}{bx}{n}

\def\sN{{\mathbb{N}}}

\newcommand{\R}{\mathbb{R}}

\usepackage{url}

\usepackage{amsmath}
\usepackage{graphicx}
\usepackage[colorlinks,linkcolor=red, anchorcolor=blue, citecolor=blue]{hyperref}

\usepackage{booktabs}
\usepackage{multirow}

\usepackage{graphicx}
\usepackage{fancyhdr}
\usepackage{tikz}
\usepackage{mathrsfs}
\usepackage{color}
\usepackage{multirow}
\usepackage{diagbox}
\usepackage{pifont}
\usepackage{mathrsfs}
\usepackage{threeparttable}
\usepackage{comment}
\usepackage[linesnumbered,ruled,vlined]{algorithm2e}
\usepackage{amsthm}
\usepackage{enumitem}
\usepackage[noabbrev,capitalise]{cleveref}
\usepackage{footnote}
\usepackage{subcaption}
\usepackage{cleveref} 
\usepackage{makecell}
\usepackage{listings}  
\makesavenoteenv{tabular}
\makesavenoteenv{table}
\usepackage{wrapfig}
\usepackage{listings}
\usepackage[toc,page]{appendix}
\usepackage{float}

\newtheorem{proposition}{Proposition}

\newcommand{\evenpairs}{\textsc{Even Pairs}}
\newcommand{\modulararithmeticsimple}{\textsc{Modular Arithmetic (Simple)}}
\newcommand{\paritycheck}{\textsc{Parity Check}}
\newcommand{\cyclenavigation}{\textsc{Cycle Navigation}}

\newcommand{\stackmanipulation}{\textsc{Stack Manipulation}}
\newcommand{\reversestring}{\textsc{Reverse String}}
\newcommand{\modulararithmeticbrackets}{\textsc{Modular Arithmetic}}
\newcommand{\solveequation}{\textsc{Solve Equation}}

\newcommand{\duplicatestring}{\textsc{Duplicate String}}
\newcommand{\missingduplicate}{\textsc{Missing Duplicate}}
\newcommand{\oddsfirst}{\textsc{Odds First}}
\newcommand{\binaryaddition}{\textsc{Binary Addition}}

\newcommand{\computesqrt}{\textsc{Compute Sqrt}}
\newcommand{\bucketsort}{\textsc{Bucket Sort}}

\newcommand{\methodShortName}{$\textrm{DAPE}_{1\times3}$\xspace}

\title{DAPE V2: Process Attention Score as Feature Map for Length Extrapolation}

\author{%
\hypersetup{
    colorlinks=true,
    linkcolor=black, 
    urlcolor=black,
    citecolor=black,
}
\textbf{Chuanyang Zheng$^1$}\thanks{Contact Email: cyzheng21@link.cuhk.edu.hk}\textbf{, Yihang Gao$^2$, Han Shi$^3$, Jing Xiong$^4$, Jiankai Sun$^1$, Jingyao Li$^1$}\\
\textbf{~Minbin Huang$^1$, Xiaozhe Ren$^3$, Michael Ng$^5$, Xin Jiang$^3$, Zhenguo Li$^3$, Yu Li$^1$} \\
  \textsuperscript{1}CUHK~~~
  \textsuperscript{2}NUS~~~
  \textsuperscript{3}Noah’s Ark Lab~~~
  \textsuperscript{4}HKU~~~
  \textsuperscript{5}HKBU~~~\\
  \url{https://github.com/chuanyang-Zheng/DAPE}
}

\iclrfinalcopy 
\begin{document}

\maketitle

\begin{abstract}

The attention mechanism is a fundamental component of the Transformer model, contributing to interactions among distinct tokens, in contrast to earlier feed-forward neural networks. In general, the attention scores are determined simply by the key-query products. However, this work's occasional trial (combining DAPE and NoPE) of including additional MLPs on attention scores without position encoding indicates that the classical key-query multiplication may limit the performance of Transformers. 
In this work, we conceptualize attention as a feature map and apply the convolution operator (for neighboring attention scores across different heads) to mimic the processing methods in computer vision. Specifically, \textbf{the main contribution of this paper is identifying and interpreting the Transformer length extrapolation problem as a result of the limited expressiveness of the naive query and key dot product, and we successfully translate the length extrapolation issue into a well-understood feature map processing problem.} 
The novel insight, which can be adapted to various attention-related models, reveals that the current Transformer architecture has the potential for further evolution.  Extensive experiments demonstrate that treating attention as a feature map and applying convolution as a processing method significantly enhances Transformer performance.

\end{abstract}

\section{Introduction}

Transformer-based models \citep{vaswani2017attention} have delivered exceptional performances across widespread applications, including language processing \citep{zhang2020pegasus,guo2021longt5,ainslie2023colt5}, computer vision \citep{chen2024pixartalpha,peebles2023scalable}, quantitative research \citep{zhou2024dont,liu2021finbert,wu2023bloomberggpt}, and scientific machine learning \citep{taylor2022galactica,geneva2022Transformers}.
However, the quadratic cost of the key-query multiplication for processing a sequence raised much concern about the modern architecture of Transformers especially for long context inputs. To address the issue of storage and computation efficiency, recent research delves into developing more efficient architectures, such as sparse structural attention \citep{xiao2024efficient,zhu2024near}, adaptive key selection \citep{xiao2024infllm,fountas2024human}, and hybrid models~\citep{lieber2024jamba}. 
While these adaptations enhance efficiency, they often involve tradeoffs with model effectiveness.

At the same time, there is another voice advocating for refining the model design for tackling complex tasks, rather than prioritizing efficiency. Positional encoding is one of the key components of the attention mechanism. 
Although the widely recognized decoder-based Transformer can implicitly incorporate the positional information of tokens, growing evidence both theoretically and empirically shows that the well-designed explicit positional encoding significantly enhances the model performances, especially in long-context tasks~\citep{su2024roformer, press2021train,zhao2023length}. 
In practice, Transformers depend on positional encoding to explicitly incorporate positional information, enabling the model to make meaningful token predictions. Without these encodings, token generation would lack the necessary contextual order, rendering the outputs nonsensical.
The well-recognized RoPE~\citep{su2024roformer}, which is adopted in LLaMA~\citep{touvron2023llama}, distinguishes the token order by rotating with different angles depending on the token position. However, it demonstrated a notable performance degradation, failing entirely when the input length is double that of the training length \citep{peng2023yarn, chen2023clex,ding2024longrope}. The undesirable performance degradation is also observed for other positional encoding methods, e.g., ALiBi~\citep{press2021train} and  Kerple~\citep{chi2022kerple} . FIRE~\citep{li2023functional} alleviates the long-context extrapolation by learnable positional encodings, trying to capture the suitable positional representation by MLPs. 
However, a common characteristic among these positional encodings is their predefined and static nature. 
Specifically, they are fixed across various tasks and models, which may lead to their inability to adapt to varying input lengths and contexts effectively. 
Recently, the data-adaptive positional encoding method, namely DAPE~\citep{zheng2024dape}, which adjusts dynamically with context, enhances the length generalization by incorporating the attention scores and positional information with a more complex mechanism. 

In this paper, we propose that precise attention scores are crucial for improving Transformer length extrapolation, and we introduce a new perspective on the attention mechanisms. Traditionally, attention scores are computed through the dot product of the query and key vectors. As illustrated in Figure \ref{fig: dape_method_analysis}, further processing these attention scores using a neural network—a general case of DAPE~\citep{zheng2024dape}—can significantly enhance the length generalization of Transformers, even in the absence of positional encoding (NoPE). Therefore, we suggest treating attention scores as feature maps. By conceptualizing attention as an image feature map (with dimensions $[B, C, W, H]$ for batch size, channel size, width, and height), we can achieve more accurate attention scores by applying techniques used in image processing.
In this work, we employ different kernel sizes (such as 1×3) to process attention, finding that the perplexity (ppl) of attention decreases significantly—from over 600 to just above 100—when trained on a sequence length of 128 and evaluated on a length of 8192.

In summary, our contributions are as follows:
\begin{enumerate}
\item We highlight that the coarse attention mechanism, which is the direct result of the query and key dot product, limits the Transformer's ability to extrapolate to longer sequences. However, Transformers can achieve good length extrapolation performance with careful processing of attention scores.
\item By treating attention scores as feature maps and refining them using image processing techniques like convolution, we can enhance the Transformer's extrapolation capabilities.
\item  We conducted extensive experiments on language tasks to support our claims and believe that these insights can significantly improve the Transformer's performance in length extrapolation.
\end{enumerate}

\section{Related Works}

\paragraph{Absolute Positional Encoding} Absolute positional encoding (APE), introduced by \cite{vaswani2017attention}, enables Transformers to incorporate positional information. Specifically, at the first layer, each position $i$ is assigned a real-valued encoding $\ve_i \in \mathbb{R}^{d}$, which can be either learnable or a fixed sinusoidal encoding~\citep{vaswani2017attention, kiyono2021shape, likhomanenko2021cape, wang2020position, liu2020learning}, and this encoding is then added to the input sequence. Although this approach is straightforward, Transformers relying on APE tend to struggle with generalizing to longer sequences~\citep{press2021train}.

\paragraph{Relative Positional Encoding} Relative positional encoding (RPE) offers an alternative for embedding positional information~\citep{shaw2018self,raffel2020exploring,press2021train}. A widely used RPE method in large language models is rotary positional encoding (RoPE)\citep{su2024roformer, chowdhery2023palm, touvron2023llama}. To address length extrapolation challenges\citep{press2021train,kazemnejad2024impact}, positional interpolation (PI) has been introduced~\citep{chen2023extending} to extend the context window. Building on this approach, models like LongLora~\citep{chen2023longlora}, LongRope~\citep{ding2024longrope}, YaRN~\citep{peng2023yarn}, and CLEX~\citep{chen2023clex} have emerged. Another notable direction involves additive positional encoding. For most additive RPE techniques, the computation of pre-softmax attention logits can be expressed using the formula:
 $\mA_{\mathrm{RPE}}(\mX) = \mX \mW_Q(\mX \mW_K)^{\top}+\mB,$
where the bias matrix $\mB \in \R^{n\times n}$ is derived from the positional encoding function $b: \sN^{2}\to\R$, with the $(i,j)$-th entry of $\mB$ defined as $b(i,j)$. Different parameterizations of $b$ give rise to various RPE variants. Methods supporting arbitrary sequence lengths include T5's RPE~\citep{raffel2020exploring}, ALiBi~\citep{press2021train}, Kerple~\citep{chi2022kerple}, Sandwich~\citep{chi2023dissecting}, and FIRE~\citep{li2023functional}. Recently, DAPE~\citep{zheng2024dape} has been introduced, employing MLPs to dynamically adjust bias values based on the input data.

\paragraph{Data-Adaptive Related Positional Encoding.} Transformer-XL~\citep{dai2019Transformer} introduced the use of learnable query and key biases for adaptive positional encodings. Data-Adaptive Positional Encoding (DAPE)\citep{zheng2024dape} extends this idea by leveraging MLPs to adjust positional encodings based on attention over the head dimension for length extrapolation, ensuring different input data receive unique positional encodings. Contextual Positional Encoding\citep{golovneva2024contextual} further refines this by conditioning position increments on specific tokens, as determined by the model, allowing positions to adapt based on context."

\section{Method}
In this section, we first review the previously developed Data-Adaptive Positional Encoding method (DAPE), which incorporates attention scores and positional information through MLPs. As a proof-of-concept, our occasional trial on DAPE without the positional information (as shown in Figure \ref{fig: dape_method_analysis}) suggests that regarding attention as a feature map and processing it with classical operators (e.g., convolution) can enhance the Transformers' behavior. As discussed in some previous works the perplexity scores come mostly from the associative recall (i.e., copy) tasks. In addition, we theoretically show by construction that the proposed method can explicitly realize the associative recall task, in contrast to the implicit conduct through positional encoding in standard Transformers. \textbf{The two key differences between DAPE \citep{zheng2024dape} and this work are: 1) Insight:} DAPE attributes length extrapolation performance gains to adaptive position encoding, while this work finds DAPE could still improve performance without position encoding so that we take a broader view, explaining that the Transformer's length extrapolation ability is limited by the expressiveness of the naive query-key dot product, which can be enhanced using image processing techniques; \textbf{2) Performance:} As shown in Figure \ref{fig: dape_method_analysis}, DAPE is designed for additive RPE and may underperform with non-additive RPE (e.g., RoPE), whereas this work suggests that increasing kernel size (e.g., with \methodShortName) may improve RoPE's performance. The \methodShortName implementation is shown in Appendix \ref{appendix: implementation}.

\subsection{Additive Relative Positional Encoding} 
For most additive relative positional encoding (ARPE) methods, the computation of pre-softmax attention logits can be unified under the following formula:
\begin{equation}
    \label{eq:rpe-attn-mat}
    \mA_{\mathrm{ARPE}}(\mX) = \mX \mW_Q(\mX \mW_K)^{\top}+\mB,
\end{equation}
where the bias matrix $\mB\in\R^{n\times n}$ is induced by the position encoding function $b: \sN^{2}\to\R$ and the $(i,j)$-th entry of $\mB$ is defined as $b(i,j)$. Various formulations and parameterizations of $b$ give rise to different variants of RPE. Examples of additive RPE include:
(1) ALiBi: $b(i,j) = -r|i-j|$, with the scaler $r>0$ as a hyper-parameter; 
(2) Kerple: $b(i,j)=-r_1 log(1+r_2|i-j|)$ with $r_1$ and $r_2$ are two learnable parameters;
(3) FIRE: $b(i,j) = f_{\theta}\left(\frac{\psi(i-j)}{\psi(\max\{L, i\})}\right)$, where the positional encoding function $f_{\theta}$ parameterized by $\theta$ is learned from data and $\psi$ is a transformation function aimed at assigning more model capacity to local positions. 

\paragraph{Data-Adaptive Position Encoding (DAPE)}
The DAPE rewrite the Equation \ref{eq:rpe-attn-mat} as the following:
\begin{equation}
    \label{eq:DAPE-attn-mat}
    \mA_{\mathrm{DAPE}}(\mX) = \mX \mW_Q(\mX \mW_K)^{\top} + f( \mX \mW_Q(\mX \mW_K)^{\top},\mB).
\end{equation}
Here, $f: \mathbb{R}^{T \times T} \times  \mathbb{R}^{T \times T} \to  \mathbb{R}^{T \times T}$ is an element-wise function and $T$ is the sequence length. Another variant of DAPE is with residual, which is the following:
\begin{equation}
\label{eq:DAPE-attn-mat_concatenation}
\mA_{\mathrm{DAPE}}(\mX) = \mX \mW_Q(\mX \mW_K)^{\top} + \mB + f(\mX \mW_Q(\mX \mW_K)^{\top},\mB).
\end{equation}
In practice, DAPE \citep{zheng2024dape} utilizes a two-layer \textit{LeakyReLU} MLP with hidden dimension $D_{\mathrm{DAPE}}$ (default value is 32) to parameterize $f(\cdot)$ due to its universal approximability \citep{leshno1993multilayer}.
All parameters are learned directly from the data during the training process. This architecture allows $f(\cdot)$ to dynamically adjust positional embeddings based on the input sequence data, ensuring that the encoding method is both adaptive and dependent on the input data.

\subsection{Special Case of DAPE: Bias is Zero}
\begin{figure}[t]
\vspace{-5pt}
\setlength{\abovecaptionskip}{0.1cm}
\centering
\includegraphics[width=0.45\textwidth]{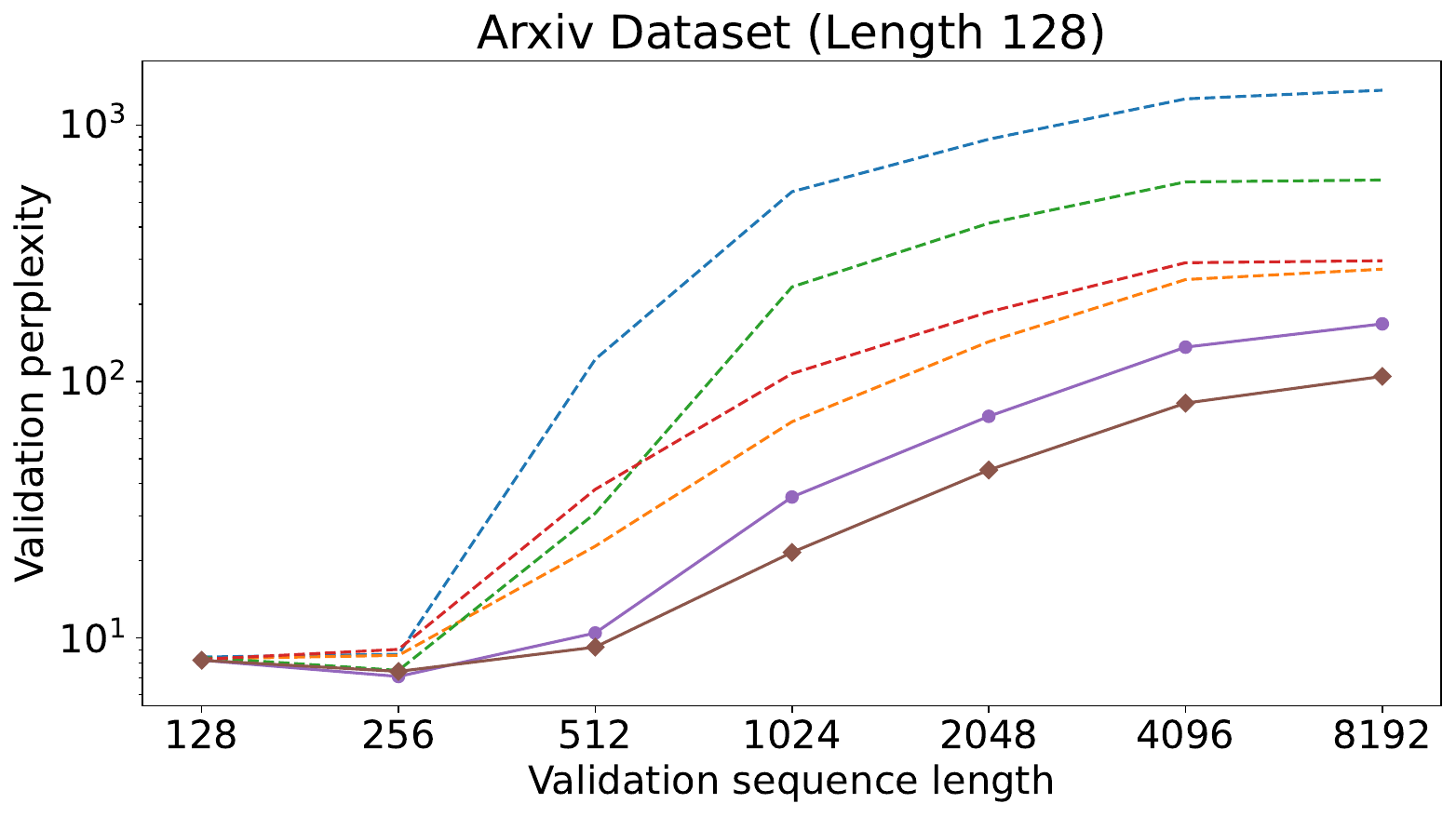}
\hspace{0in}
\includegraphics[width=0.45\textwidth]{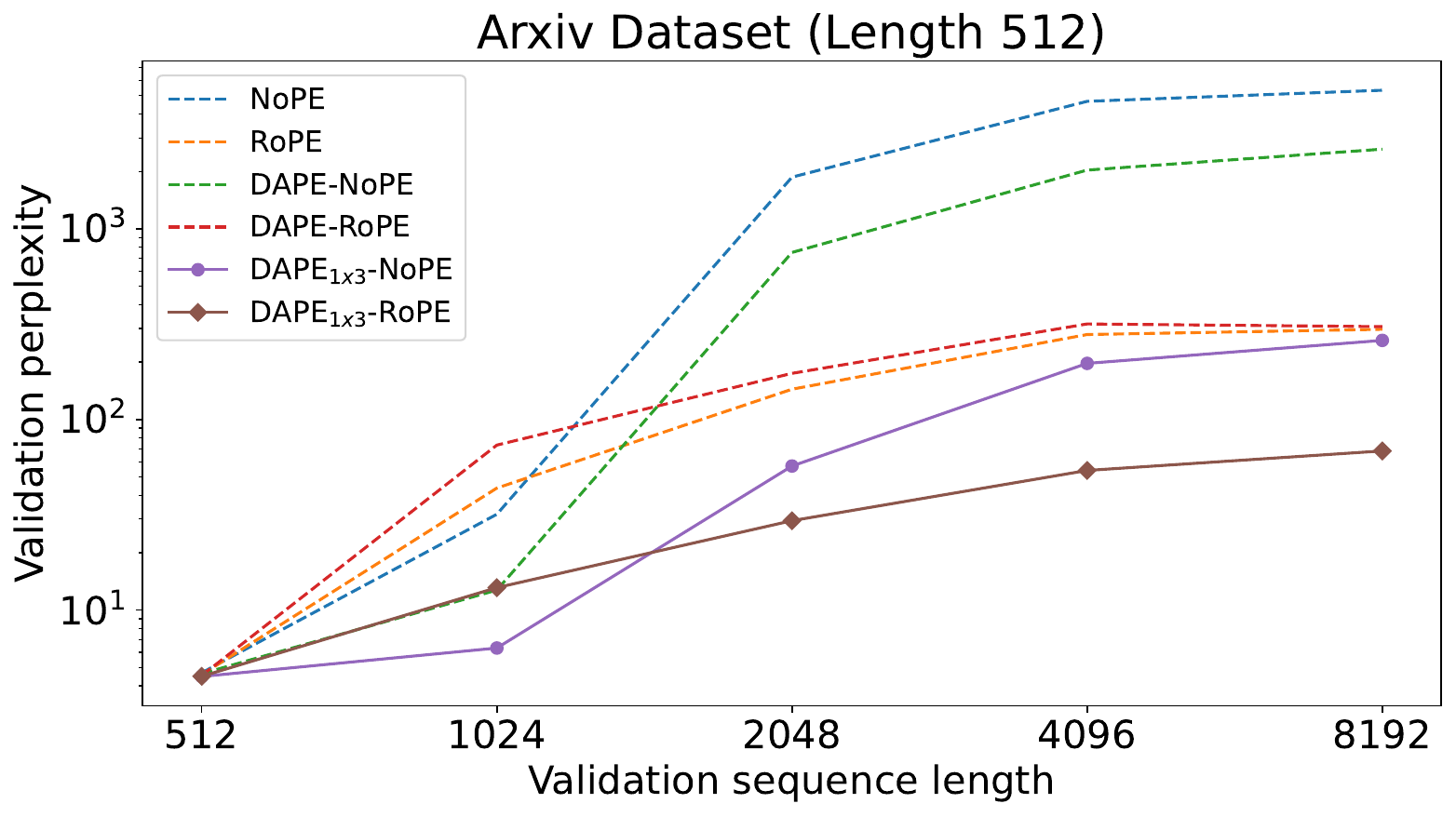}
\caption{
\small
\textbf{The result of DAPE \citep{zheng2024dape} (equivalent to kernel $1\times1$ in our explanation) and \methodShortName (kernel $1\times3$ by this work), with baseline NoPE and RoPE.} The model is trained with length 128 and length 512 respectively. The \methodShortName denotes that we use $H \times 1\times3$  convolutions kernel size on the attention score with shape $[B, H, T, T]$. \textbf{We find that DAPE can even improve the performance of NoPE (without biased position encoding), suggesting that the explanation in \cite{zheng2024dape}, which attributes the improvement to adaptive position encoding, may have a more general underlying cause.}
}
\label{fig: dape_method_analysis}
\vspace{-10pt}
\end{figure}

DAPE was originally designed to dynamically adjust the positional encoding by incorporating input data information. Generally, any additive positional encoding method that includes positional information can be represented as the matrix $\bm{B}$ in the DAPE model, as outlined in Equation \ref{eq:DAPE-attn-mat}. Notably, No Positional Encoding (NoPE)~\citep{kazemnejad2024impact} is a special case of additive RPE that assigns zero value to the matrix $\bm{B}$. The mathematical formulation of DAPE equipped with NoPE is given by:
\begin{equation}
    \label{eq:DAPE-attn-mat_nope}
    \mA_{\mathrm{DAPE}}(\mX) = \mX \mW_Q(\mX \mW_K)^{\top} + f( \mX \mW_Q(\mX \mW_K)^{\top}).
\end{equation}
\paragraph{The result of DAPE-NoPE (the \cite{zheng2024dape} only combine DAPE with ALiBi, Kerple and FIRE but not with NoPE or RoPE).} Compared with the standard Transformer architecture, DAPE-NoPE introduces additional MLPs post the key-query multiplication and prior to the softmax operator. 
As shown in Figure \ref{fig: dape_method_analysis}, experimental evidence suggests that DAPE with NoPE significantly outperforms the basic NoPE, prompting a reconsideration of the behaviors of standard Transformers. 
The additional MLPs (i.e., denoted as $f(\cdot)$ in Equation \ref{eq:DAPE-attn-mat_nope}) facilitate information sharing across attention heads and complicate the attention calculation with nonlinear transformation beyond the simple key-query multiplication. 
This leads to a critical question: \textit{Is the current Transformer architecture, particularly the attention mechanism, sufficiently expressive for real-world language tasks?} 
Although numerous studies aim to enhance efficiency by reducing computation and storage in standard Transformers, these often come at the cost of effectiveness, potentially hindering the evolution of next-generation Transformer models.
Motivated by these insights and observations, we enhance the Transformer's expressiveness and behavior by regarding attention as a feature map and applying convolutional operations, akin to those used in computer vision.

\paragraph{The result of DAPE-RoPE.}Building on the hypothesis that DAPE enhances Transformer performance by processing pre-softmax scores with MLPs, we explore its applicability to non-additive positional encoding methods, specifically RoPE~\citep{su2024roformer}. In the DAPE-RoPE configuration, RoPE first computes the classic attention scores of key-query multiplication with RoPE, which are then refined using the MLPs described in Equation \ref{eq:DAPE-attn-mat_nope}. The visualized results of the validation perplexity for DAPE-RoPE and other positional encoding methods are presented in Figure \ref{fig: dape_method_analysis}. The results indicate that DAPE-RoPE may degrade the performance, while  \methodShortName-RoPE (with kernel size $1\times3$, propsoed by this work) not only improves overall performance but also excels in length extrapolation tasks, particularly at larger sequence lengths. This finding substantiates the effectiveness of \methodShortName-RoPE, confirming its superior performance compared to standard RoPE, attributing to the additionally introduced convolution operations to the attention scores.

\subsection{DAPE V2: Process Attention Scores as Feature Maps}
As discussed above, improving Transformer performance necessitates refining the processing of attention score computation beyond the conventional key-query multiplication. 
We propose regarding the pre-softmax attention scores as feature maps (4-dimensional tensors) and applying convolutional operators. 
This approach facilitates enhanced communication across neighboring tokens and heads, drawing parallels to popular techniques used in computer vision.  
This novel method aims to leverage the spatial relationships within tokens, potentially unlocking new aspects of model capabilities.

\paragraph{Rethink the DAPE formulation.}  In DAPE~\citep{zheng2024dape}, MLPs are utilized to process and integrate attention and biases. Notably, these MLP operations can be equated to convolution operations with $1 \times 1$ kernel ~\citep{krizhevsky2012imagenet,simonyan2014very,he2016deep}, a stride of one, and no padding. Consequently, we can reformulate the DAPE in Equation \ref{eq:DAPE-attn-mat_concatenation} as the following: 
\begin{equation}
    \label{eq:DAPE-attn-mat_nope_conv}
    \mA_{\mathrm{DAPE}}(\mX) = \mX \mW_Q(\mX \mW_K)^{\top} + \mB+Conv(tril(( \mX \mW_Q(\mX \mW_K)^{\top},\mB)).
\end{equation}
Under such formulation, DAPE employs convolution operation to process the pre-softmax attention scores of key-query multiplication. The $\texttt{tril}(\cdot)$ returns the lower triangular part of the matrix and the other elements of the result tensor out are set to 0. The resulting attention tensor has a shape of \( [B, H, T, T] \), where the four dimensions correspond to the batch size, number of heads, and the context length for both the query and key. This mirrors the structure of an image feature tensor with shape \( [B, C, H, W] \), where the dimensions represent the batch size, number of channels, image height, and image width, respectively. This structural similarity underscores the feasibility of considering attention scores as a tensor of feature mappings, where popular and effective convolution operations can be leveraged for refined processing.

\paragraph{Process attention with more powerful convolution operation.}
In computer vision, the limitations of $1 \times 1$ kernels for processing image features are well-recognized. To improve upon the attention scores processed by these kernels (e.g., DAPE), we introduce $1 \times k$ kernels with a stride of 1 and padding of $k-1$. This approach allows for wider and deeper convolution across key dimensions and heads without information leakage, as we ensure the attention scores remain lower-triangular. This mechanism is visualized in Appendix \ref{more DAPE visualization}.
The use of $1 \times k$ kernels suggests a targeted convolution along the key dimensions across heads. In general, while extending this to include the query dimensions as a standard kernel is theoretically possible, it would significantly increase computational demands. Our forthcoming analysis demonstrates that Transformers modified with $1 \times k$ convolution are adept at associative recall tasks (i.e., the copy task), validating the benefits of integrating convolution in attention calculation. We left as a future work investigating the performances and the soundness of general convolution kernels, such as square sizes. \textbf{The key contribution of this work is providing a novel insight that suggests applying convolution operations and processing attention as feature maps to improve Transformers' performances.}

\paragraph{Realizing associate recall tasks through convolution.} As pointed out in some previous works \citep{arora2023zoology}, the perplexity scores of Transformers mostly result from the performances on associate recall tasks (i.e., the copy tasks). Numerous studies have explored the mechanism of associative recall within Transformers, both from theoretical perspectives and experimental validations \citep{arora2023zoology,bietti2024birth,golovneva2024contextual}. Here, we theoretically prove that the proposed model can realize the associative recall tasks. Notably, this capability is achieved independently of positional encodings, marking a significant advancement in the flexibility and applicability of the proposed architecture. By integrating convolutional operations, we enable the model to handle associative tasks more effectively, leveraging spatial relationships inherent in the data, similar to methods used in image processing.
To explain the associative recall mechanism, \citep{bietti2024birth} proved that the first layer of the Transformer is responsible for the previous token mechanism through the positional encoding. More specifically, given a sequence of input tokens $\bm{X} = [\bm{x}_1, \bm{x}_2, \cdots, \bm{x}_{N}]$ with corresponding orthogonal positional encoding vectors $[\bm{p}_1, \bm{p}_2, \cdots, \bm{p}_{N}]$, the first layer primarily facilitates the copying of the previous token to the current token (e.g., $\bm{x}_{i}+ \bm{W}_{V}^{1} \bm{x}_{i-1}$, where $\bm{W}_{V}^{1}$ is the value matrix at the first layer of the Transformer). The input tokens are combined with positional encodings $\bm{x}_{i}+\bm{p}_{i}$ and the key-query weight matrix is defined as $\bm{W}_{K}^{1 \top}\bm{W}_{Q}^{1} = \sum_{i=1}^{N} \bm{p}_{i-1} \bm{p}_{i}^{\top}$. The orthogonality of positional encoding vectors and the special choices of the key-query matrix ensure that attention scores predominantly focus on the previous token. 
In contrast to this implicit mechanism in standard Transformers, our proposed method leverages a convolution operation to explicitly realize associative recall. This approach not only simplifies the process but also enhances its effectiveness by directly manipulating the spatial relationships within tokens and attention scores.
Consider a scenario where the word ``Hakuna" is consistently followed by ``Matata" within a lengthy paragraph.  Without the loss of generality, we assume that $\bm{x}_1$ and $\bm{x}_2$ represent the tokens of “Hakuna” and “Matata” respectively, and $\bm{x}_{N} = \bm{x}_1$ implies that the N-th token in the sequence is “Hakuna”. Then we expect that the Transformer can predict and output the next token $\bm{x}_{N+1}$ as “Matata”. For simplicity, we consider a one-head Transformer without positional encoding. We employ a convolution operation with a kernel size of $1 \times 2$ and weights $[-1, 1]$. 
Note that the convolution is linear and processing the attention scores along the key dimensions is effectively equivalent to applying convolutions directly to the key vectors themselves. Consequently, the key vector of $\bm{x}_2$ can be expressed as $\bm{W}_{K}^{1}\left(\bm{x}_2 - \bm{x}_1 \right)$ and the query vector for $\bm{x}_{N}$ admits $\bm{W}_{Q}^{1}\bm{x}_{N}$. By configuring the matrix $\bm{W}_{K}^{1 \top} \bm{W}_{Q}^{1}$ to be $- \bm{I}$, the attention mechanism after the convolution predominantly allocates the attention values of $\bm{x}_{N}$ to the token  $\bm{x}_{2}$. This ensures that the token values of $\bm{x}_2$ are effectively copied to $\bm{x}_{N}$, resulting in the model outputting ``Matata" following ``Hakuna".

\begin{proposition}
    Transformers incorporating convolution operations can perform associative recall tasks without the need for positional encoding.
\end{proposition}

\paragraph{Comparisons with hybrid models of convolution and Transformers.} Recent developments in hybrid architectures have seen the integration of convolutional and Transformer models to capitalize on the strengths of both. For instance, 
\cite{fu2022hungry} introduced the FlashConv layer, which combines the efficiency of State Space Models (SSMs) with the capabilities of attention-based models. Similarly,  \cite{arora2023zoology} developed a gated convolution layer, noted for its effectiveness in addressing associative recall tasks. These models typically stack convolution layers directly with standard Transformer layers, resulting in modifications to the token values through convolution. In contrast, our model adopts a distinctive approach by applying convolution along the key dimension during the computation of attention scores. This method preserves the original token values while still leveraging the convolution's benefits for processing attention.

\section{Experiment}
\paragraph{Baselines.}
We evaluate the proposed \methodShortName against several well-established baselines, including NoPE~\citep{kazemnejad2024impact}, RoPE~\citep{su2024roformer}, T5's Bias~\citep{raffel2020exploring}, ALiBi~\citep{press2021train}, Kerple~\citep{chi2022kerple}, FIRE~\citep{li2023functional}, CoPE~\citep{golovneva2024contextual}, and DAPE~\citep{zheng2024dape}. As our kernels are applied across all heads, we simplify by omitting the kernel size description at the head dimension. For example, \methodShortName indicates the use of a $H \times 1 \times 3$ convolution kernel size on the attention scores, with a shape of $[B, H, T, T]$.

\paragraph{Datasets.}
Our analysis is based on training language models using the Arxiv and Books3 datasets, commonly employed benchmarks for assessing model performance~\citep{press2021train,chi2022kerple,li2023functional,ding2024longrope}. We begin our evaluation by processing entire sequences and comparing the zero-shot perplexity of the last 256 tokens across various input lengths. In addition to perplexity, we also leverage downstream datasets with randomized positional encoding~\citep{ruoss2023randomized} to further assess \methodShortName.
\paragraph{Experiment settings.}
Initially, we compare \methodShortName with other baselines at training lengths of 128, 512, and 1024, using 125M decoder-only Transformers~\citep{brown2020language}, with model configurations detailed in Appendix \ref{model configuration details}. Subsequently, we evaluate the performance of different training lengths using the same number of training tokens but with larger model sizes (350M and 2.7B). We also explore the impact of the convolutional hidden dimension $D_{\text{DAPE}}$, the effect of information leakage, and the influence of varying kernel sizes. Additionally, we examine the computational efficiency of \methodShortName, focusing on processing times. Lastly, we evaluate \methodShortName on algorithmic reasoning datasets using accuracy metrics. Compared to DAPE~\citep{zheng2024dape}, \methodShortName demonstrates a more pronounced attention sink~\citep{xiao2024efficient}, as visualized in Appendix~\ref{more DAPE visualization}.

\subsection{Compare with Baselines}
\begin{figure}[htbp]
\setlength{\abovecaptionskip}{0.1cm}
\centering
\includegraphics[width=0.45\textwidth]{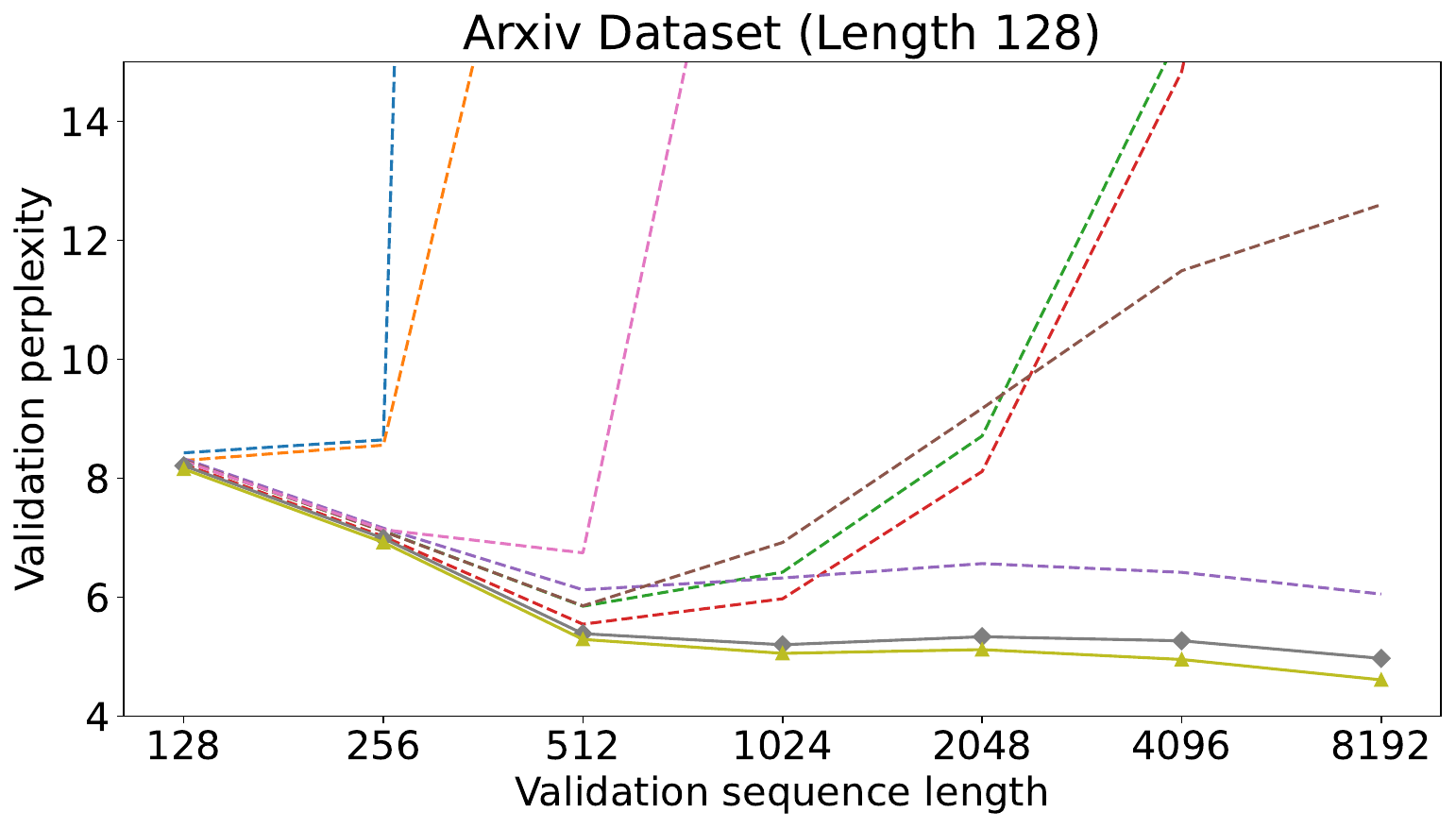}
\hspace{0in}
\includegraphics[width=0.45\textwidth]{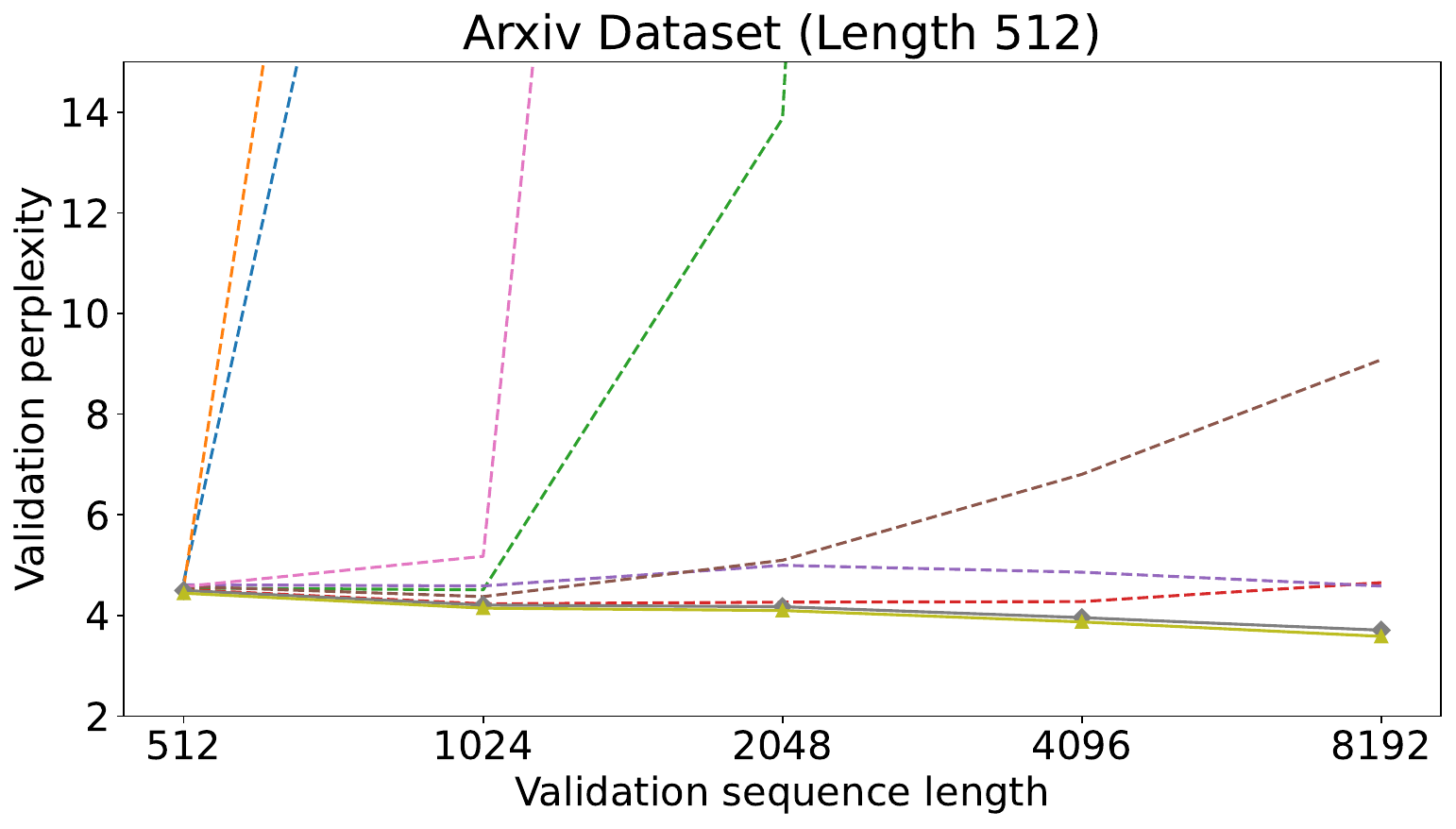}
\centering
\includegraphics[width=0.45\textwidth]{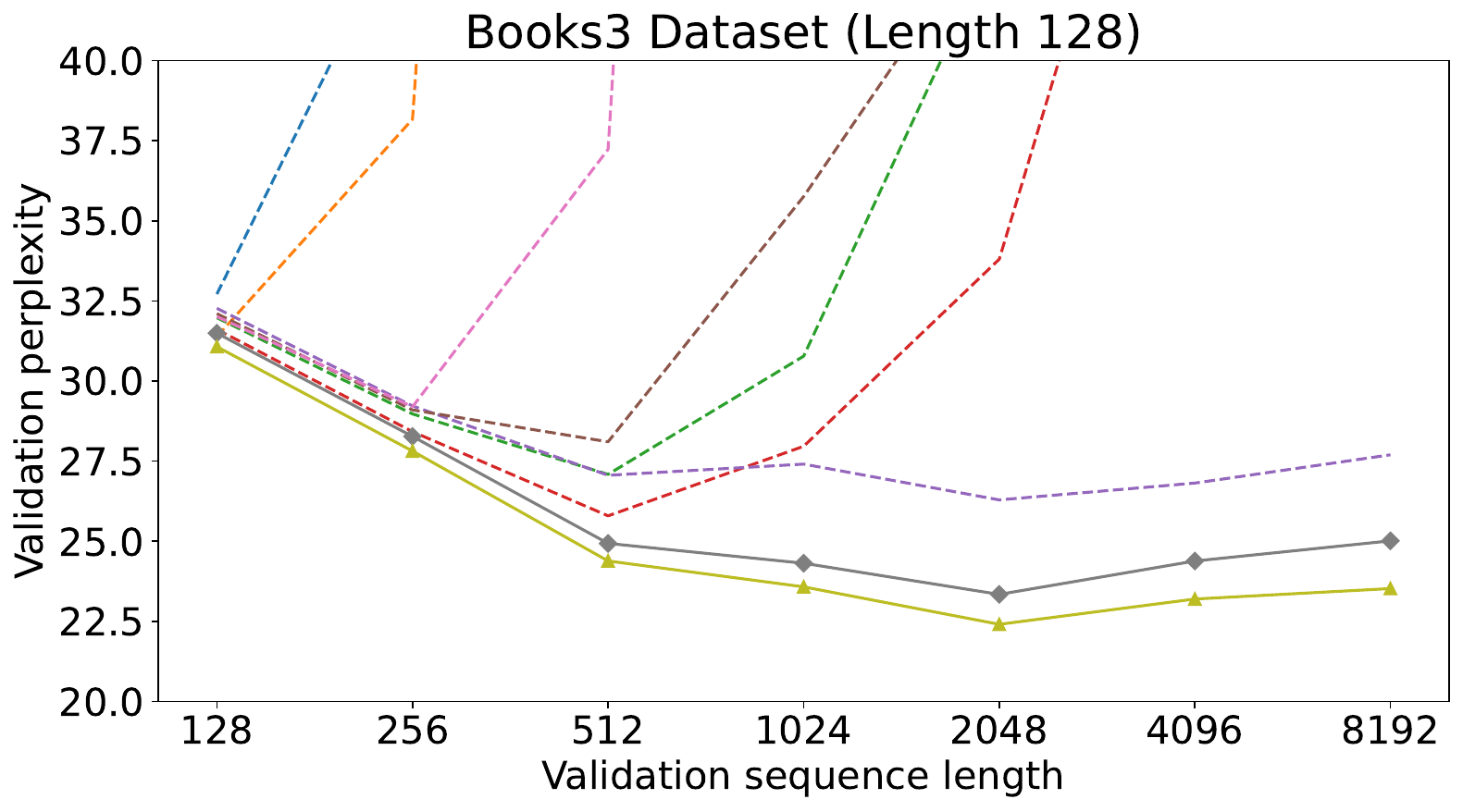}
\hspace{0in}
\includegraphics[width=0.45\textwidth]{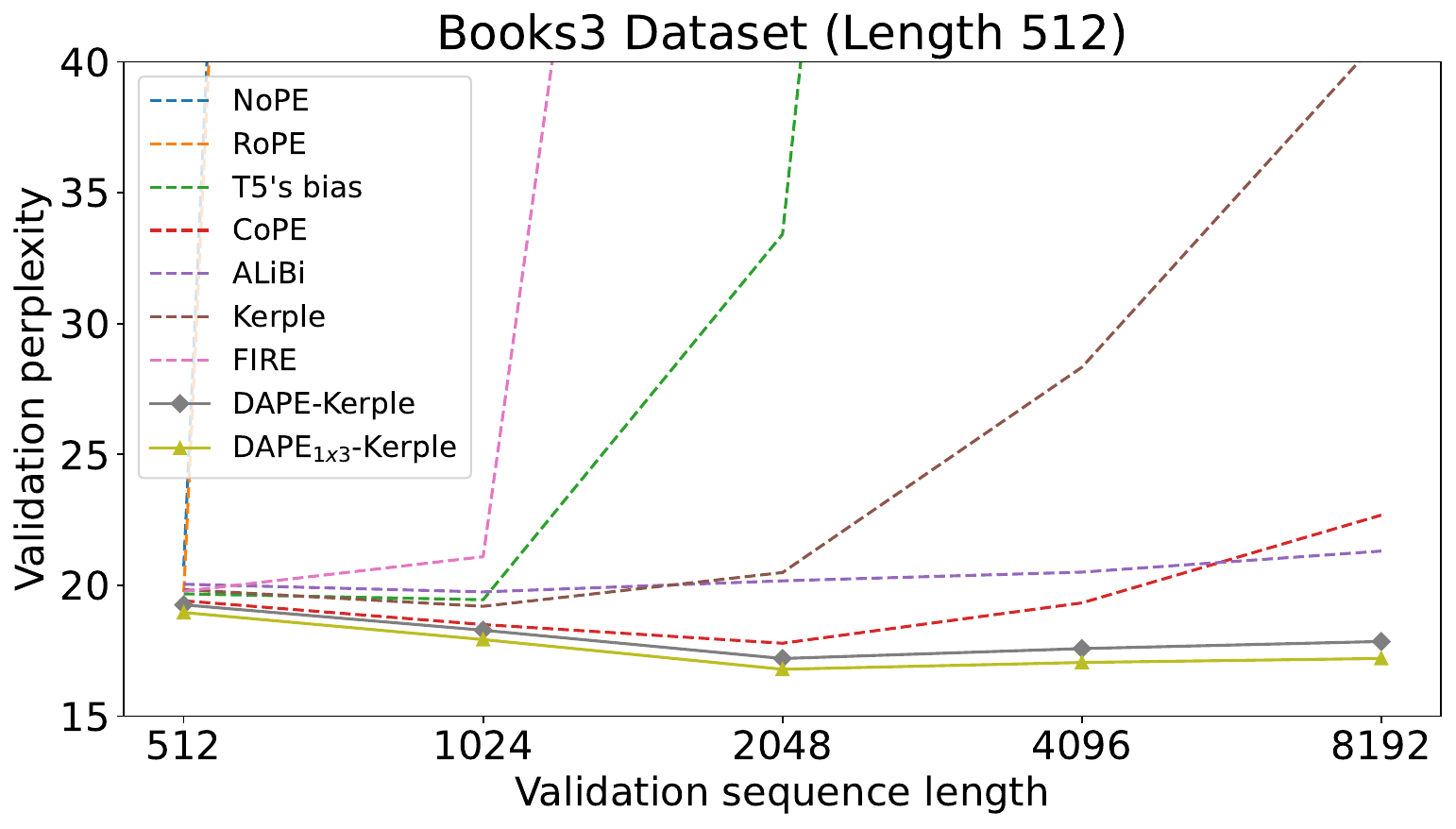}
\caption{
\small
\textbf{Comparisons with baselines:} performance with training lengths 128 and 512 on Arxiv and Books3 datasets.
}
\label{fig: compare_with_baseline}
\vspace{-12pt}
\end{figure}

\paragraph{\methodShortName-Kerple improves performance within training length, proving its ability to process the entire sequence.} According to Figure \ref{fig: compare_with_baseline}, the proposed \methodShortName-Kerple demonstrates superior performance across various training and evaluation lengths. Specifically, \methodShortName-Kerple achieves the best performance where the training length is 128 or 512 and the evaluation length ranges from 128 to 8192. This performance consistency is observed across both the arXiv and Books datasets.
For instance, on the arXiv dataset with a training length of 512, \methodShortName-Kerple achieves a perplexity score of 4.44. This score surpasses those of other methods, such as DAPE-Kerple with a perplexity of 4.49, CoPE with 4.51, Kerple with 4.57, and RoPE with 4.57. These results indicate that \methodShortName-Kerple has a more robust modeling capability within the training length compared to the other methods evaluated. The Appendix \ref{appendix: length 1024} also presents the performance of different methods with training length 1024. The improvements are not only significant but also consistent, reinforcing the efficacy of the \methodShortName-Kerple approach in handling various training lengths effectively.

\paragraph{\methodShortName-Kerple improves performance beyond training length.} The advantages of \methodShortName-Kerple extend beyond the training length. When the training length is set to 128 and the evaluation length is extended to 8192, \methodShortName-Kerple achieves a perplexity score of 4.60 on the arXiv dataset and 23.52 on the Books3 dataset. These scores are significantly better than those achieved by DAPE-Kerple, which records perplexity scores of 4.97 and 25.01 on the arXiv and Books3 datasets, respectively. Similarly, CoPE performs poorly with perplexity scores of 29.86 on the arXiv dataset and 90.66 on the Books3 dataset under the same conditions.
Furthermore, when the training duration is increased to 512, \methodShortName-Kerple continues to deliver the best performance, further validating its superior generalization capabilities. These findings highlight the scalability and robustness of \methodShortName-Kerple, which is attributed to the introduced convolution operator, making it a promising approach for diverse data scenarios and lengths.

\subsection{Performance with Same Training tokens and Different Training Length}
\begin{figure}[htbp]
\setlength{\abovecaptionskip}{0.1cm}
\centering
\includegraphics[width=0.45\textwidth]{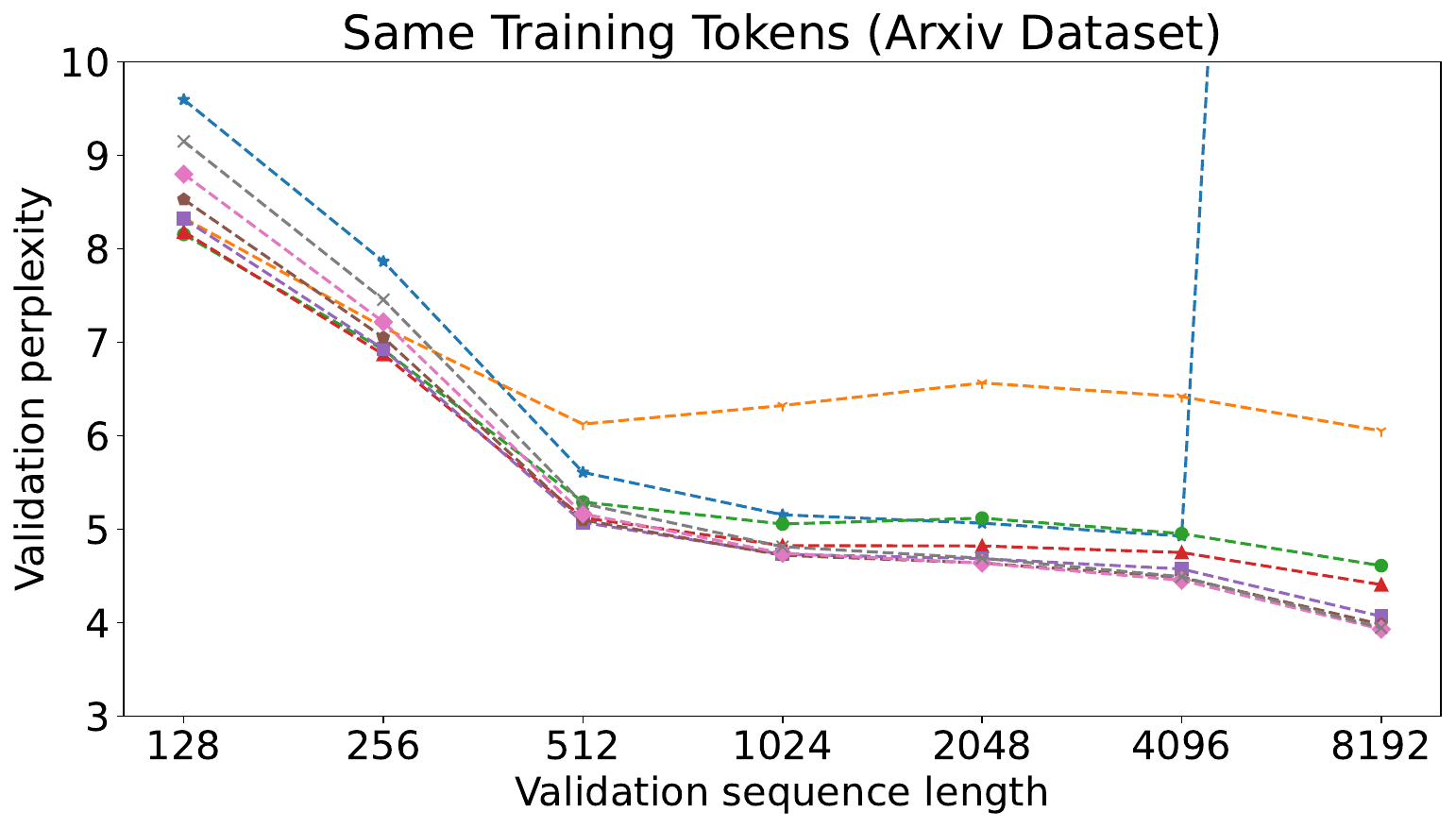}
\hspace{0in}
\includegraphics[width=0.45\textwidth]{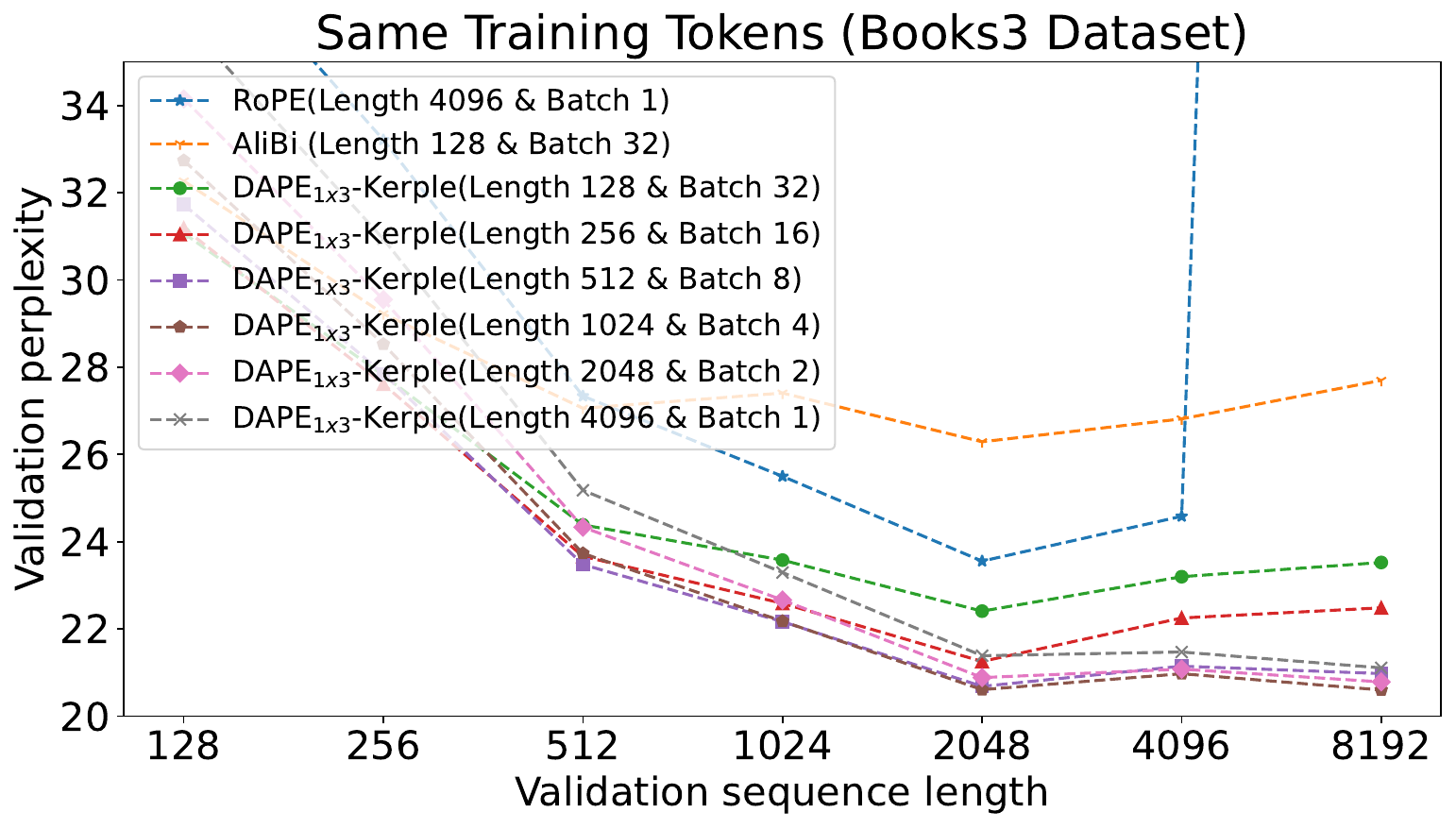}
\caption{
\small
\textbf{The performance with same training tokens and different training length}. With the same training tokens, \methodShortName with training length 512 could even achieve better performance than RoPE with training length 4096.
}
\label{fig: same_tokens}
\vspace{-10pt}
\end{figure}
\paragraph{Compared to RoPE, with the same training tokens, \methodShortName-Kerple with a training length of 128 achieves performance comparable to RoPE with a training length of 4096, for varying evaluation length.} As shown in Figure \ref{fig: same_tokens}, for \methodShortName-Kerple trained with a length of 128, it achieves a perplexity (ppl) of 8.15 at an evaluation length of 128 and 4.95 at an evaluation length of 4096 on the arXiv dataset. In comparison, RoPE trained with a length of 4096 achieves a ppl of 9.59 at an evaluation length of 128 and 4.92 at an evaluation length of 4096. Similarly, on the Books3 dataset, \methodShortName-Kerple trained with a length of 128 achieves a ppl of 31.07 at an evaluation length of 128 and 23.19 at an evaluation length of 4096, while RoPE trained with a length of 4096 achieves 38.36 and 24.58, respectively. This suggests the superiority of the proposed \methodShortName with the introduced convolution operators among heads and neighboring tokens.

\paragraph{With the same training tokens, compared to \methodShortName with longer training lengths, \methodShortName with shorter training lengths can achieve comparable performance, indicating that \methodShortName enhances the model's understanding of text structure.} On the arXiv dataset, \methodShortName-Kerple with training lengths of 512 demonstrates performance close to that of training with a length of 4096 when the evaluation length is 4096. Moreover, the performance curves for training lengths of 1024, and 2048 are almost identical. This trend is also observed with the Books3 dataset. These results indicate that \methodShortName-Kerple effectively helps the model comprehend text structure, enabling it to extend to longer lengths.

\paragraph{Transformers may overfit their training length: training on longer sequences may decrease performance when testing on shorter sequences.} On the arXiv dataset, \methodShortName-Kerple with a training length of 128 achieves the best performance when the evaluation length is 128. Similarly, \methodShortName-Kerple with training lengths of 256, 512, 1024, and 2048 achieves the best performance at evaluation lengths of 256, 512, 1024, and 2048, respectively. Also, on evaluation 128, the RoPE with training length 4096 and batch size 1 also achieves worse performance than the RoPE with training length 128 and batch size 32. This suggests that training on longer sequences may worsen a Transformer's performance at shorter sequence lengths.

\paragraph{\methodShortName can reduce the training time cost via larger batch size and shorter training length, achieving comparable performance compared to trained on longer length.} The cost of \methodShortName is $\mathcal{O}(B \cdot (h \cdot d \cdot T^2 + h \cdot D_{\text{DAPE}} \cdot T^2))$, where $B$, $h$, $d$, $T$ and $D_{\text{DAPE}}$ are the batch size, attention hidden dimension, attention head number, sequence length and DAPE hidden dimension. By reducing the training length from $T$ to $\frac{T}{K}$ and increasing the batch size from $B$ to $B \cdot K$ with the same training tokens, the cost becomes $\mathcal{O}(B \cdot K \cdot (h \cdot d \cdot (\frac{T}{K})^2 + h \cdot D_{\text{DAPE}} \cdot (\frac{T}{K})^2))$, which simplifies to $\mathcal{O}(\frac{B \cdot (h \cdot d \cdot T^2 + h \cdot D_{\text{DAPE}} \cdot T^2)}{K})$. For example, when the training length is 128 and the batch size is 32, the time cost of one step is 40.30ms. The time cost of length 256 (batch 16), length 512 (batch 8), length 1024 (batch 4), and length 2048 (batch 2) are 42.61ms, 50.38ms, 79.36ms, and 120.14ms.  This reduction demonstrates the potential for significant training time savings.

\subsection{The Effect of Larger Model Size}
\begin{figure}[htbp]
\vspace{-5pt}
\setlength{\abovecaptionskip}{0.1cm}
\centering
\includegraphics[width=0.45\textwidth]{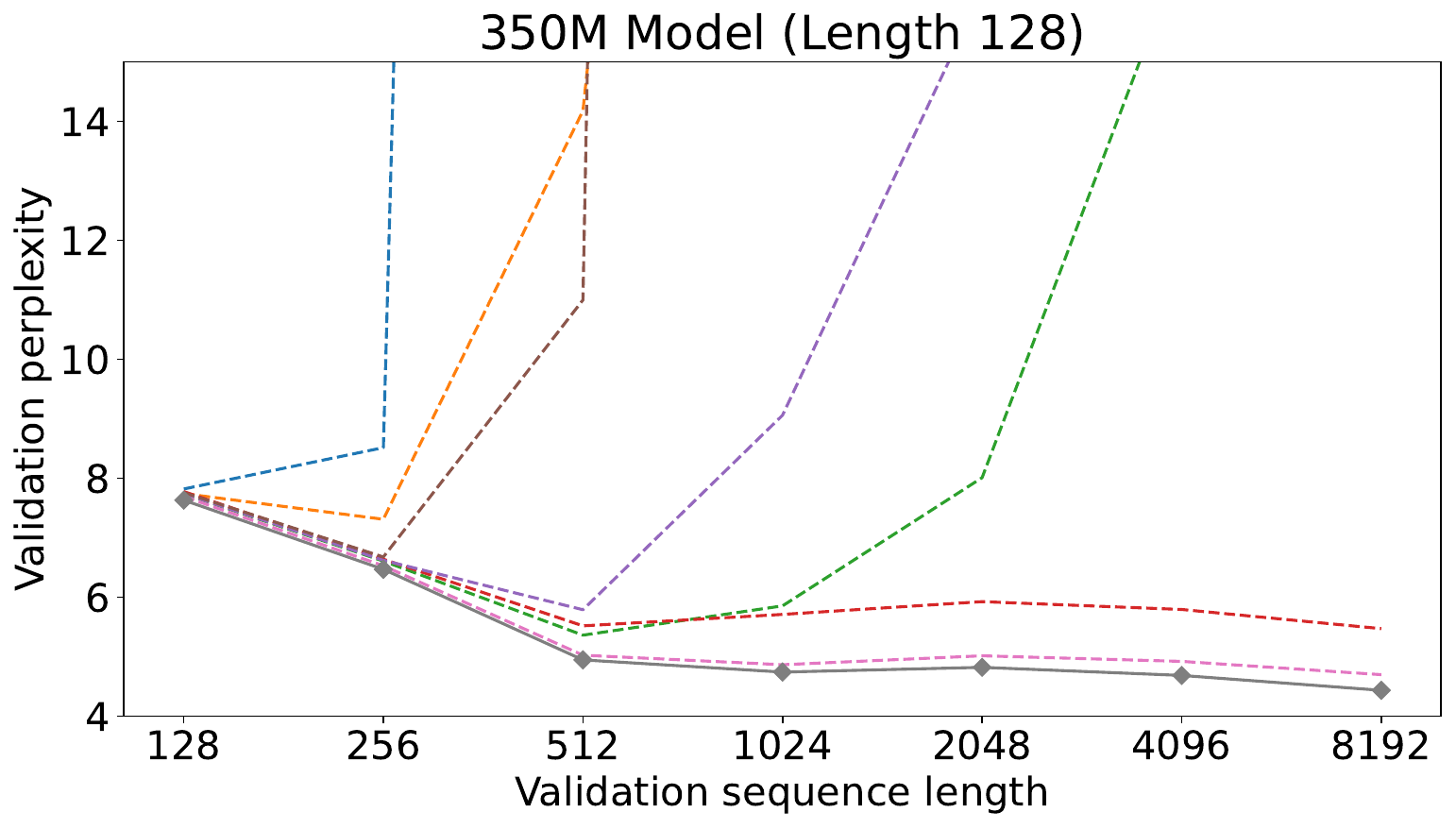}
\hspace{0in}
\includegraphics[width=0.45\textwidth]{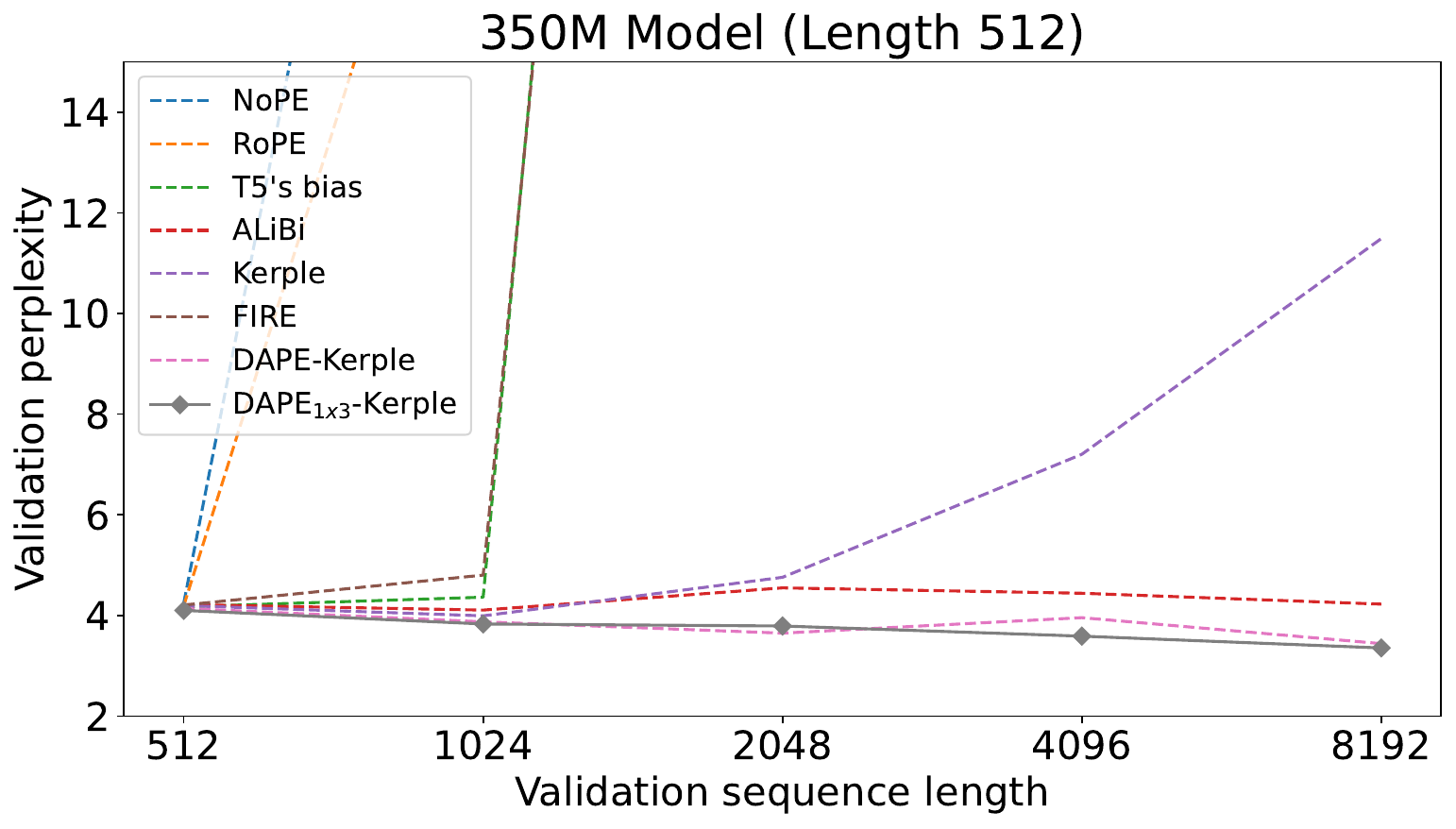}
\caption{
\small
\textbf{The Effect of Larger Model Size 350M}. We show the results with training length 128 and training length 512 on Arxiv dataset.
}
\label{fig: 350M model size}
\vspace{-12pt}
\end{figure}
\paragraph{\methodShortName performs well with larger model sizes, such as 350M and 2.7B.} As illustrated in Figure \ref{fig: 350M model size}, the proposed \methodShortName shows superior performance at varying evaluation lengths with a model size of 350M. For a training length of 128, \methodShortName-Kerple achieves a perplexity (ppl) of 7.63 at an evaluation length of 128 and 4.43 at an evaluation length of 8192, compared to DAPE's 7.69 and 4.69, respectively. Similarly, for a training length of 512, \methodShortName-Kerple achieves a ppl of 4.10 at an evaluation length of 128 and 3.35 at an evaluation length of 8192, whereas DAPE achieves 4.14 and 3.44, respectively. We also present the 2.7B model size result in Appendix \ref{appendix: large model size result}. Therefore, the proposed \methodShortName demonstrates excellent performance with larger model sizes, showing the potential of including the proposed processing techniques in existing large language models.

\subsection{The Effect of \methodShortName}
\begin{figure}[!htbp]
\setlength{\abovecaptionskip}{0.1cm}
\centering
\includegraphics[width=0.48\textwidth]{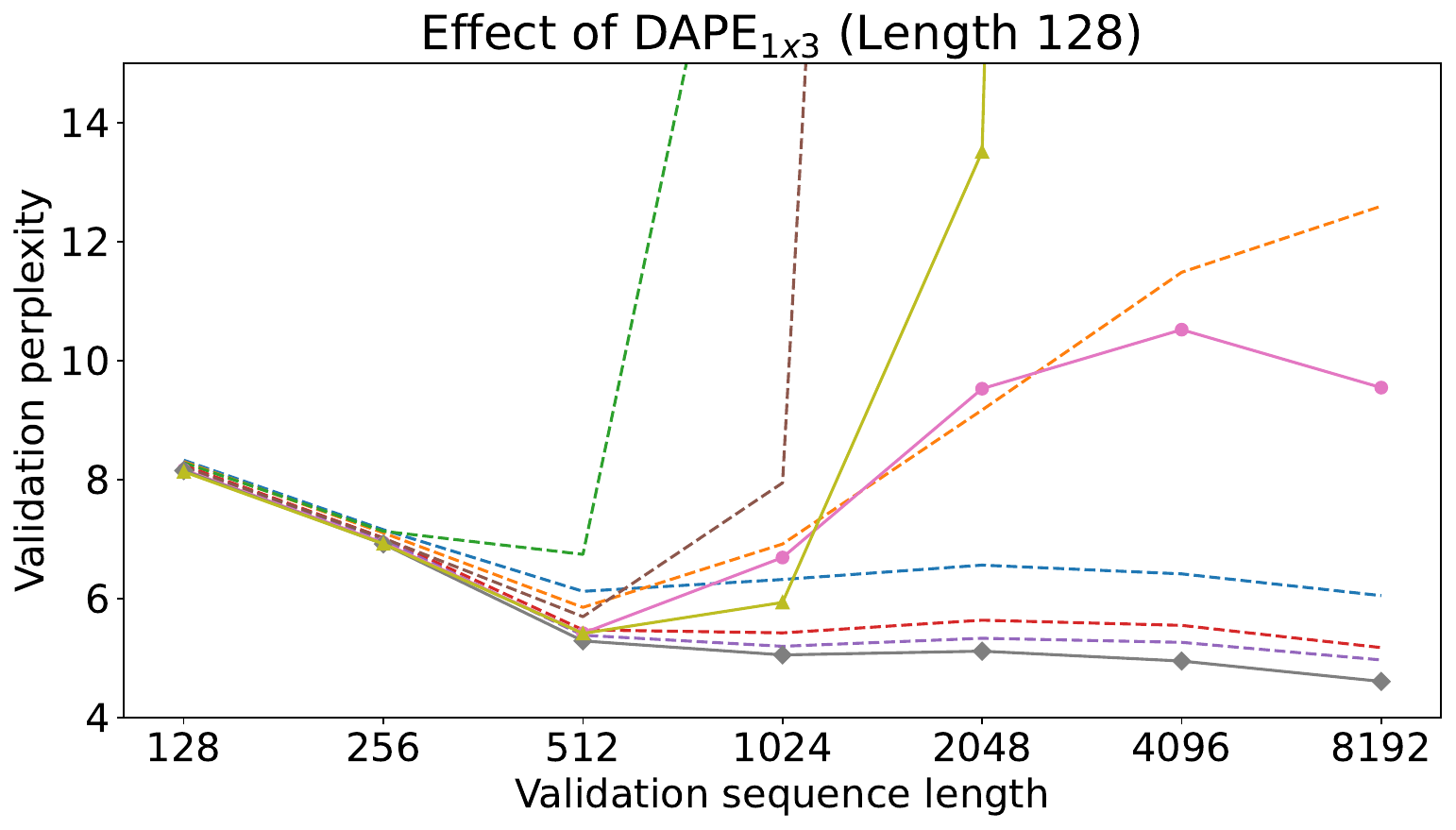}
\hspace{0in}
\includegraphics[width=0.48\textwidth]{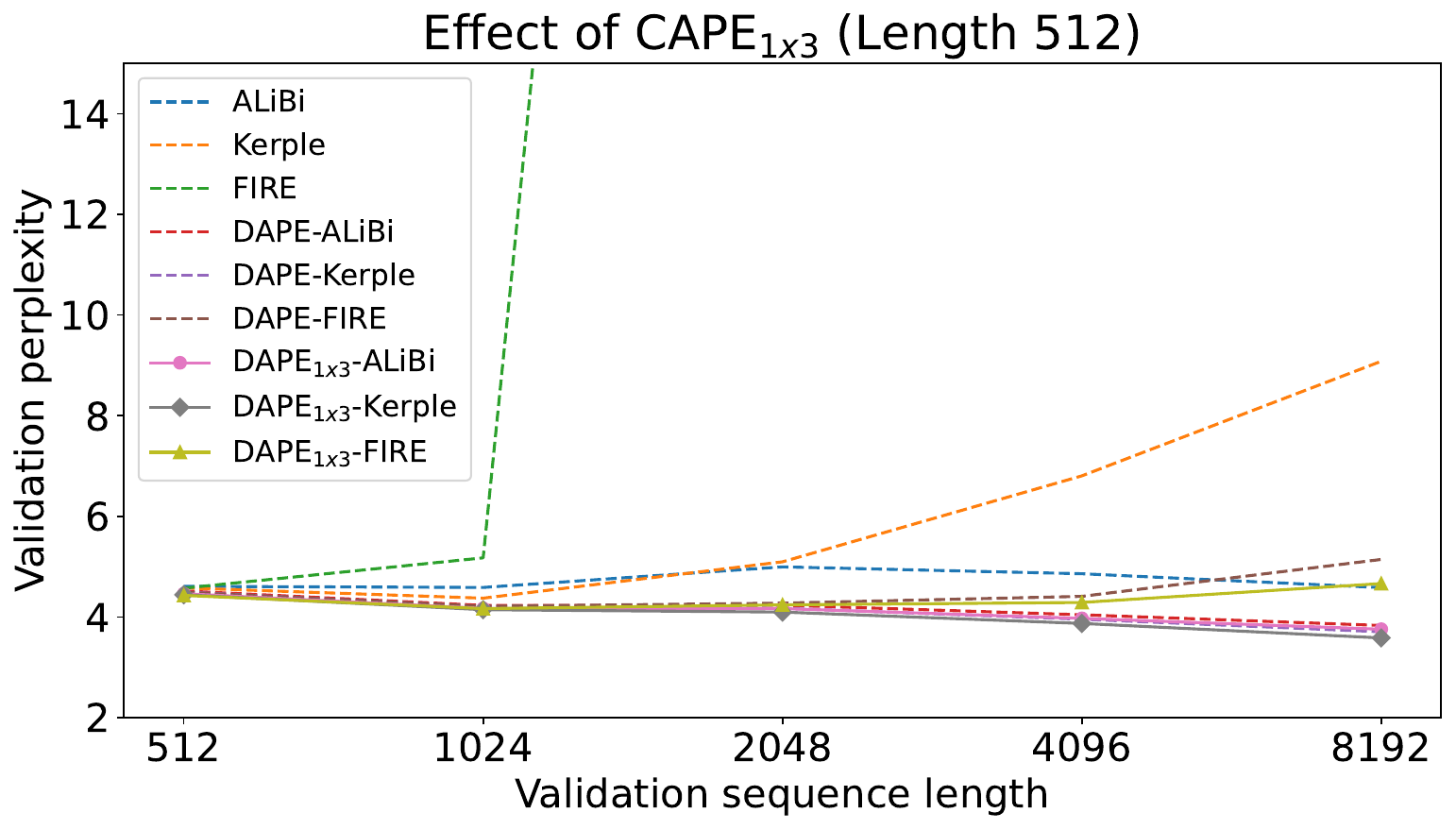}
\caption{
\small
\textbf{The effect of \methodShortName}. Whatever the baseline is ALiBi, Kerple or FIRE, the proposed \methodShortName can all improve their performance. The Figure \ref{fig: dape_method_analysis} also proves that the proposed \methodShortName is effective for NoPE and RoPE.
}
\label{fig: effect_of_dapev2}
\vspace{-10pt}
\end{figure}
\paragraph{For Additive Positional Encoding, \methodShortName enhances performance within and beyond the training length.} As demonstrated in Figure \ref{fig: effect_of_dapev2}, for varying additive positional encoding such as ALiBi, Kerple, and FIRE, their incorporations with \methodShortName (i.e., \methodShortName-ALiBi, \methodShortName-Kerple, and \methodShortName-FIRE) consistently improve performance. Furthermore, regardless of the specific additive positional encoding used, the proposed \methodShortName (configured with a kernel size of $1 \times 3$) outperforms the standard DAPE method (which employs a kernel size of $1 \times 1$). Also, as shown in Figure \ref{fig: dape_method_analysis}, \methodShortName imrpoves the performance of NoPE, both within and beyond the training length
These results highlight the robustness and scalability of \methodShortName, suggesting its broad applicability in enhancing additive positional encoding frameworks.

\paragraph{For Non-Additive Positional Encoding, \methodShortName also improves performance within and beyond the training length.} As illustrated in Figure \ref{fig: dape_method_analysis}, \methodShortName enhances the performance of RoPE, both within and beyond the training length. In contrast, naive DAPE reduces the performance of RoPE, with training lengths of 128 and 512. This indicates that the proposed \methodShortName is a versatile and widely applicable method with the potential to be applied to various position encoding techniques on the language modeling task.

\subsection{The Performance of \methodShortName with Information Leakage}
\begin{wrapfigure}{r}{0.5\textwidth}
    \centering
    \vspace{-1em}
    \includegraphics[width=\linewidth]{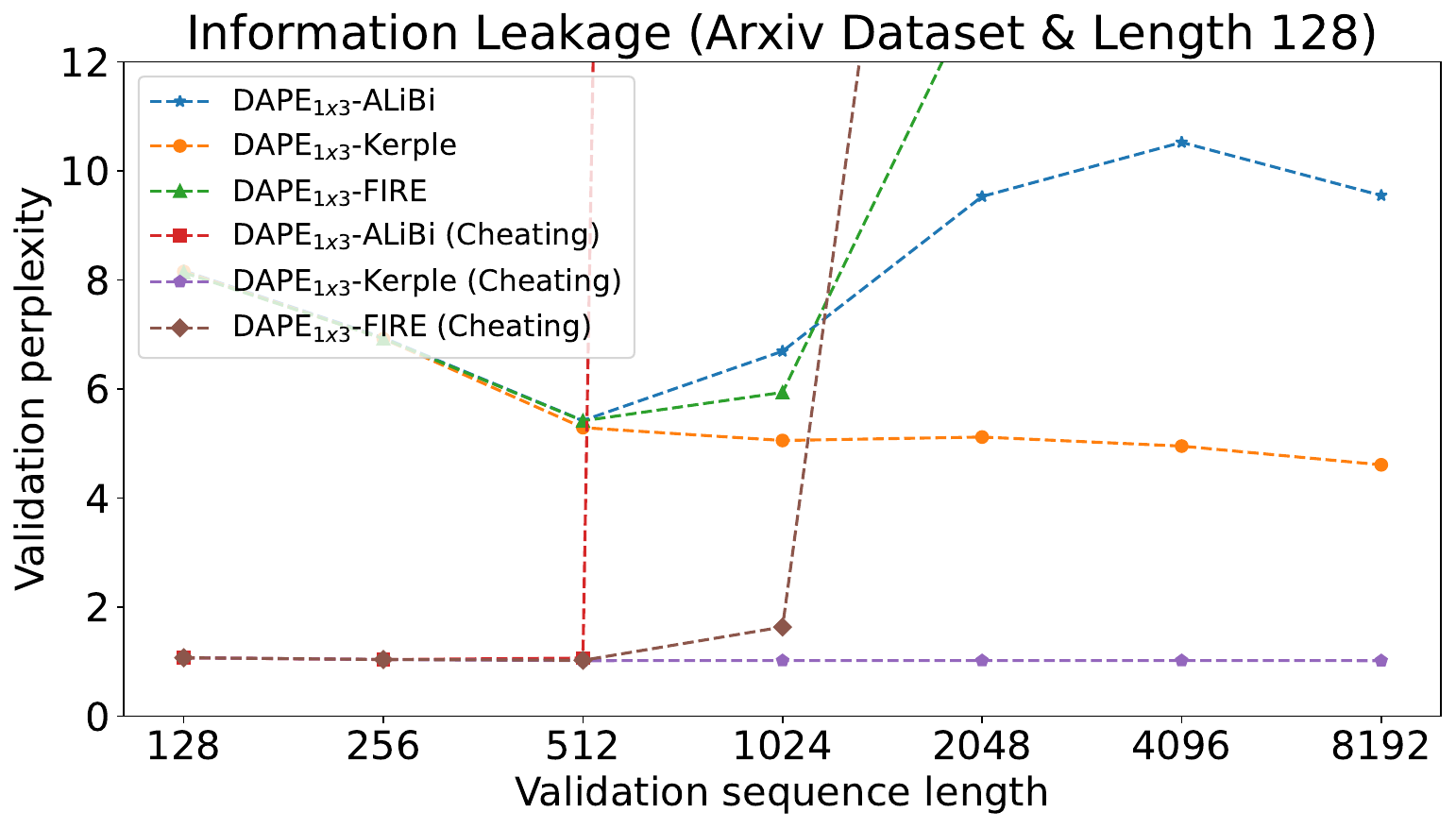}
    \vspace{-1.2em}
    \caption{\textbf{Result with information leakage.}}
    \vspace{-0.5em}
    \label{fig: information_leakage}
\end{wrapfigure}
\paragraph{The \methodShortName can utilize attention data, which is supported by almost zero loss (perplexity is 1) under information leakage.} To prevent the information leakage, we use the $torch.tril$ before \methodShortName to make the attention score lower-triangular matrix. For the cheating version, we do not use the \texttt{torch.tril}. As shown in Figure \ref{fig: information_leakage}, whatever \methodShortName-ALiBi, \methodShortName-Kerple or \methodShortName-FIRE, their cheating version can all achieve about zero loss within evaluation length 1024. Furthermore, the \methodShortName-Kerple can even aachievezero loss when the evaluation length is extended to 8096. This suggest that the proposed \methodShortName can really realize and utilize the information of attention score.
\subsection{Compare DAPE and \methodShortName with Approximate Computational Cost}

\paragraph{\methodShortName achieves even better performance at a lower computational cost.} As shown in Appendix \ref{appendix: approximate cost}, when the training length is set to 128, \methodShortName-Kerple with $D_{\text{DAPE}}$ as 10 achieves a perplexity (ppl) of 8.16 at an evaluation length of 128 and 4.74 at an evaluation length of 8192. This performance is notably better than that of DAPE-Kerple with $D_{\text{DAPE}}$ as 64, which achieves perplexities of 8.21 and 4.87, respectively. Moreover, when the training length is extended to 512 and the evaluation length is smaller or equal to 4096, \methodShortName-Kerple with $D_{\text{DAPE}}$ as 10 continues to surpass the performance of DAPE-Kerple with $D_{\text{DAPE}}$ as 64. Also,  \methodShortName-Kerple with $D_{\text{DAPE}}$ as 21 always achieves better performance than DAPE-Kerple with $D_{\text{DAPE}}$ as 64. This demonstrates that \methodShortName not only maintains its performance advantage across different training lengths but also requires a lower computational cost.

\subsection{The Performance with Different Kernel Sizes}
\paragraph{Different experiment settings may have different optimal kernel sizes.}  Appendix \ref{appendix: different kernel size} shows the performance of DAPE with various kernel sizes, including DAPE (equivalent to a $1\times1$ kernel size), $\textrm{\text{DAPE}}_{1\times3}$, $\textrm{DAPE}_{1\times5}$, and $\textrm{DAPE}_{1\times7}$. For the Arxiv dataset, larger kernel sizes consistently achieve better performance, evaluating with training lengths of 128 or 512. However, for the Books3 dataset, $\textrm{DAPE}_{1\times3}$ performs best when the training length is 128 and evaluated at 8192, whereas $\textrm{DAPE}_{1\times5}$ performs best at the same evaluation level when the training length is 512. These results suggest that the optimal kernel size may vary depending on the experimental setting, ranging from $1\times1$ to larger kernel sizes. Although larger kernel sizes contribute to stronger expressiveness from intuition, we conjecture that the performance degradation for overly large kernel sizes results from optimization challenges.

\subsection{The Performance on CHE Benchmark with Accuracy Evaluation Metrics}

\paragraph{Different tasks have different optimal kernel sizes, as shown in Appendix \ref{appendix: che benchmark result} and Appendix \ref{appendix: different kernel size}.} For example, on \missingduplicate{} task, the \methodShortName-Kerple improves the 87.57 of DAPE-Kerple to 99.65. However, on the \stackmanipulation task, the  \methodShortName-Kerple decreases the 72.04 of DAPE-Kerple to 68.18. Also, as shown in Appendix \ref{appendix: different kernel size}, the larger kernel size does not always lead to better performance. Overall, larger kernel size provides a potential way to improve the Transformer length extrapolation performance, and we usually could find a suitable kernel size (ranging from 1×1 to larger
kernel sizes) to achieve better performance than without further processing attention score. 

\paragraph{The large kernel size performance improvement is related to the baseline bias matrix.} As shown in Appendix \ref{appendix: che benchmark result}, the best performance is usually achieved by further processing attention scores via kernel size 1 or 3. Moreover, on 11 permutation-variant tasks, the \methodShortName-Kerple achieves better performance on 8 of 11 tasks compared to Kerple. And the \methodShortName-FIRE achieves better performance on 6 of 11 tasks compared to FIRE. This suggests that the large kernel size performance improvement is related to the baseline bias matrix.

\subsection{The Time Cost}

\paragraph{As the model size increases, the additional computational cost ratio gradually decreases.} As shown in Appendix \ref{appendix: time cost}, when the model size is 350M, the time cost for Kerple is 189.91 ms, while DAPE-Kerple takes 224.22 ms, and \methodShortName-Kerple requires 252.84 ms. Compared to \methodShortName-Kerple, the time cost ratios for Kerple and DAPE-Kerple are 0.7511 and 0.8868, respectively. As the model size increases from 350M to 2.7B and 6.7B, the time cost ratio for Kerple rises from 0.7511 to 0.8205 and 0.8918, respectively. Similarly, the time cost ratio for DAPE-Kerple increases from 0.8868 to 0.9361 and 0.9677. Therefore, as the model size increases, the time cost ratio also increases, indicating that the additional computational cost decreases progressively.

\section{Conclusion}
In this paper, we point out that the key of Transformer length extrapolation is the better and more accurate attention score. Therefore, we develop and analyze \methodShortName by processing the attention score as feature maps via convolution operation. 
Theoretically, we show that the associative recall tasks, which account for the most perplexity scores, can be realized by the proposed Transformer with convolution, in contrast to the vanilla Transformer.
We conducted comprehensive experiments on Arxiv, Books3, and CHE to validate the effectiveness of the proposed method,
where the proposed method exhibits significant superiority.

\nocite{*}

\bibliography{iclr2025_conference}
\bibliographystyle{iclr2025_conference}

\newpage
\appendix

\section{Compare with Baseline on Arxiv Dataset with Training Length 1024}
\label{appendix: length 1024}
\begin{table}[htb]
  \caption{The performance (ppl) on Arxiv dataset with training length 1024, compared to baselines.}
  \centering
  \resizebox{0.8\textwidth}{!}{
\begin{tabular}{ccccc}
\toprule
\textbf{Method} &   1024 & 2048 & 4096 &8192 \\ \midrule
NoPE \citep{kazemnejad2024impact} & 4.16&42.27&1854.73&17167.32 \\ \midrule 
RoPE \citep{su2024roformer}          & 4.07&  	86.20& 	237.67& 	256.12 \\ \midrule
T5's bias \citep{raffel2020exploring}    & 4.03& 	4.28& 	13.07& 	79.55  \\ \midrule
ALiBi \citep{press2021train}     & 4.09& 	4.53& 	4.45& 	4.22  \\ \midrule
Kerple \citep{chi2022kerple}       & 4.06& 	4.09& 	4.68& 	6.951  \\ \midrule
FIRE \citep{li2023functional}     & 4.06& 	9.21& 	236.18& 	440.60  \\ \midrule
DAPE-Kerple \citep{zheng2024dape}   & 3.98& 	3.91	& 3.68& 	3.41 \\  \midrule
\methodShortName-Kerple    & 3.93& 	3.86	&3.61& 	3.37   \\ 
    \bottomrule
  \end{tabular}
  }
  \label{table:length 1024}
  \vspace{-10pt}
\end{table}

\section{Large Model Size } 
\label{appendix: large model size result}
\begin{table}[htb]
  \caption{The performance (ppl) under large model size 2.7B on Books3 dataset.}
  \centering
  \resizebox{0.8\textwidth}{!}{
\begin{tabular}{cccccc}
\toprule
\textbf{Method} & 512 & 1024 & 2048 & 4096  \\ \midrule
RoPE          & 21.01&  	25.00& 	48.13& 	160.59 \\ \midrule
T5's bias     & 21.10& 	21.88& 	23.59& 	33.23  \\ \midrule
Kerple        & 21.14& 	22.08& 	23.38& 	27.21  \\ \midrule
DAPE-Kerple   & 20.52& 	21.01	& 20.23& 	19.67  \\  \midrule
\methodShortName-Kerple (kernel size 1x3)   & 20.16& 	20.54	&19.80& 	19.02   \\ 
    \bottomrule
  \end{tabular}
  }
  \label{table:model size 2.7B}
  \vspace{-10pt}
\end{table}
\section{Compare DAPE and \methodShortName with Approximate Computational Cost}
\label{appendix: approximate cost}
\begin{figure}[htbp]
\setlength{\abovecaptionskip}{0.1cm}
\centering
\includegraphics[width=0.45\textwidth]{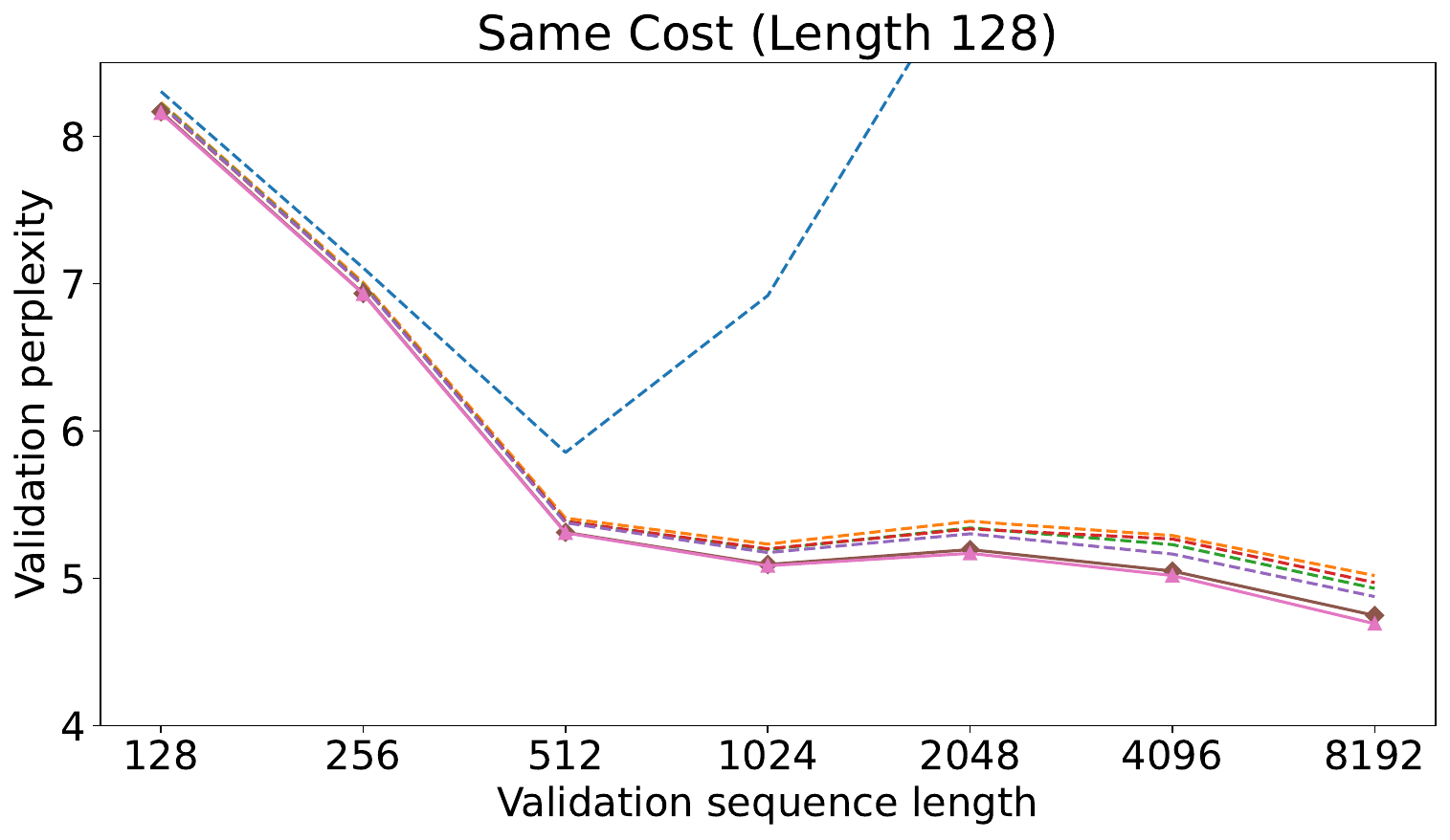}
\hspace{0in}
\includegraphics[width=0.45\textwidth]{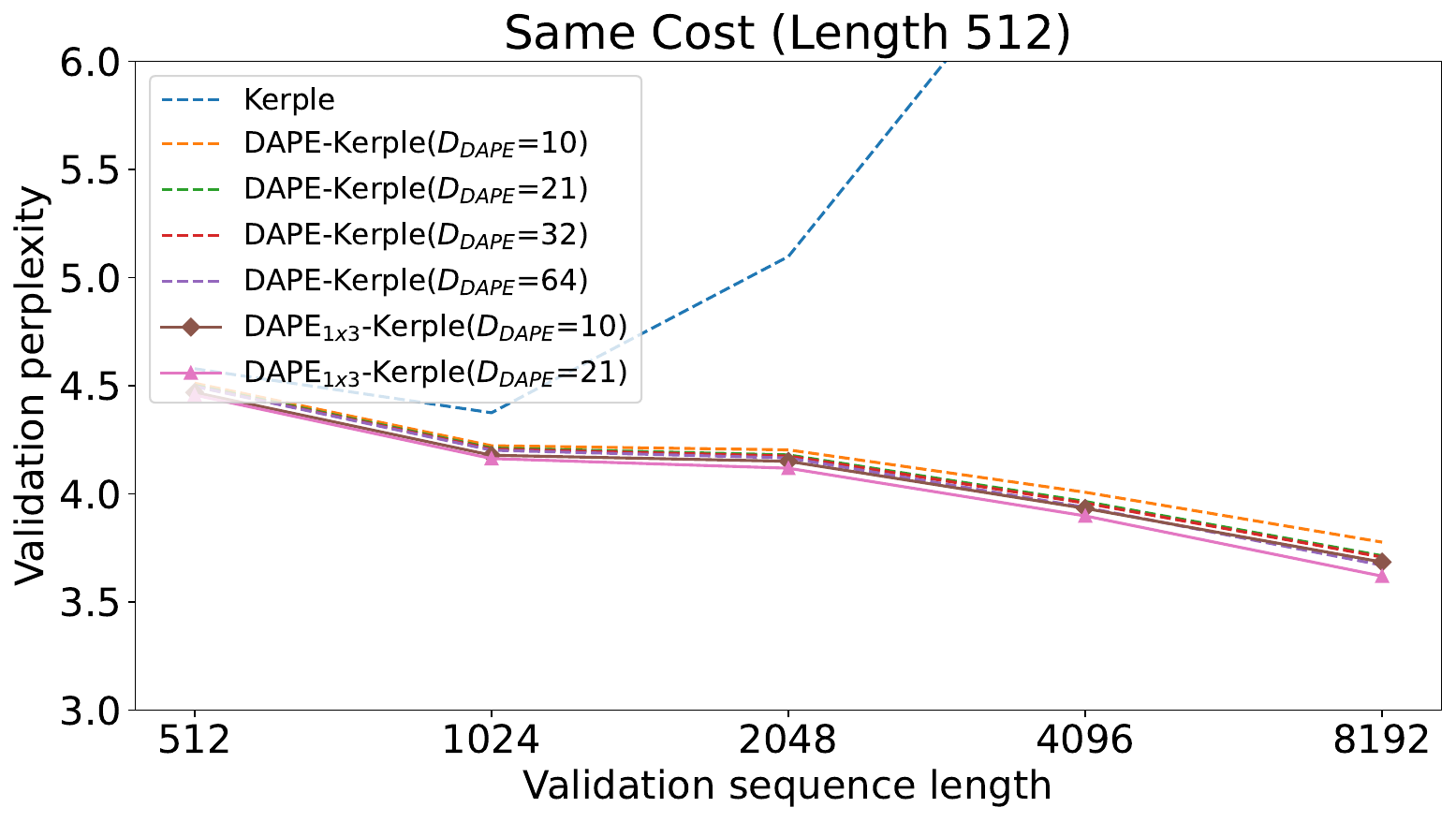}
\caption{
\small
\textbf{Compare \methodShortName and DAPE with the approximately same cost on Arxiv Dataset.} We compare the \methodShortName and DAPE with approximate cost and different $D_{\text{DAPE}}$. As the kernel size of is \methodShortName $1 \times 3$, the proposed \methodShortName is the triple computation cost of DAPE, with the same $D_{\text{DAPE}}$.
}
\label{fig: same cost}
\vspace{-10pt}
\end{figure}
\clearpage
\section{The Performance with Different Kernel Size} 
\label{appendix: different kernel size}
\begin{table}[htb]
  \caption{The performance with different kernel sizes, with training length 128  and evaluation from length 128 to 8192. For different datasets and training length, the optimal kernel size may not always be the largest one, especially when the evaluation length is larger.}
  \centering
  \resizebox{\textwidth}{!}{
\begin{tabular}{ccccccccc}
\toprule
\textbf{Dataset}&\textbf{Method} & 128&256&512 & 1024 & 2048 & 4096 & 8192 \\ \midrule
\multirow{5}{*}{Arxiv}&Kerple          & 8.30 & 7.10 & 5.85 & 6.91 & 9.17 & 11.48 & 12.59 \\ \cmidrule{2-9}
&DAPE-Kerple (Kernel Size 1x1)          & 8.21 & 6.98 & 5.38 & 5.20 & 5.33 & 5.26 & 4.97 \\ \cmidrule{2-9}
&$\textrm{DAPE}_{1\times3}$-Kerple (Kernel Size 1x3) &8.15 & 6.92 & 5.29 & 5.05 & 5.11 & 4.95 & 4.60  \\ \cmidrule{2-9}
& $\textrm{DAPE}_{1\times5}$-Kerple (Kernel Size 1x5)       & 8.13 & 6.91 & 5.27 & 5.04 & 5.10 & 4.91 & 4.57  \\ \cmidrule{2-9}
& $\textrm{DAPE}_{1\times7}$-Kerple (Kernel Size 1x7)       & \textbf{8.12}& \textbf{6.89}& \textbf{5.26}&  \textbf{5.02}&  \textbf{5.09}&  \textbf{4.91}&  \textbf{4.57}  \\ \midrule
\multirow{5}{*}{Books3}&Kerple          & 32.10 & 29.09 & 28.10 & 35.75 & 44.68 & 56.39 & 66.23 \\ \cmidrule{2-9}
&DAPE-Kerple (Kernel Size 1x1)          & 31.49 & 28.27 & 24.93 & 24.31 & 23.34 & 24.38 & 25.01 \\ \cmidrule{2-9}
&$\textrm{DAPE}_{1\times3}$-Kerple (Kernel Size 1x3) &31.07 & 27.81 & 24.38 & 23.57 & 22.40 & 23.19 & \textbf{23.52}  \\ \cmidrule{2-9}
& $\textrm{DAPE}_{1\times5}$-Kerple (Kernel Size 1x5)       & 31.02 & 27.79 & 24.36 & 23.57 & 22.41 & 23.32 & 23.71  \\ \cmidrule{2-9}
& $\textrm{DAPE}_{1\times7}$-Kerple (Kernel Size 1x7)       &\textbf{30.98} & \textbf{27.76} & \textbf{24.31} & \textbf{23.47} & \textbf{22.30} & \textbf{23.00} & 23.57 \\
    \bottomrule
  \end{tabular}
  }
  \label{table:different kernel size_128}
\end{table}

\begin{table}[htp]
  \caption{The performance with different kernel size, with training length 512 and evaluation from length 512 to 8192. For different datasets and training length, the optimal kernel size may not always be the largest one, especially when the evaluation length is larger.}
  \centering
  \resizebox{0.9\textwidth}{!}{
\begin{tabular}{ccccccc}
\toprule
\textbf{Dataset}&\textbf{Method} & 512 & 1024 & 2048 & 4096 & 8192 \\ \midrule
\multirow{5}{*}{Arxiv}&Kerple          & 4.57&4.37& 5.09& 6.80& 9.08 \\ \cmidrule{2-7}
&DAPE-Kerple (Kernel Size 1x1)          & 4.49& 4.20&4.17& 3.95& 3.70 \\ \cmidrule{2-7}
&$\textrm{DAPE}_{1\times3}$-Kerple (Kernel Size 1x3) &4.44& 4.14&4.09& 3.87& 3.58  \\ \cmidrule{2-7}
& $\textrm{DAPE}_{1\times5}$-Kerple (Kernel Size 1x5)       & 4.44& 4.14& 4.10& 3.85& 3.59  \\ \cmidrule{2-7}
& $\textrm{DAPE}_{1\times7}$-Kerple (Kernel Size 1x7)       & \textbf{4.43}& \textbf{4.13}& \textbf{4.08}& \textbf{3.85}& \textbf{3.57}  \\ \midrule
\multirow{5}{*}{Books3}&Kerple          & 19.83& 19.19& 20.48&28.33& 40.94 \\ \cmidrule{2-7}
&DAPE-Kerple (Kernel Size 1x1)          & 19.25& 18.28& 17.20& 17.58& 17.85 \\ \cmidrule{2-7}
&$\textrm{DAPE}_{1\times3}$-Kerple (Kernel Size 1x3) &18.95& 17.92& 16.79& 17.05& 17.20  \\ \cmidrule{2-7}
&$\textrm{DAPE}_{1\times5}$-Kerple (Kernel Size 1x5)       & 18.89& 17.87 &16.76 &17.09& \textbf{17.10}  \\ \cmidrule{2-7}
& $\textrm{DAPE}_{1\times7}$-Kerple (Kernel Size 1x7)       &\textbf{18.86}& \textbf{17.82}& \textbf{16.70}& \textbf{17.01}& 17.16 \\
    \bottomrule
  \end{tabular}
  }
  \label{table:different kernel size_512}
\end{table}

\newpage

\section{The Performance of \methodShortName on CHE Benchmark}
\label{appendix: che benchmark result}
\begin{table}[htbp]
    \caption{Train on length 40 with 200k steps, and test from lengths 41 to 500.
        The random accuracy is 50\%, except for \modulararithmeticsimple{}, \cyclenavigation{}, \bucketsort{}, \solveequation{} and \modulararithmeticbrackets{}, where it is 20\%.
        $\dagger$$\dagger$$\dagger$ denotes permutation-invariant tasks, which are expected to be solved without positional information. The dataset comes from \cite{choromanski2020rethinking}, with experiment setting from Randomized PE\citep{ruoss2023randomized}.
    }
    \begin{center}
        \resizebox{\linewidth}{!}{\begin{tabular}{llccccccccccccc}
    \toprule
    & & \multicolumn{5}{c}{\textbf{Baseline}} && \multicolumn{3}{c}{\textbf{DAPE (Kernel Size 1)}} && \multicolumn{3}{c}{\textbf{DAPE (Kernel Size 3)}} \\
    \cmidrule{3-7} \cmidrule{9-11} \cmidrule{13-15} 
    \textbf{Level} & \textbf{Task}  & \textbf{RoPE} & \textbf{Relative} &\textbf{ALiBi} & \textbf{Kerple} &   \textbf{FIRE} && \textbf{ALiBi} & \textbf{Kerple} &   \textbf{FIRE} &&  \textbf{ALiBi} & \textbf{Kerple} & \textbf{FIRE} \\
    \midrule
    \multirow{4}{*}{R}
    & \evenpairs{}  &99.98 & 96.60 & 73.52   & 57.50 & 73.86 & & 99.99 & 99.58& \textbf{100}&&99.99&\textbf{100}&\textbf{100}  \\
    & \modulararithmeticsimple{}  &  21.35& 20.84 & 20.02 & 21.79 & 21.09 & & 23.58 & \textbf{24.47} & 24.46&&21.48&23.90&23.43  \\
    & \paritycheck{}$\dagger$$\dagger$$\dagger$ &  50.05 &50.09 & 50.09 & 50.07 & 50.97 & & 50.30 & 50.07 & 50.04&& 50.13& \textbf{52.51}&50.11 \\
    & \cyclenavigation{}$\dagger$$\dagger$$\dagger$  & 27.63 & 26.95 & 24.64& 29.47 & 28.41 & & 22.99 & \textbf{34.53} & 27.54&&24.43&24.32&24.34  \\
    \midrule
    \multirow{4}{*}{DCF}
    & \stackmanipulation{}  & 61.49& 64.73 & 66.42 & 66.93 & 69.33 & & 68.18 & \textbf{72.04}  & 70.90&&58.90&68.18&60.90  \\
    & \reversestring{}  &65.23 & 65.59 & 71.09 &  71.54 & 65.89 & & 73.37 & 70.74 & 76.40&&56.61&\textbf{81.84}&70.11  \\
    & \modulararithmeticbrackets{}& 31.25 & 31.74 & 30.56& 24.79 & 30.92 & & 31.34 & \textbf{32.37} &  31.50&&29.46&26.13&27.00  \\
    & \solveequation{} & 21.85 & 22.93 & 19.92 & 21.15 & 22.06 & & 20.03 & 22.49 & 22.42&&20.26&\textbf{23.95}&23.62  \\
    \midrule
    \multirow{7}{*}{CS}
    & \duplicatestring{}  & 64.97 & 67.66 & 65.13 & 66.72 & 69.03 & & 70.84 & \textbf{72.95} & 72.71&&52.96&57.03&66.01 \\
    & \missingduplicate{}  &   63.37 & 72.34 & 74.21 & 79.06& 79.27 & & 83.41 & 87.57& 89.17&&59.33&\textbf{99.65}& 74.83 \\
    & \oddsfirst{} & 61.00 & 61.57 & 59.88 & 62.59 & 63.28 && 63.78 & \textbf{67.08} & 66.34&&57.35&56.87&56.57  \\
    & \binaryaddition{}&   55.59 & 56.96 & 54.72 & 56.35 &  55.70 & & 59.71 &  \textbf{60.88} & 56.62 && 57.49 &55.32 & 57.86  \\
    & \computesqrt{}   & 51.88 & 51.63 & 50.63 & 51.11 & 50.80 & & 51.64 & 51.33 &  \textbf{52.46} && 52.08 & 51.76&51.93  \\
    & \bucketsort{}$\dagger$$\dagger$$\dagger$ & 98.12 & 99.31 & 98.45 &  99.38 & \textbf{99.57} & & 99.38 & 98.81 & 99.37 && 96.61& 99.06& 98.56   \\
    \bottomrule
    \end{tabular}
    }
    \end{center}
\label{table: algorithmic reasoning dataset}
\end{table}

\section{\methodShortName Time Cost}
\label{appendix: time cost}
\begin{table}[htbp]
  \caption{The time cost (millisecond) under different testing lengths, with $D_{\textbf{\text{DAPE}}}$ as 32 and default batch size 1, with training length 512.}
  \centering
  \resizebox{\textwidth}{!}{
\begin{tabular}{ccccccc}
\toprule
\textbf{Method} & \textbf{350M Total} & \textbf{Ratio} & \textbf{2.7B Total} & \textbf{Ratio} & \textbf{6.7B Total} & \textbf{Ratio} \\ \midrule
RoPE \citep{su2024roformer}          & 210.01& 0.8306& 472.63& 1.0472& 635.57& 0.8564 \\ \midrule
T5's bias \citep{raffel2020exploring}     & 355.16& 1.4046& 537.62& 1.1912&808.85& 1.0899 \\ \midrule
ALiBi \citep{press2021train}         & 172.60& 0.6826& 325.95& 0.7222& 596.77& 0.8041 \\ \midrule
Kerple  \citep{chi2022kerple}       & 189.91& 0.7511& 370.32& 0.8205& 661.82& 0.8918\\ \midrule
FIRE \citep{li2023functional}         & 248.13& 0.9813& 432.63& 0.9586&797.68& 1.0748 \\ \midrule
$\text{DAPE}$-Kerple \citep{zheng2024dape}   & 224.22& 0.8868& 422.48& 0.9361&717.46& 0.9667 \\  \midrule
$\text{DAPE}_{1\times3}$-Kerple   &252.84& 1.0000&451.29& 1.0000&742.10& 1.0000 \\ 
    \bottomrule
  \end{tabular}
  }
  \label{table:time_cost}
\end{table}

\newpage
\section{Model Configuration} 
\label{model configuration details} 
All experiments are conducted on 8 GPUs. The 125M and 350M model configuration is the following.

\begin{table}[!ht]
    \centering
    \setlength{\tabcolsep}{3pt}
    \label{model configuration}
    \caption{\textbf{Model Configurations.}}
    \begin{tabular}{c c c c c}
    \toprule
    & & \textbf{125M} & & \textbf{350M} \\ \midrule
    Training sequence length & & $512$ & & $512$\\
    Batch size & & 32 $\times$ 8  & & 32 $\times$ 8 \\
    Numer of iterations & & $50$k & & $50$k \\
    Dropout prob. & & $0.0$ & & $0.0$ \\
    Attention dropout prob. & & $0.0$ & & $0.0$ \\
    Attention head && 12 && 16  \\
    Feature dimension && 768 && 1024\\
    Layer number && 12 && 24 \\
    Optimizer & & Adam & & Adam\\
    Optimizer parameter betas & & [0.9, 0.95] && [0.9, 0.95] \\
    Learning rate & & $6\mathrm{e}-4$  & & $3\mathrm{e}-4$ \\
    Precision & & float16 & & float16 \\ 
    \bottomrule
    \end{tabular}
    \label{tab:model_configs}
\end{table}

\section{Data-Adaptive Related Position Encoding Performance Comparison} 
\label{appendix: different context related methods}
\begin{table}[htb]
  \caption{The performance comparison between data-related position encoding, with dataset Books3 and training length 128.}
  \centering
  \resizebox{0.9\textwidth}{!}{
\begin{tabular}{cccccccc}
\toprule
\textbf{Method} & 128&256&512 & 1024 & 2048 & 4096 & 8192 \\ \midrule
Transformer-XL & 31.57 & 28.49& 26.07&26.98&27.90&32.76&41.12\\ \midrule
CoPE & 31.61& 28.41 &25.79& 27.96& 33.80&54.08& 90.66 \\ \midrule
$\textrm{DAPE}$-Kerple (Kernel Size 1x1)         & 31.49 & 28.27 & 24.93 & 24.31 & 23.34 & 24.38 & 25.01 \\ \midrule
$\textrm{DAPE}_{1\times3}$-Kerple (Kernel Size 1x3) &31.07 & 27.81 & 24.38 & 23.57 & 22.40 & 23.19 & \textbf{23.52}  \\
    \bottomrule
  \end{tabular}
  }
  \label{table:contex-related position encoding performance comparsion`}
\end{table}

\clearpage
\newpage
\section{\methodShortName Visualization}
\label{more DAPE visualization}
The model is trained with \methodShortName-Kerple on length 512. Compared to DAPE~\citep{zheng2024dape}, it seems that the \methodShortName presents a more obvious attention sink~\citep{xiao2024efficient}.
\subsection{Visualization on length 512}
\begin{figure}[htbp]
\setlength{\abovecaptionskip}{0.1cm}
\centering
\includegraphics[width=0.32\textwidth]{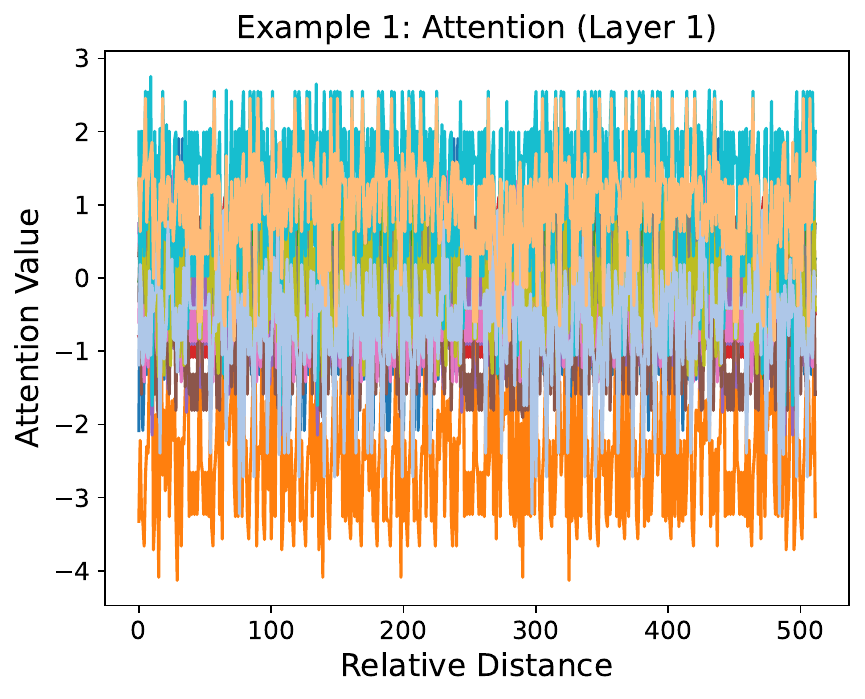}
\hspace{0in}
\includegraphics[width=0.32\textwidth]{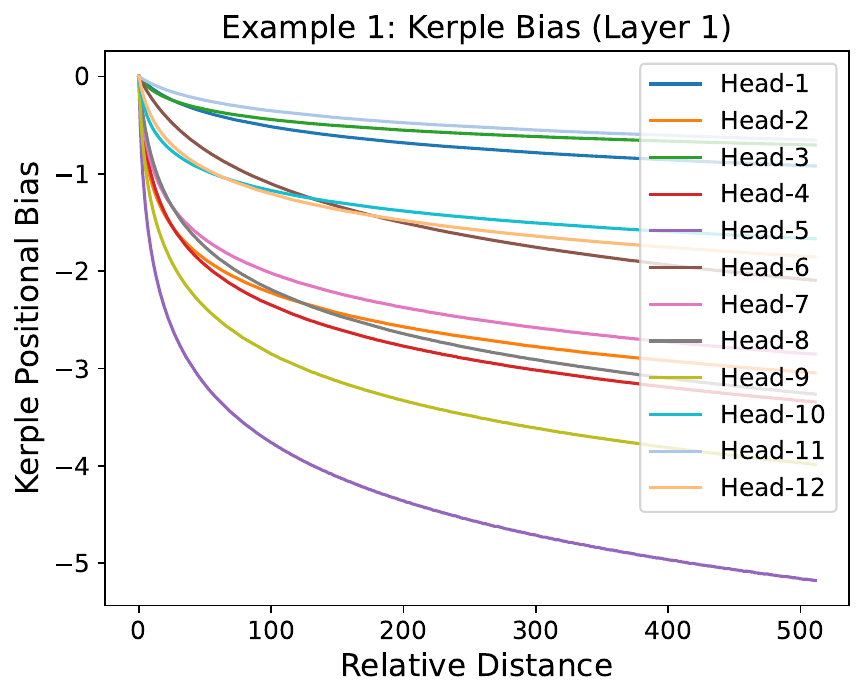}
\hspace{0in}
\includegraphics[width=0.32\textwidth]{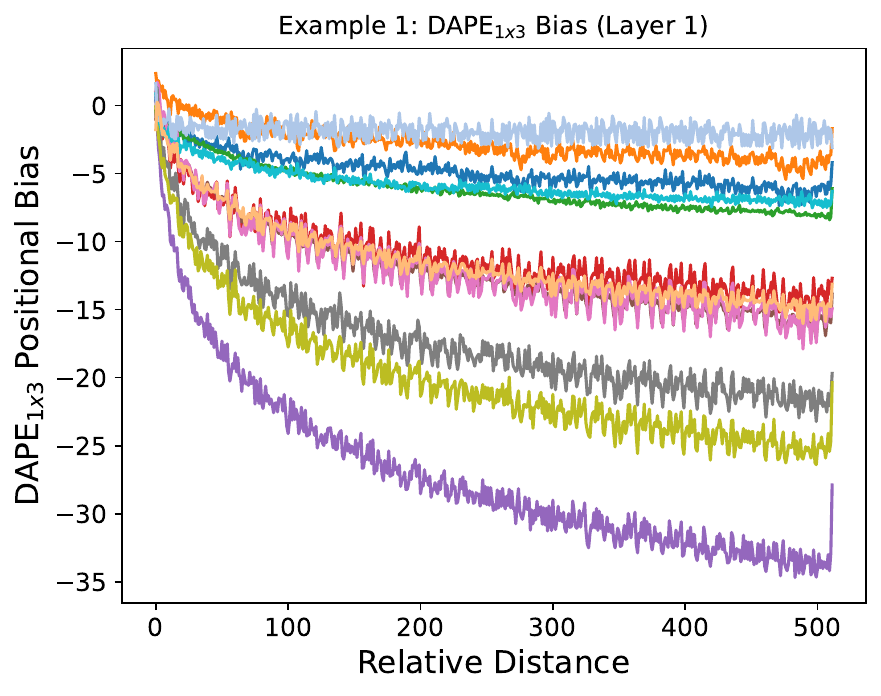}
\hspace{0in}

\includegraphics[width=0.32\textwidth]{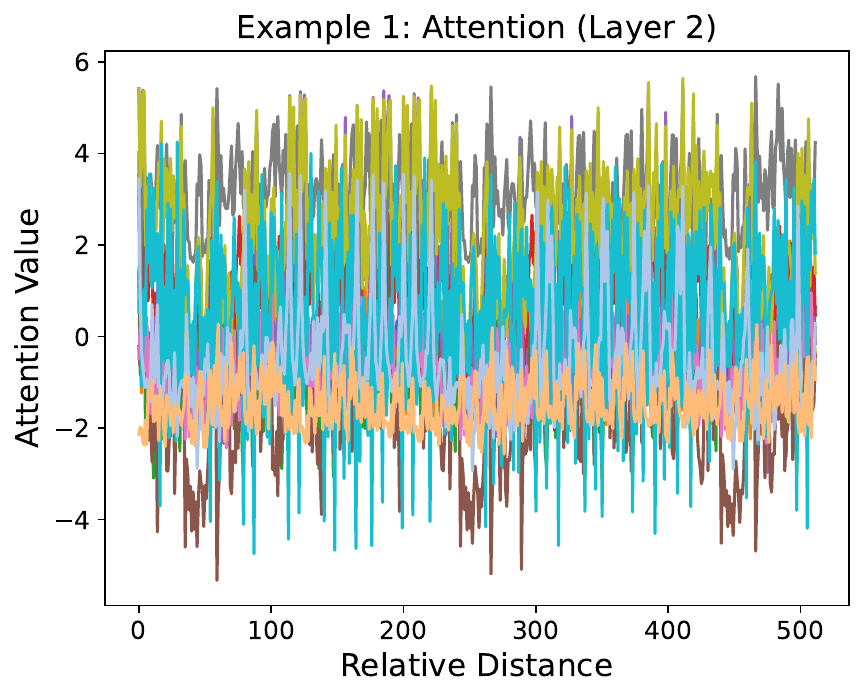}
\hspace{0in}
\includegraphics[width=0.32\textwidth]{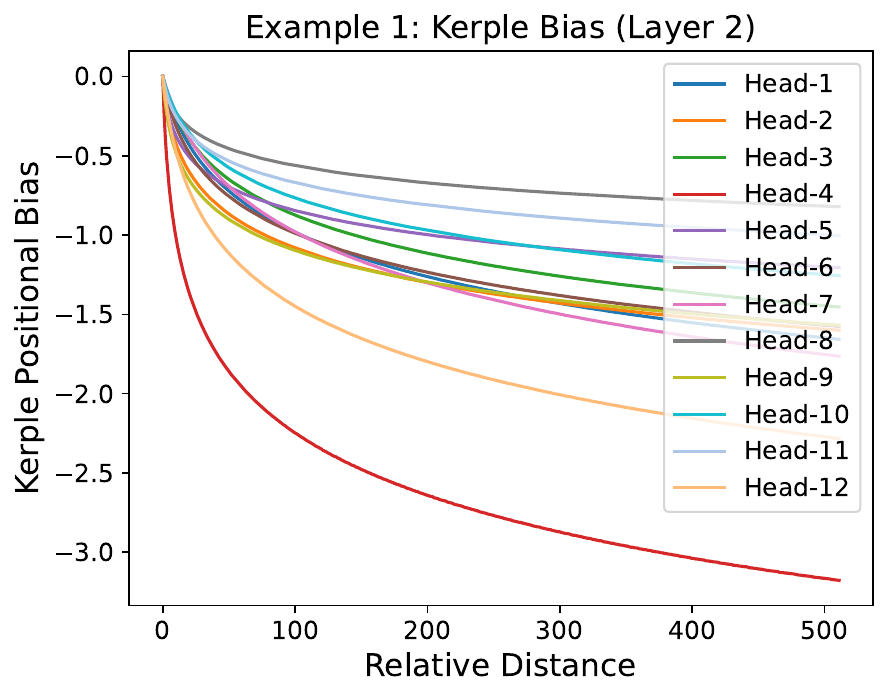}
\hspace{0in}
\includegraphics[width=0.32\textwidth]{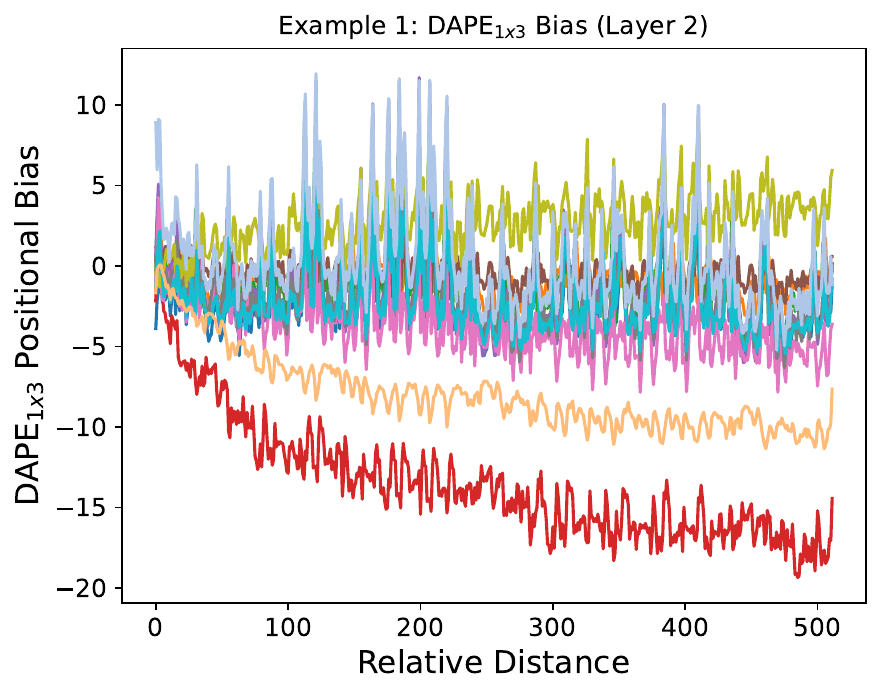}
\hspace{0in}

\includegraphics[width=0.32\textwidth]{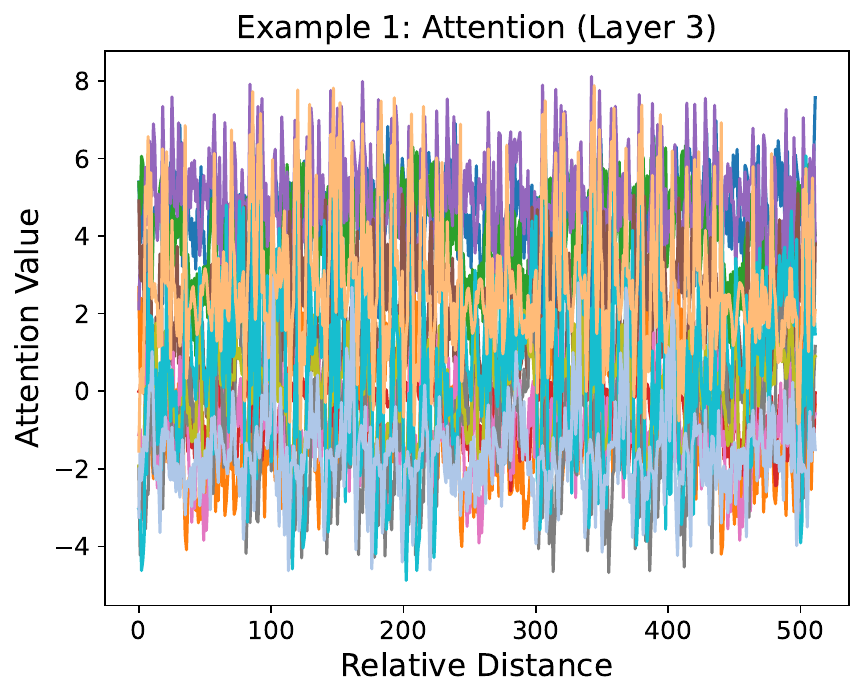}
\hspace{0in}
\includegraphics[width=0.32\textwidth]{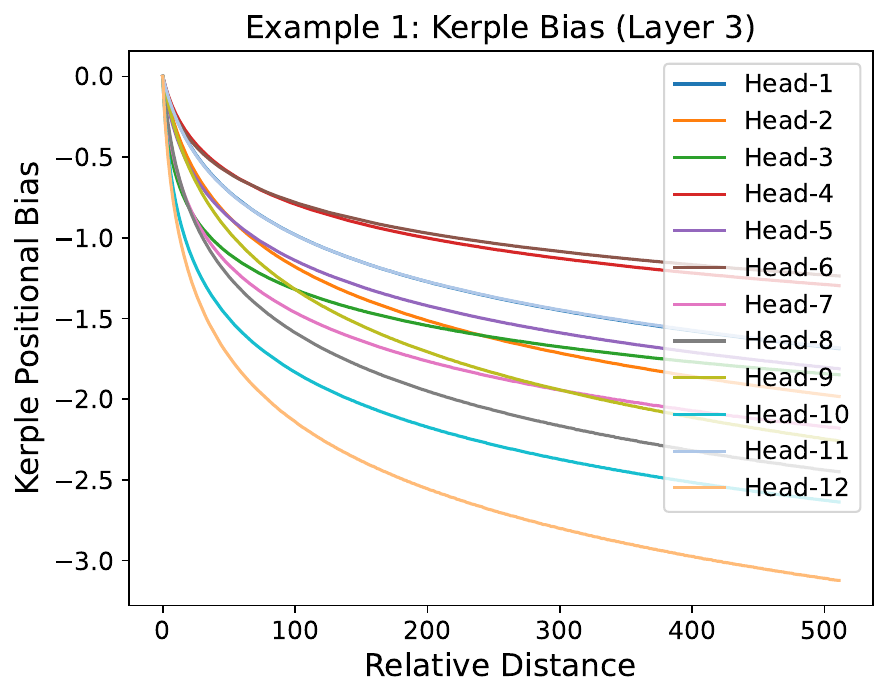}
\hspace{0in}
\includegraphics[width=0.32\textwidth]{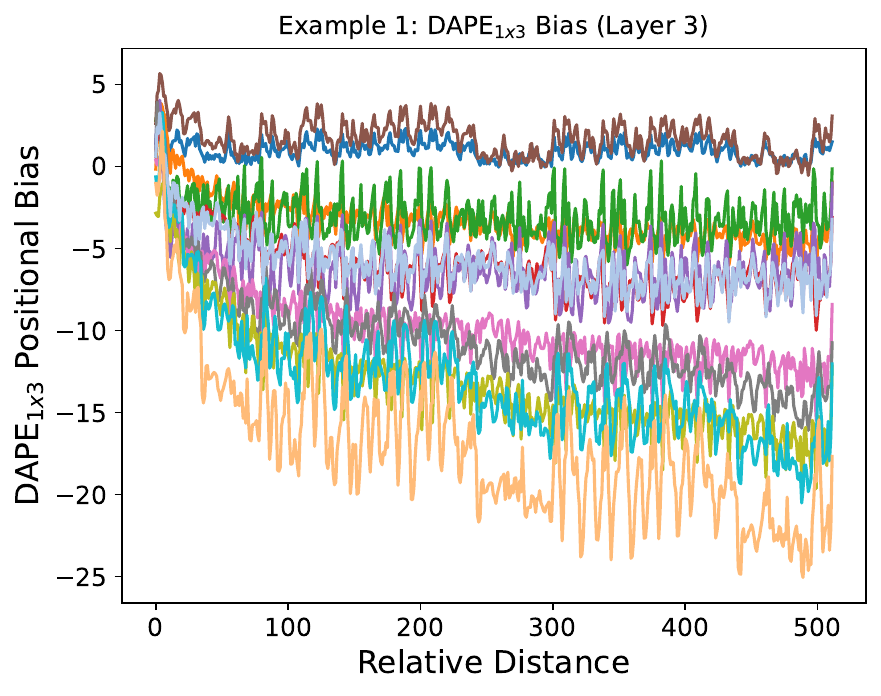}
\hspace{0in}

\includegraphics[width=0.32\textwidth]{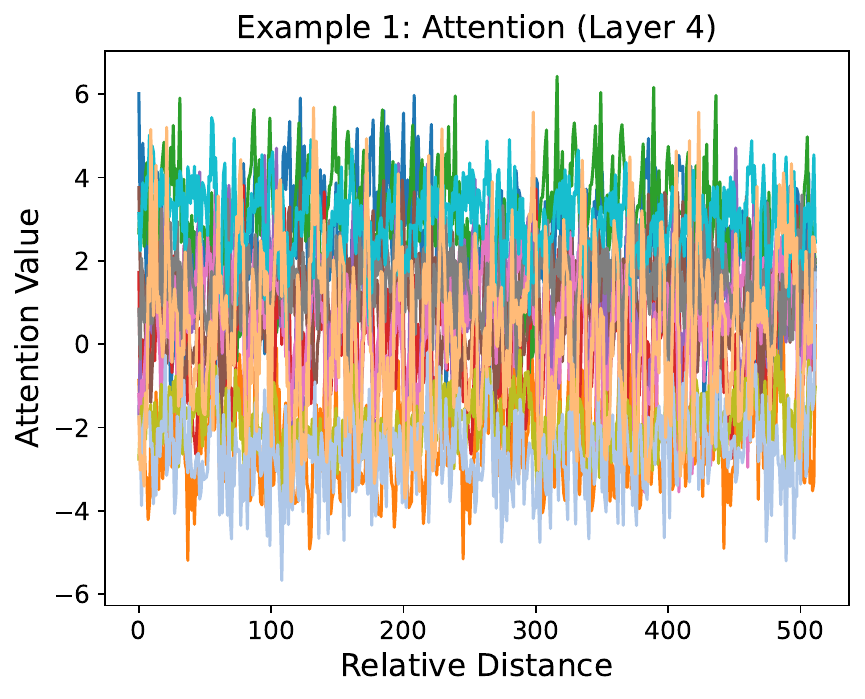}
\hspace{0in}
\includegraphics[width=0.32\textwidth]{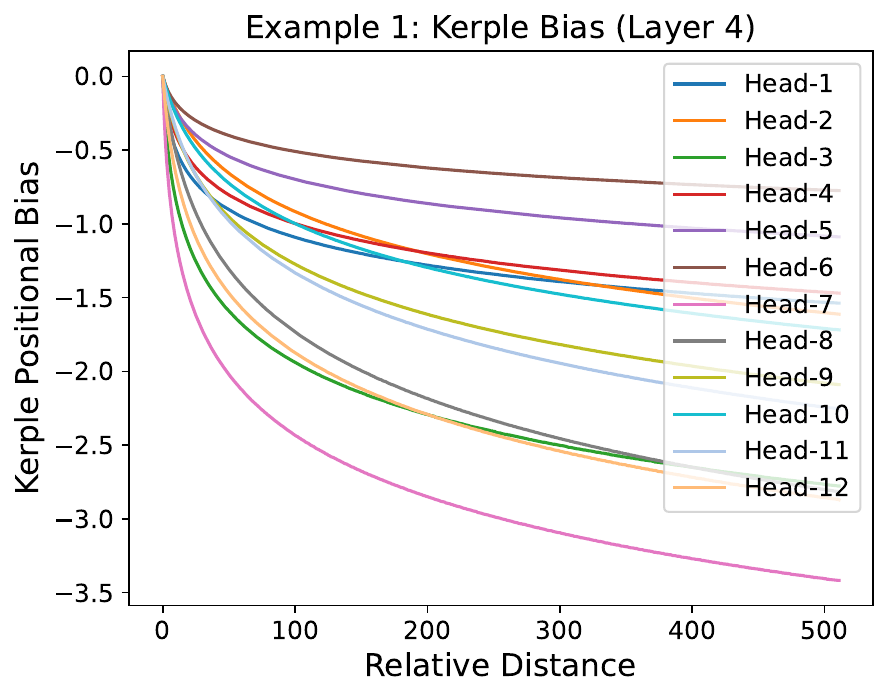}
\hspace{0in}
\includegraphics[width=0.32\textwidth]{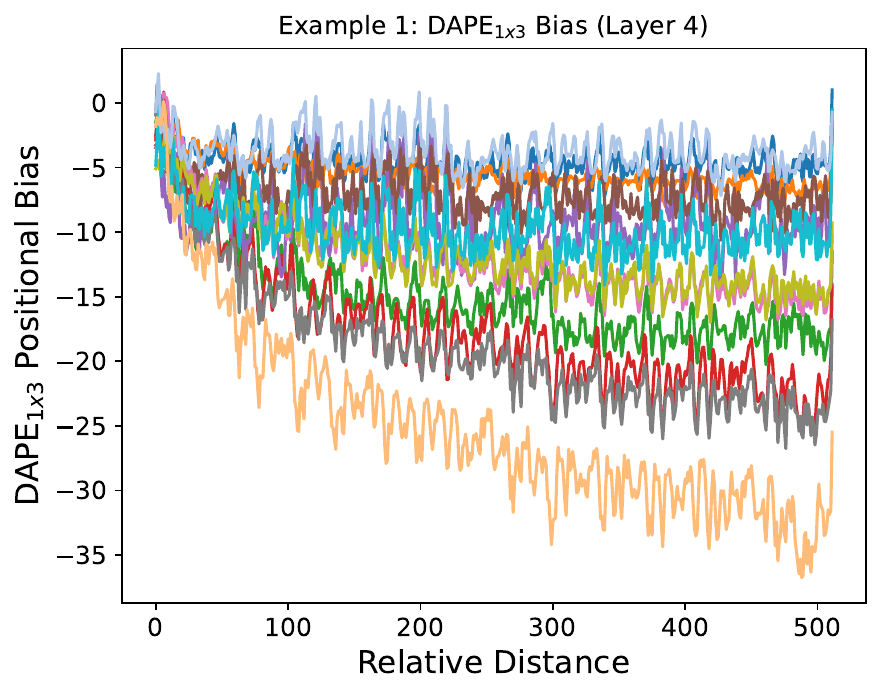}
\hspace{0in}

\caption{
\small
\textbf{Evaluation Length 512 Example 1: Part 1. From Left to Right: (1) The Attention is $\mX \mW_Q(\mX \mW_K)^{\top}$; (2) The Kerple bias is $\mB$; (3) The \methodShortName (with Kerple) bias is $f( \mX \mW_Q(\mX \mW_K)^{\top},\mB)$.
}
}
\end{figure}

\begin{figure}[htbp]
\setlength{\abovecaptionskip}{0.1cm}
\centering
\includegraphics[width=0.32\textwidth]{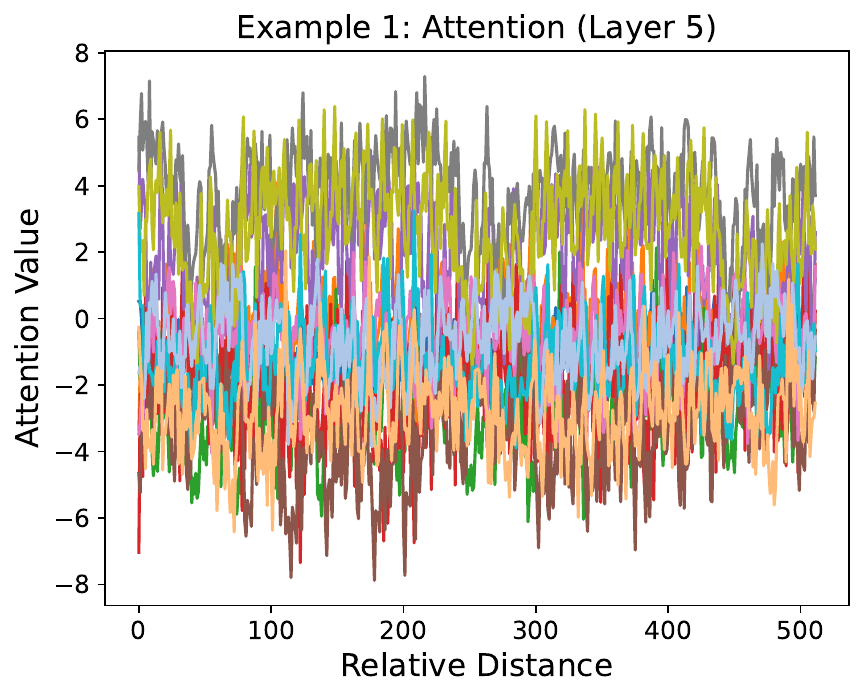}
\hspace{0in}
\includegraphics[width=0.32\textwidth]{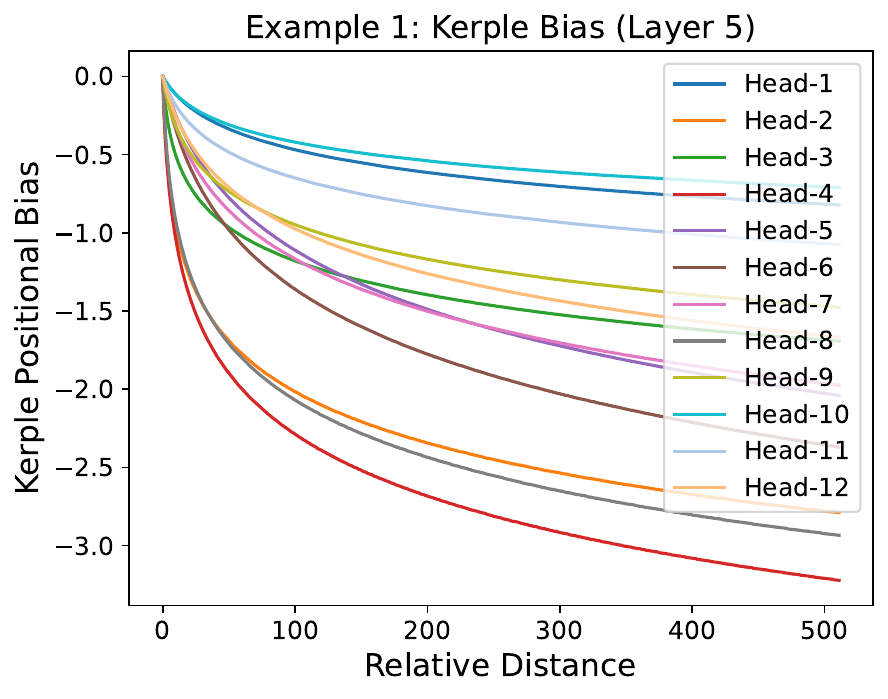}
\hspace{0in}
\includegraphics[width=0.32\textwidth]{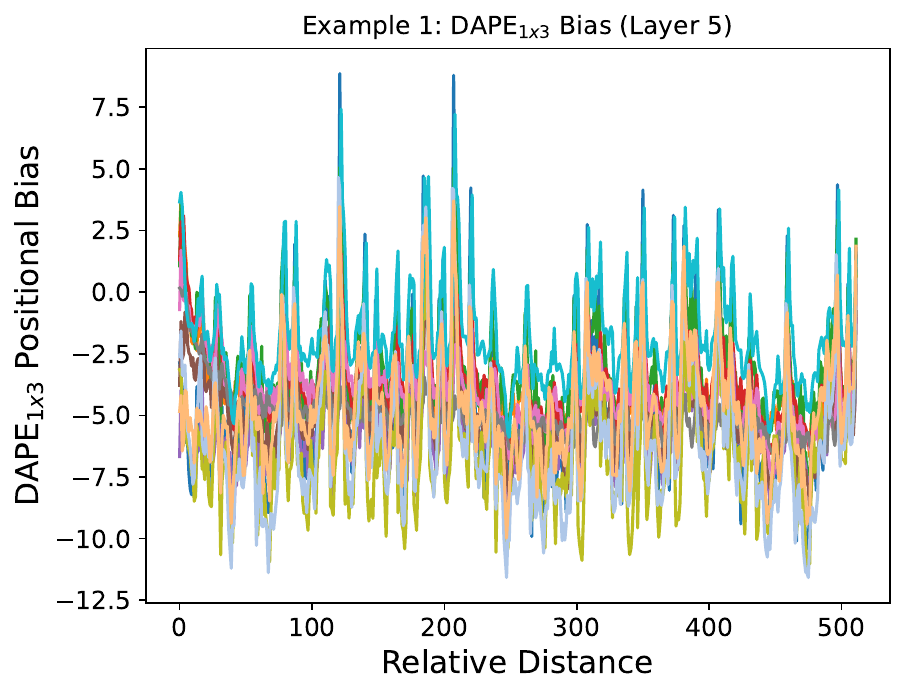}
\hspace{0in}
\hspace{0in}

\includegraphics[width=0.32\textwidth]{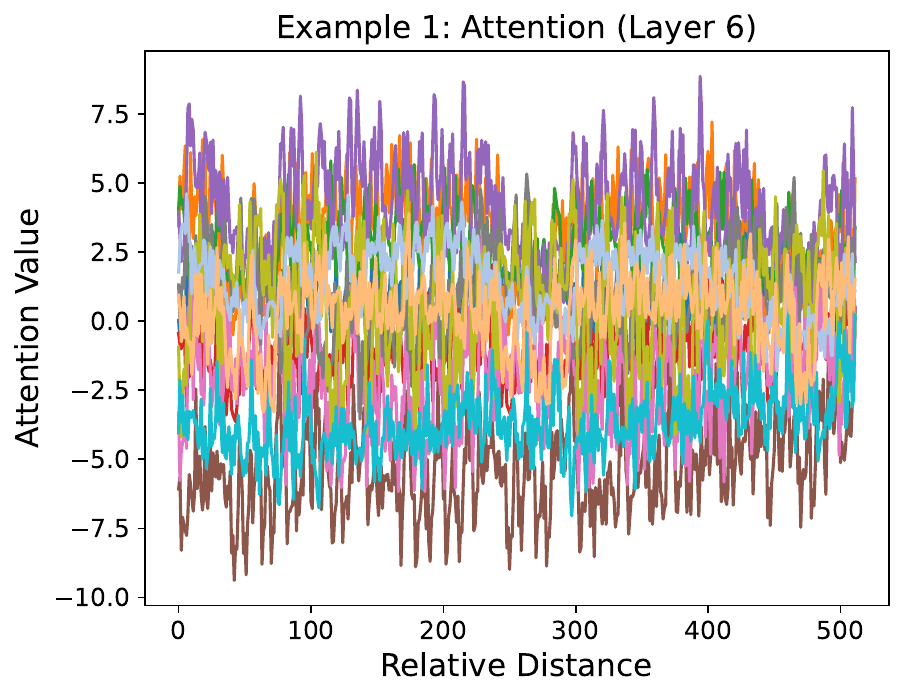}
\hspace{0in}
\includegraphics[width=0.32\textwidth]{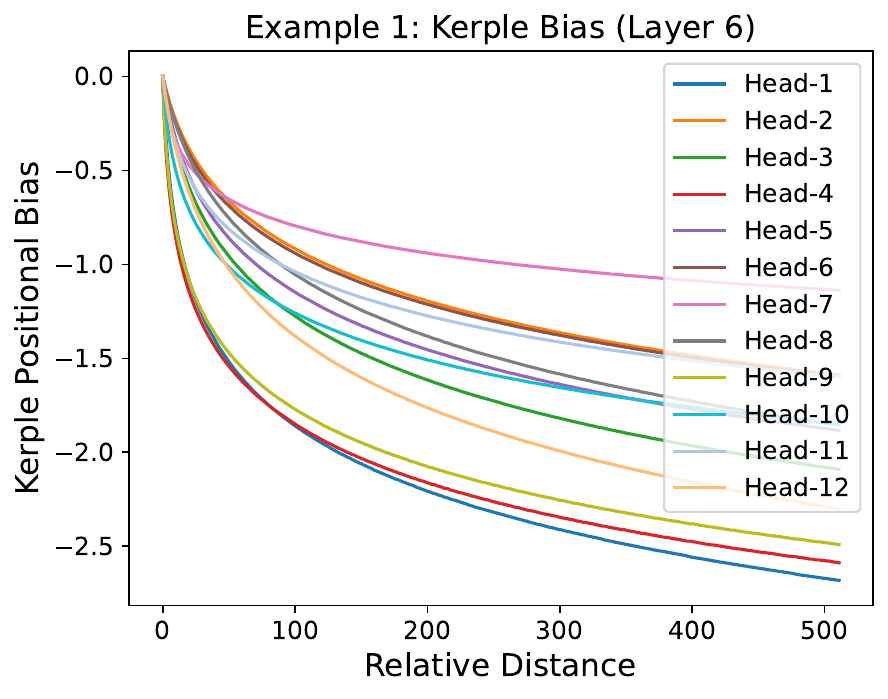}
\hspace{0in}
\includegraphics[width=0.32\textwidth]{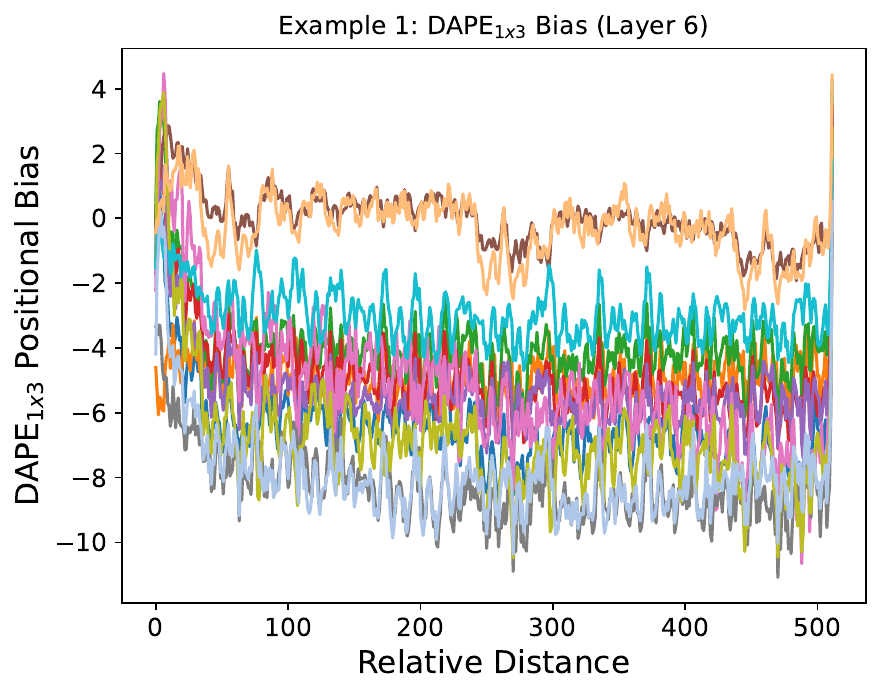}
\hspace{0in}

\includegraphics[width=0.32\textwidth]{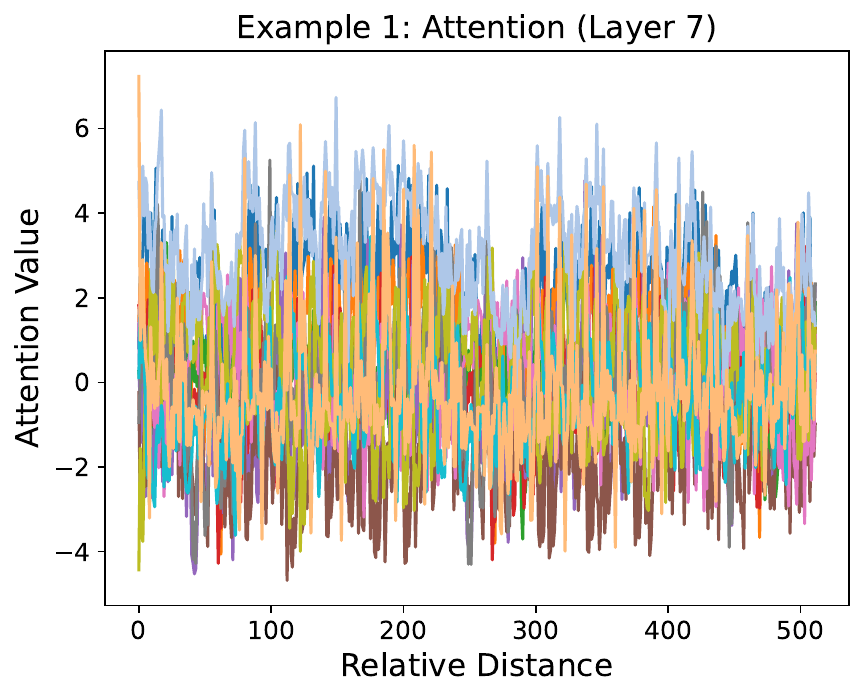}
\hspace{0in}
\includegraphics[width=0.32\textwidth]{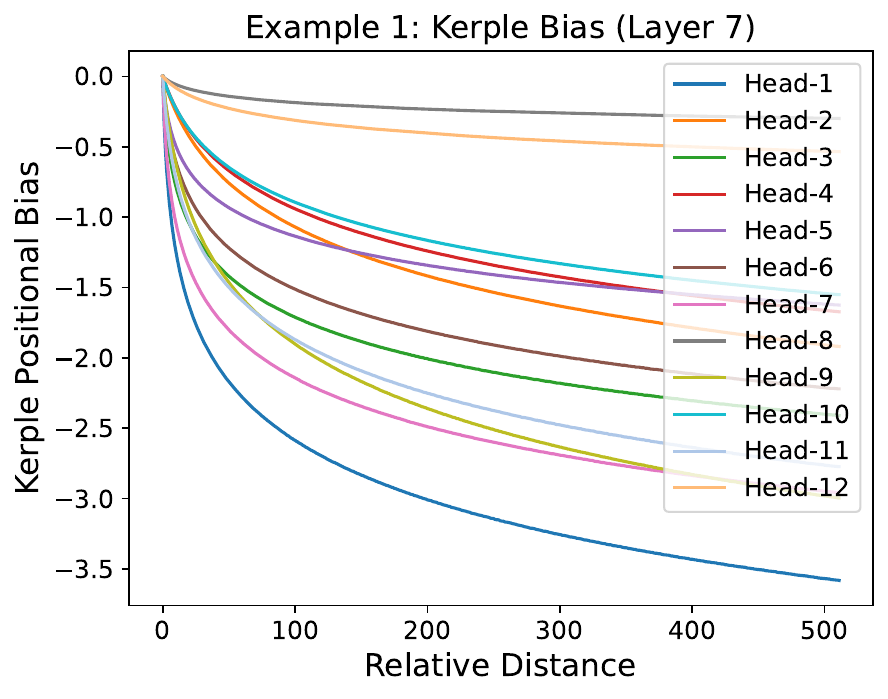}
\hspace{0in}
\includegraphics[width=0.32\textwidth]{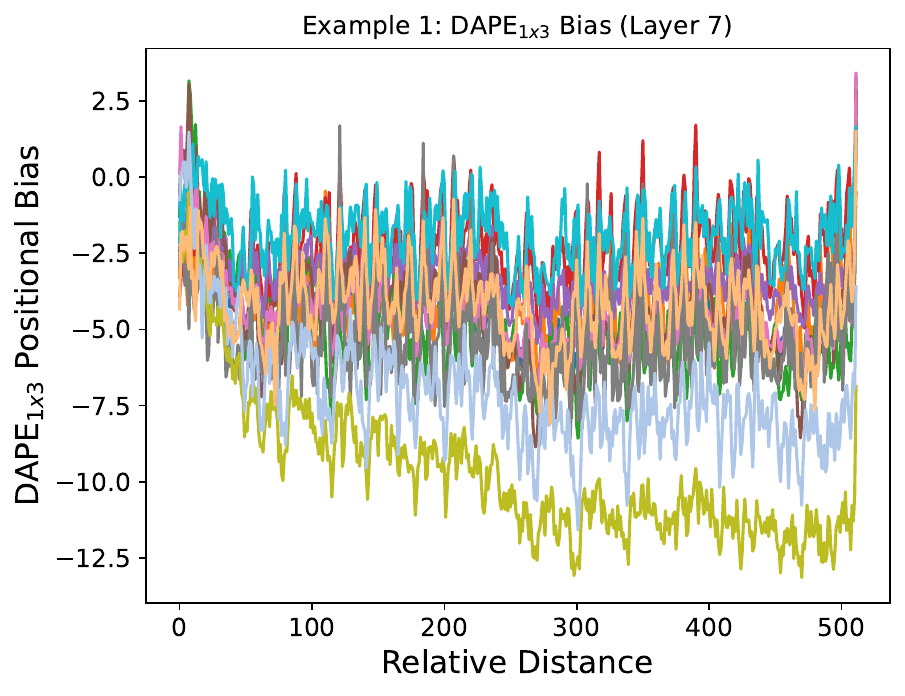}
\hspace{0in}

\includegraphics[width=0.32\textwidth]{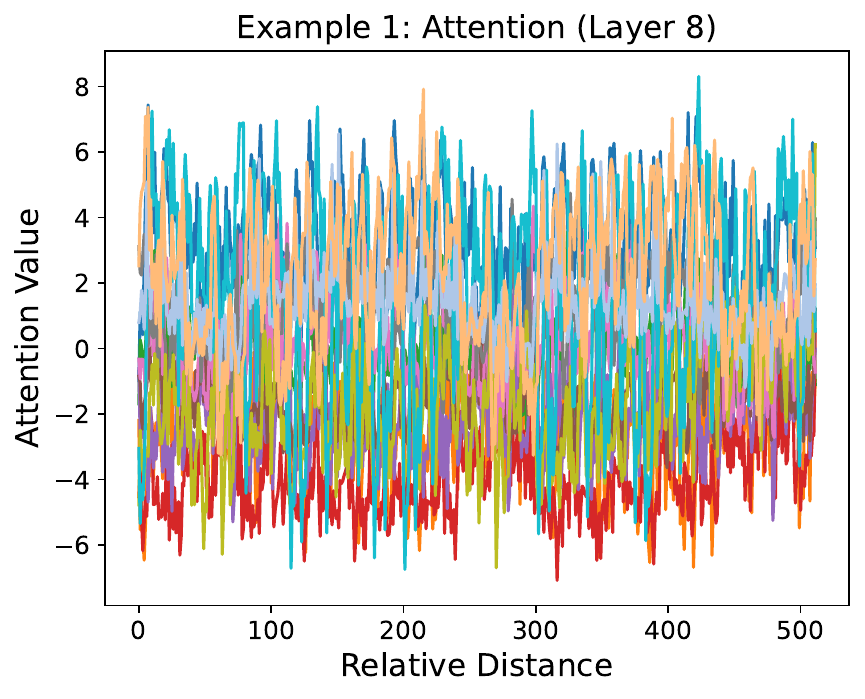}
\hspace{0in}
\includegraphics[width=0.32\textwidth]{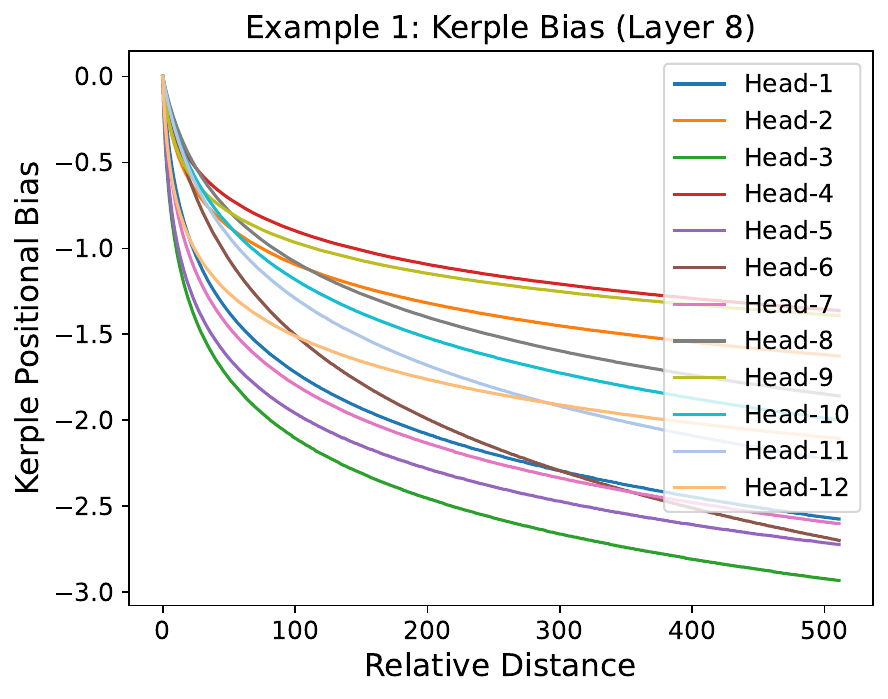}
\hspace{0in}
\includegraphics[width=0.32\textwidth]{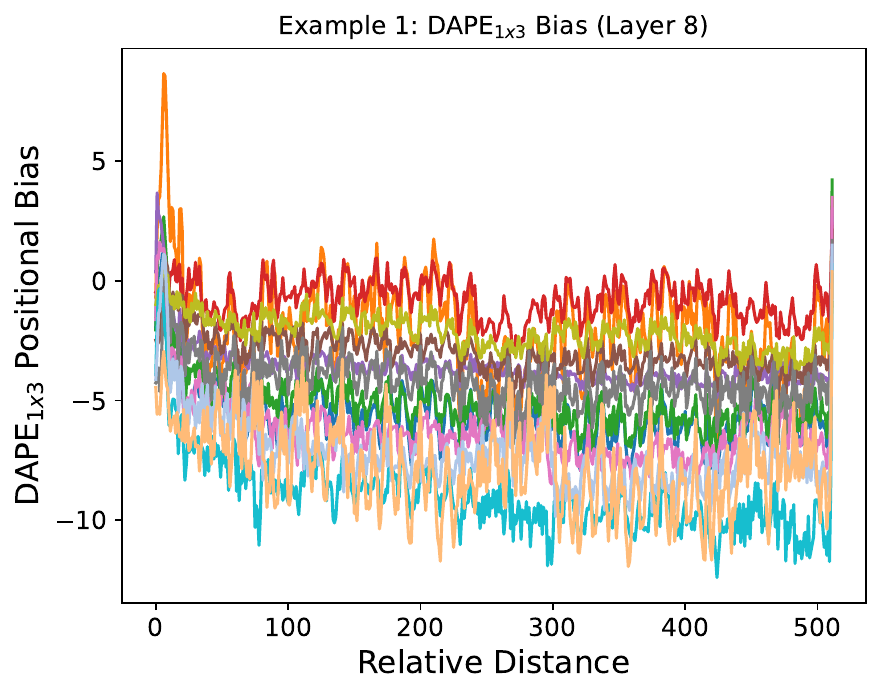}
\hspace{0in}

\includegraphics[width=0.32\textwidth]{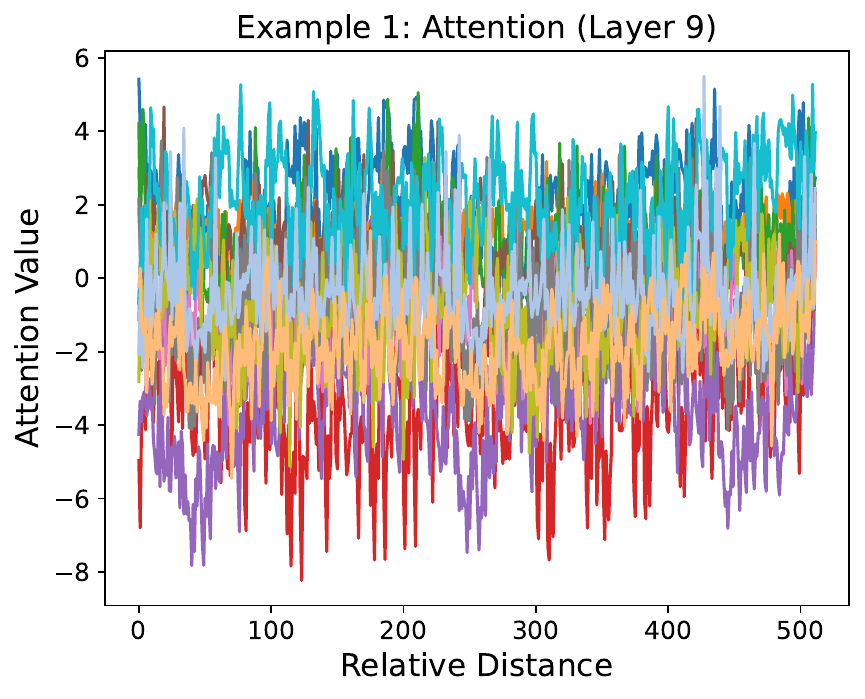}
\hspace{0in}
\includegraphics[width=0.32\textwidth]{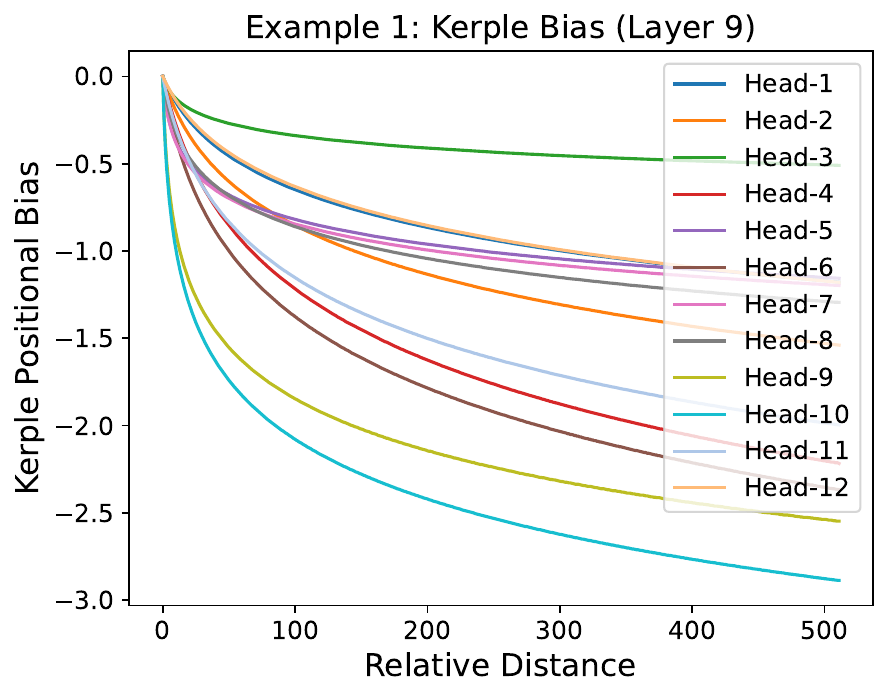}
\hspace{0in}
\includegraphics[width=0.32\textwidth]{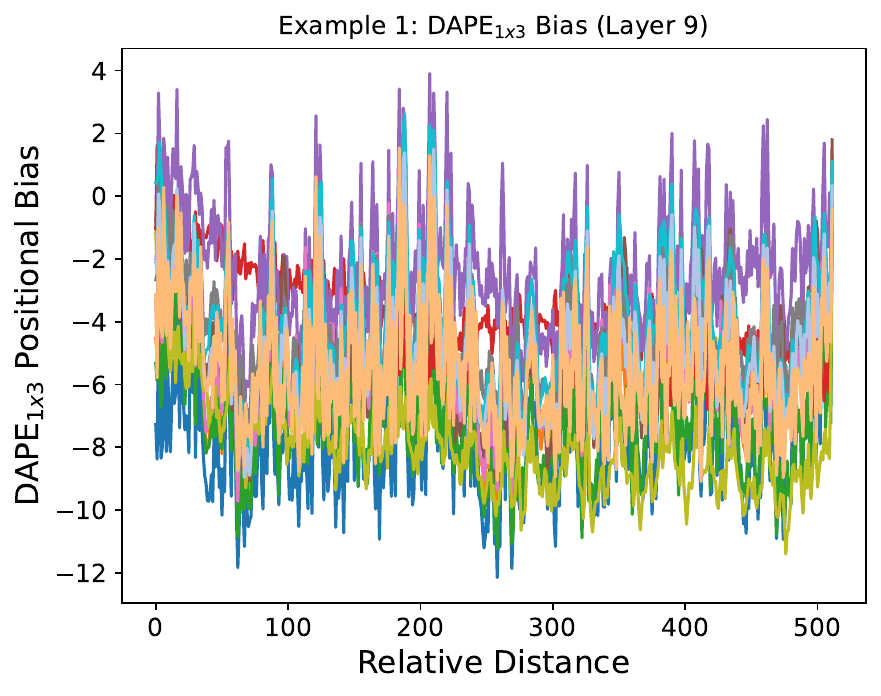}

\hspace{0in}
\caption{
\small
\textbf{Evaluation Length 512 Example 1: Part 2. From Left to Right: (1) The Attention is $\mX \mW_Q(\mX \mW_K)^{\top}$; (2) The Kerple bias is $\mB$; (3) The \methodShortName (with Kerple) bias is $f( \mX \mW_Q(\mX \mW_K)^{\top},\mB)$.
}
}
\end{figure}

\newpage

\begin{figure}[ht]
\setlength{\abovecaptionskip}{0.1cm}
\centering
\includegraphics[width=0.32\textwidth]{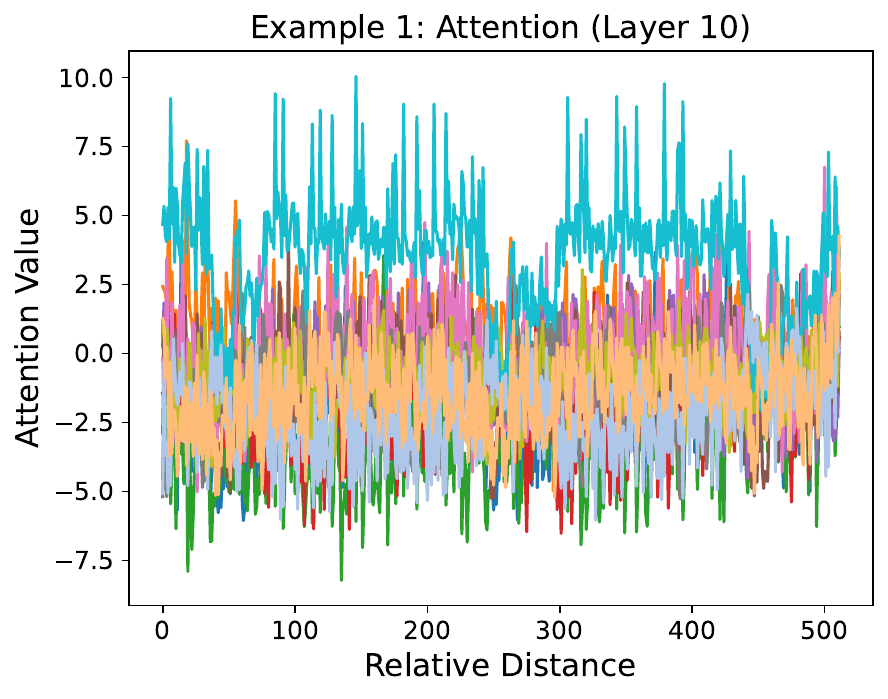}
\hspace{0in}
\includegraphics[width=0.32\textwidth]{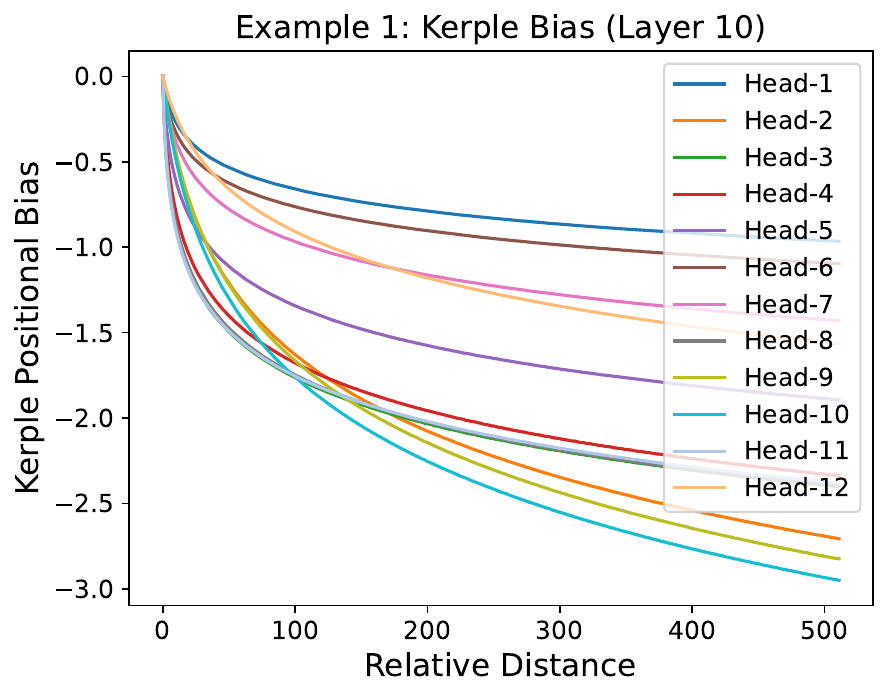}
\hspace{0in}
\includegraphics[width=0.32\textwidth]{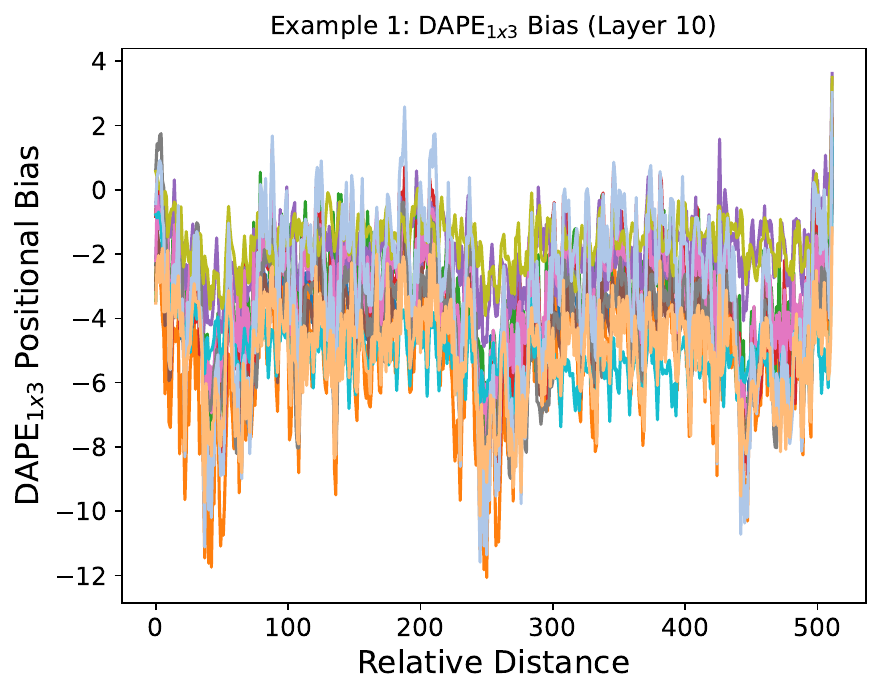}

\includegraphics[width=0.32\textwidth]{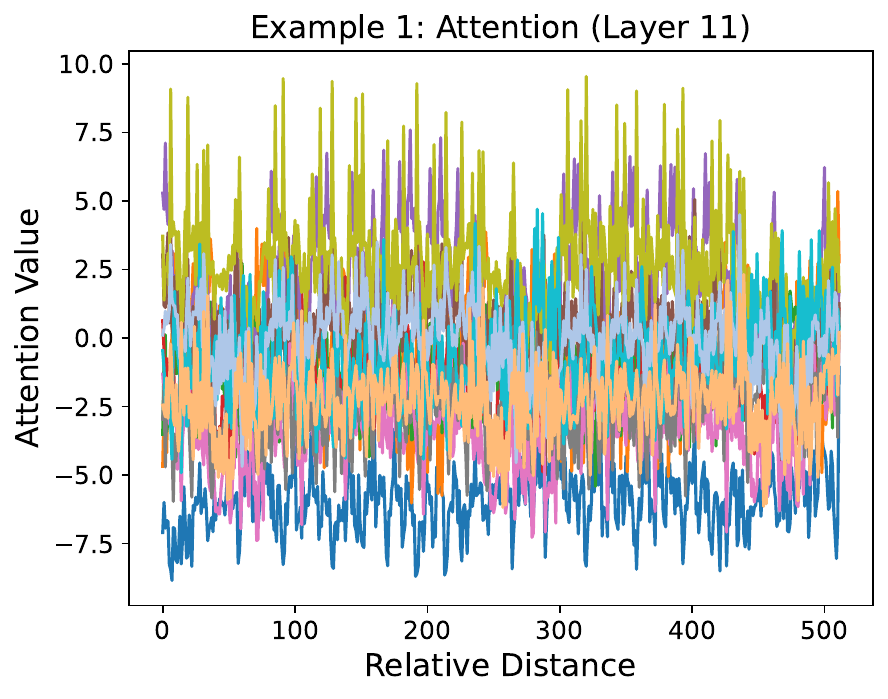}
\hspace{0in}
\includegraphics[width=0.32\textwidth]{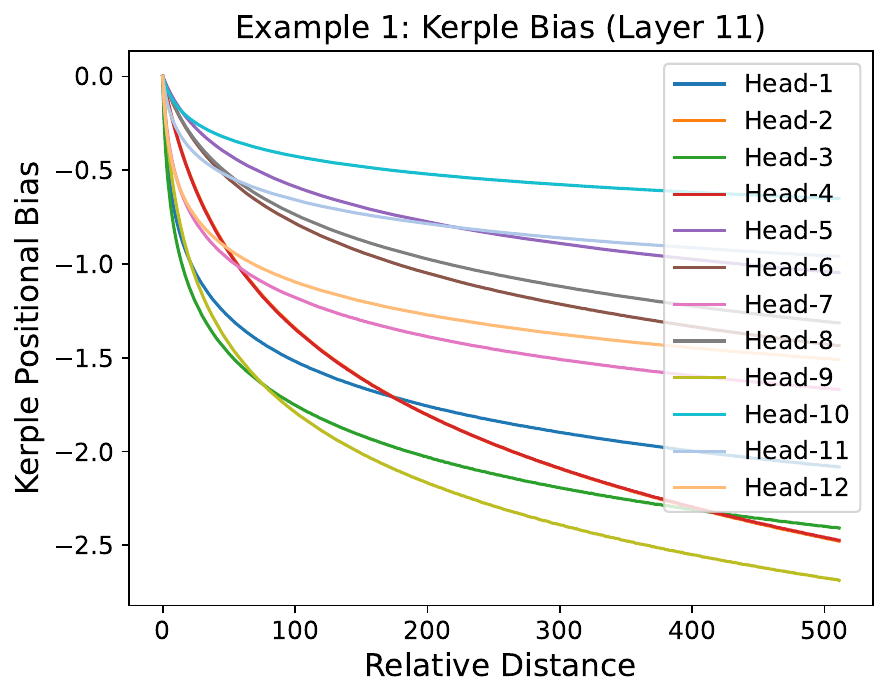}
\hspace{0in}
\includegraphics[width=0.32\textwidth]{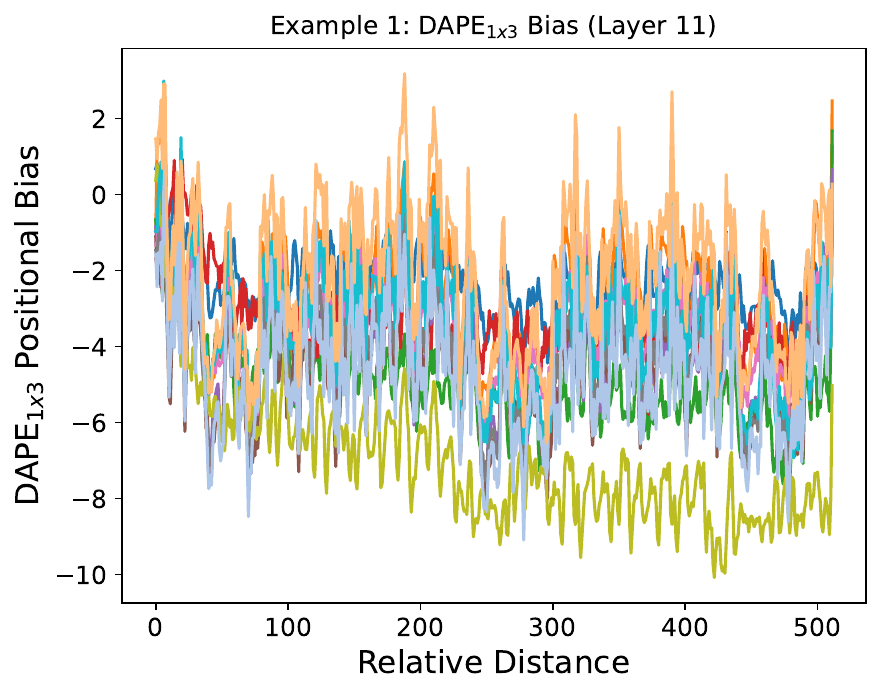}
\hspace{0in}

\includegraphics[width=0.32\textwidth]{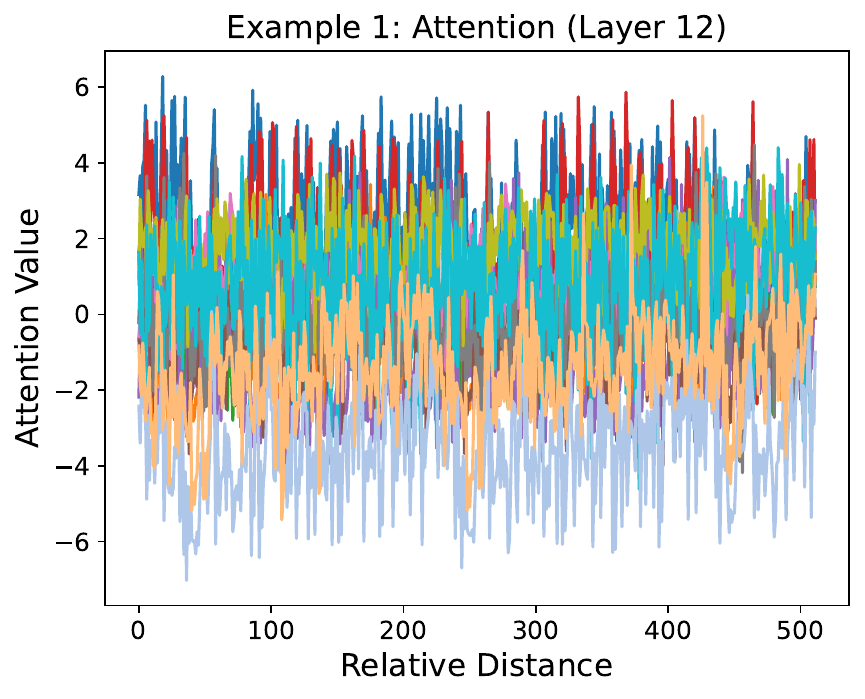}
\hspace{0in}
\includegraphics[width=0.32\textwidth]{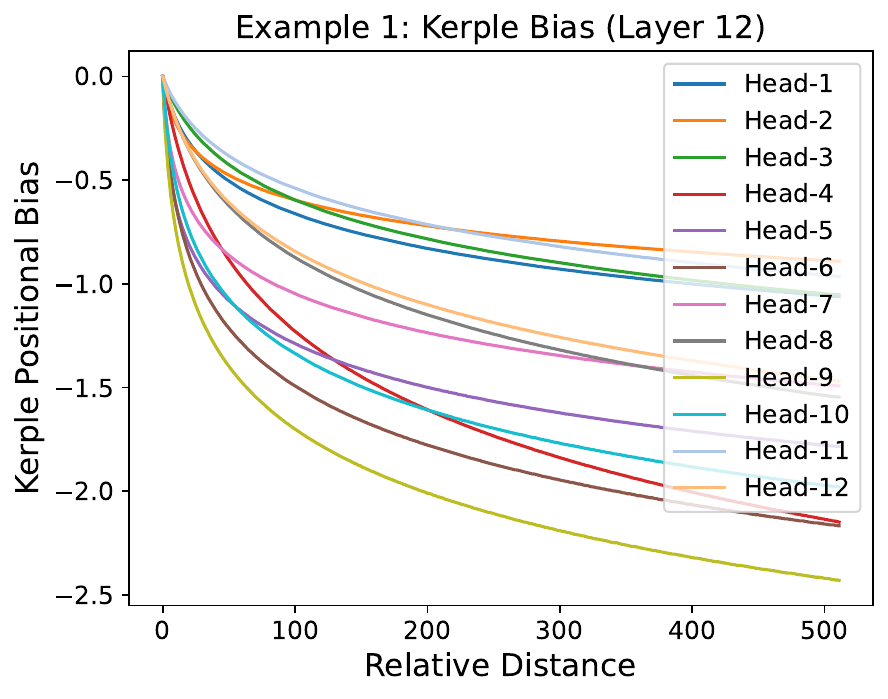}
\hspace{0in}
\includegraphics[width=0.32\textwidth]{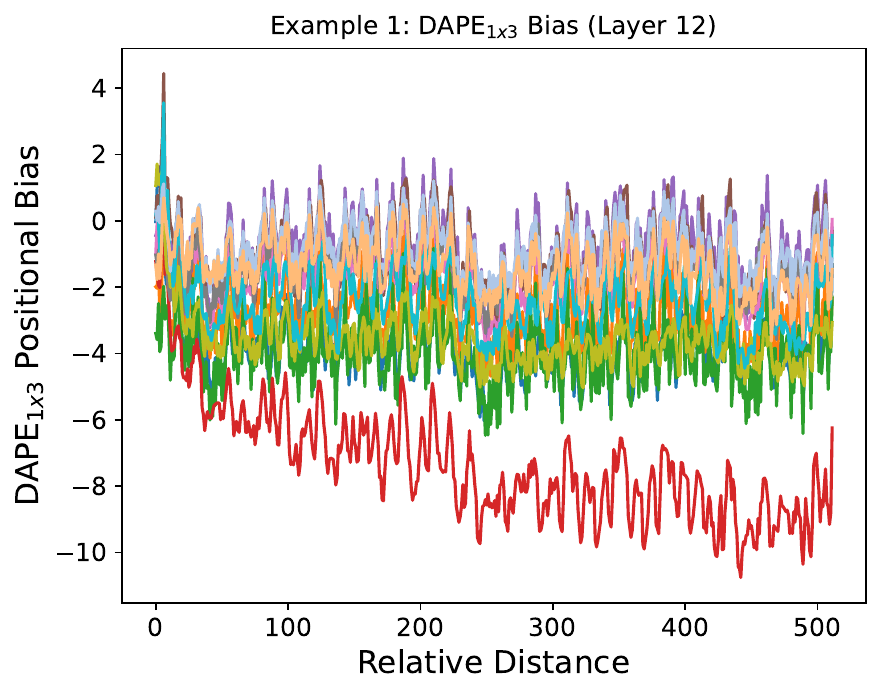}
\caption{
\small
\textbf{Evaluation Length 512 Example 1: Part 3. From Left to Right: (1) The Attention is $\mX \mW_Q(\mX \mW_K)^{\top}$; (2) The Kerple bias is $\mB$; (3) The \methodShortName (with Kerple) bias is $f( \mX \mW_Q(\mX \mW_K)^{\top},\mB)$.
}
}
\end{figure}

\newpage

\begin{figure}[htbp]
\setlength{\abovecaptionskip}{0.1cm}
\centering
\includegraphics[width=0.32\textwidth]{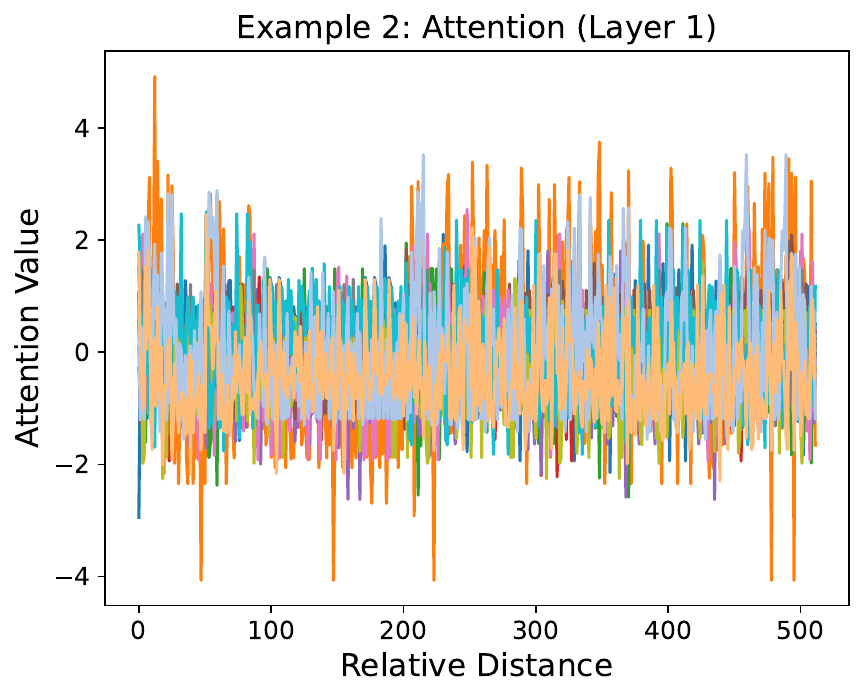}
\hspace{0in}
\includegraphics[width=0.32\textwidth]{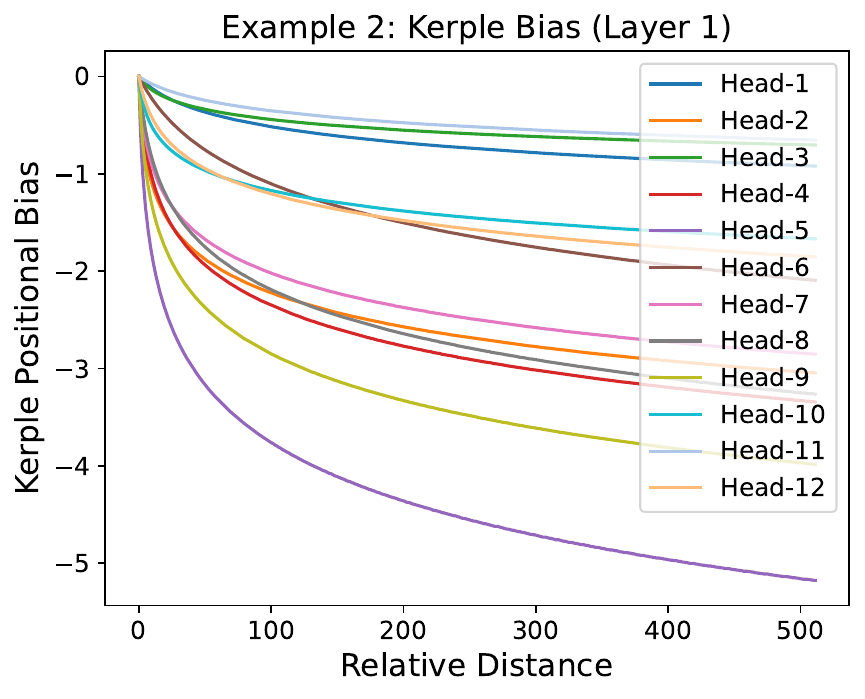}
\hspace{0in}
\includegraphics[width=0.32\textwidth]{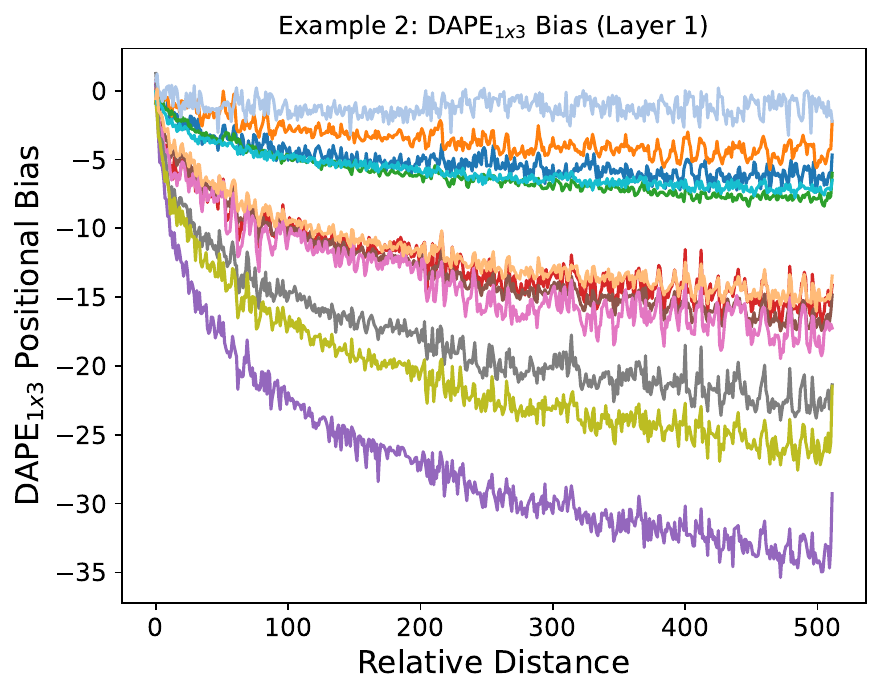}
\hspace{0in}

\includegraphics[width=0.32\textwidth]{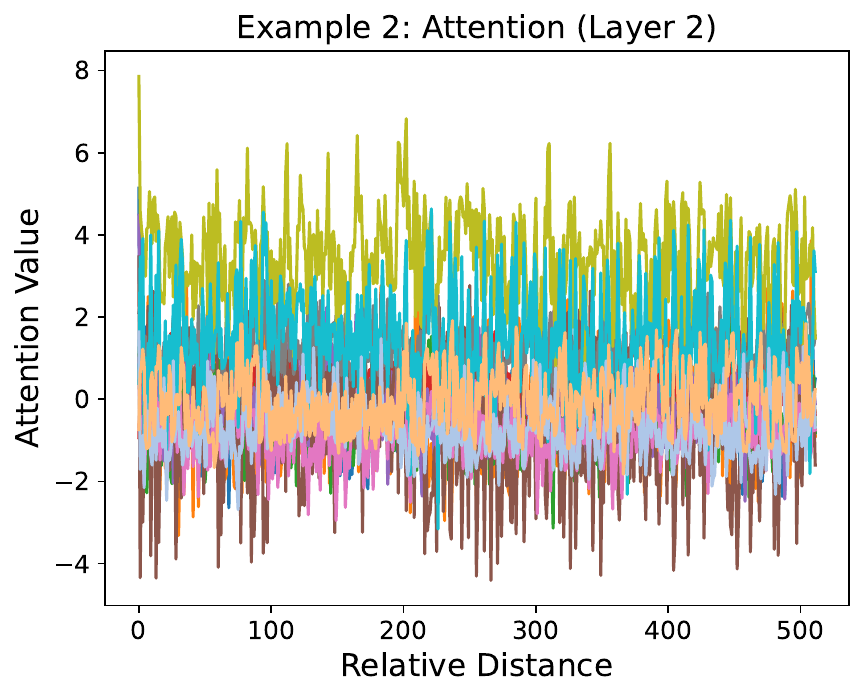}
\hspace{0in}
\includegraphics[width=0.32\textwidth]{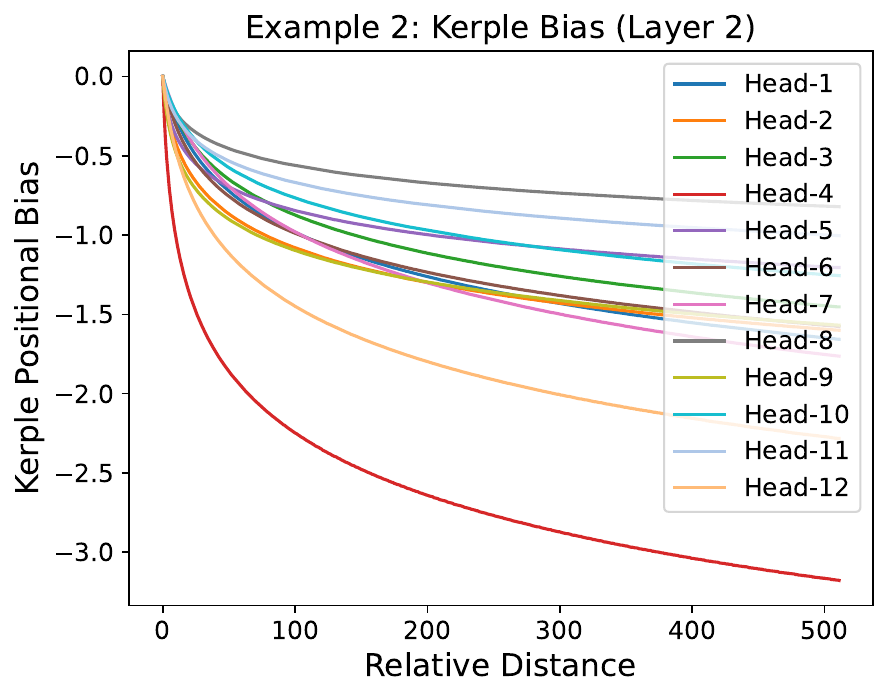}
\hspace{0in}
\includegraphics[width=0.32\textwidth]{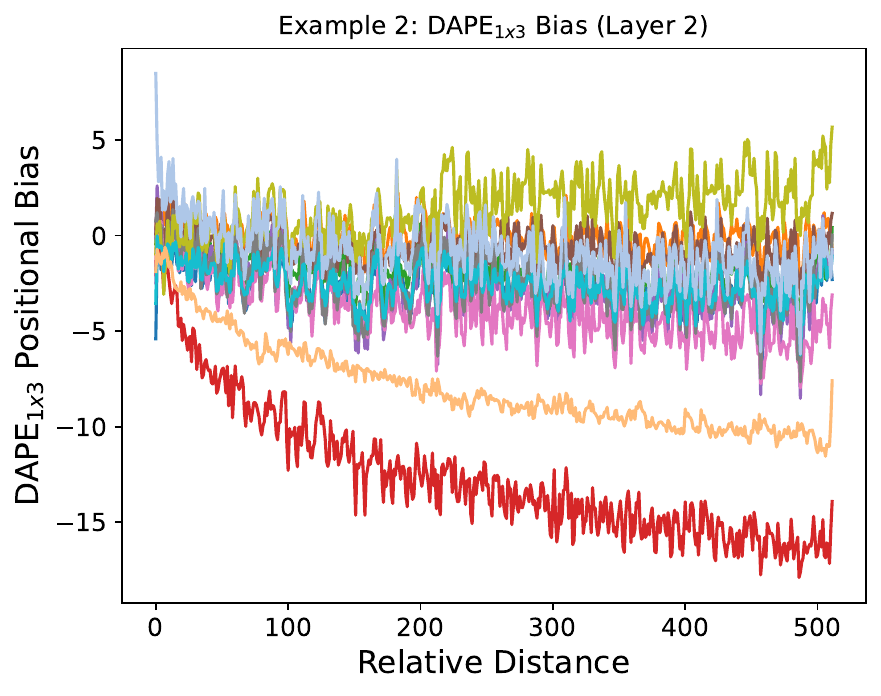}
\hspace{0in}

\includegraphics[width=0.32\textwidth]{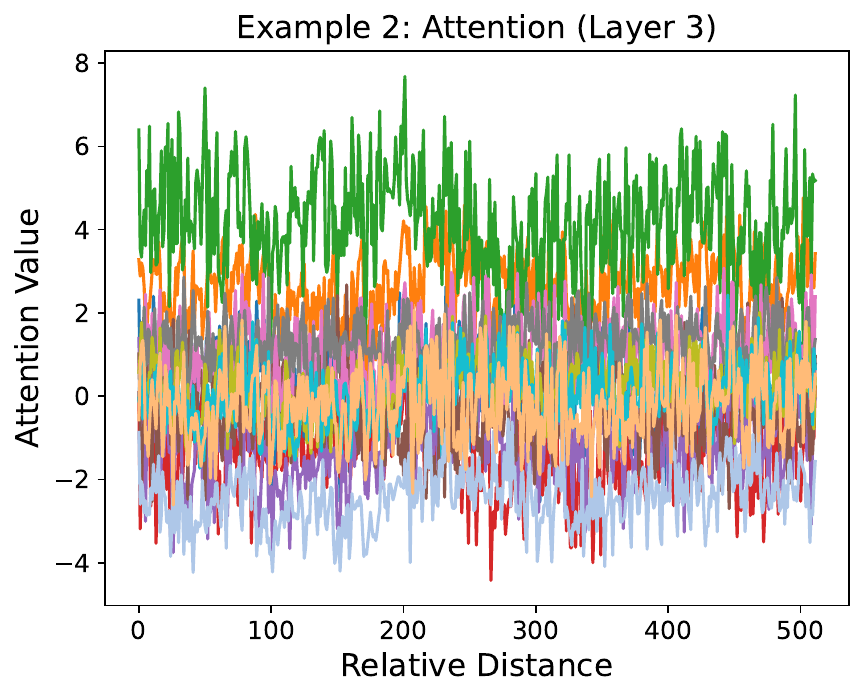}
\hspace{0in}
\includegraphics[width=0.32\textwidth]{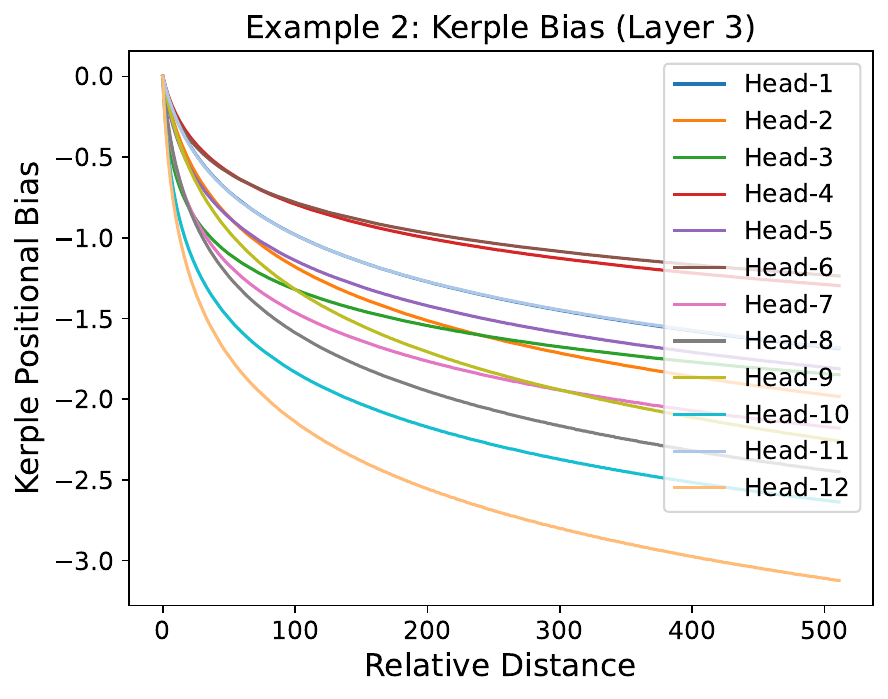}
\hspace{0in}
\includegraphics[width=0.32\textwidth]{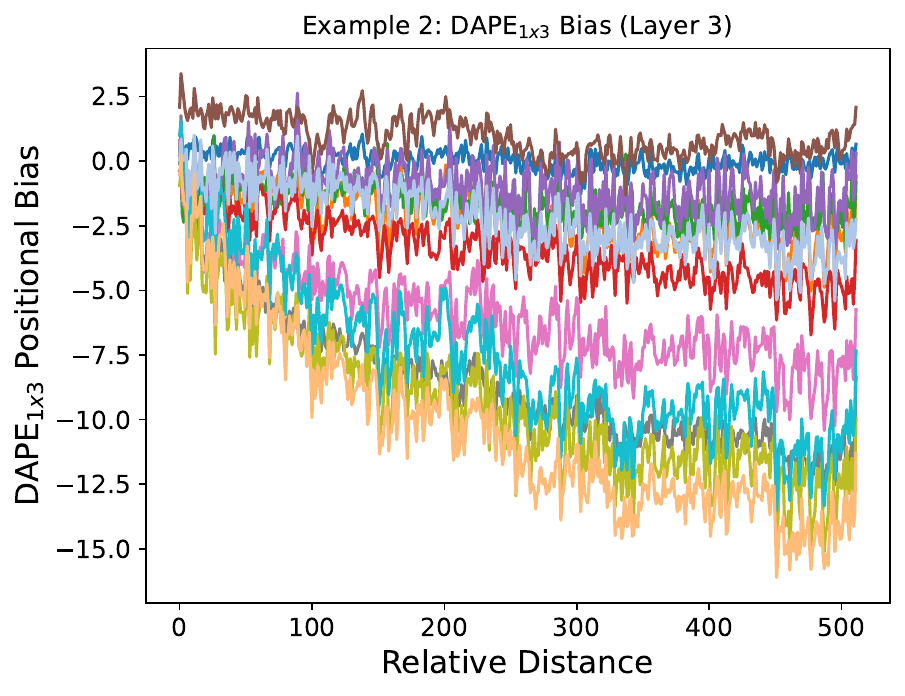}
\hspace{0in}

\includegraphics[width=0.32\textwidth]{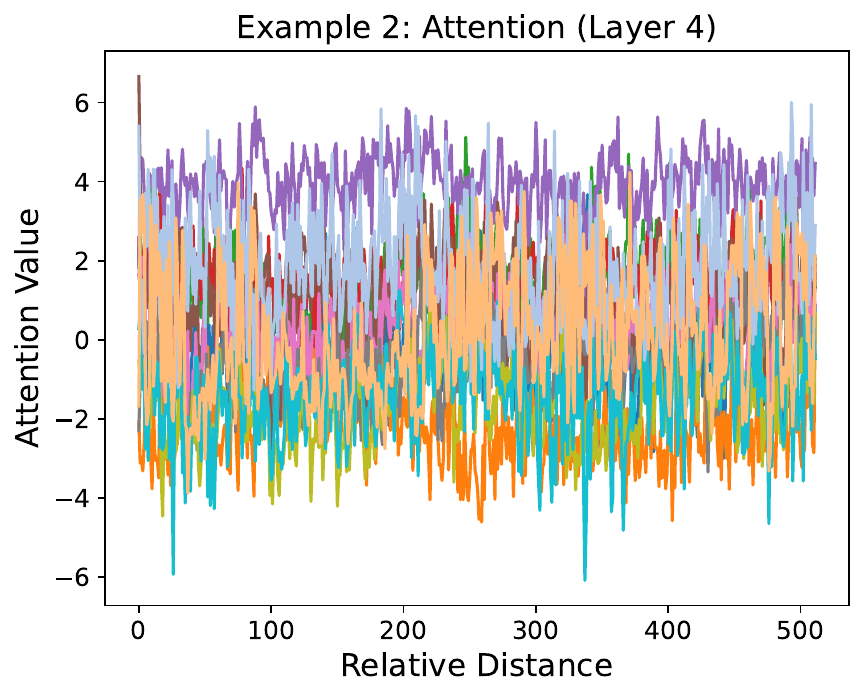}
\hspace{0in}
\includegraphics[width=0.32\textwidth]{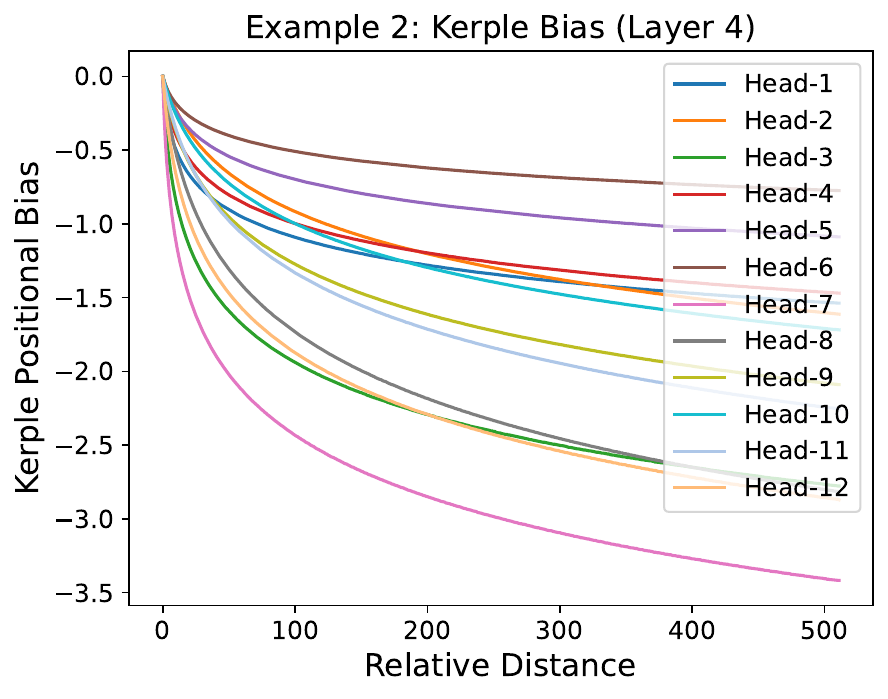}
\hspace{0in}
\includegraphics[width=0.32\textwidth]{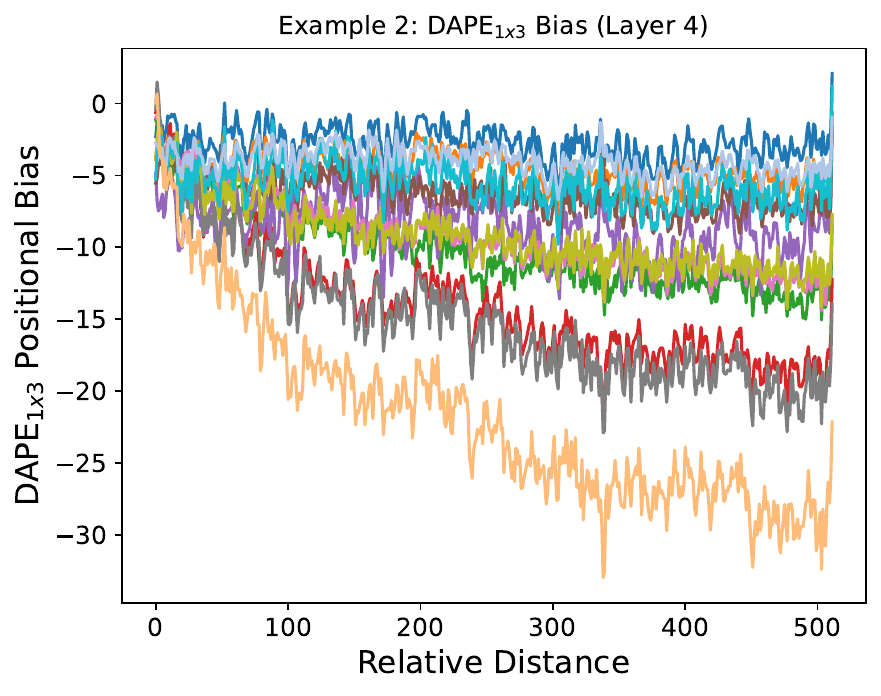}
\hspace{0in}

\includegraphics[width=0.32\textwidth]{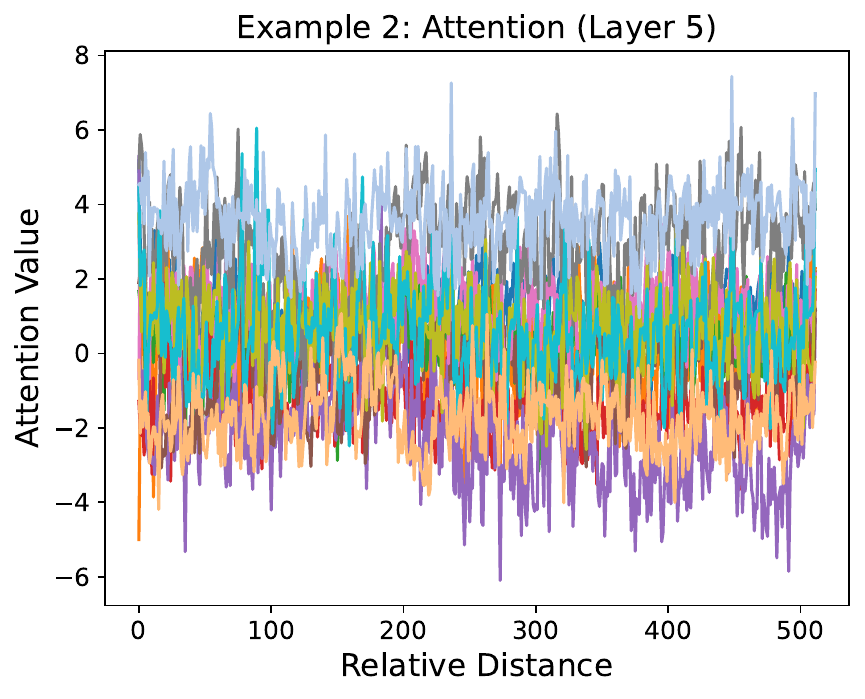}
\hspace{0in}
\includegraphics[width=0.32\textwidth]{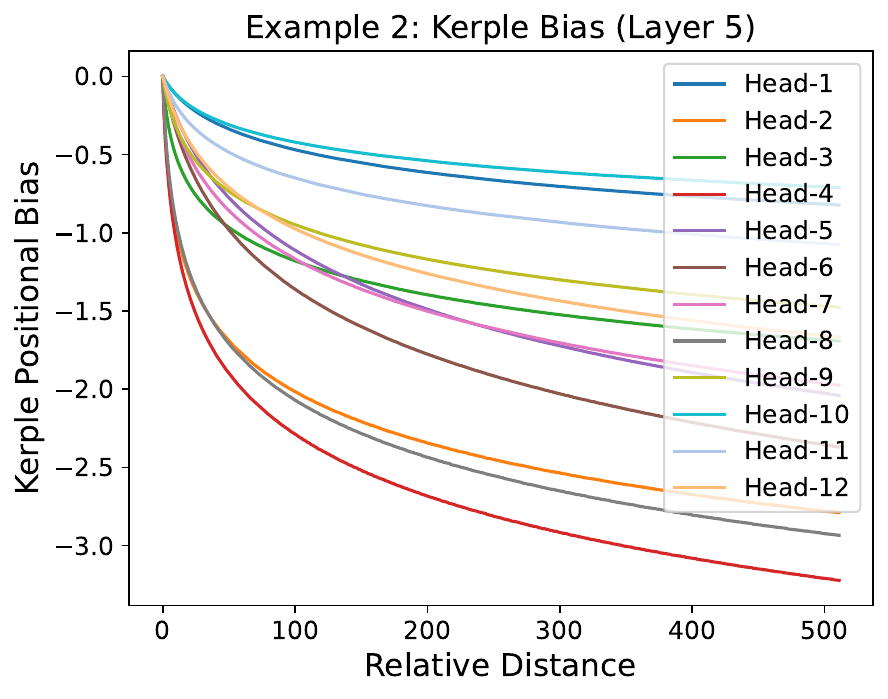}
\hspace{0in}
\includegraphics[width=0.32\textwidth]{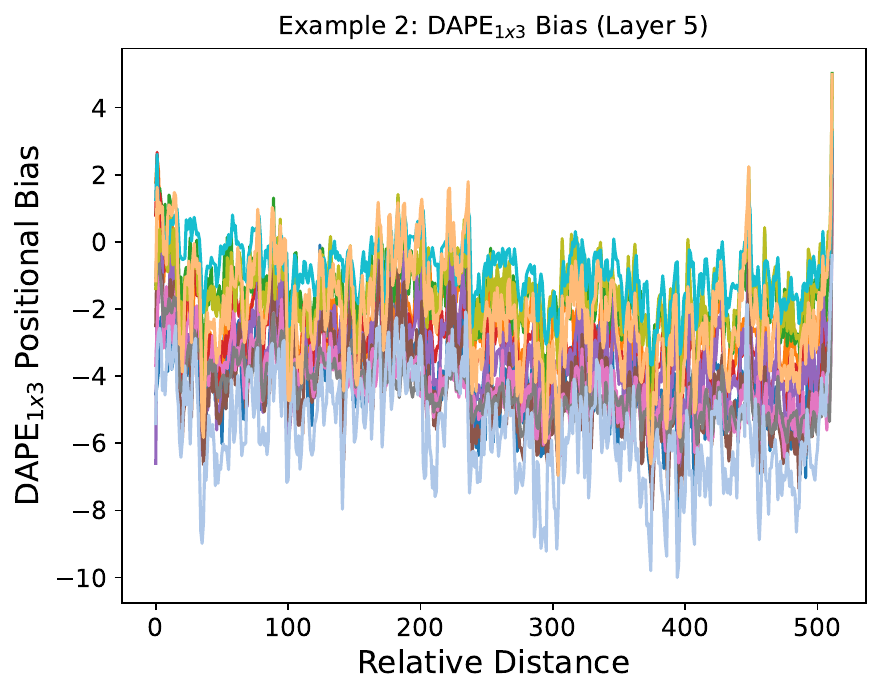}
\hspace{0in}

\includegraphics[width=0.32\textwidth]{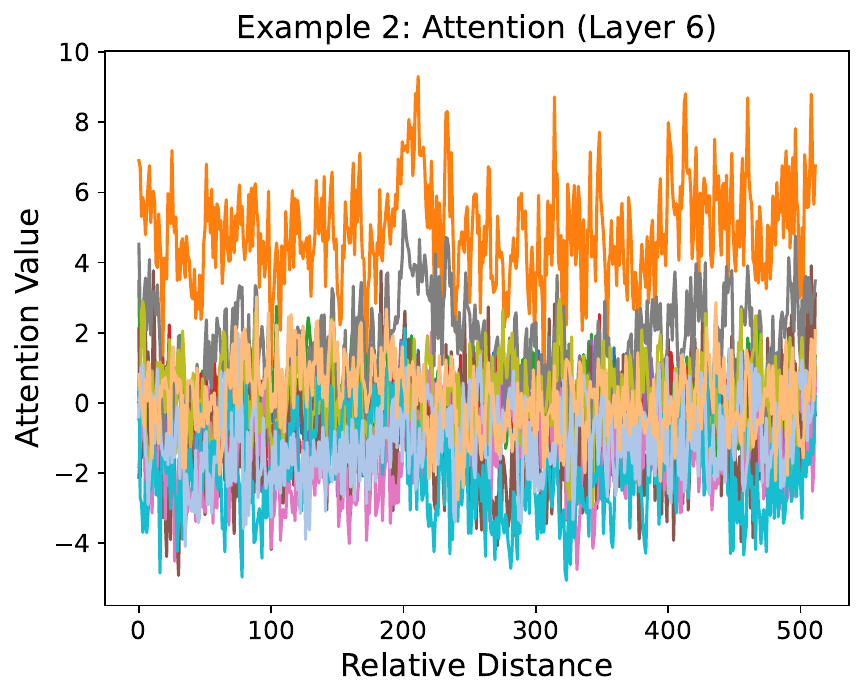}
\hspace{0in}
\includegraphics[width=0.32\textwidth]{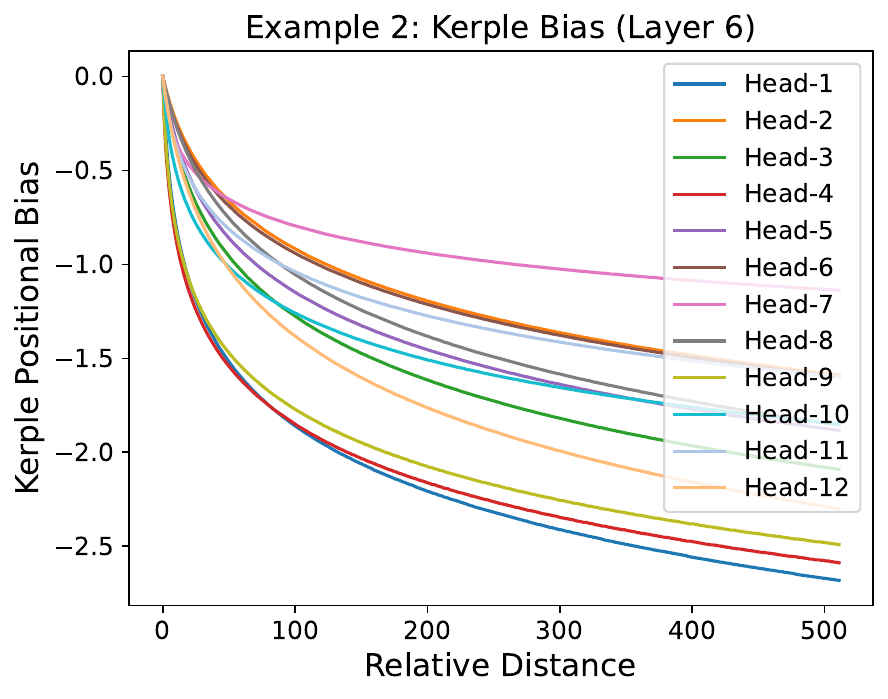}
\hspace{0in}
\includegraphics[width=0.32\textwidth]{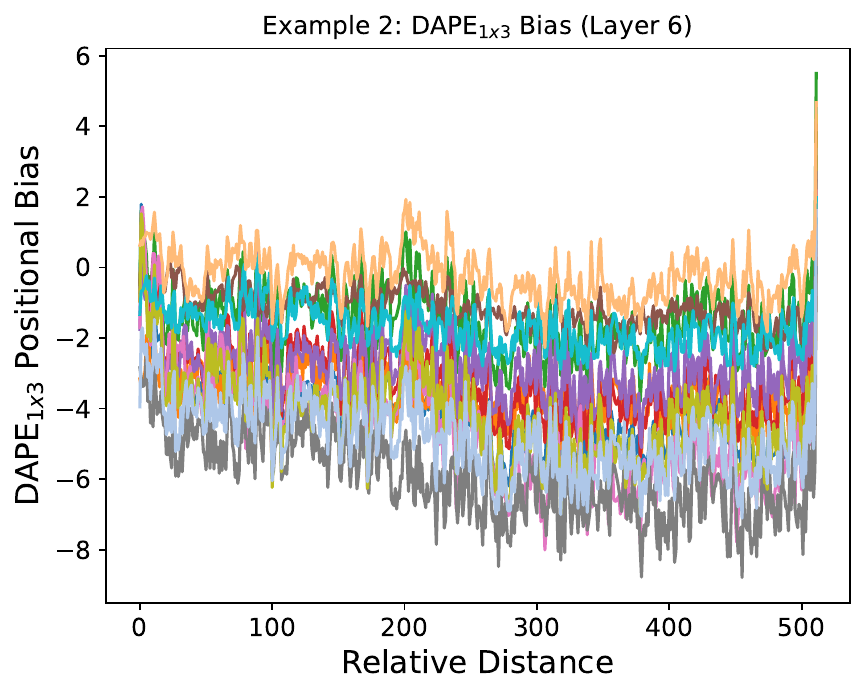}
\hspace{0in}

\hspace{0in}
\caption{
\small
\textbf{Evaluation Length 512 Example 2: Part 1. From Left to Right: (1) The Attention is $\mX \mW_Q(\mX \mW_K)^{\top}$; (2) The Kerple bias is $\mB$; (3) The \methodShortName (with Kerple) bias is $f( \mX \mW_Q(\mX \mW_K)^{\top},\mB)$.
}
}
\end{figure}

\begin{figure}[htbp]
\setlength{\abovecaptionskip}{0.1cm}
\centering

\includegraphics[width=0.32\textwidth]{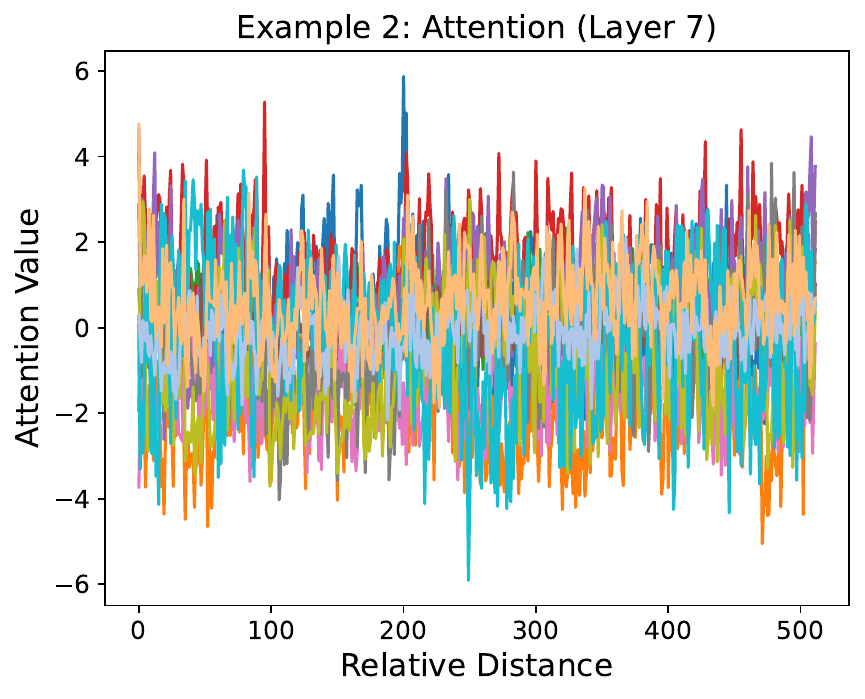}
\hspace{0in}
\includegraphics[width=0.32\textwidth]{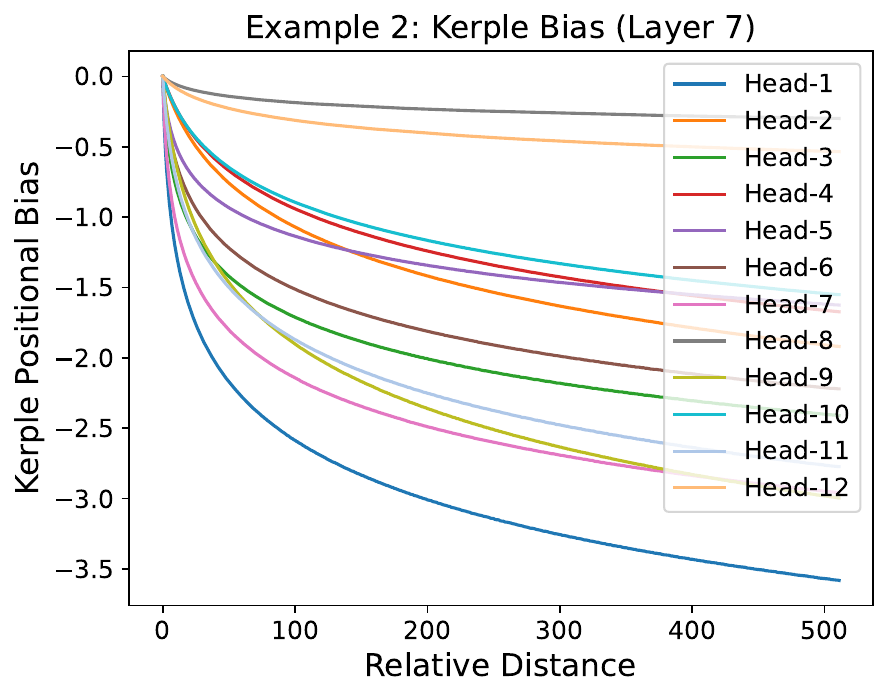}
\hspace{0in}
\includegraphics[width=0.32\textwidth]{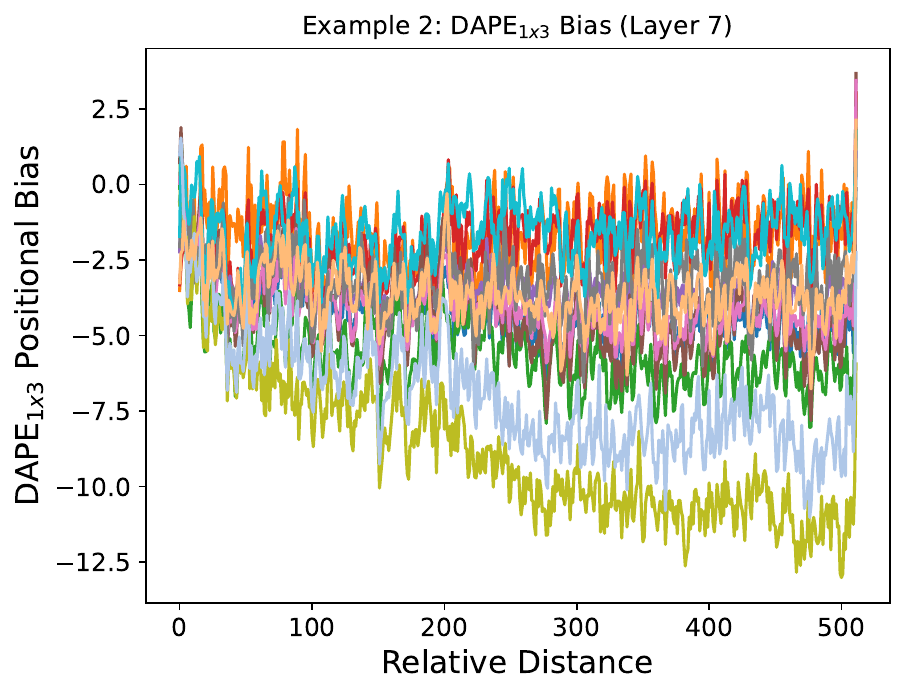}
\hspace{0in}

\includegraphics[width=0.32\textwidth]{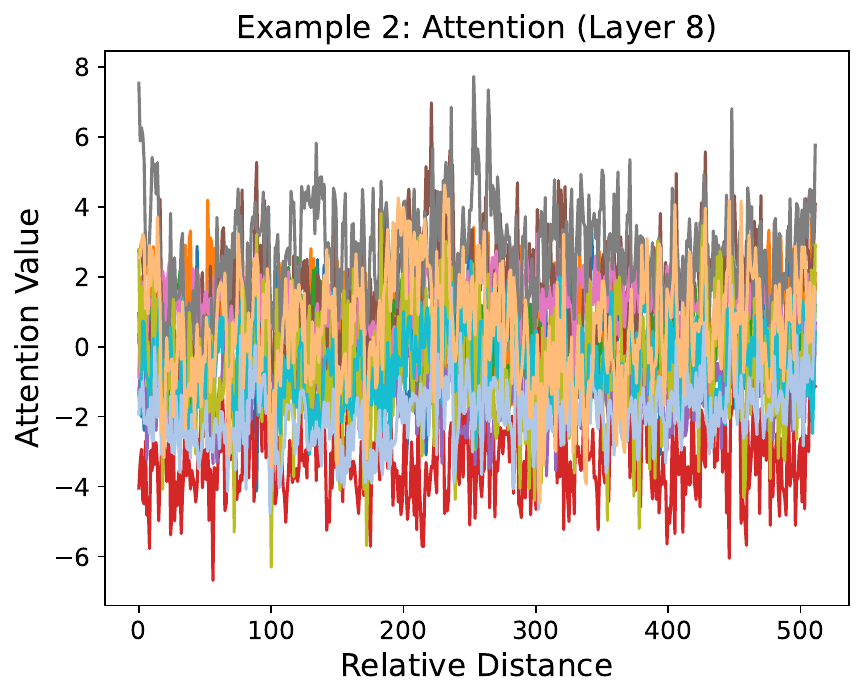}
\hspace{0in}
\includegraphics[width=0.32\textwidth]{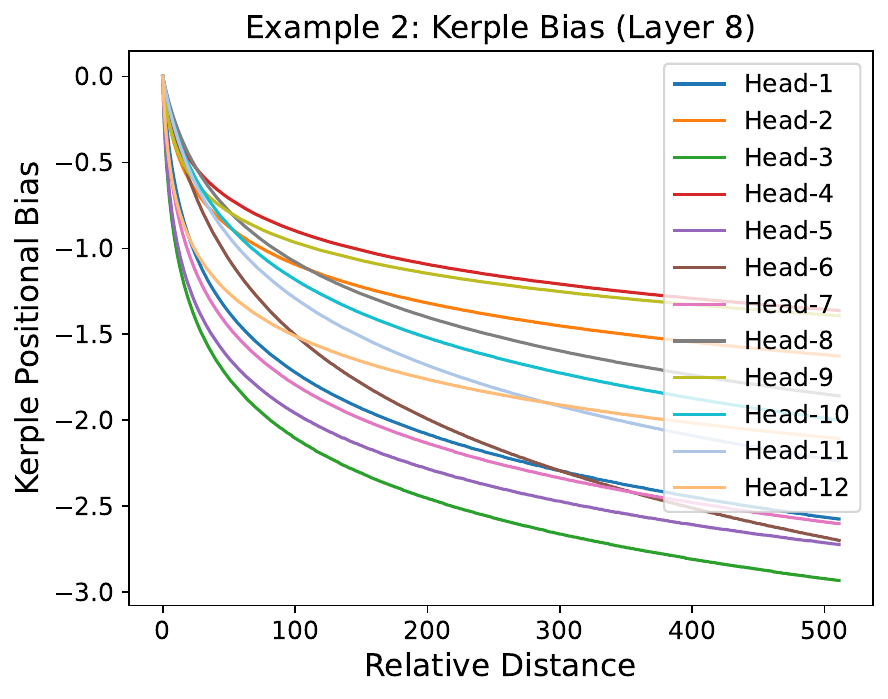}
\hspace{0in}
\includegraphics[width=0.32\textwidth]{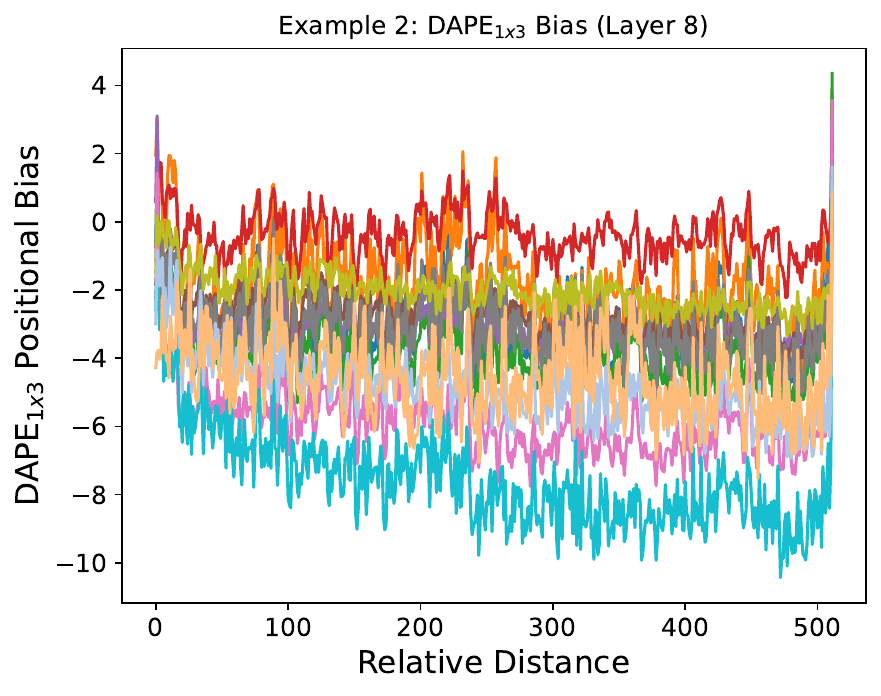}
\hspace{0in}

\includegraphics[width=0.32\textwidth]{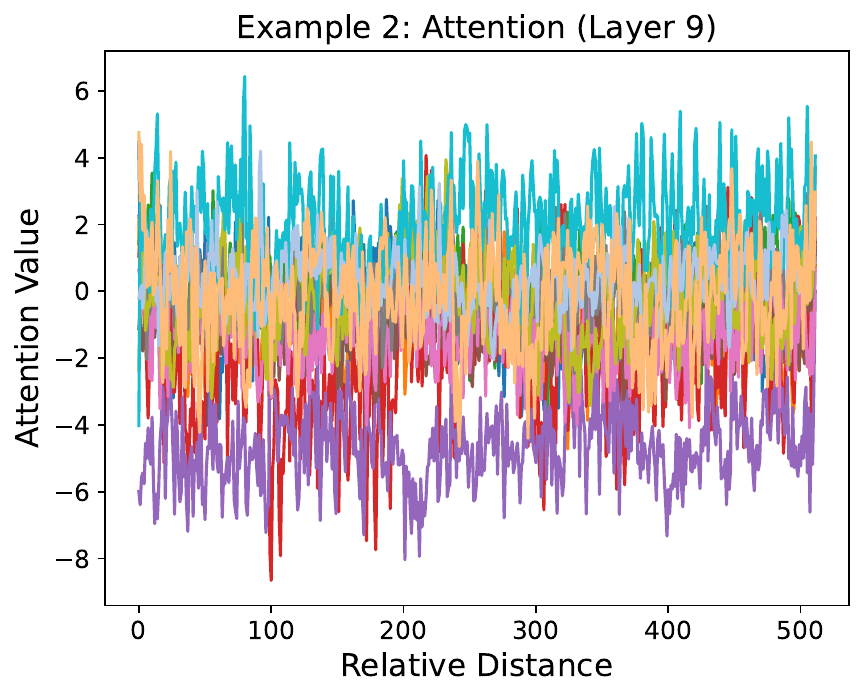}
\hspace{0in}
\includegraphics[width=0.32\textwidth]{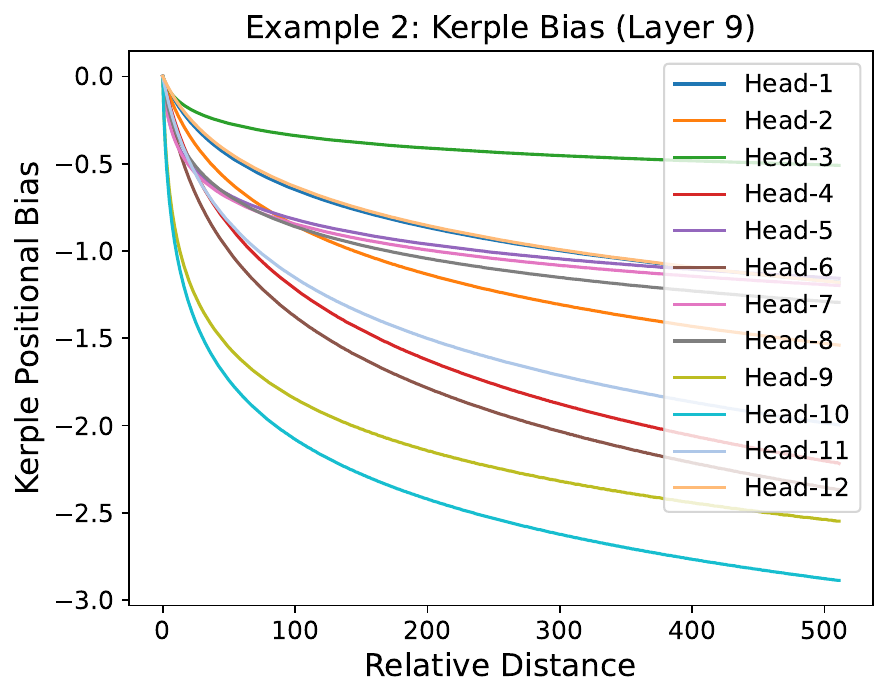}
\hspace{0in}
\includegraphics[width=0.32\textwidth]{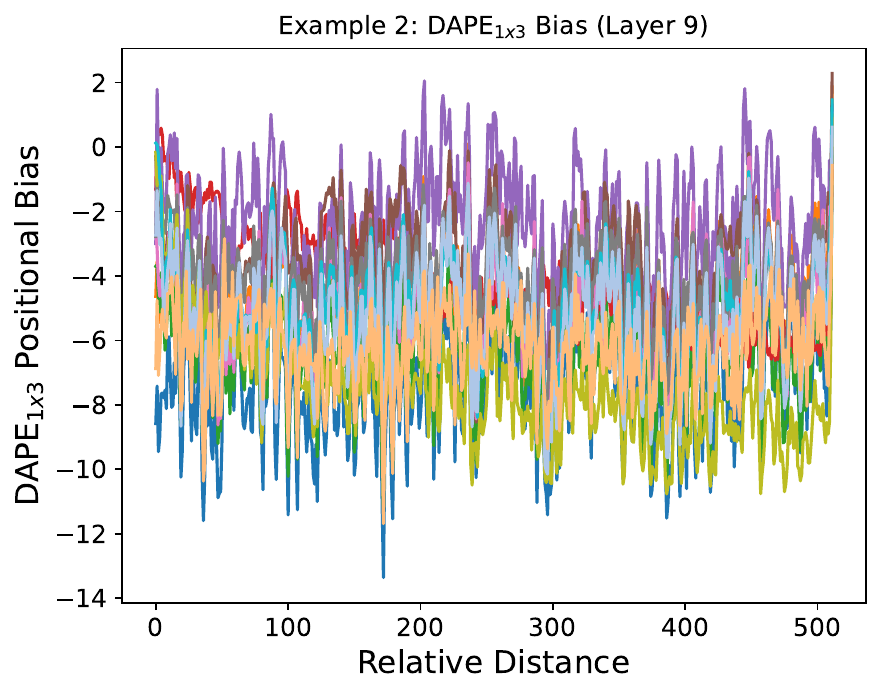}

\includegraphics[width=0.32\textwidth]{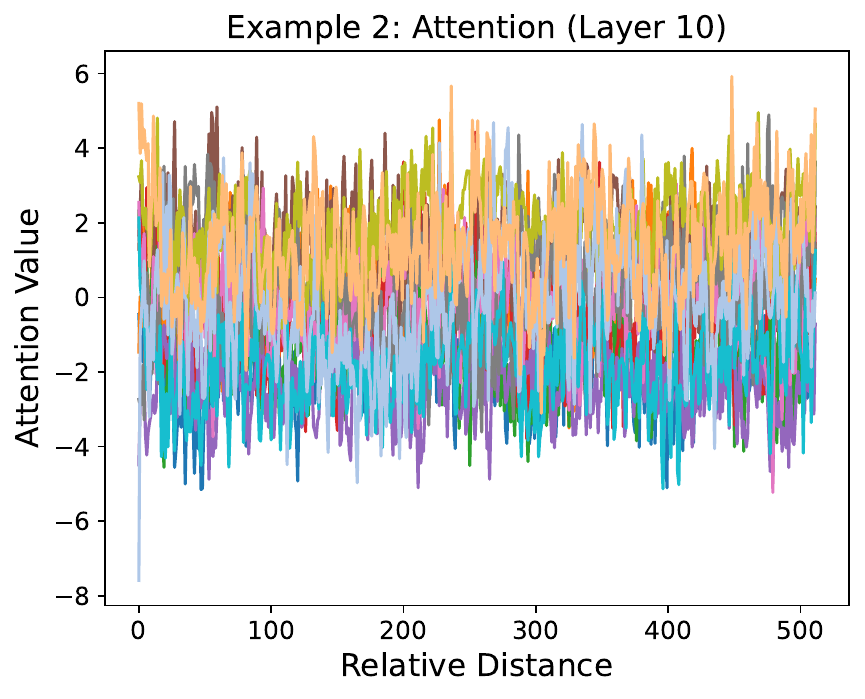}
\hspace{0in}
\includegraphics[width=0.32\textwidth]{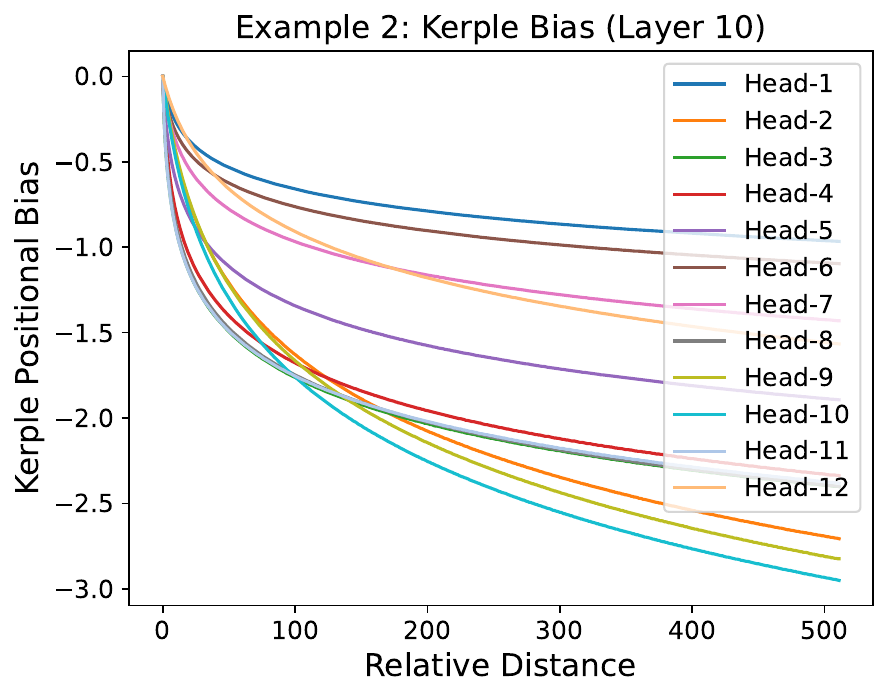}
\hspace{0in}
\includegraphics[width=0.32\textwidth]{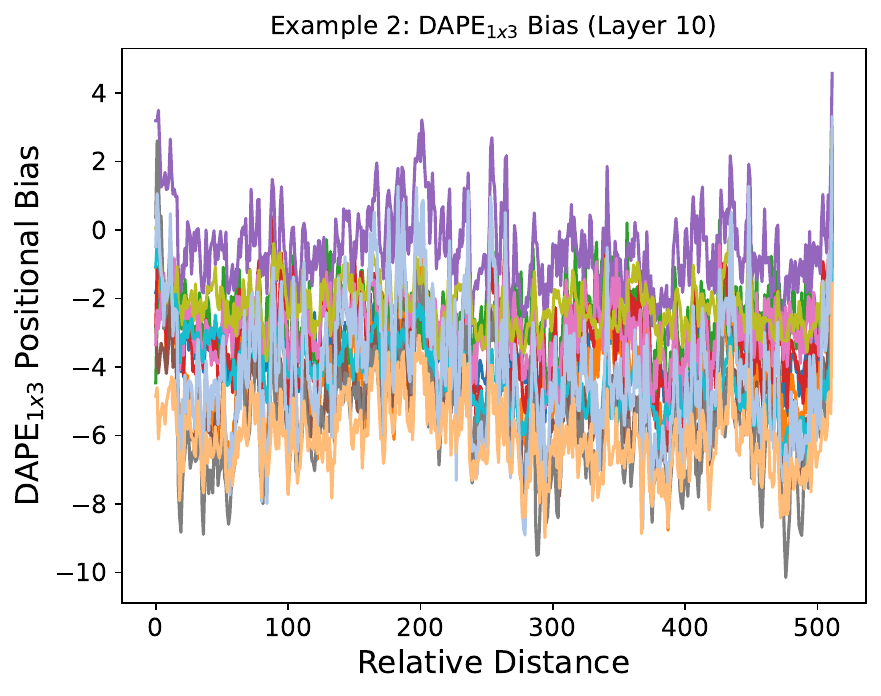}

\includegraphics[width=0.32\textwidth]{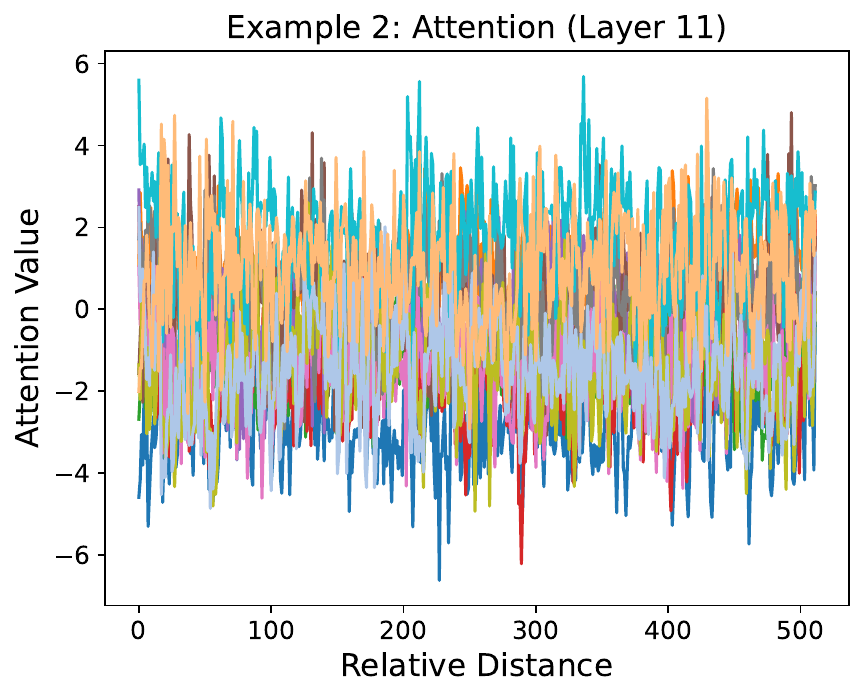}
\hspace{0in}
\includegraphics[width=0.32\textwidth]{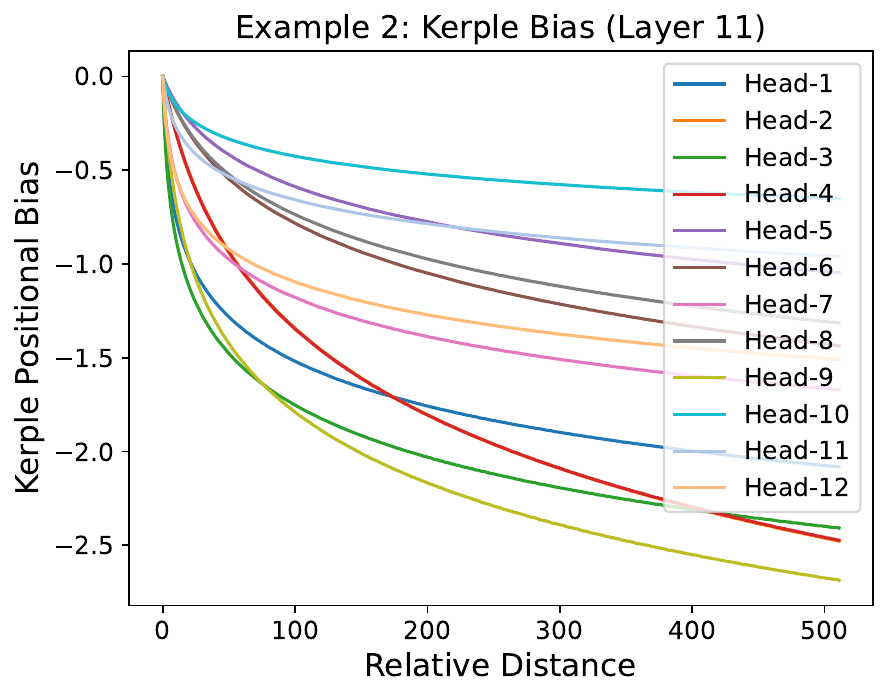}
\hspace{0in}
\includegraphics[width=0.32\textwidth]{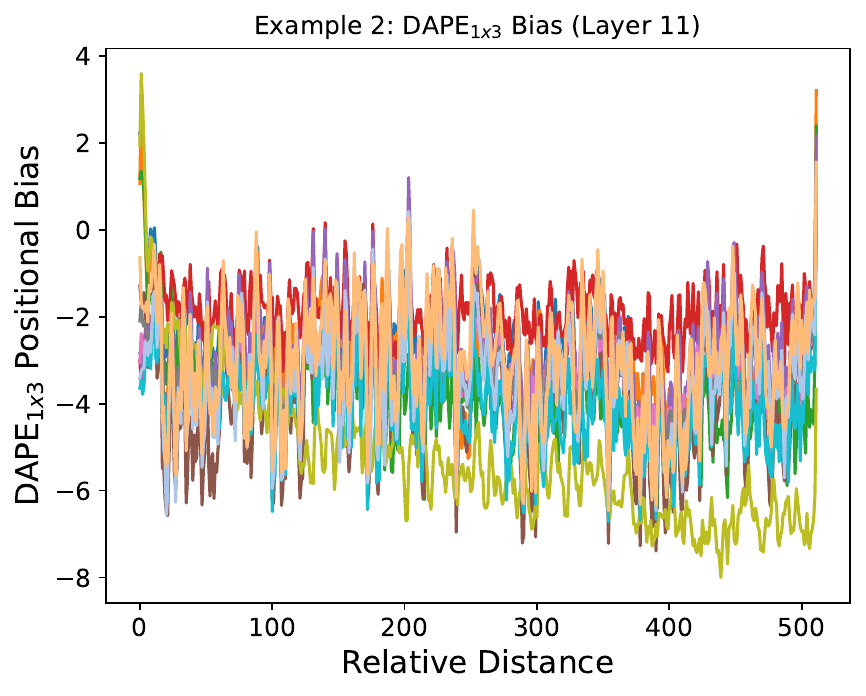}
\hspace{0in}

\includegraphics[width=0.32\textwidth]{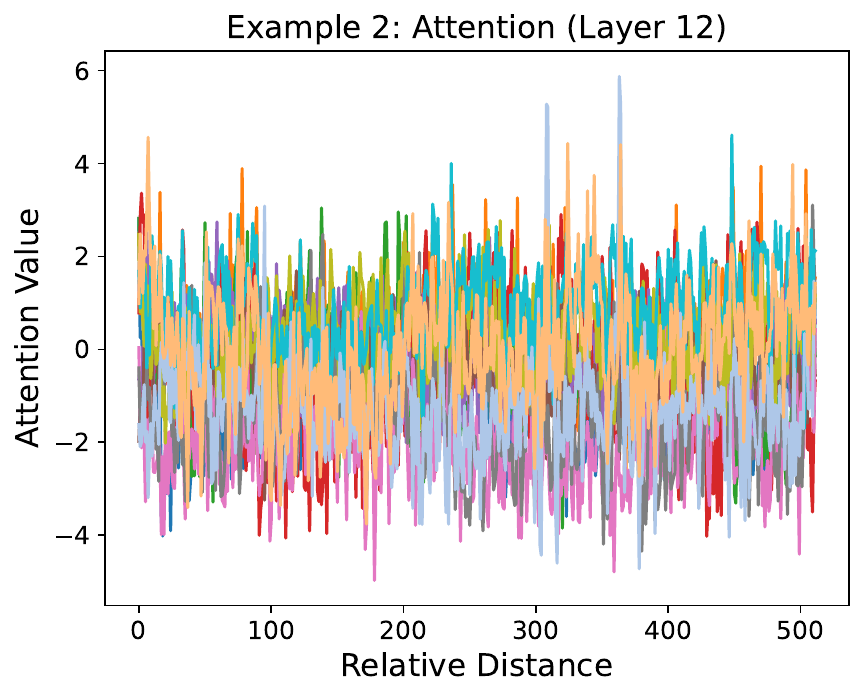}
\hspace{0in}
\includegraphics[width=0.32\textwidth]{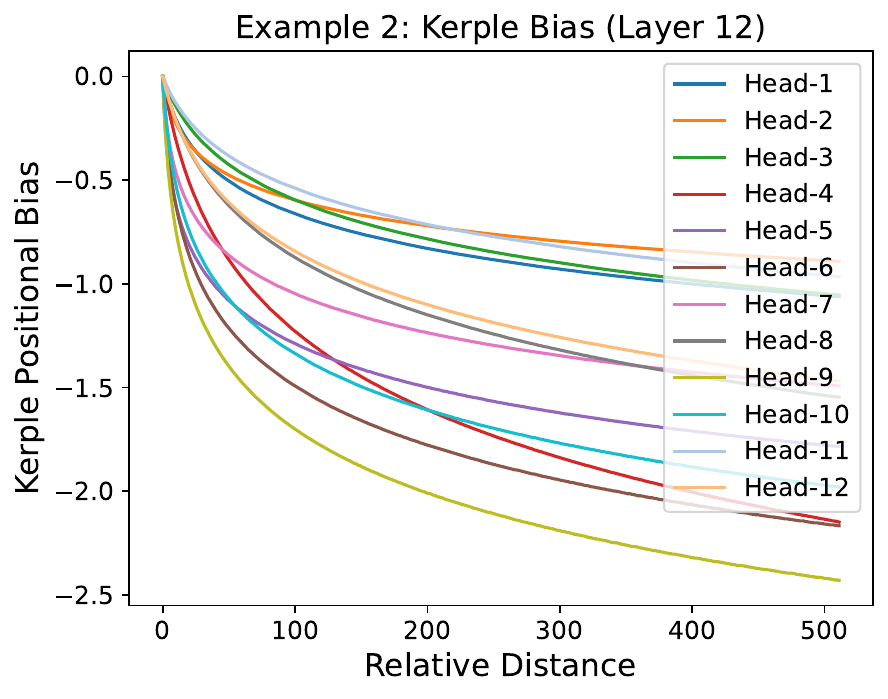}
\hspace{0in}
\includegraphics[width=0.32\textwidth]{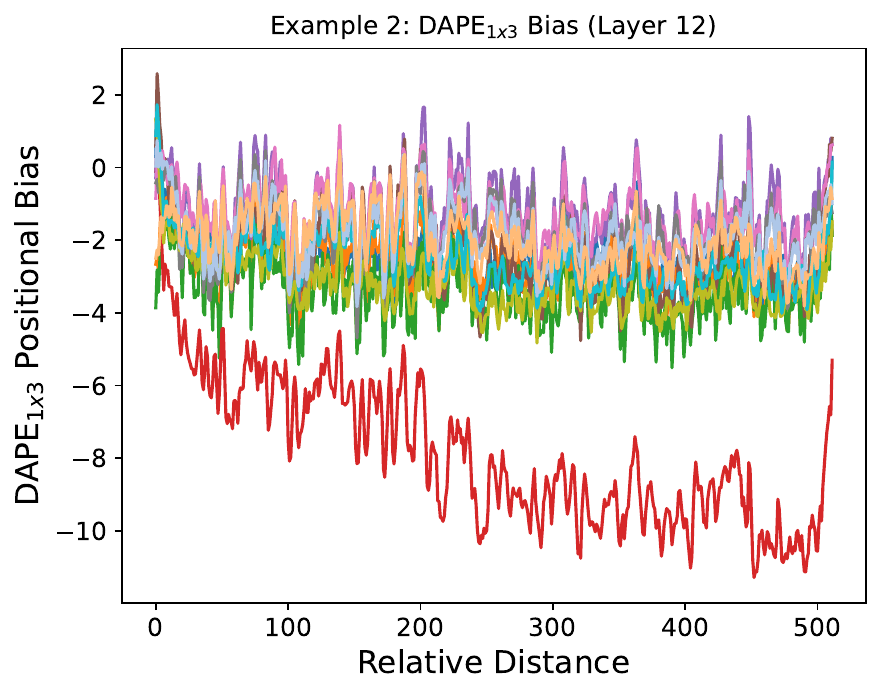}

\hspace{0in}
\caption{
\small
\textbf{Evaluation Length 512 Example 2: Part 2. From Left to Right: (1) The Attention is $\mX \mW_Q(\mX \mW_K)^{\top}$; (2) The Kerple bias is $\mB$; (3) The \methodShortName (with Kerple) bias is $f( \mX \mW_Q(\mX \mW_K)^{\top},\mB)$.
}
}
\end{figure}

\clearpage
\newpage

\subsection{Visualization on length 2048}
\begin{figure}[htbp]
\setlength{\abovecaptionskip}{0.1cm}
\centering
\includegraphics[width=0.32\textwidth]{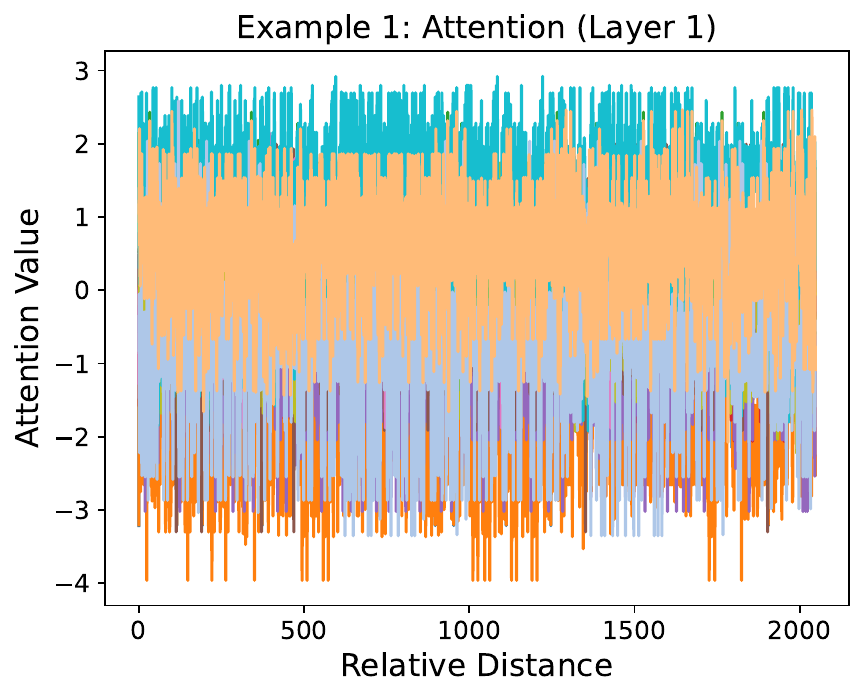}
\hspace{0in}
\includegraphics[width=0.32\textwidth]{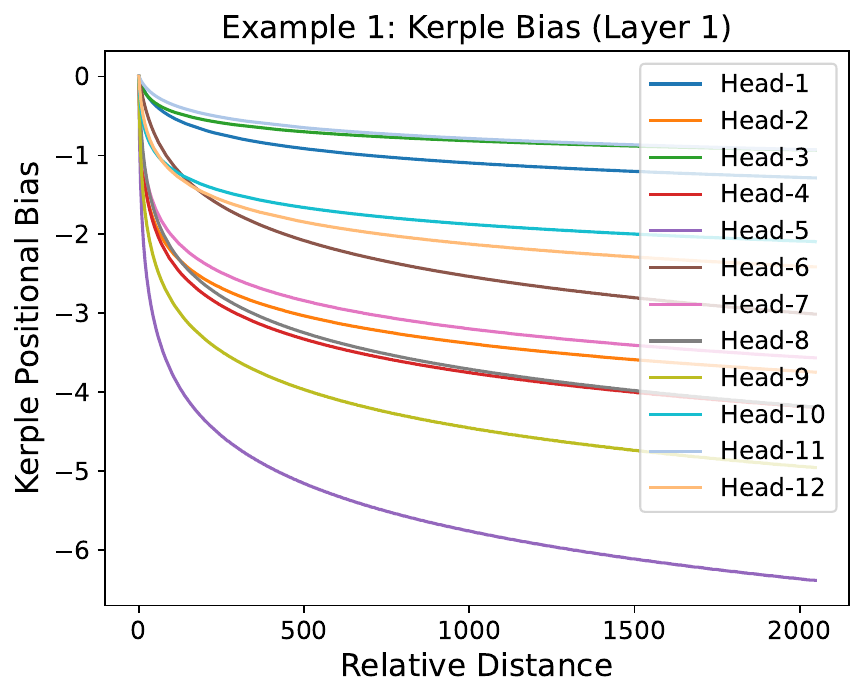}
\hspace{0in}
\includegraphics[width=0.32\textwidth]{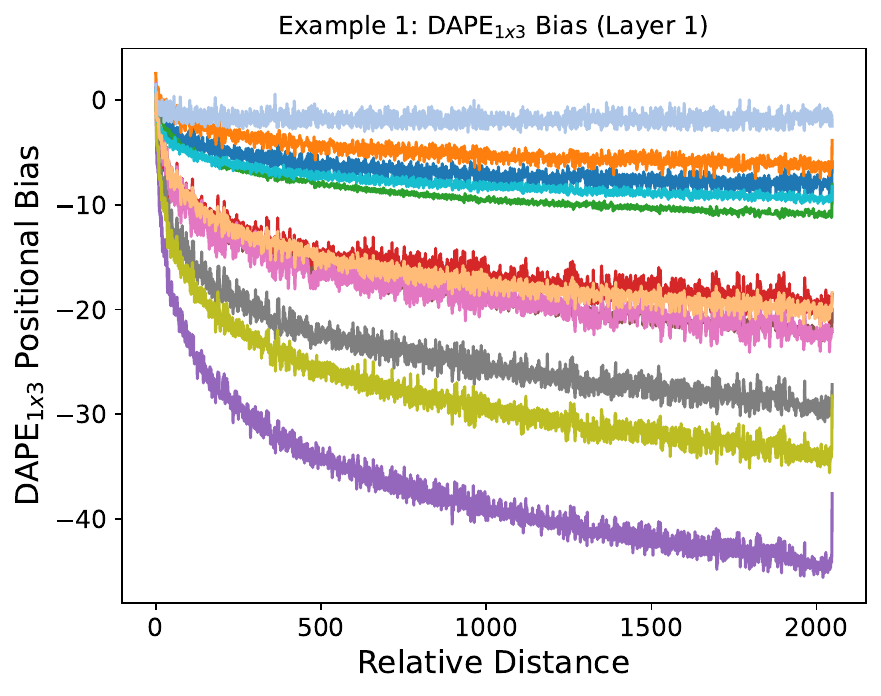}
\hspace{0in}

\includegraphics[width=0.32\textwidth]{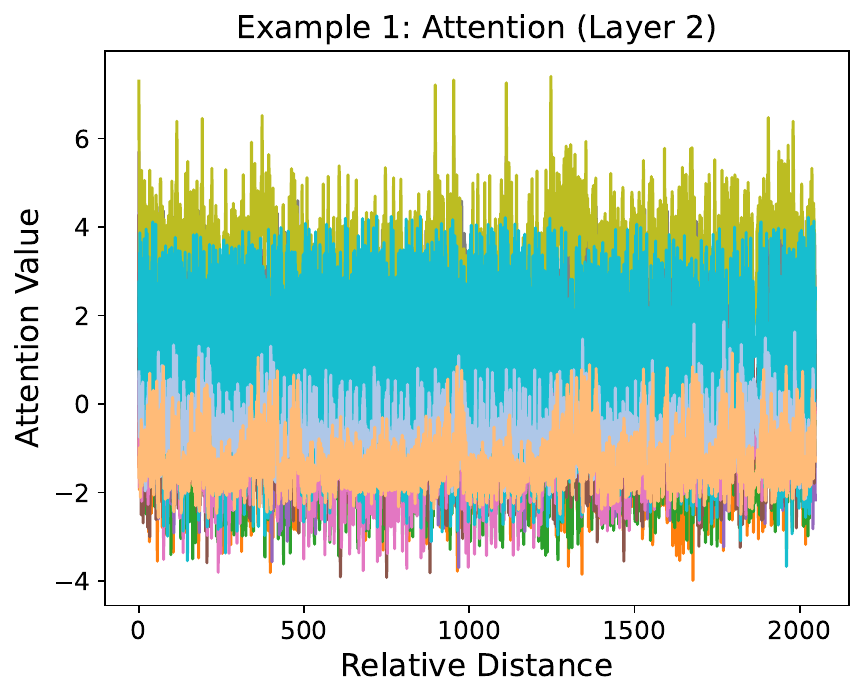}
\hspace{0in}
\includegraphics[width=0.32\textwidth]{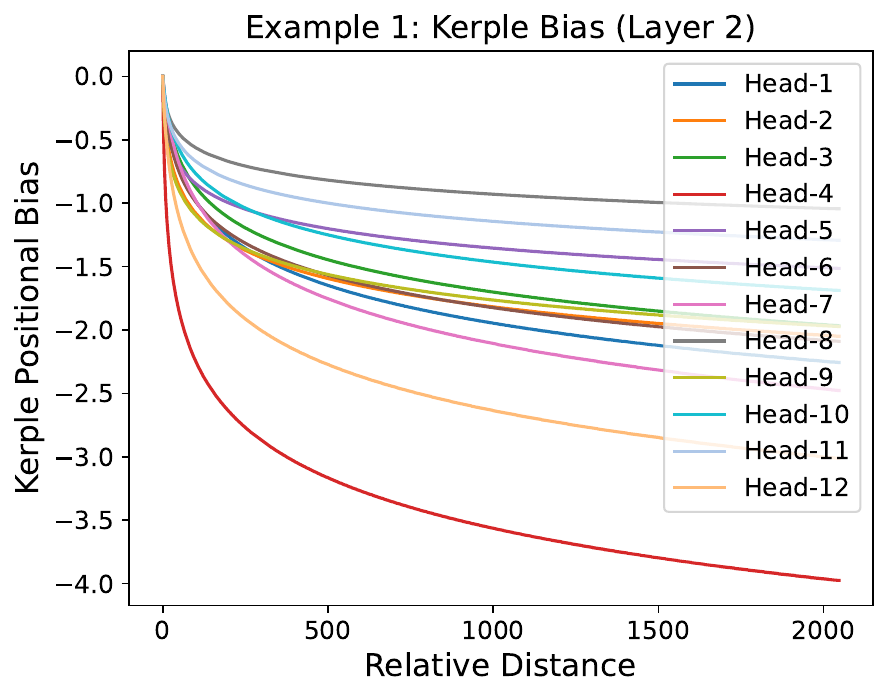}
\hspace{0in}
\includegraphics[width=0.32\textwidth]{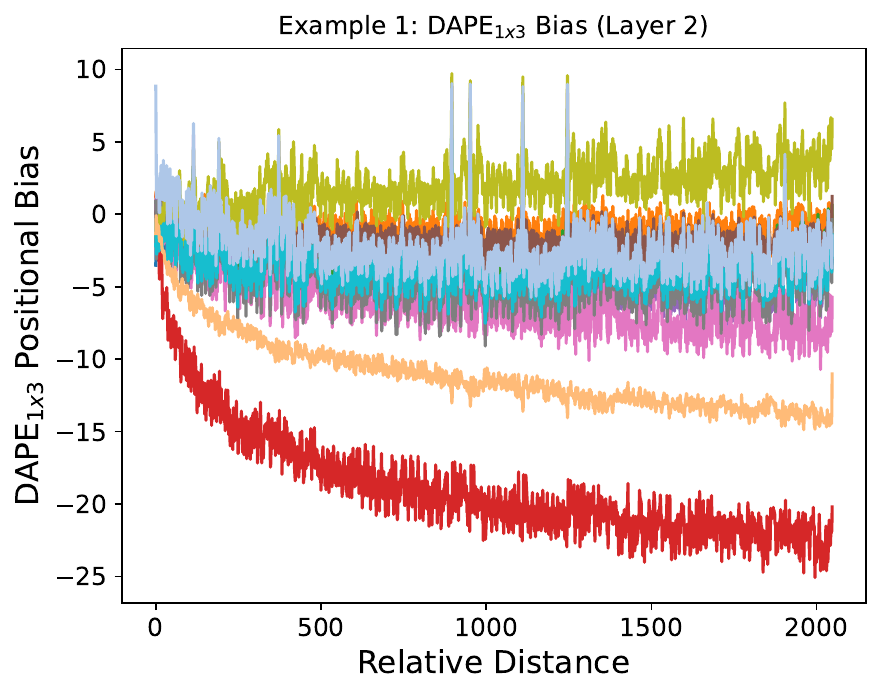}
\hspace{0in}

\includegraphics[width=0.32\textwidth]{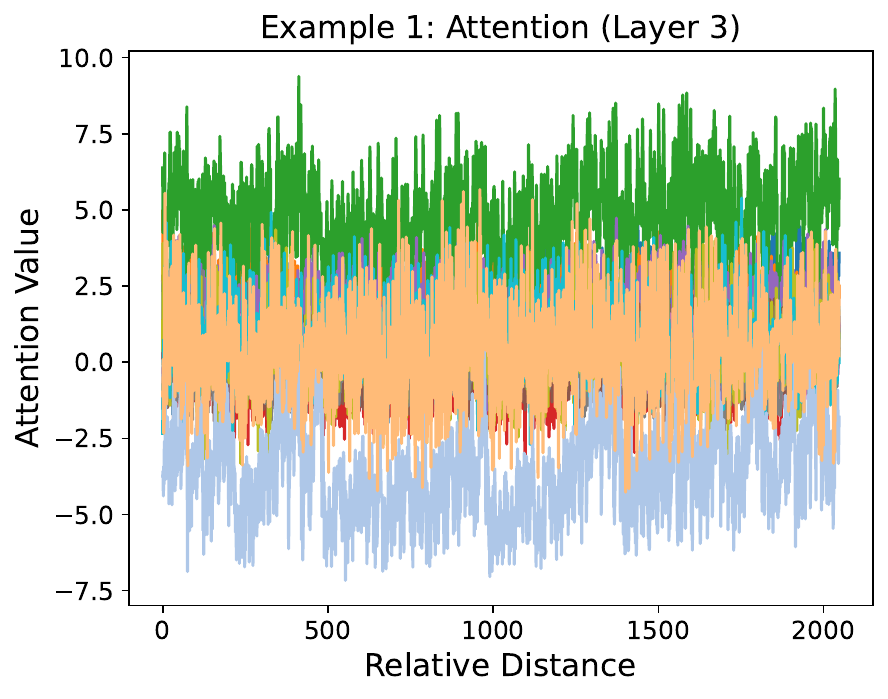}
\hspace{0in}
\includegraphics[width=0.32\textwidth]{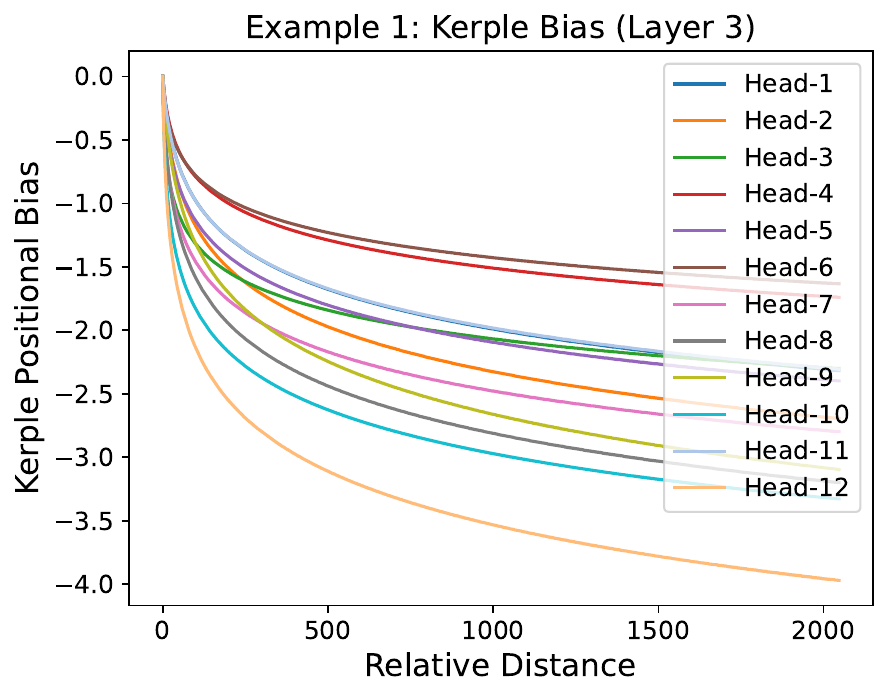}
\hspace{0in}
\includegraphics[width=0.32\textwidth]{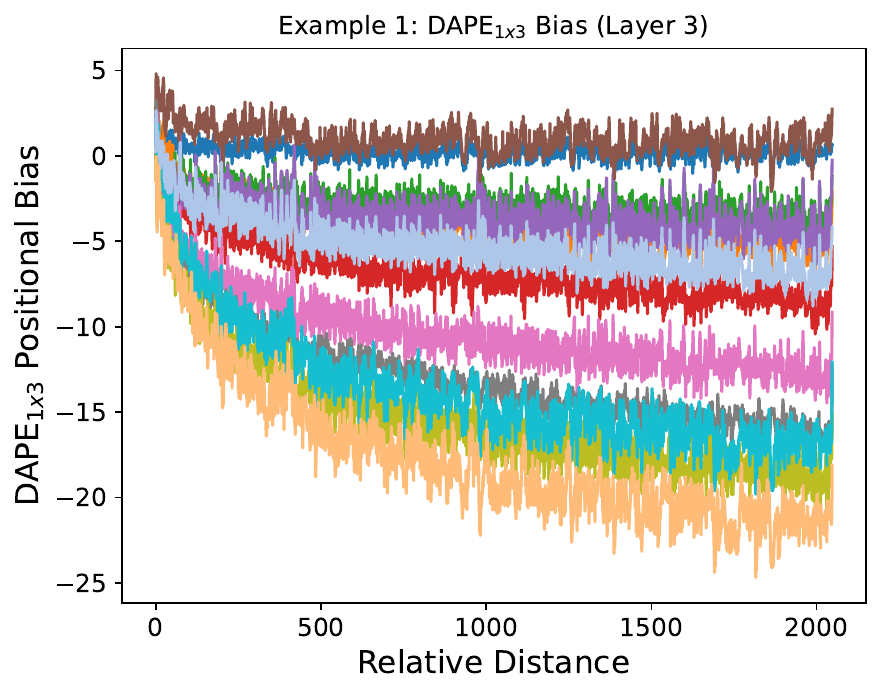}
\hspace{0in}

\includegraphics[width=0.32\textwidth]{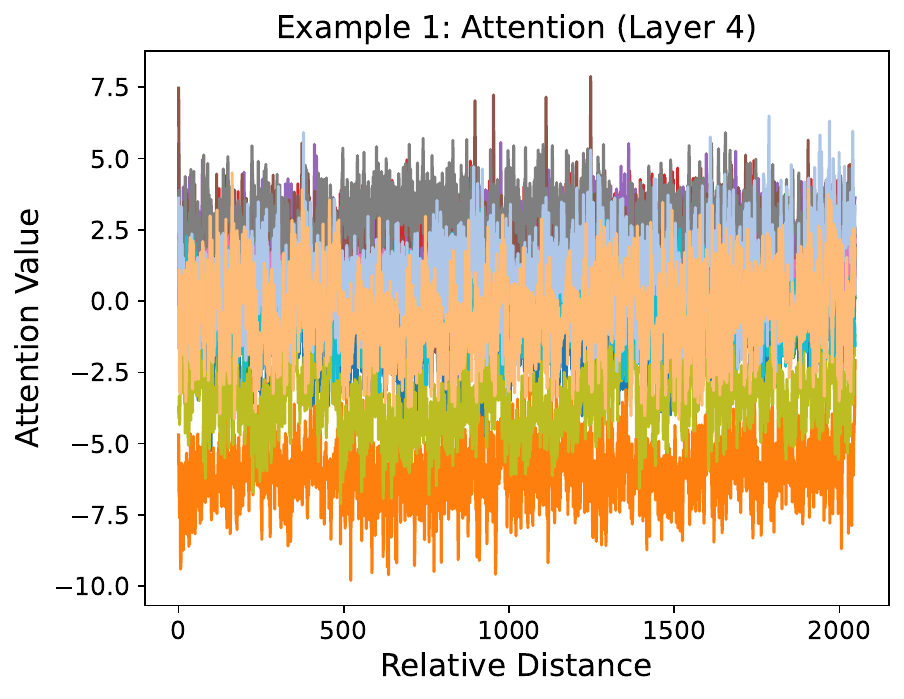}
\hspace{0in}
\includegraphics[width=0.32\textwidth]{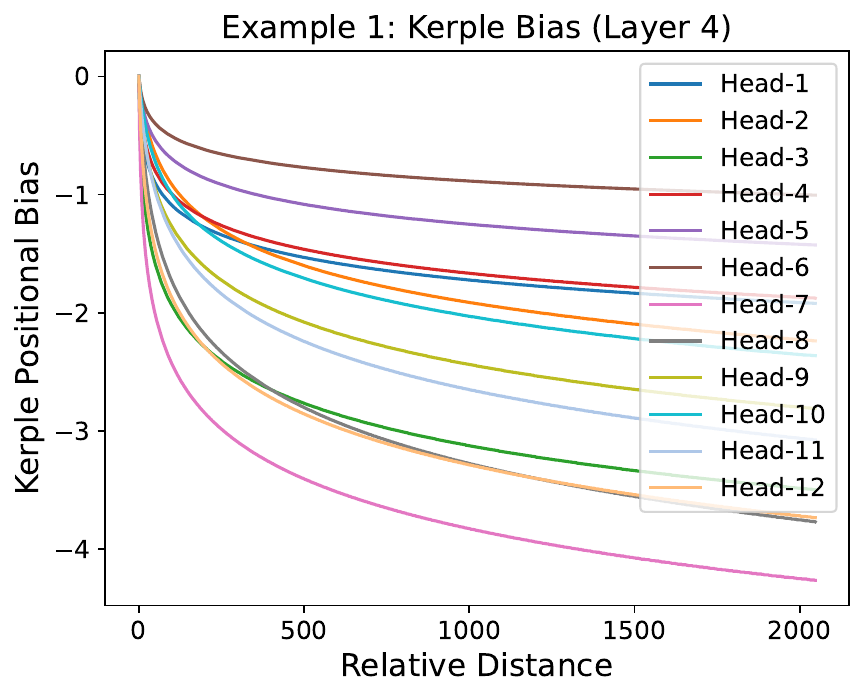}
\hspace{0in}
\includegraphics[width=0.32\textwidth]{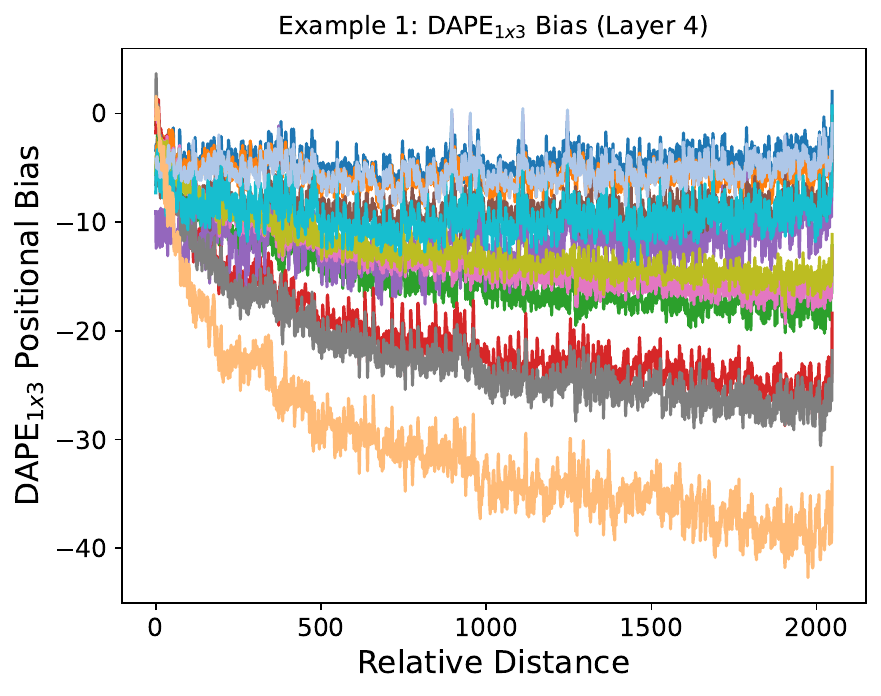}
\hspace{0in}

\includegraphics[width=0.32\textwidth]{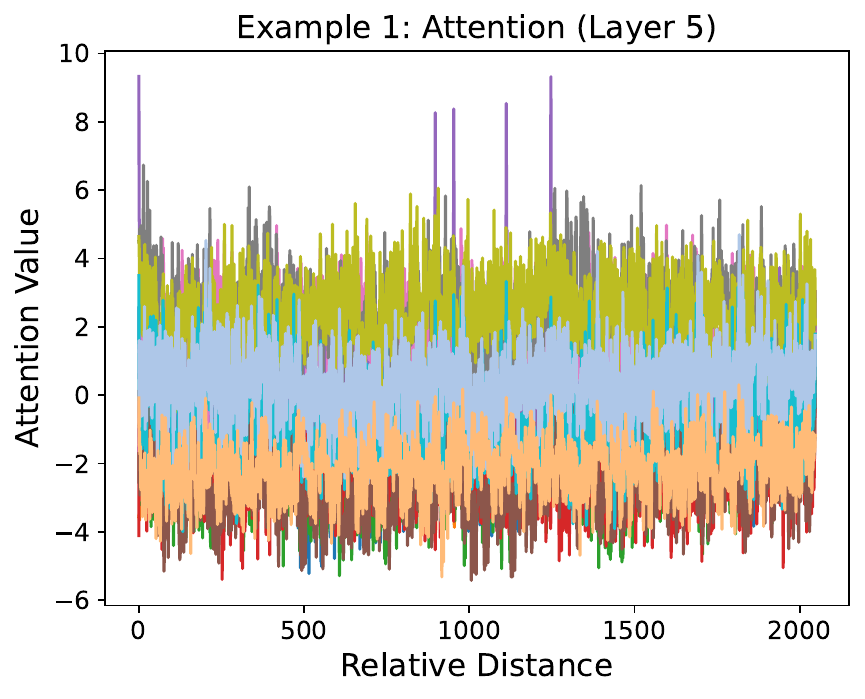}
\hspace{0in}
\includegraphics[width=0.32\textwidth]{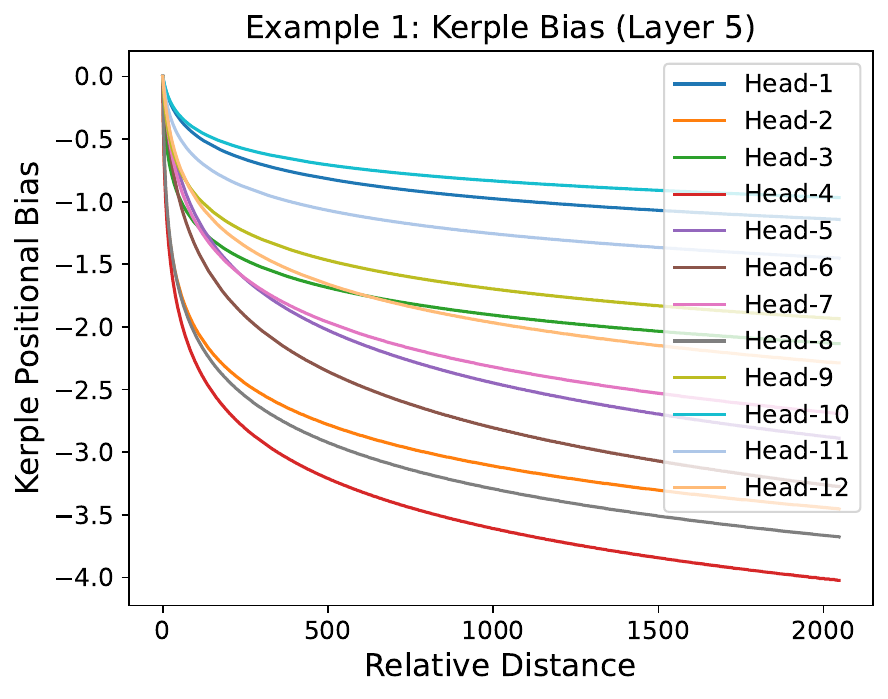}
\hspace{0in}
\includegraphics[width=0.32\textwidth]{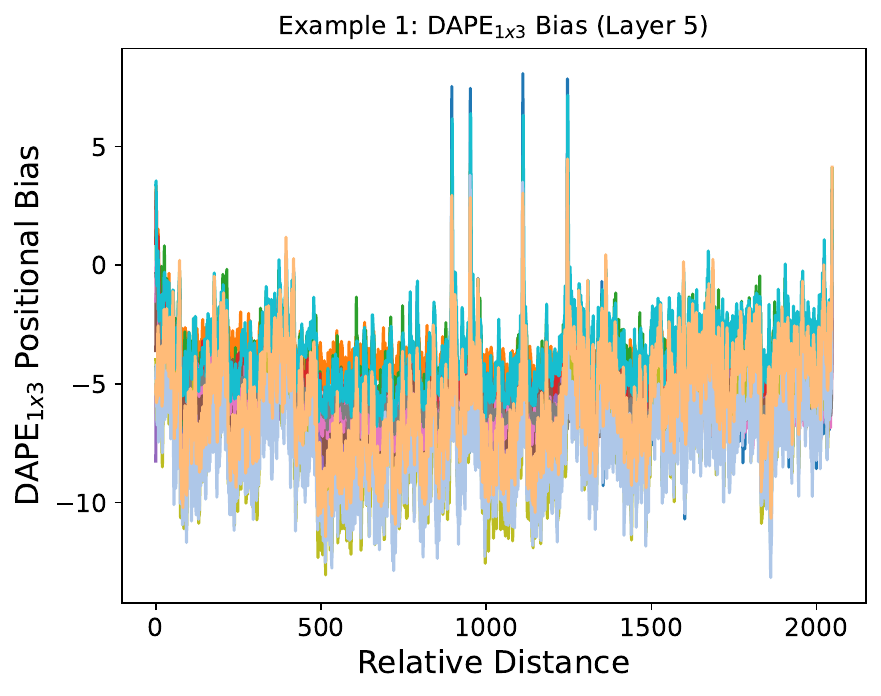}
\hspace{0in}

\hspace{0in}
\caption{
\small
\textbf{Evaluation Length 2048 Example 1: Part 1. From Left to Right: (1) The Attention is $\mX \mW_Q(\mX \mW_K)^{\top}$; (2) The Kerple bias is $\mB$; (3) The \methodShortName (with Kerple) bias is $f( \mX \mW_Q(\mX \mW_K)^{\top},\mB)$.
}
}
\end{figure}

\begin{figure}[htbp]
\setlength{\abovecaptionskip}{0.1cm}
\centering

\includegraphics[width=0.32\textwidth]{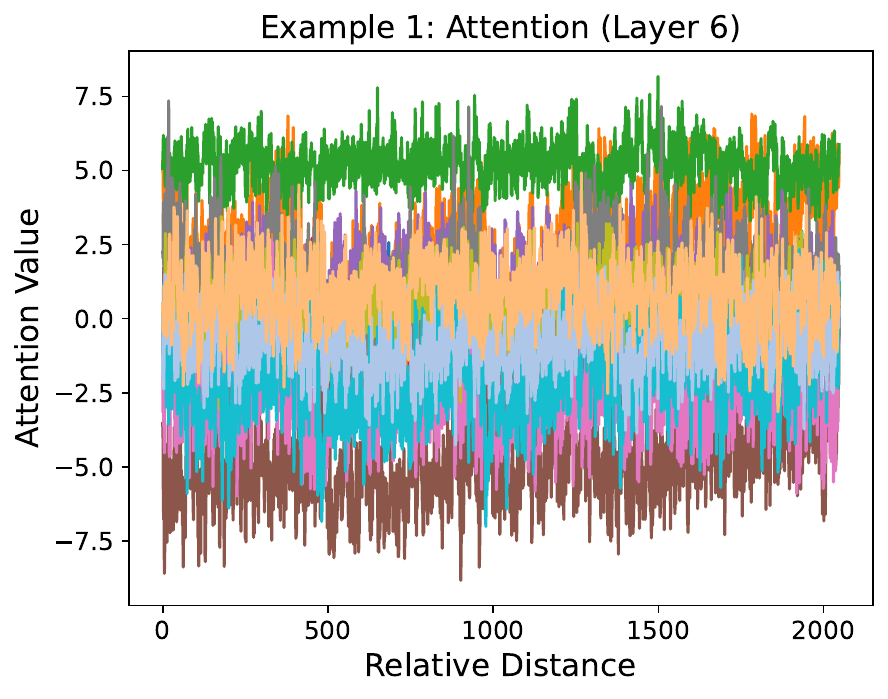}
\hspace{0in}
\includegraphics[width=0.32\textwidth]{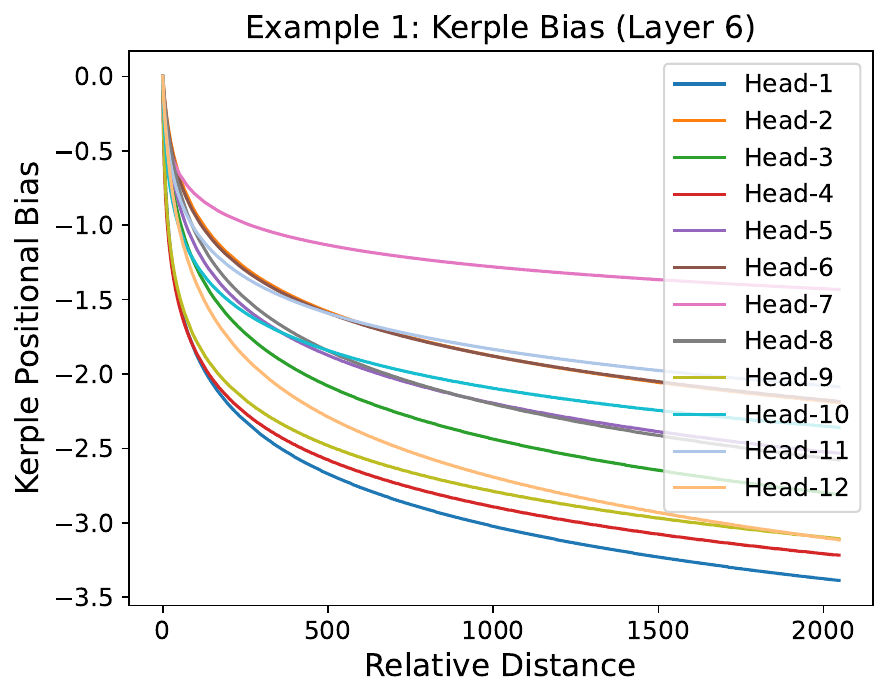}
\hspace{0in}
\includegraphics[width=0.32\textwidth]{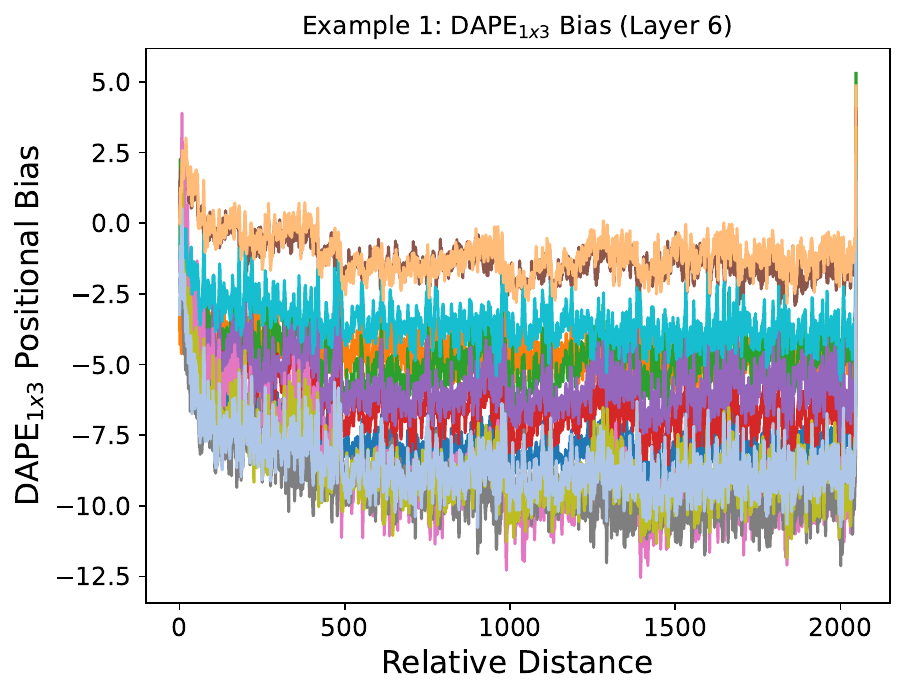}
\hspace{0in}

\includegraphics[width=0.32\textwidth]{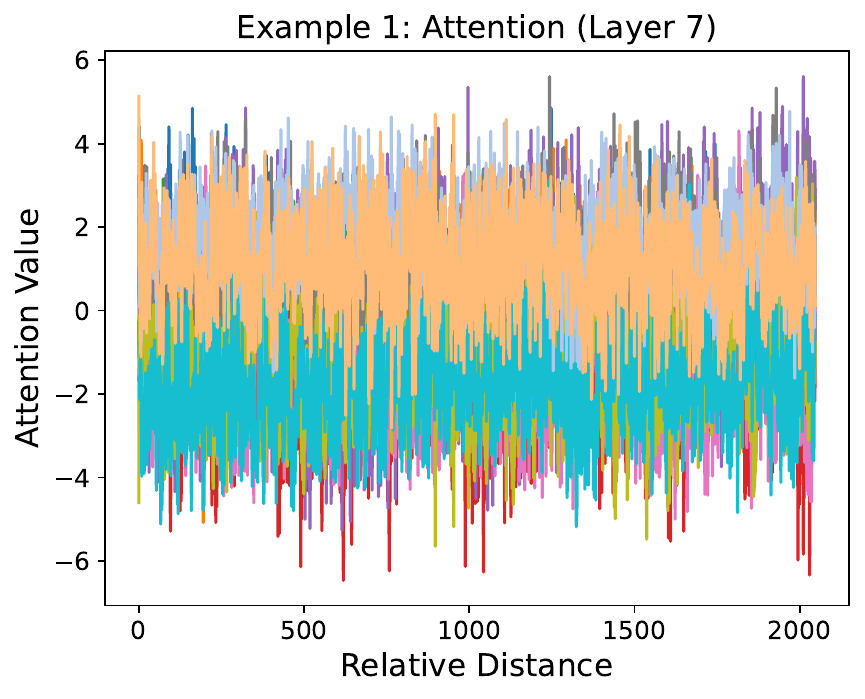}
\hspace{0in}
\includegraphics[width=0.32\textwidth]{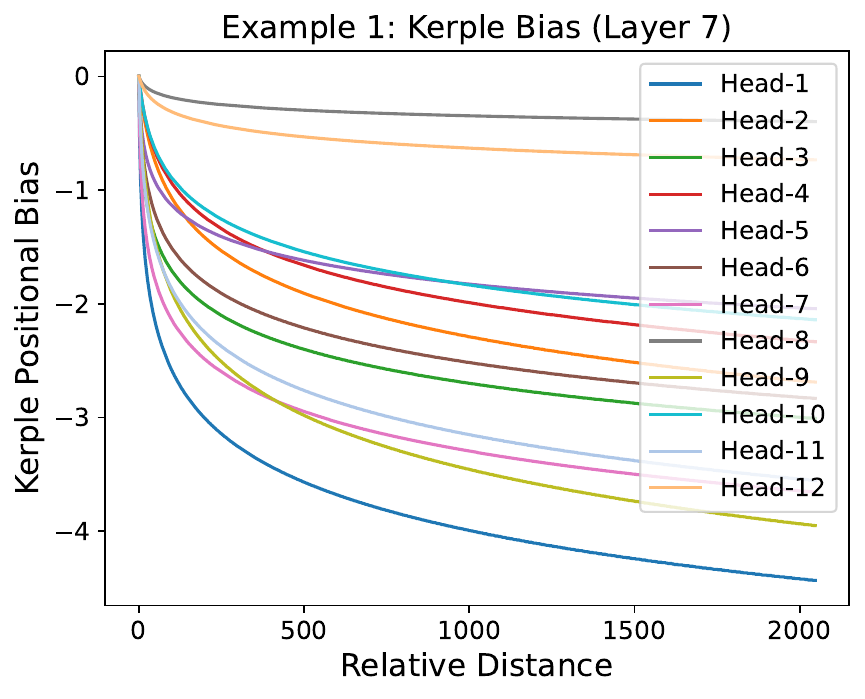}
\hspace{0in}
\includegraphics[width=0.32\textwidth]{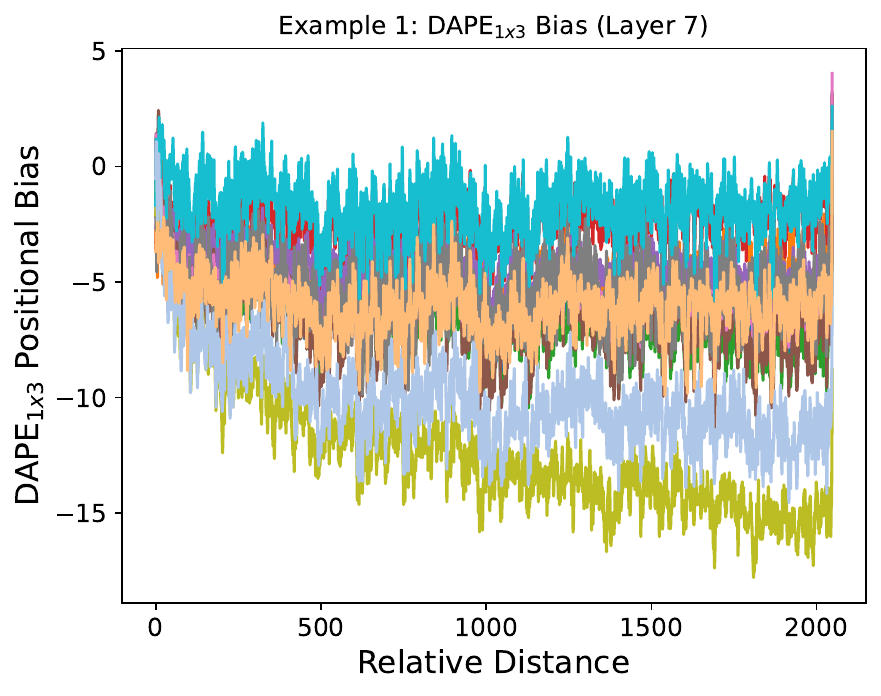}
\hspace{0in}

\includegraphics[width=0.32\textwidth]{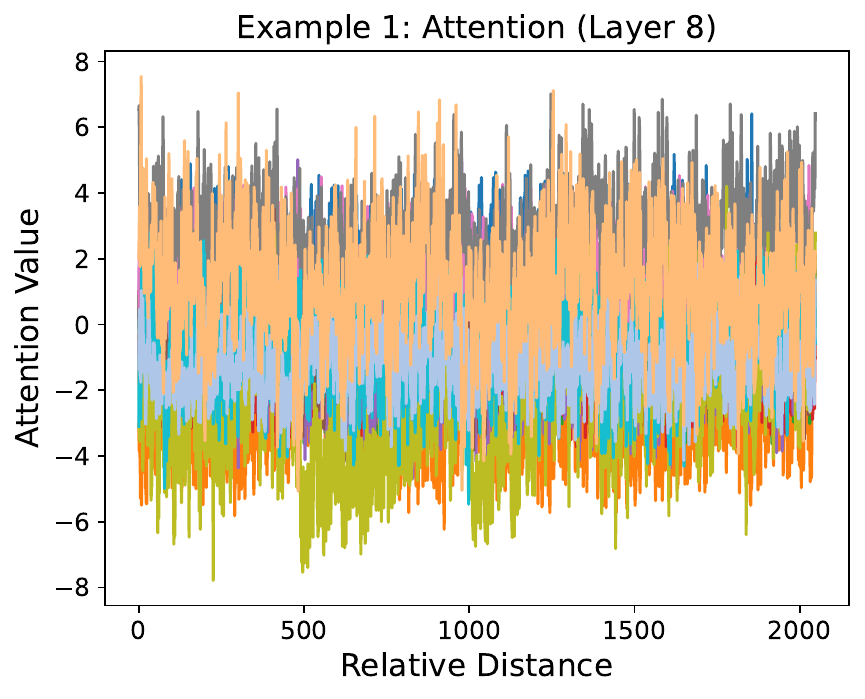}
\hspace{0in}
\includegraphics[width=0.32\textwidth]{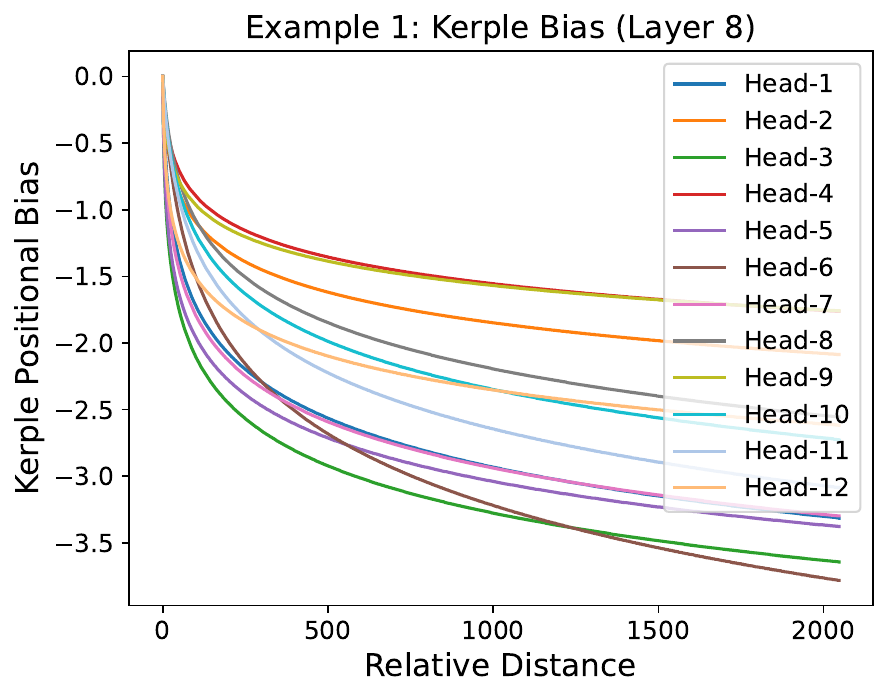}
\hspace{0in}
\includegraphics[width=0.32\textwidth]{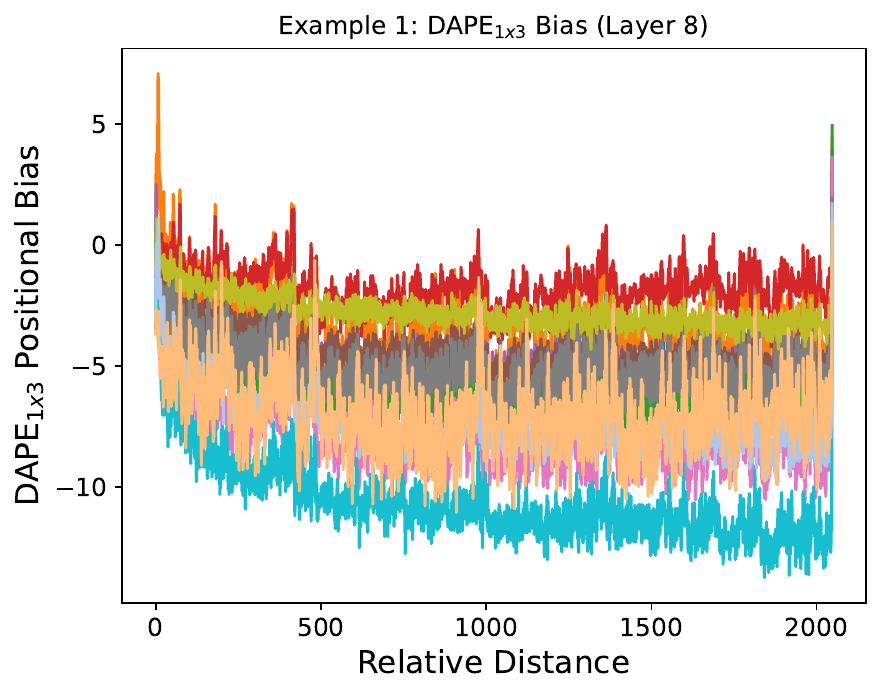}
\hspace{0in}

\includegraphics[width=0.32\textwidth]{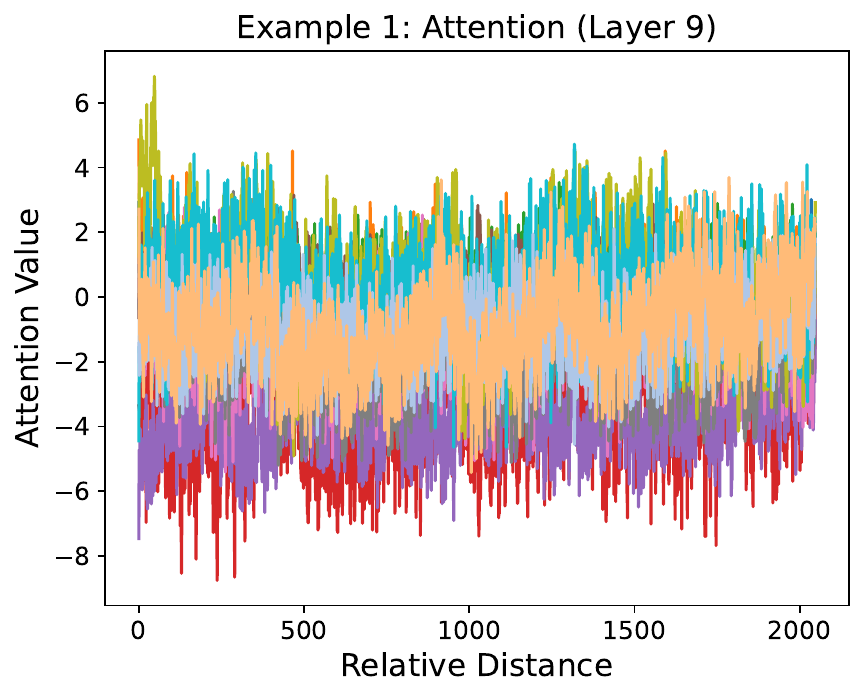}
\hspace{0in}
\includegraphics[width=0.32\textwidth]{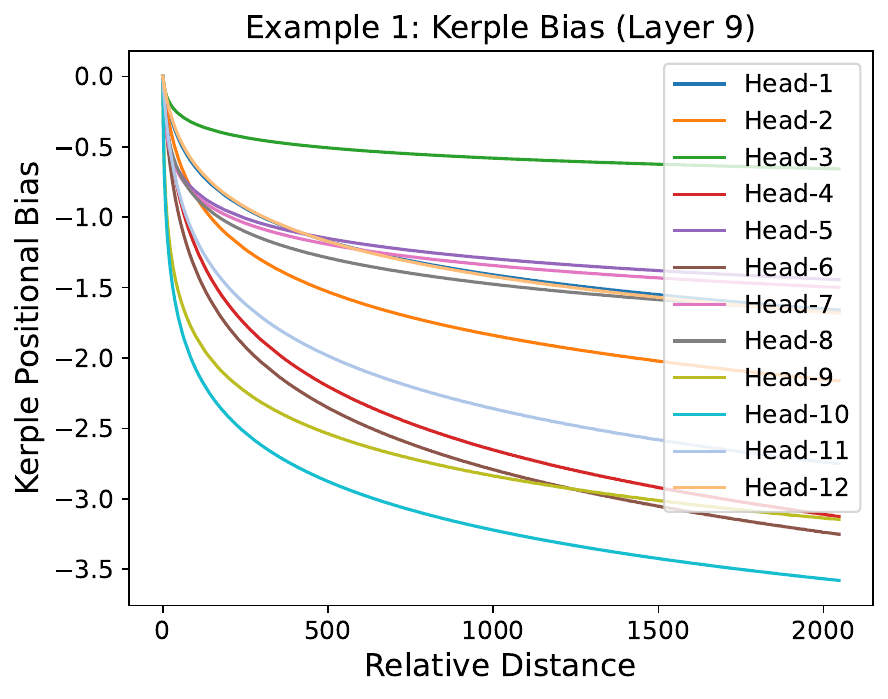}
\hspace{0in}
\includegraphics[width=0.32\textwidth]{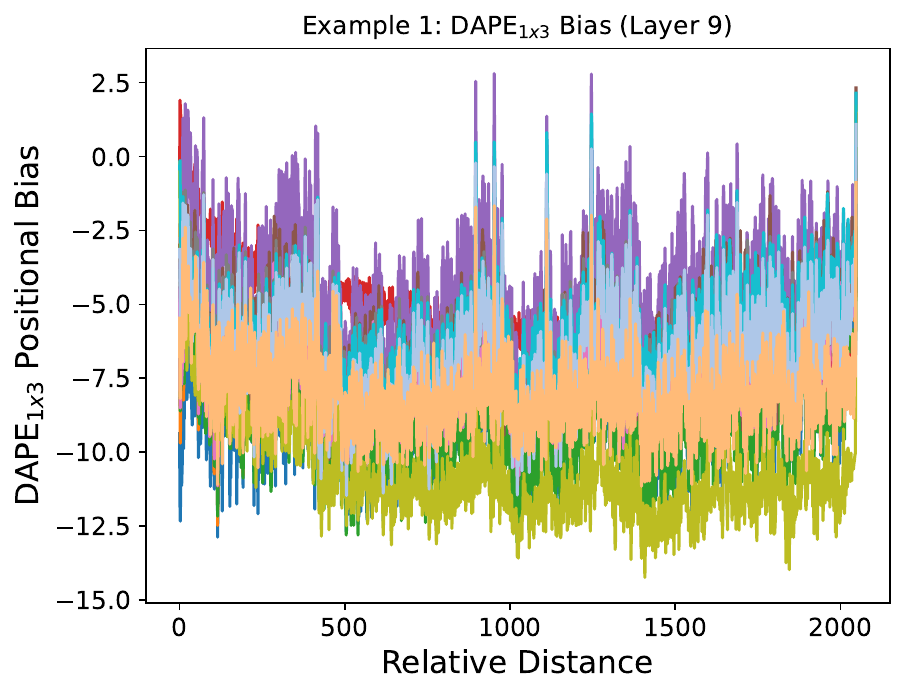}

\includegraphics[width=0.32\textwidth]{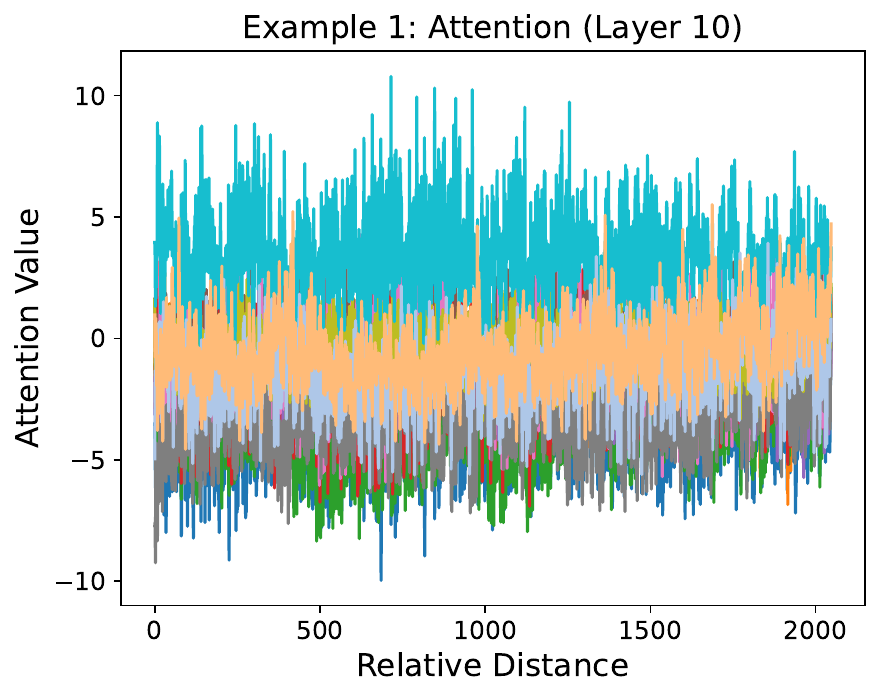}
\hspace{0in}
\includegraphics[width=0.32\textwidth]{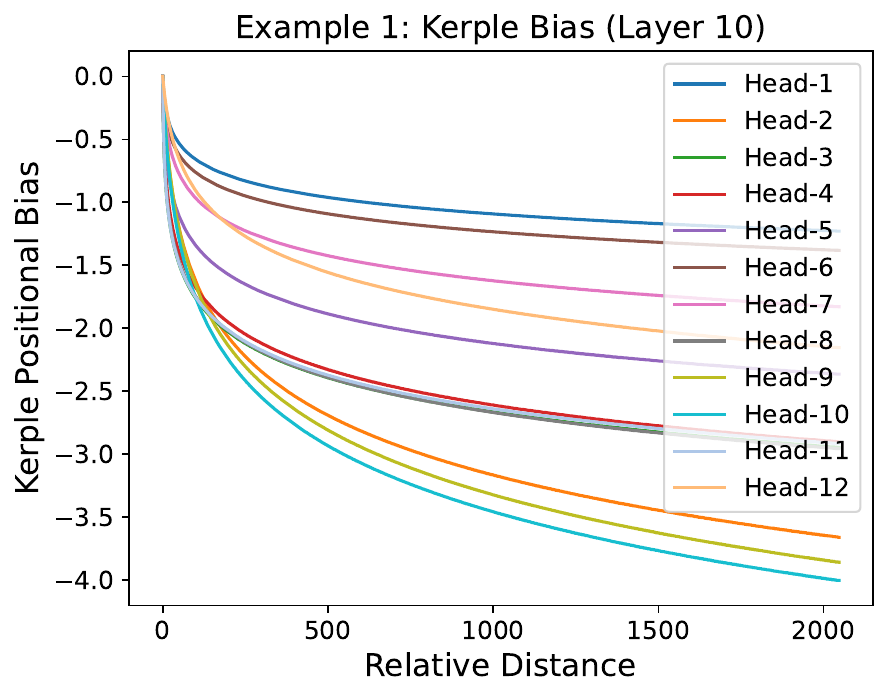}
\hspace{0in}
\includegraphics[width=0.32\textwidth]{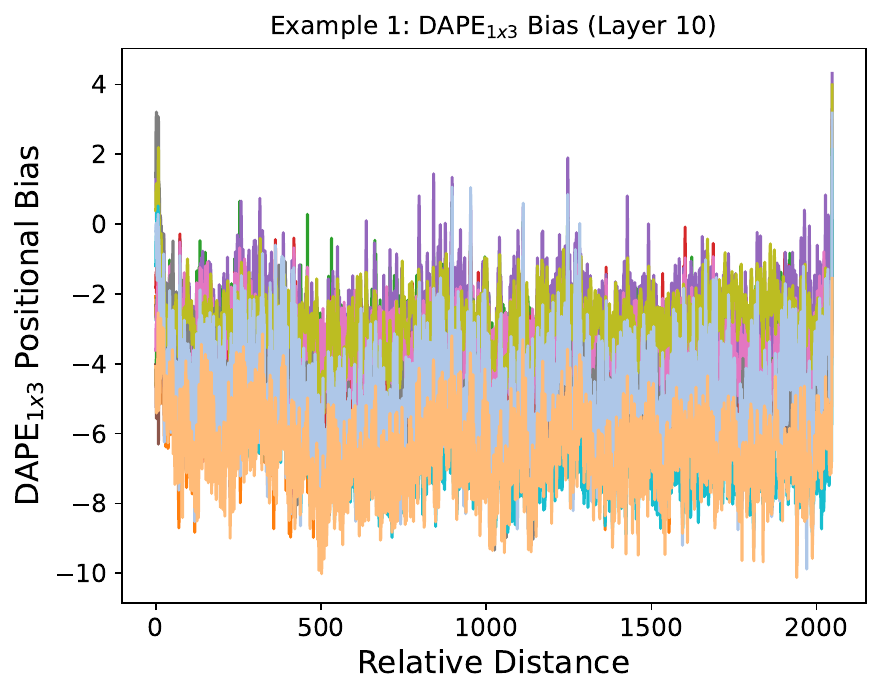}

\hspace{0in}
\caption{
\small
\textbf{Evaluation Length 2048 Example 1: Part 2. From Left to Right: (1) The Attention is $\mX \mW_Q(\mX \mW_K)^{\top}$; (2) The Kerple bias is $\mB$; (3) The \methodShortName (with Kerple) bias is $f( \mX \mW_Q(\mX \mW_K)^{\top},\mB)$.
}
}
\end{figure}

\newpage
\begin{figure}[htbp]
\setlength{\abovecaptionskip}{0.1cm}
\centering

\includegraphics[width=0.32\textwidth]{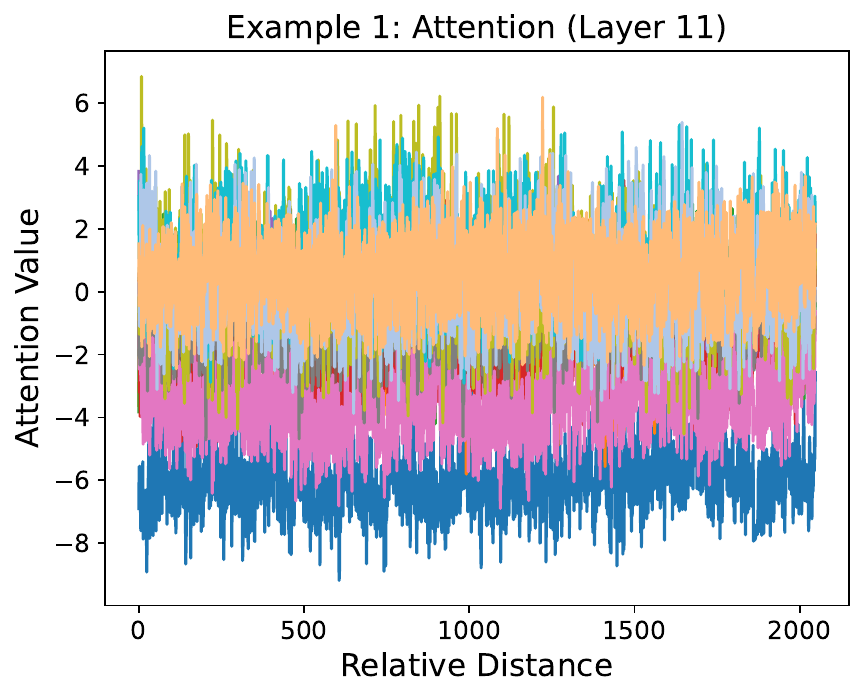}
\hspace{0in}
\includegraphics[width=0.32\textwidth]{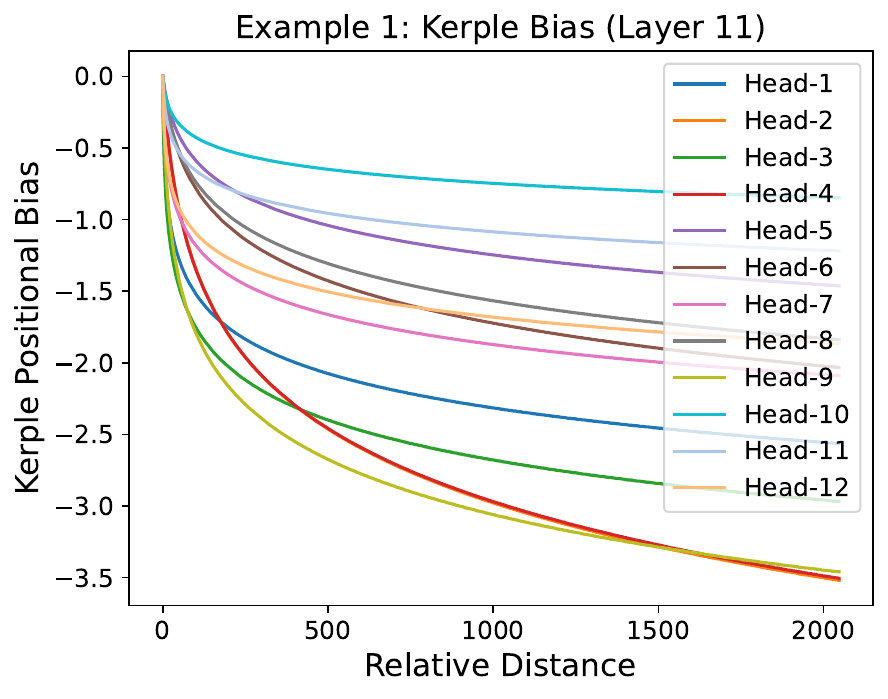}
\hspace{0in}
\includegraphics[width=0.32\textwidth]{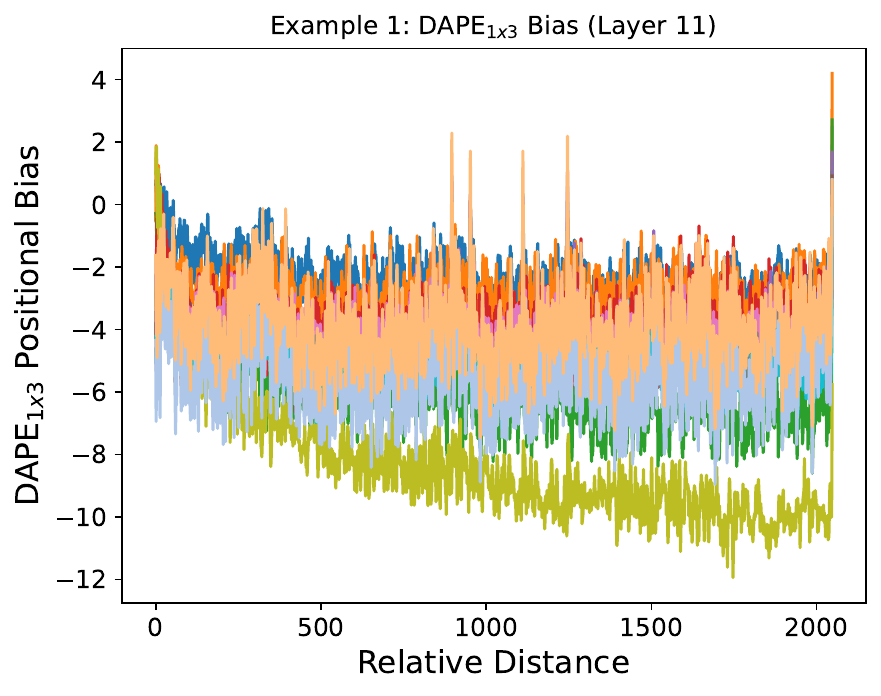}
\hspace{0in}

\includegraphics[width=0.32\textwidth]{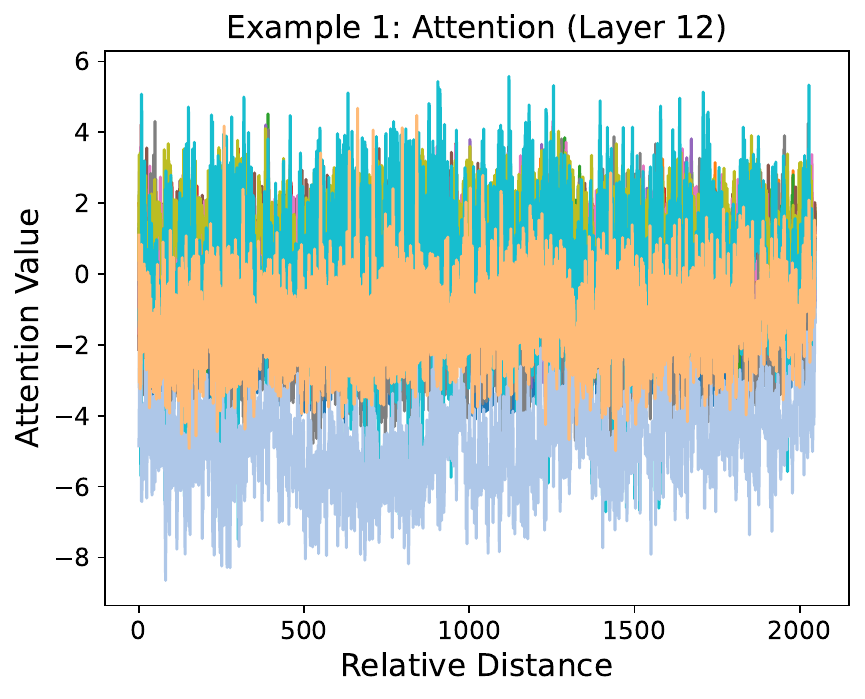}
\hspace{0in}
\includegraphics[width=0.32\textwidth]{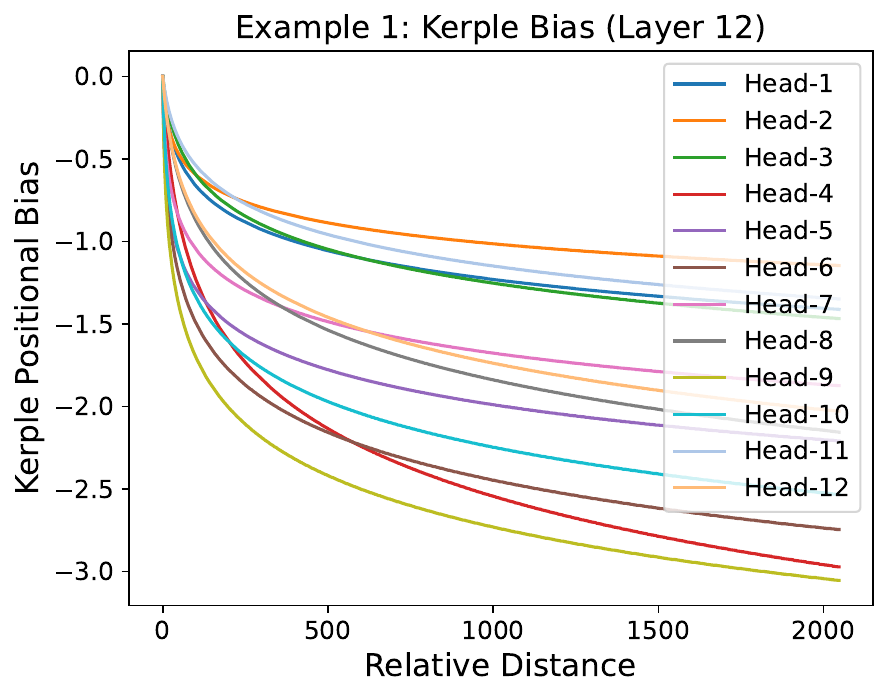}
\hspace{0in}
\includegraphics[width=0.32\textwidth]{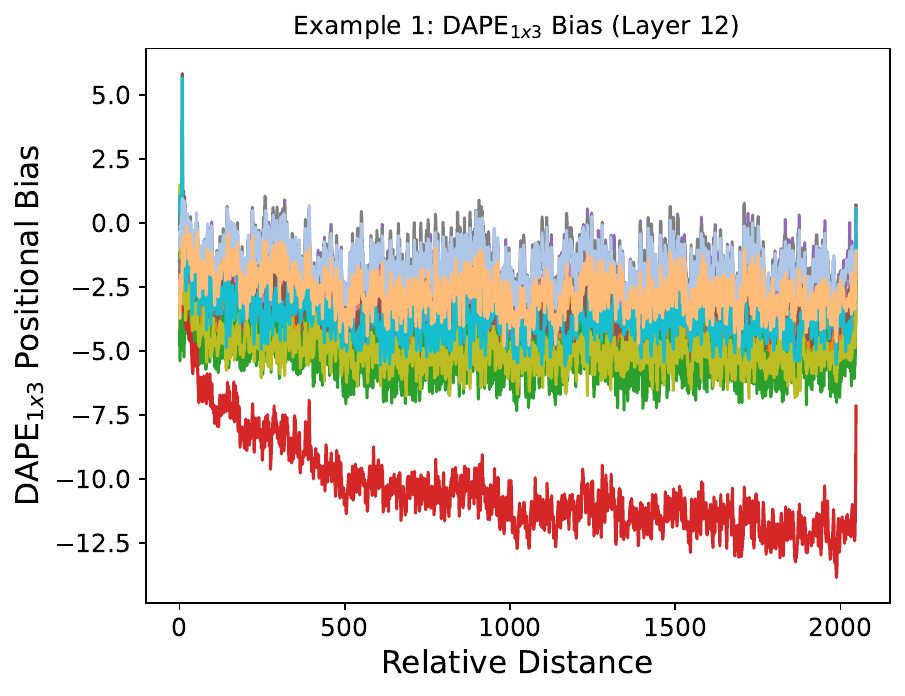}
\caption{
\small
\textbf{Evaluation Length 2048 Example 1: Part 3. From Left to Right: (1) The Attention is $\mX \mW_Q(\mX \mW_K)^{\top}$; (2) The Kerple bias is $\mB$; (3) The \methodShortName (with Kerple) bias is $f( \mX \mW_Q(\mX \mW_K)^{\top},\mB)$.
}
}
\end{figure}

\newpage

\begin{figure}[htbp]
\setlength{\abovecaptionskip}{0.1cm}
\centering
\includegraphics[width=0.32\textwidth]{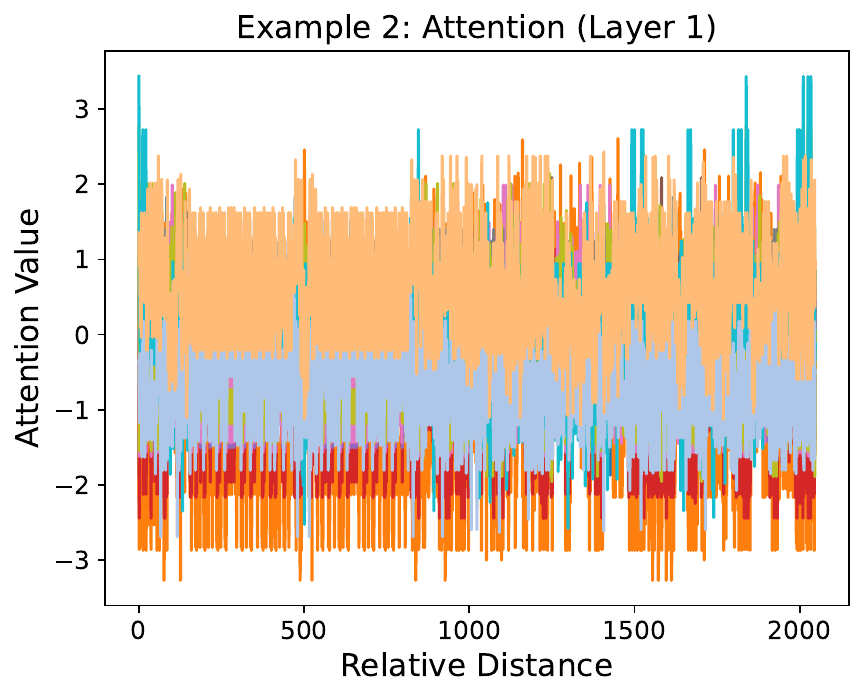}
\hspace{0in}
\includegraphics[width=0.32\textwidth]{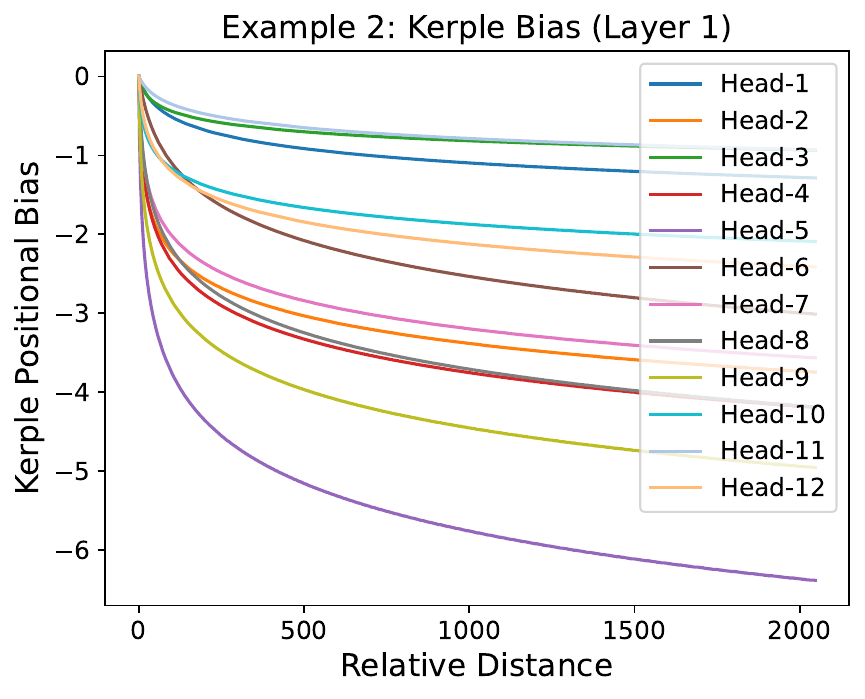}
\hspace{0in}
\includegraphics[width=0.32\textwidth]{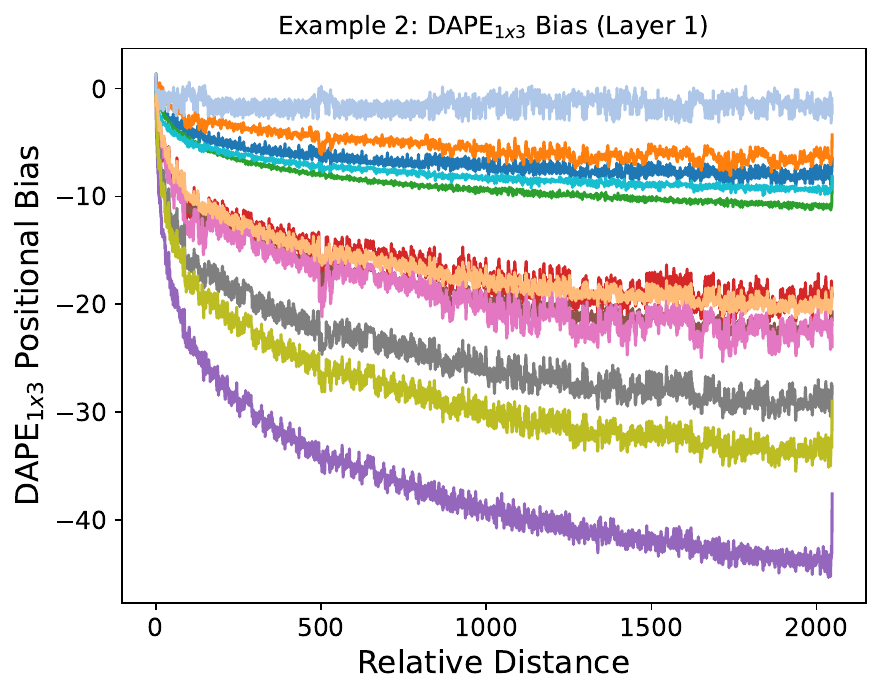}
\hspace{0in}

\includegraphics[width=0.32\textwidth]{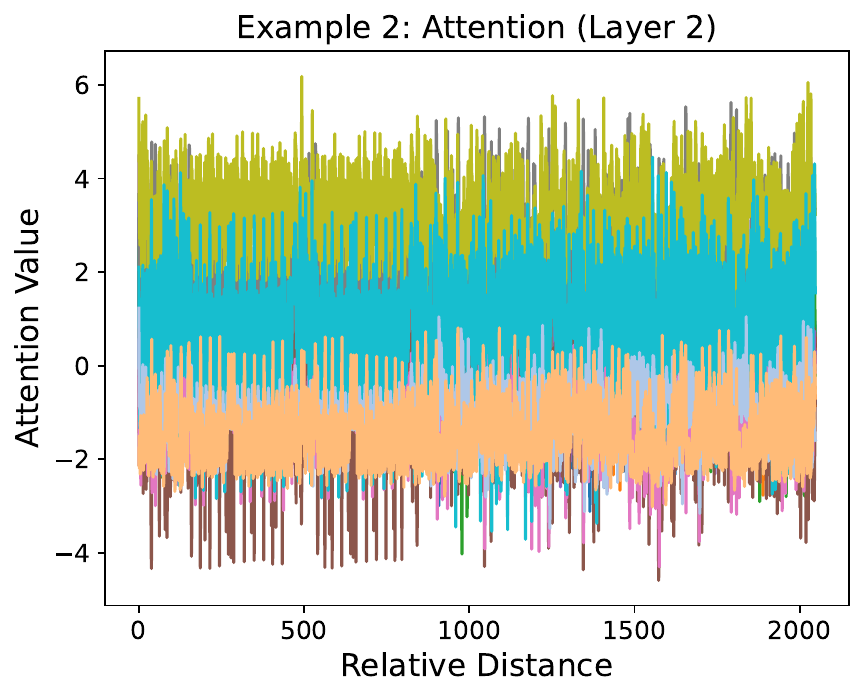}
\hspace{0in}
\includegraphics[width=0.32\textwidth]{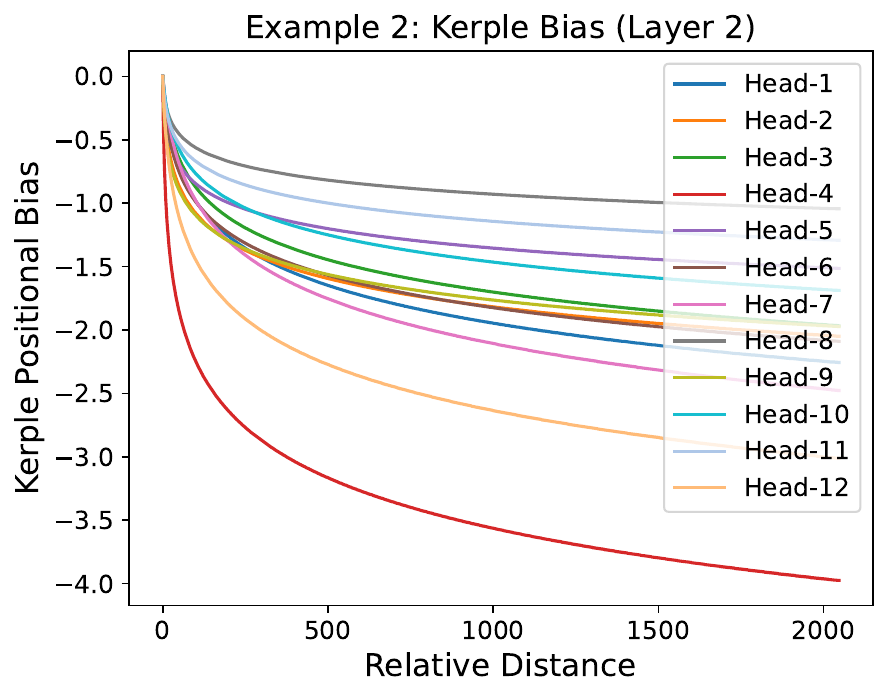}
\hspace{0in}
\includegraphics[width=0.32\textwidth]{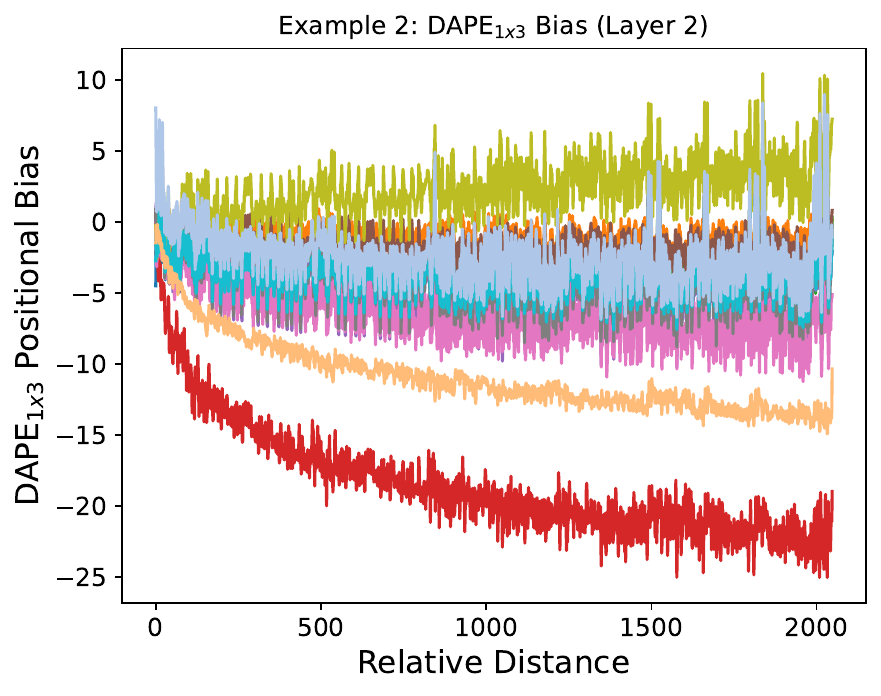}
\hspace{0in}

\includegraphics[width=0.32\textwidth]{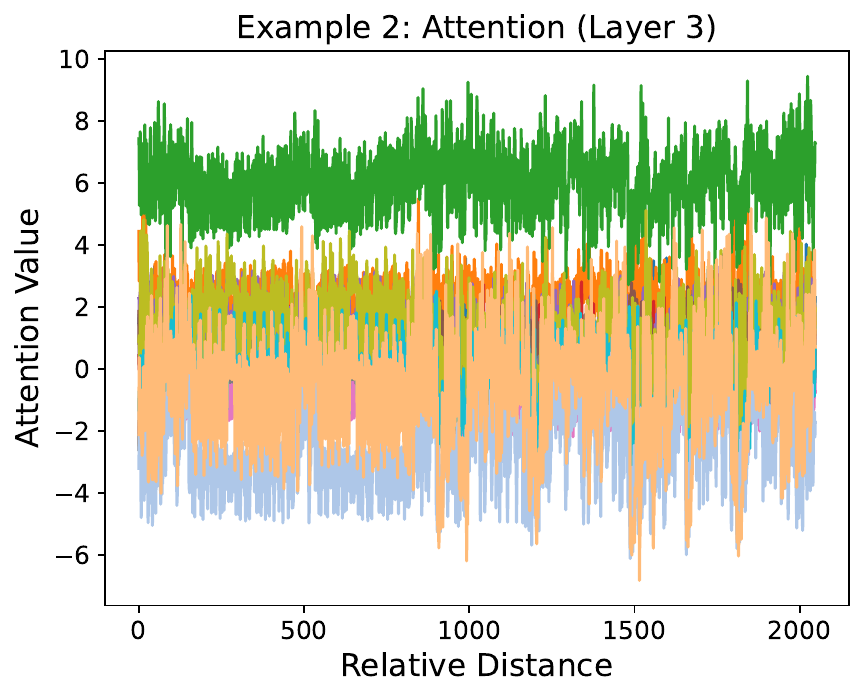}
\hspace{0in}
\includegraphics[width=0.32\textwidth]{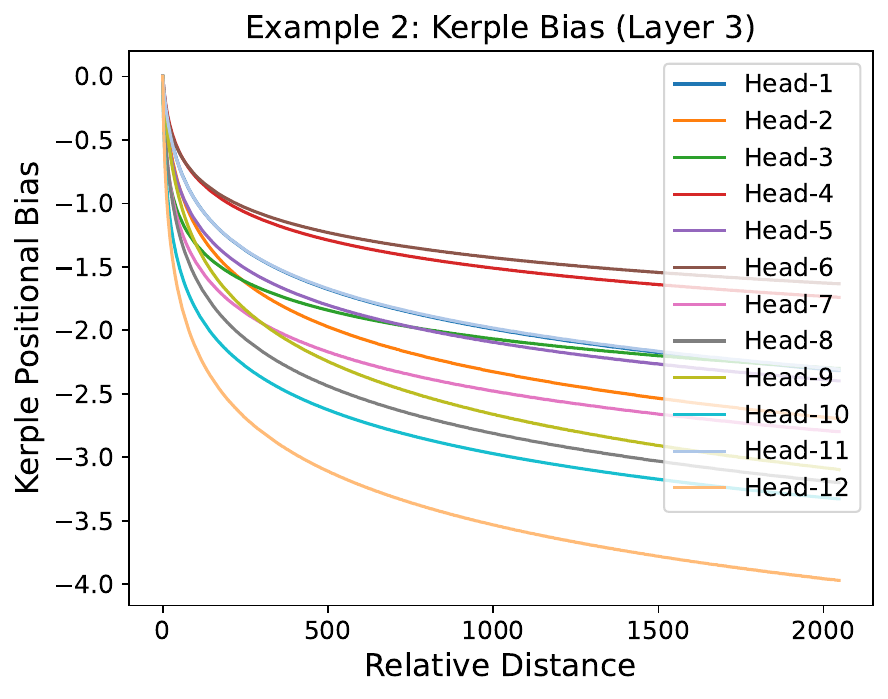}
\hspace{0in}
\includegraphics[width=0.32\textwidth]{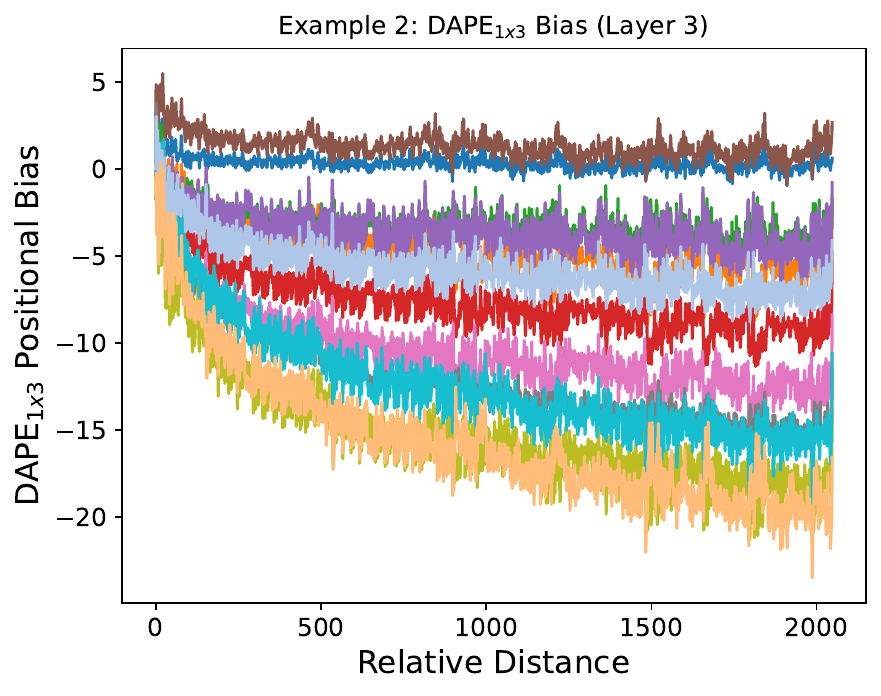}
\hspace{0in}

\includegraphics[width=0.32\textwidth]{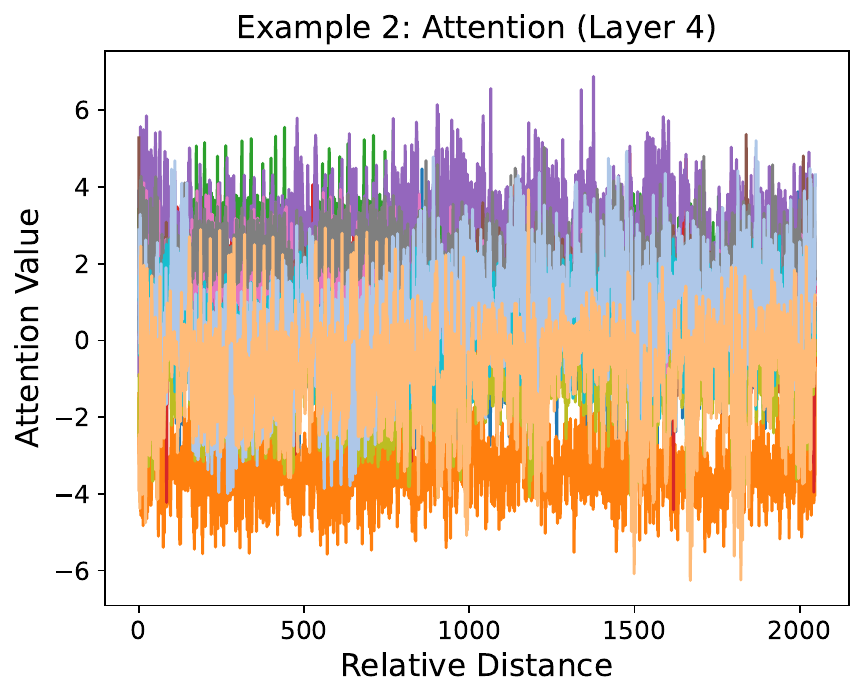}
\hspace{0in}
\includegraphics[width=0.32\textwidth]{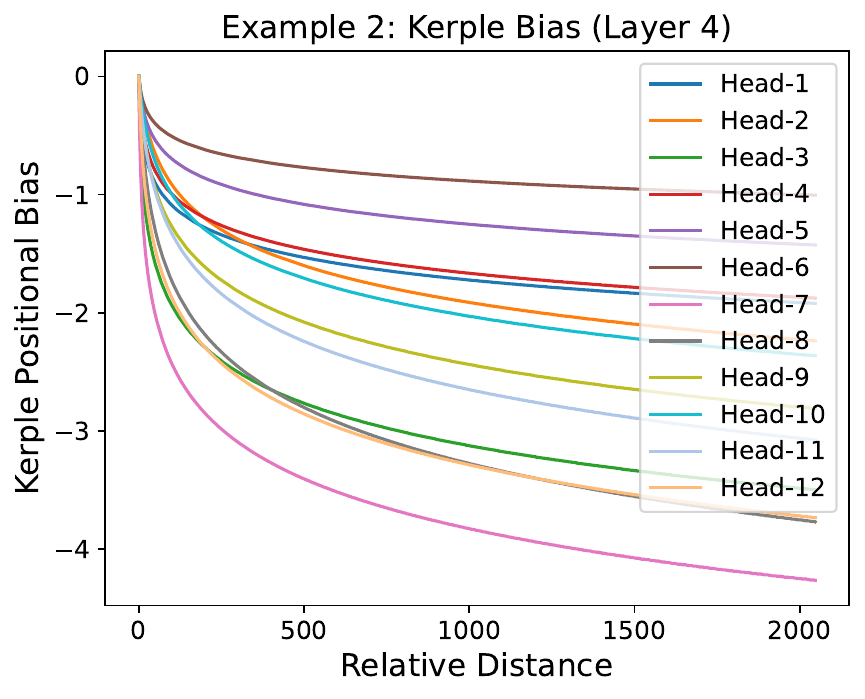}
\hspace{0in}
\includegraphics[width=0.32\textwidth]{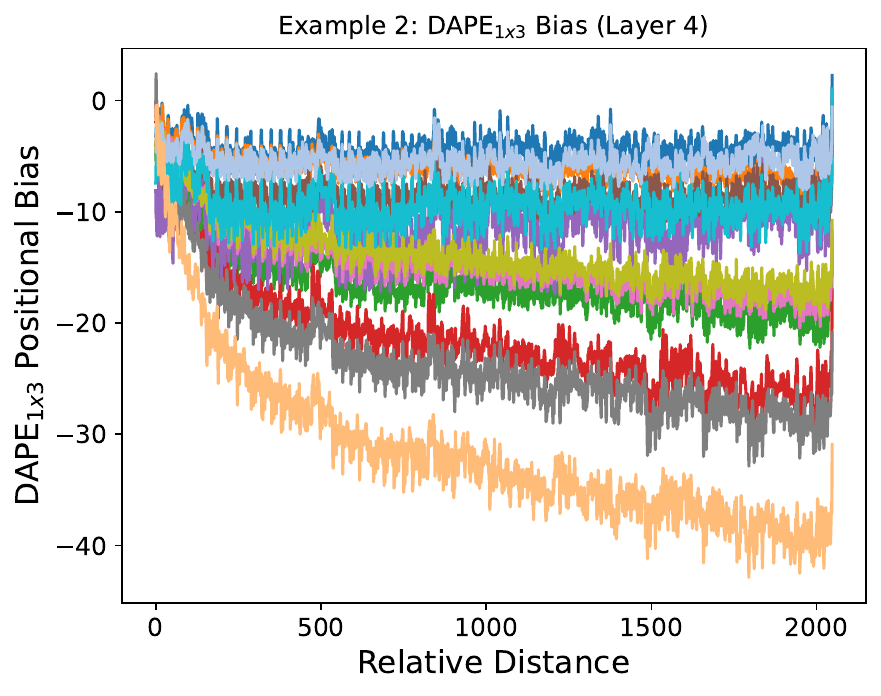}
\hspace{0in}

\includegraphics[width=0.32\textwidth]{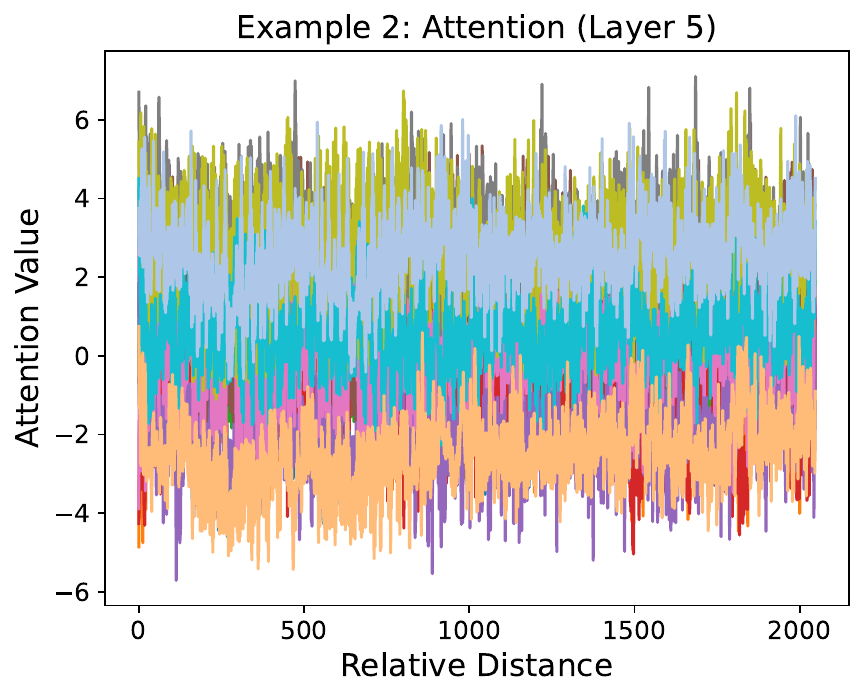}
\hspace{0in}
\includegraphics[width=0.32\textwidth]{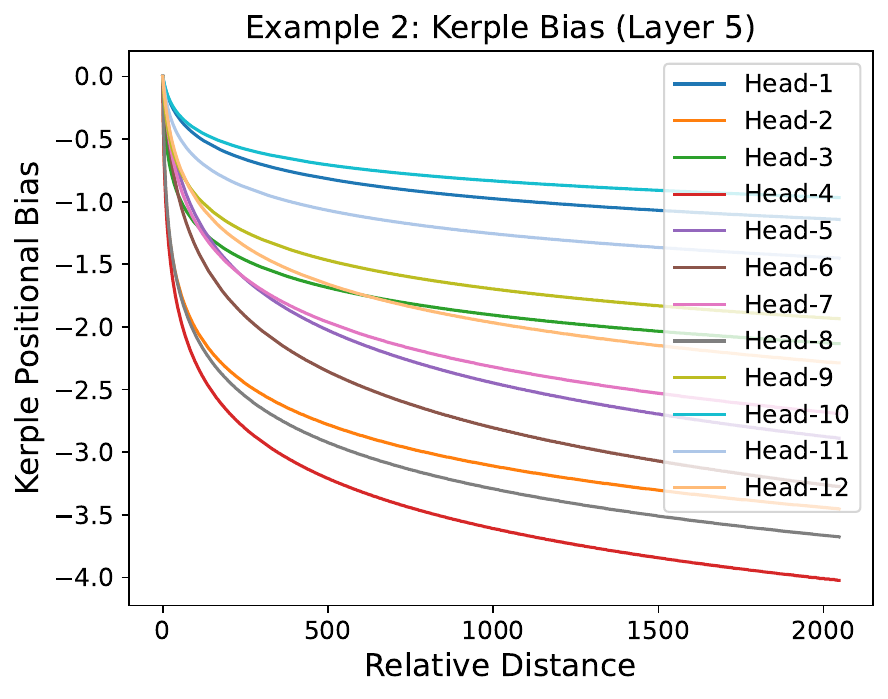}
\hspace{0in}
\includegraphics[width=0.32\textwidth]{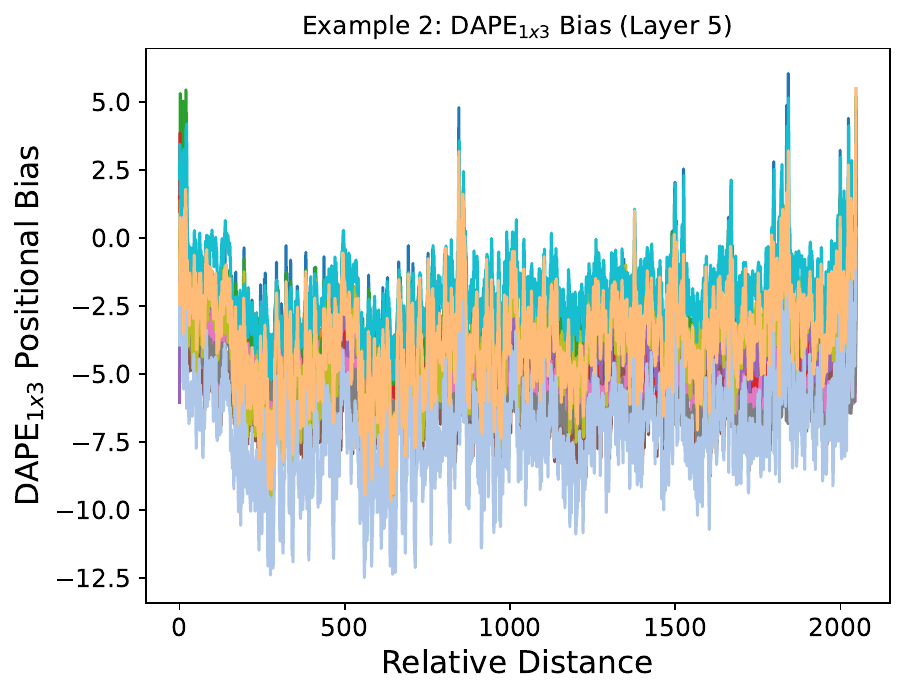}
\hspace{0in}

\includegraphics[width=0.32\textwidth]{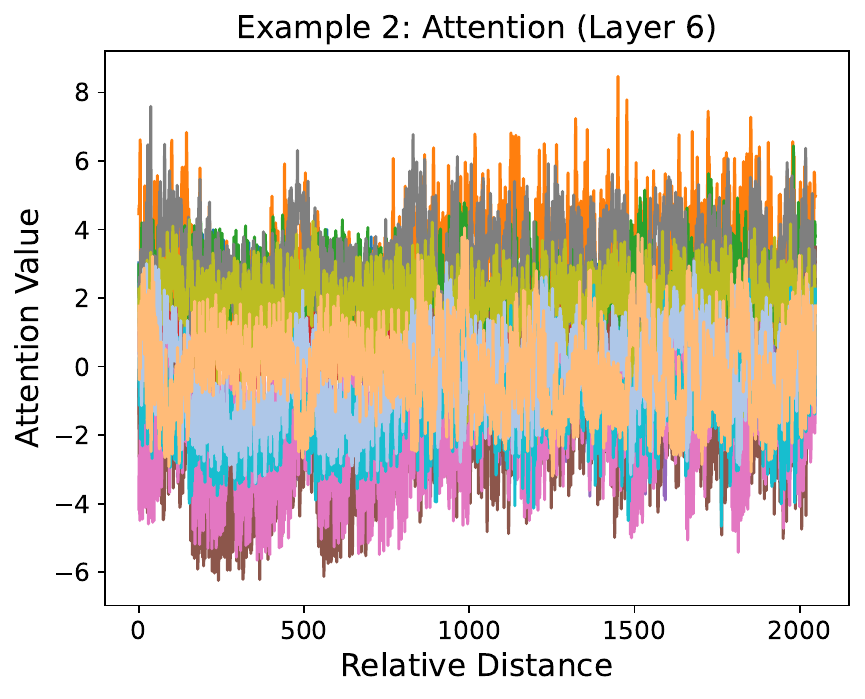}
\hspace{0in}
\includegraphics[width=0.32\textwidth]{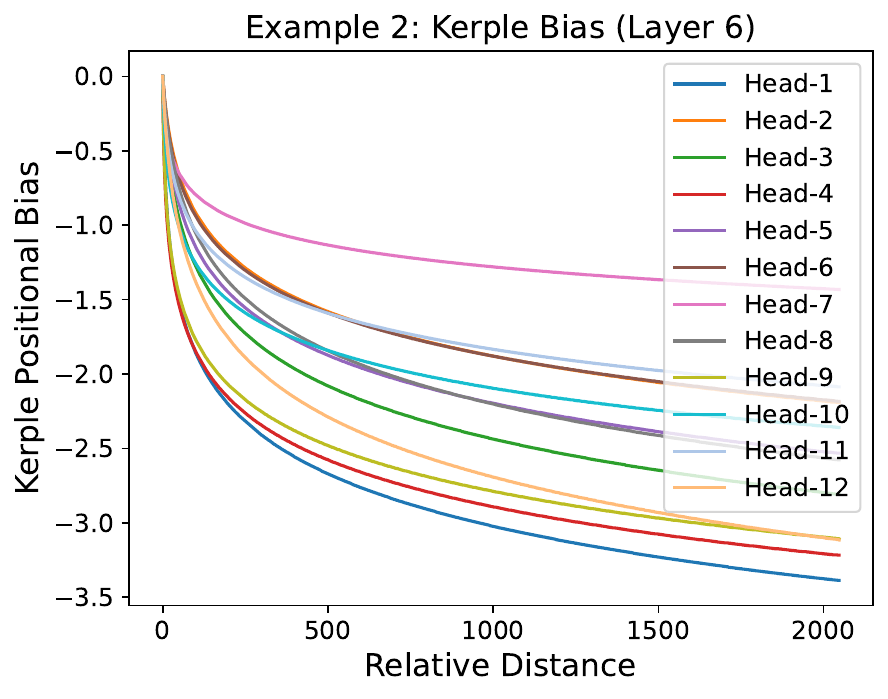}
\hspace{0in}
\includegraphics[width=0.32\textwidth]{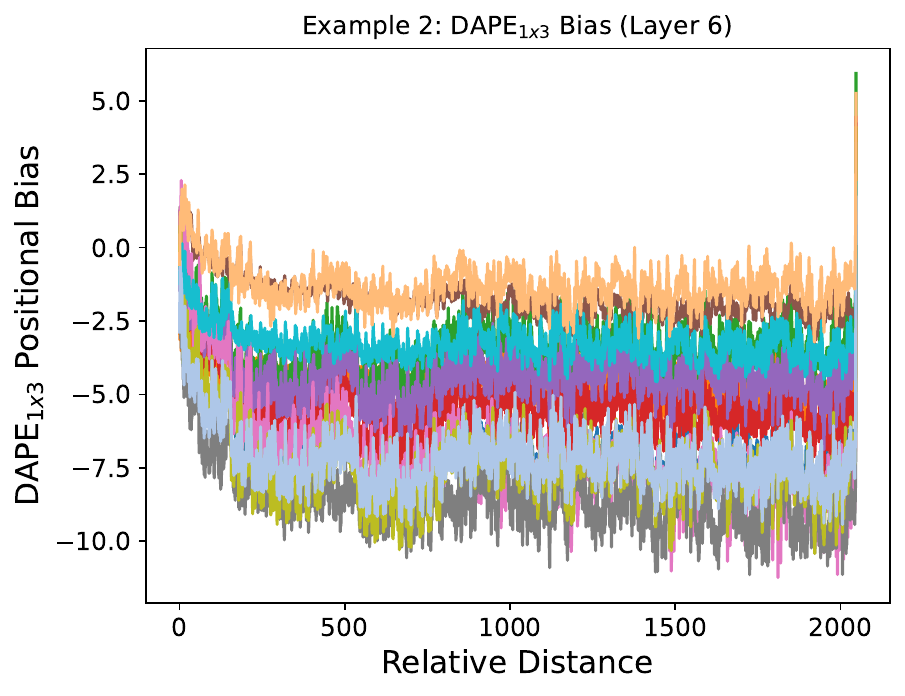}
\hspace{0in}

\hspace{0in}
\caption{
\small
\textbf{Evaluation Length 2048 Example 2: Part 1. From Left to Right: (1) The Attention is $\mX \mW_Q(\mX \mW_K)^{\top}$; (2) The Kerple bias is $\mB$; (3) The \methodShortName (with Kerple) bias is $f( \mX \mW_Q(\mX \mW_K)^{\top},\mB)$.
}
}
\end{figure}

\begin{figure}[htbp]
\setlength{\abovecaptionskip}{0.1cm}
\centering

\includegraphics[width=0.32\textwidth]{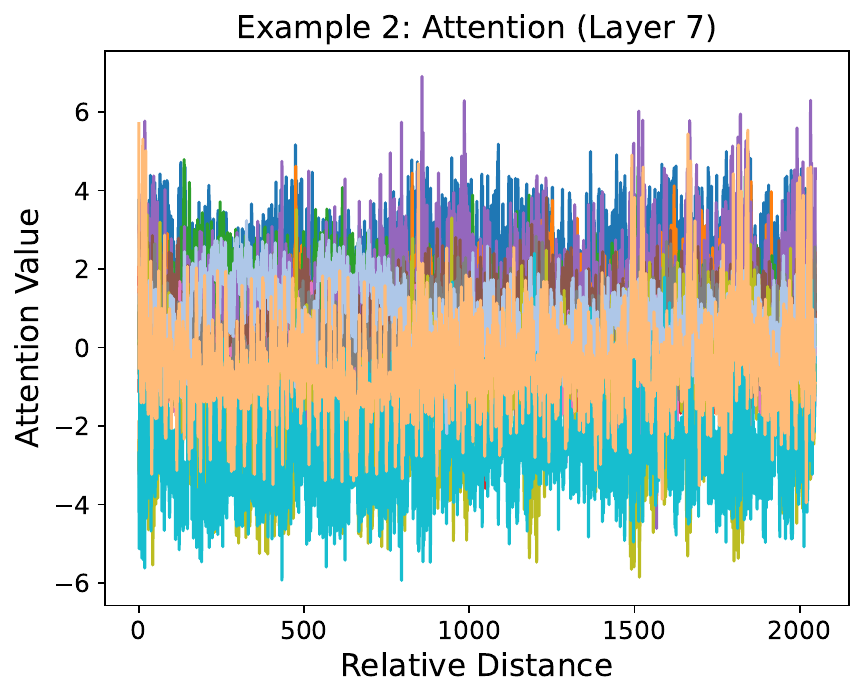}
\hspace{0in}
\includegraphics[width=0.32\textwidth]{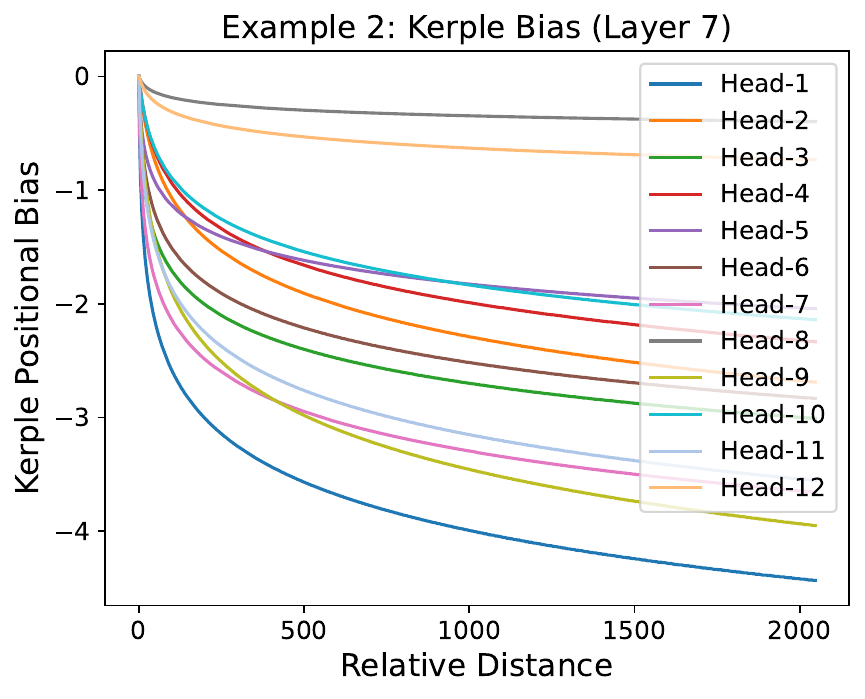}
\hspace{0in}
\includegraphics[width=0.32\textwidth]{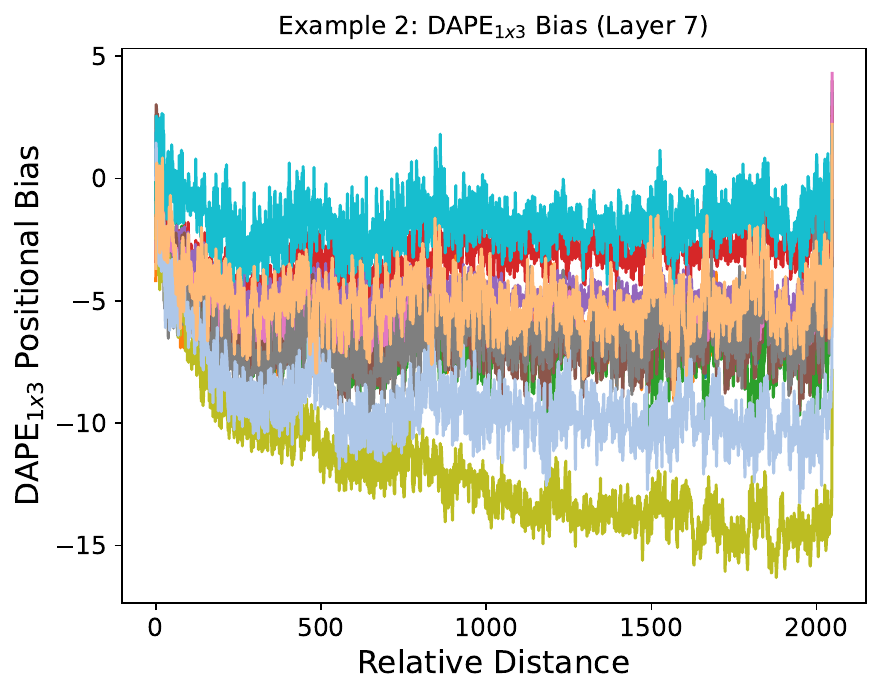}
\hspace{0in}

\includegraphics[width=0.32\textwidth]{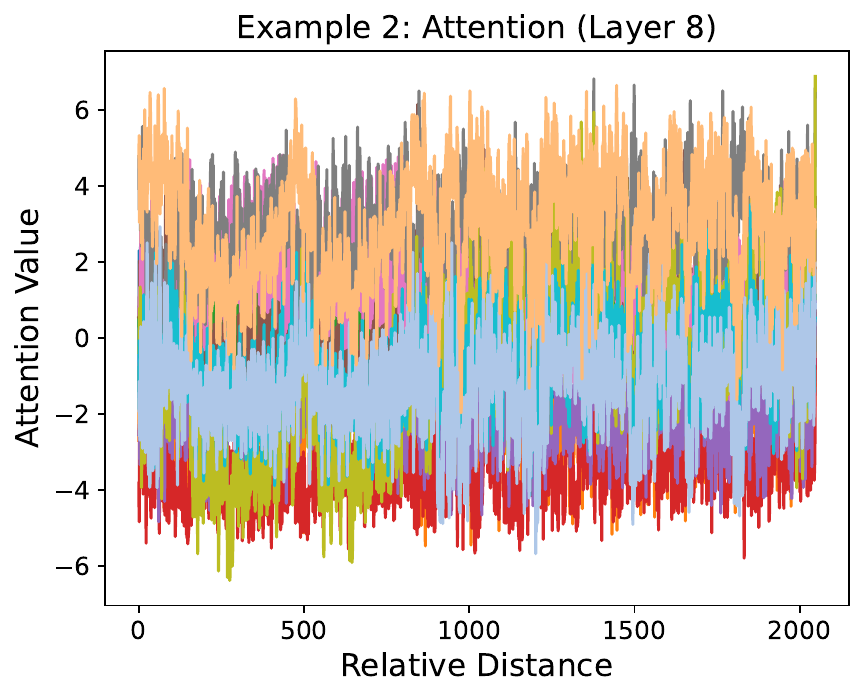}
\hspace{0in}
\includegraphics[width=0.32\textwidth]{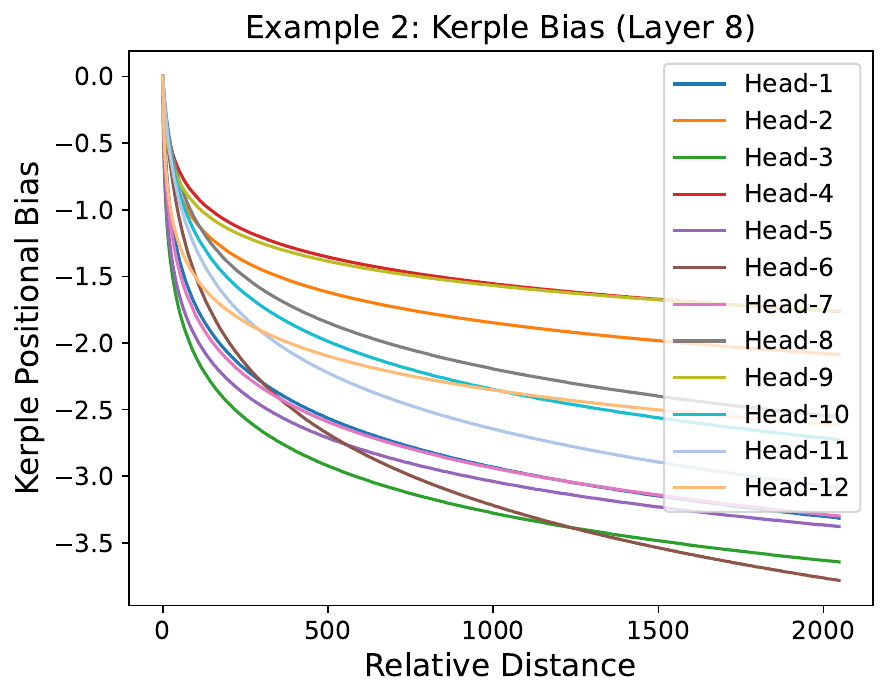}
\hspace{0in}
\includegraphics[width=0.32\textwidth]{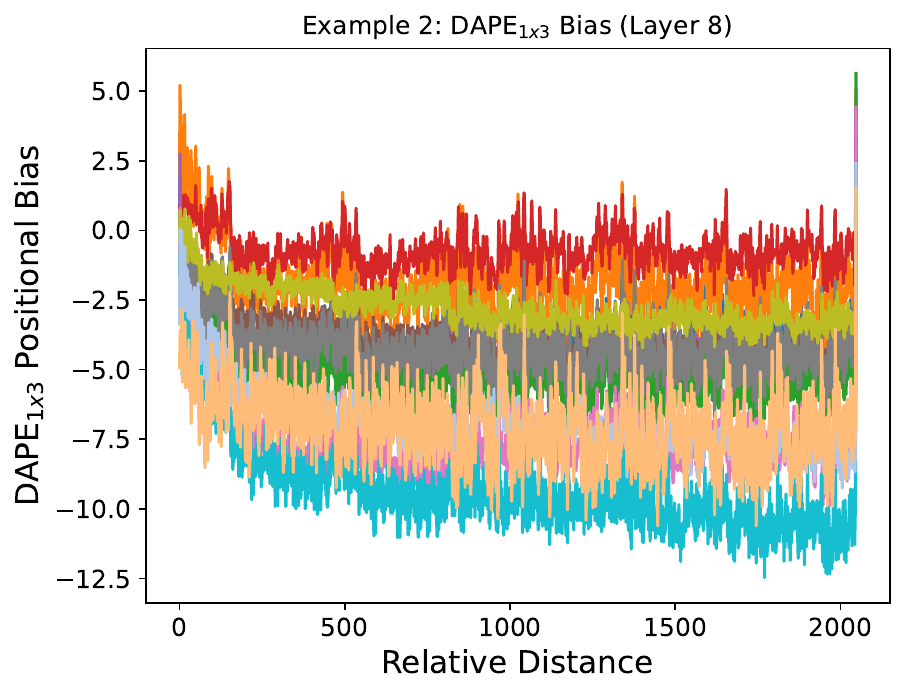}
\hspace{0in}

\includegraphics[width=0.32\textwidth]{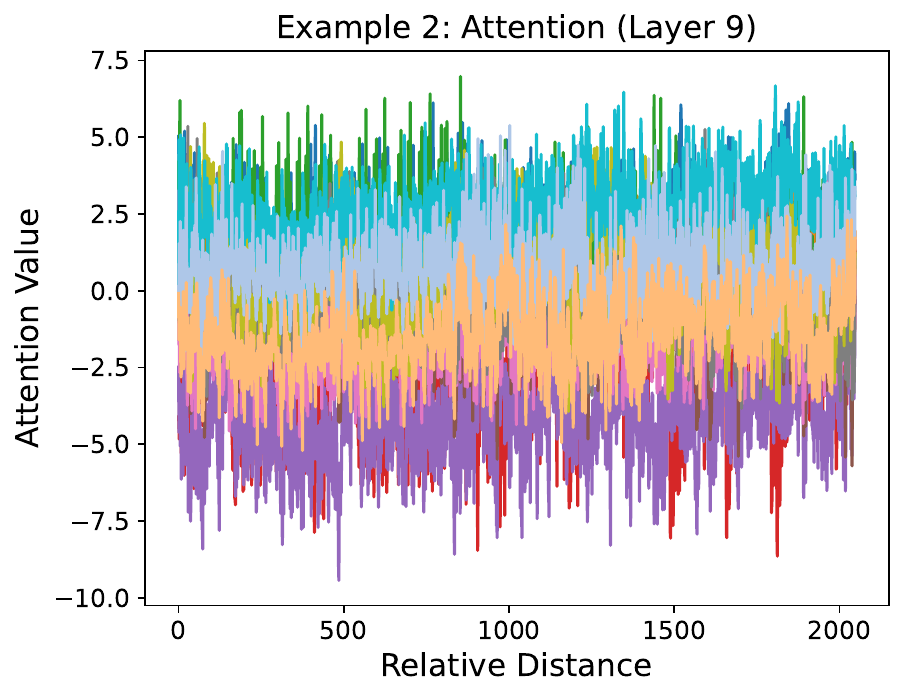}
\hspace{0in}
\includegraphics[width=0.32\textwidth]{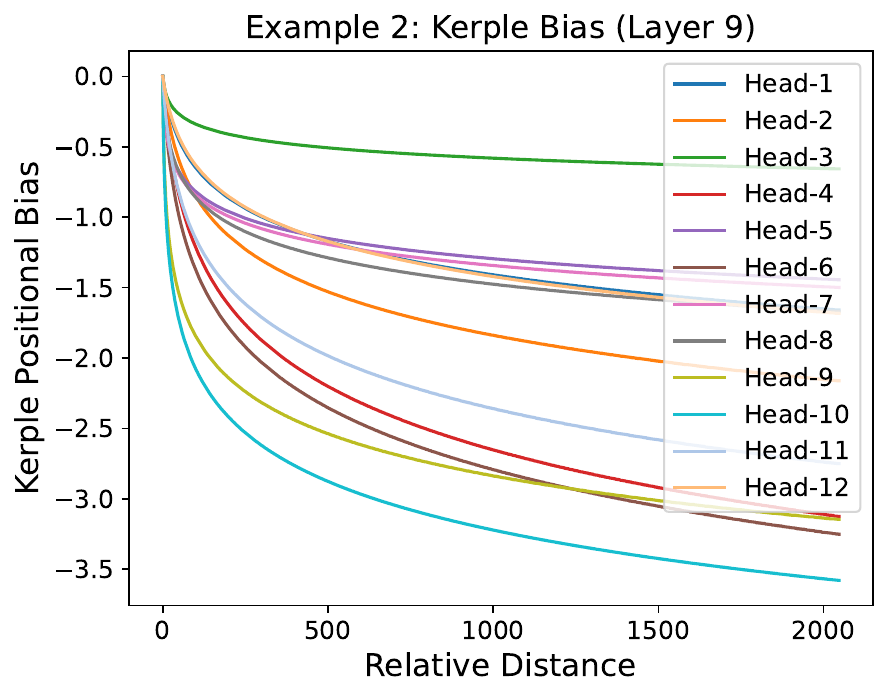}
\hspace{0in}
\includegraphics[width=0.32\textwidth]{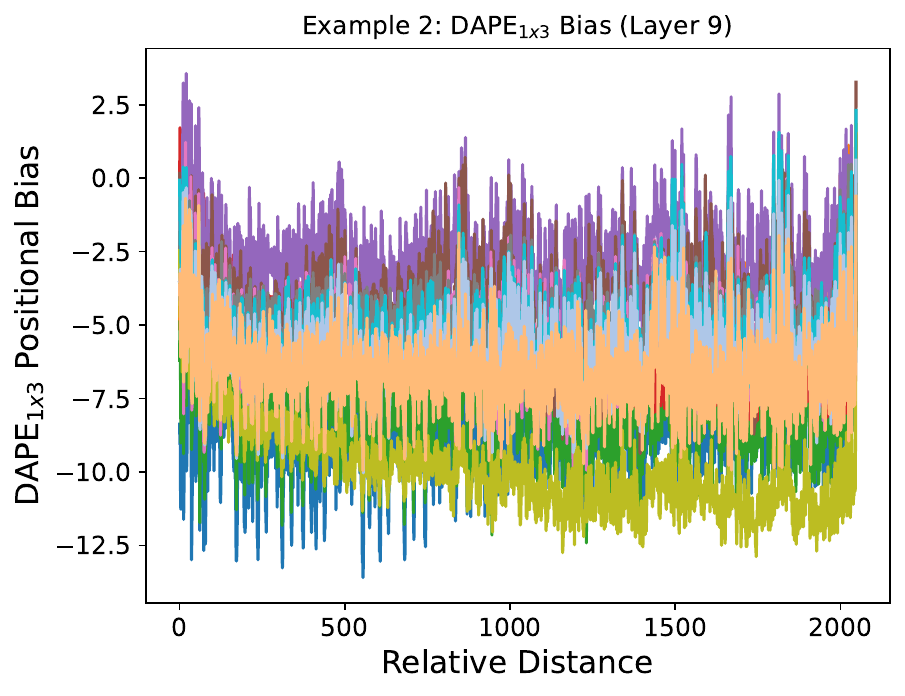}

\includegraphics[width=0.32\textwidth]{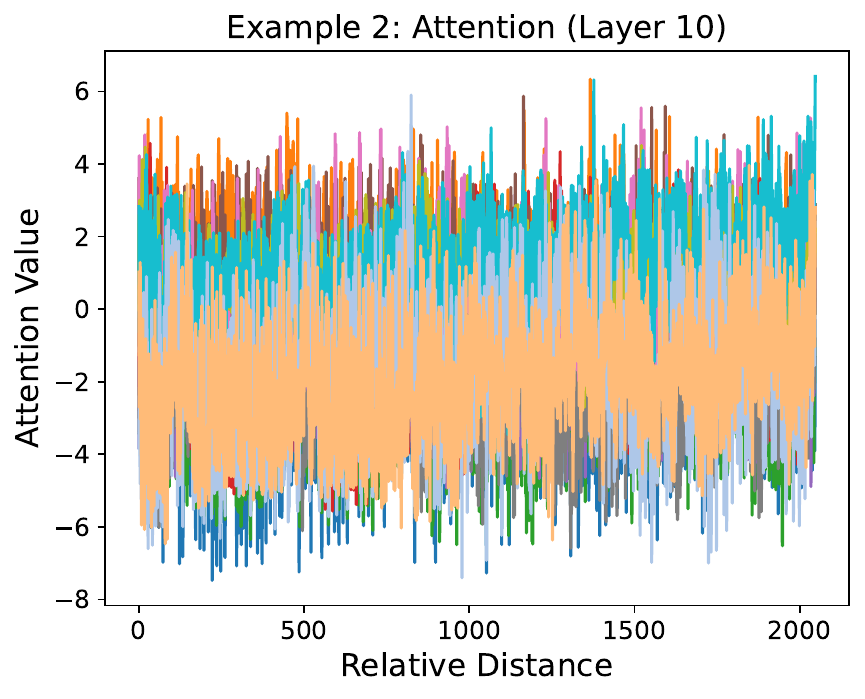}
\hspace{0in}
\includegraphics[width=0.32\textwidth]{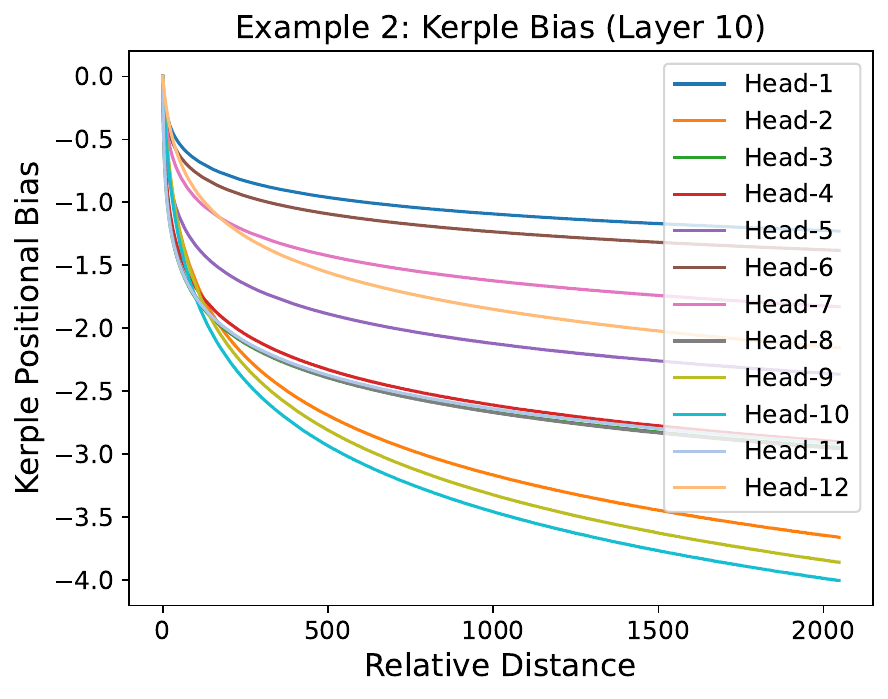}
\hspace{0in}
\includegraphics[width=0.32\textwidth]{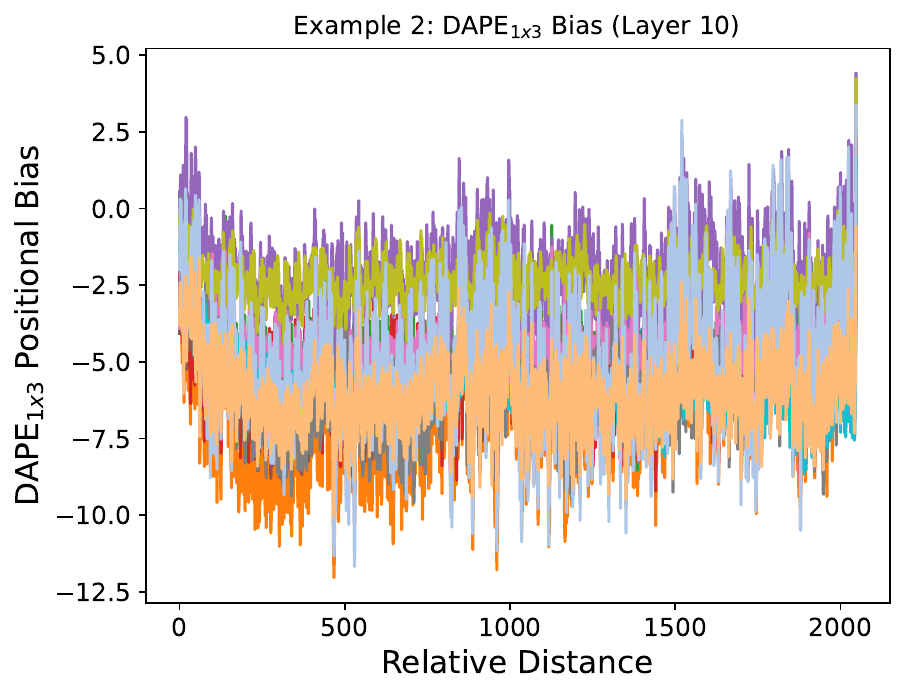}

\includegraphics[width=0.32\textwidth]{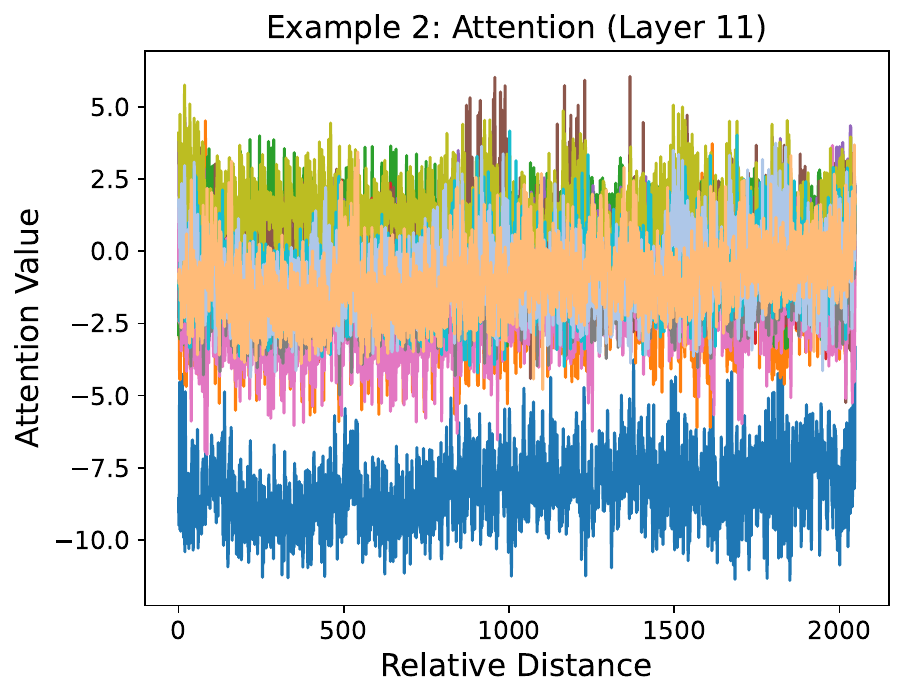}
\hspace{0in}
\includegraphics[width=0.32\textwidth]{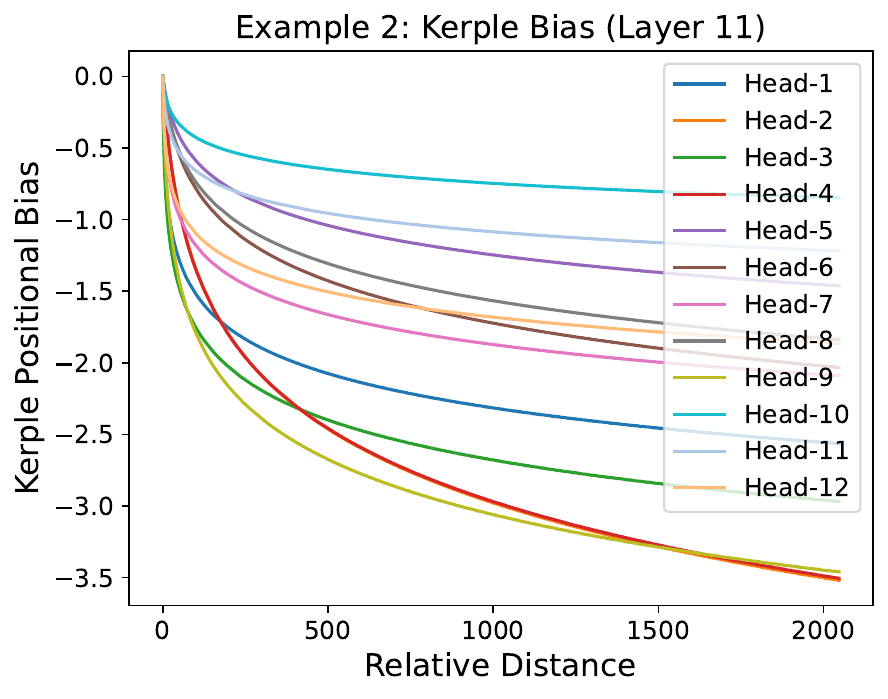}
\hspace{0in}
\includegraphics[width=0.32\textwidth]{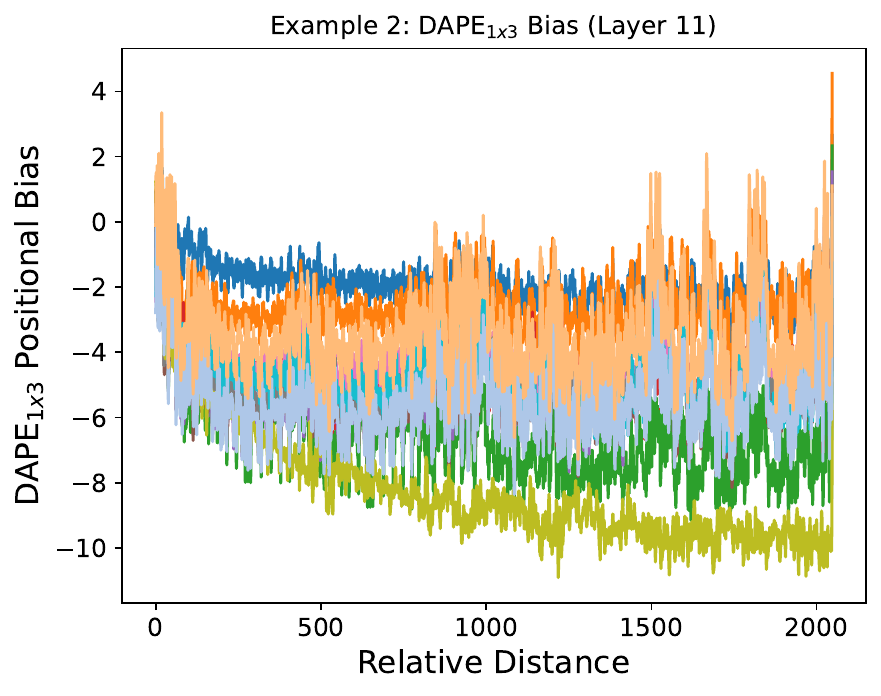}
\hspace{0in}

\includegraphics[width=0.32\textwidth]{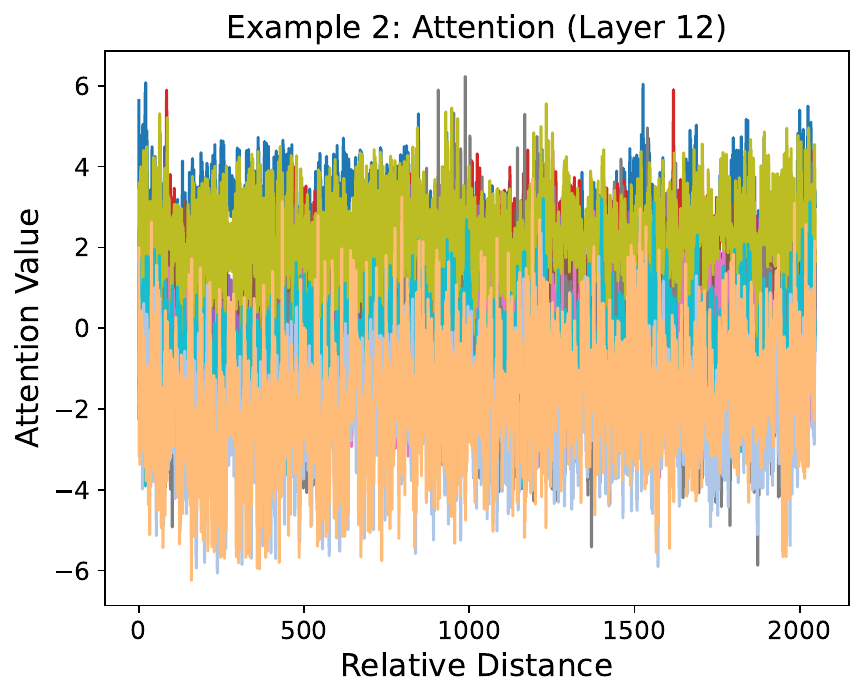}
\hspace{0in}
\includegraphics[width=0.32\textwidth]{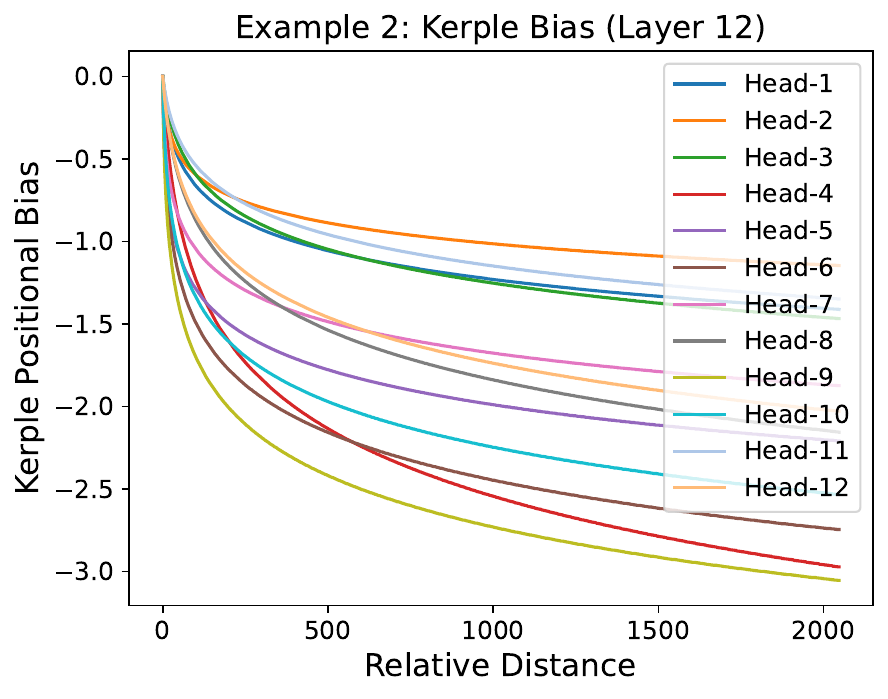}
\hspace{0in}
\includegraphics[width=0.32\textwidth]{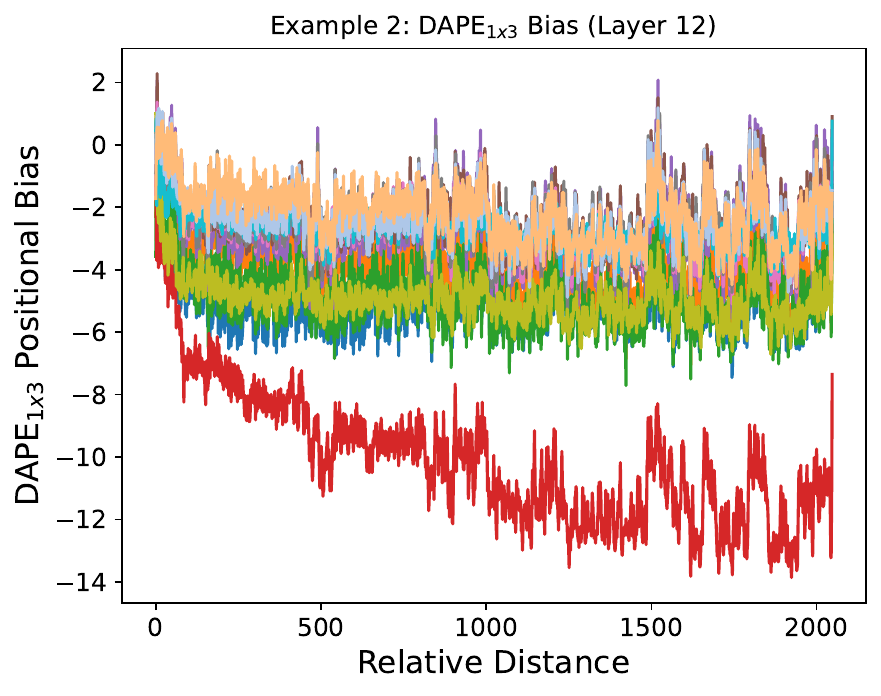}

\hspace{0in}
\caption{
\small
\textbf{Evaluation Length 2048 Example 2: Part 2. From Left to Right: (1) The Attention is $\mX \mW_Q(\mX \mW_K)^{\top}$; (2) The Kerple bias is $\mB$; (3) The \methodShortName (with Kerple) bias is $f( \mX \mW_Q(\mX \mW_K)^{\top},\mB)$.
}
}
\end{figure}

\clearpage
\newpage

\subsection{Visualization on length 8192}
\begin{figure}[htbp]
\setlength{\abovecaptionskip}{0.1cm}
\centering
\includegraphics[width=0.32\textwidth]{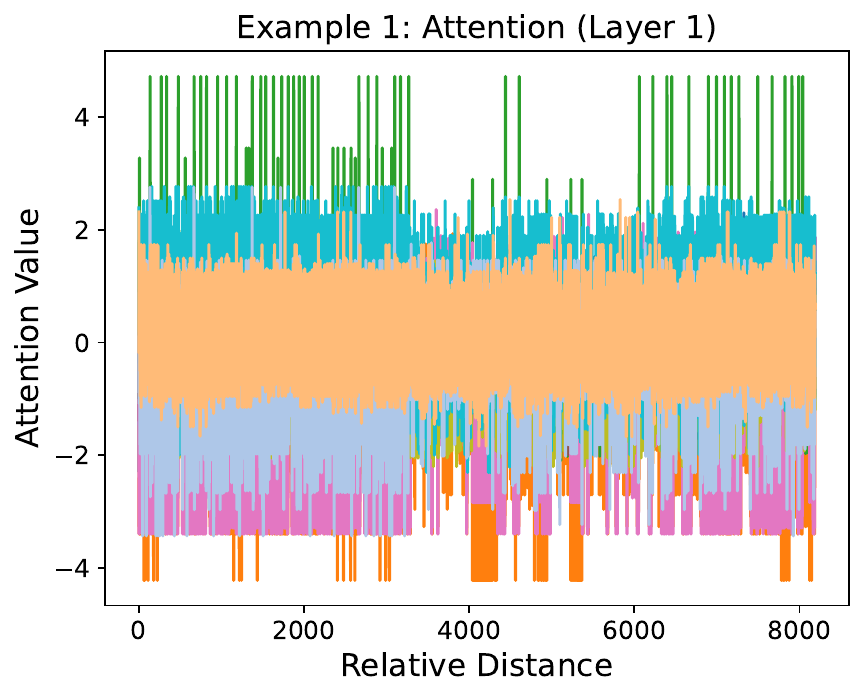}
\hspace{0in}
\includegraphics[width=0.32\textwidth]{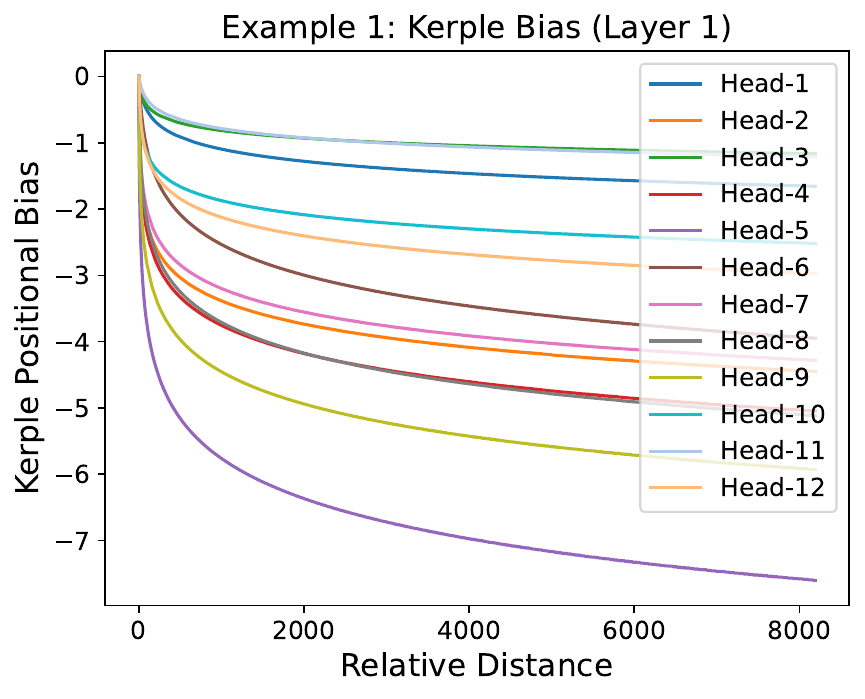}
\hspace{0in}
\includegraphics[width=0.32\textwidth]{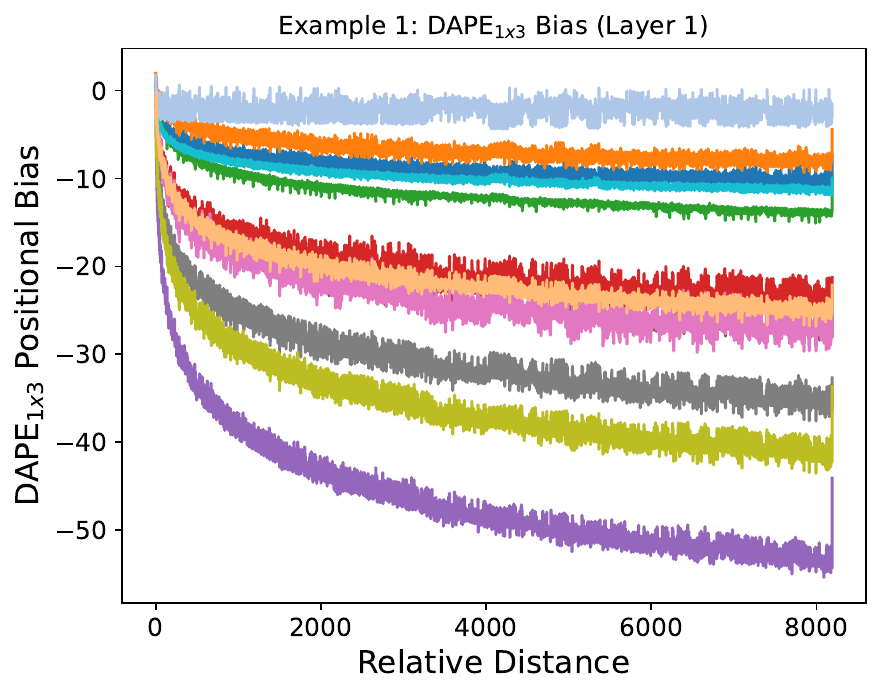}
\hspace{0in}

\includegraphics[width=0.32\textwidth]{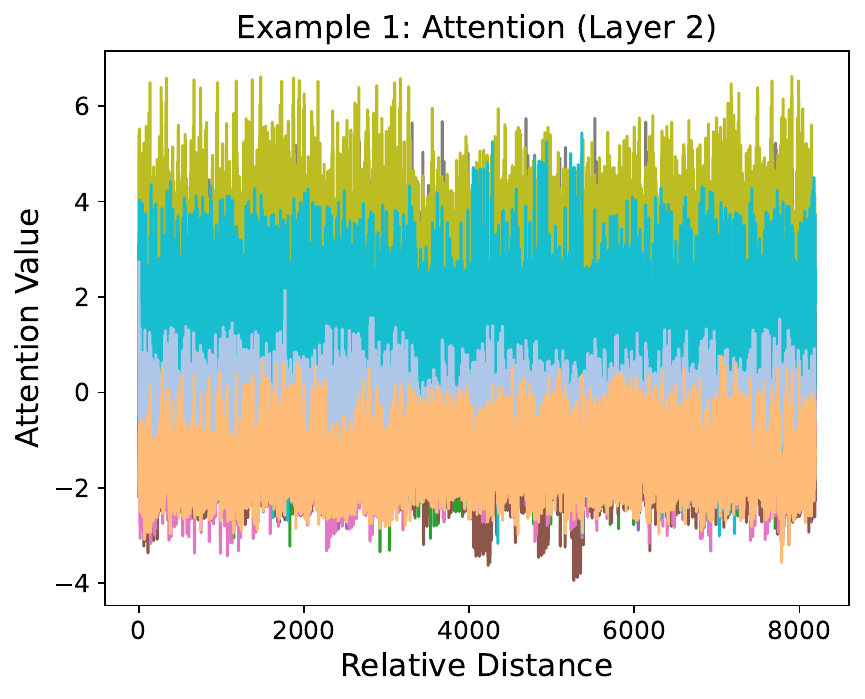}
\hspace{0in}
\includegraphics[width=0.32\textwidth]{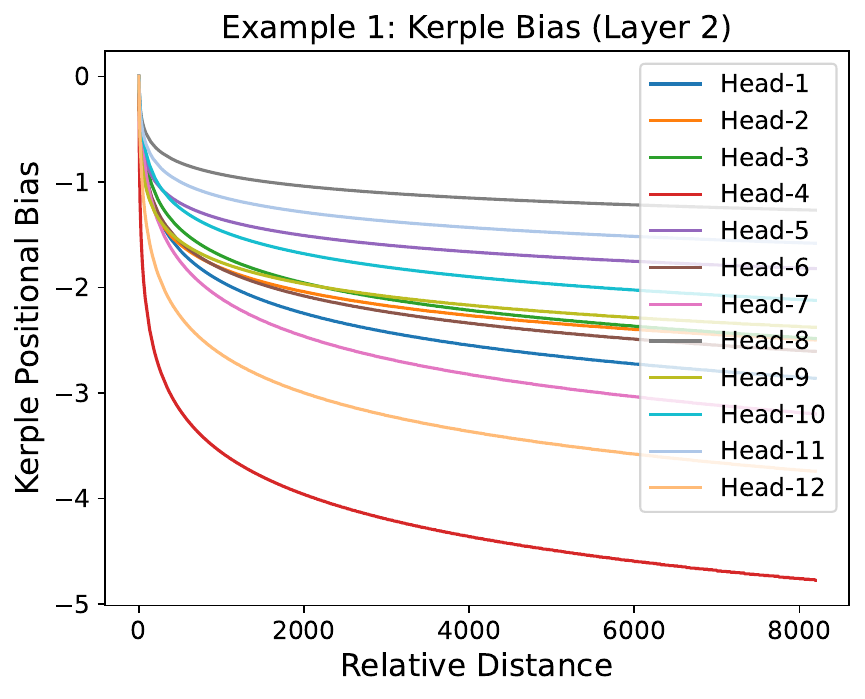}
\hspace{0in}
\includegraphics[width=0.32\textwidth]{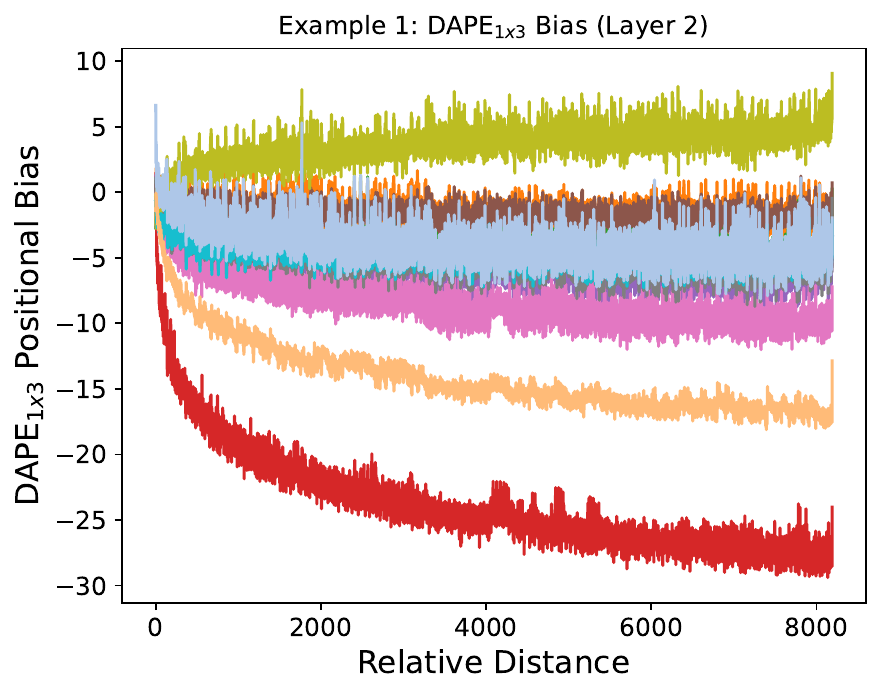}
\hspace{0in}

\includegraphics[width=0.32\textwidth]{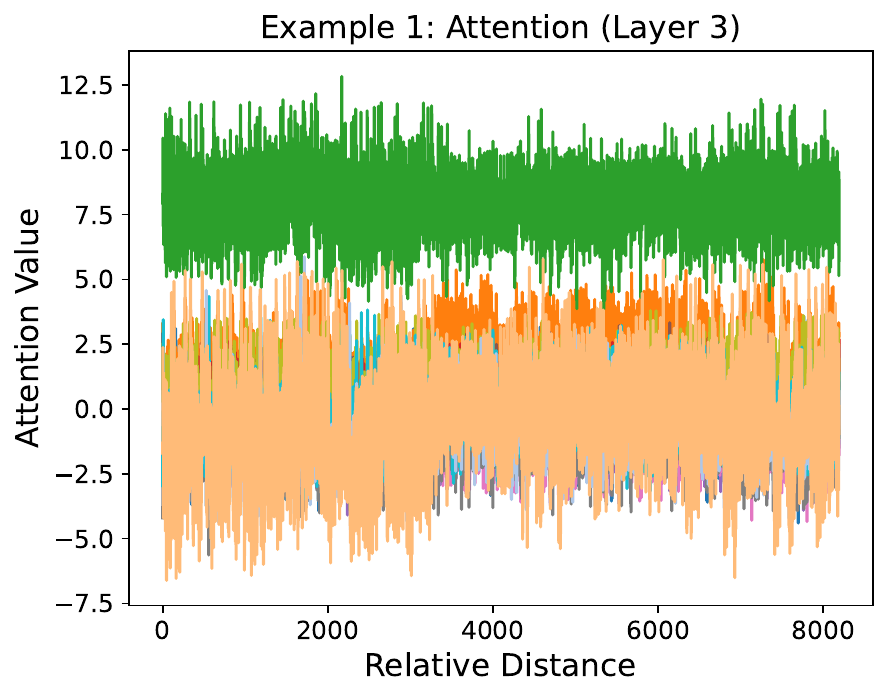}
\hspace{0in}
\includegraphics[width=0.32\textwidth]{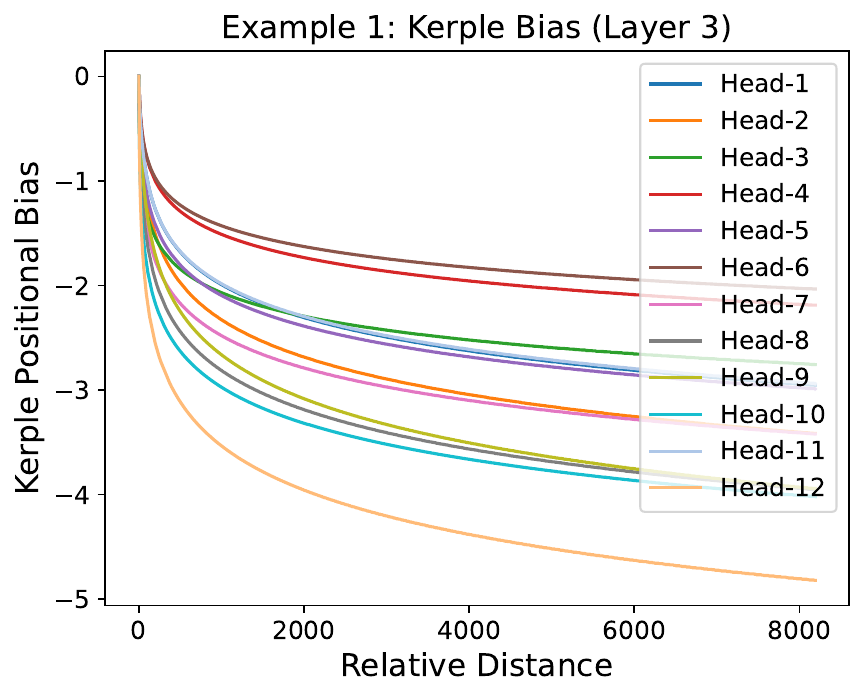}
\hspace{0in}
\includegraphics[width=0.32\textwidth]{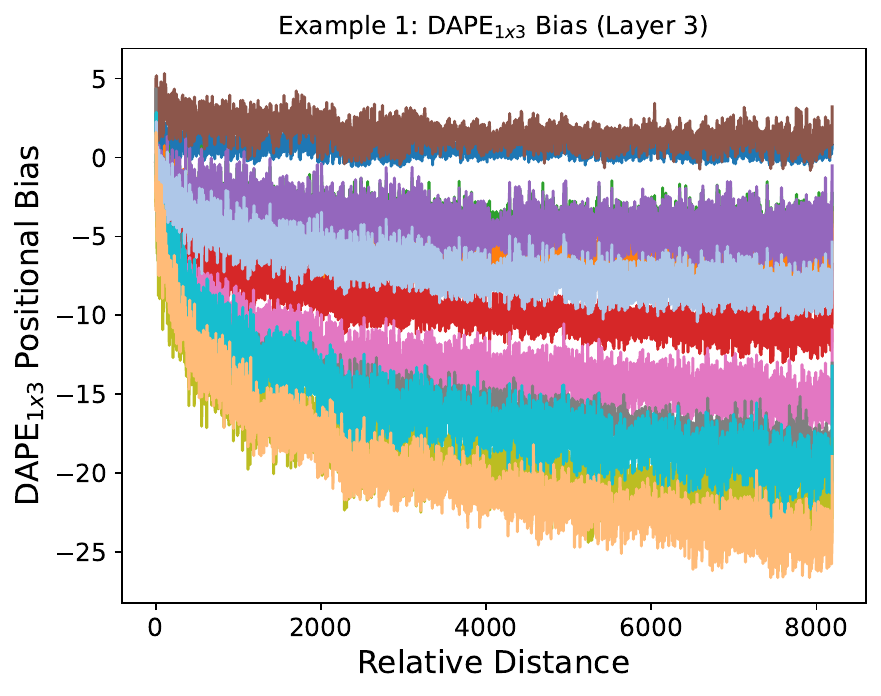}
\hspace{0in}

\includegraphics[width=0.32\textwidth]{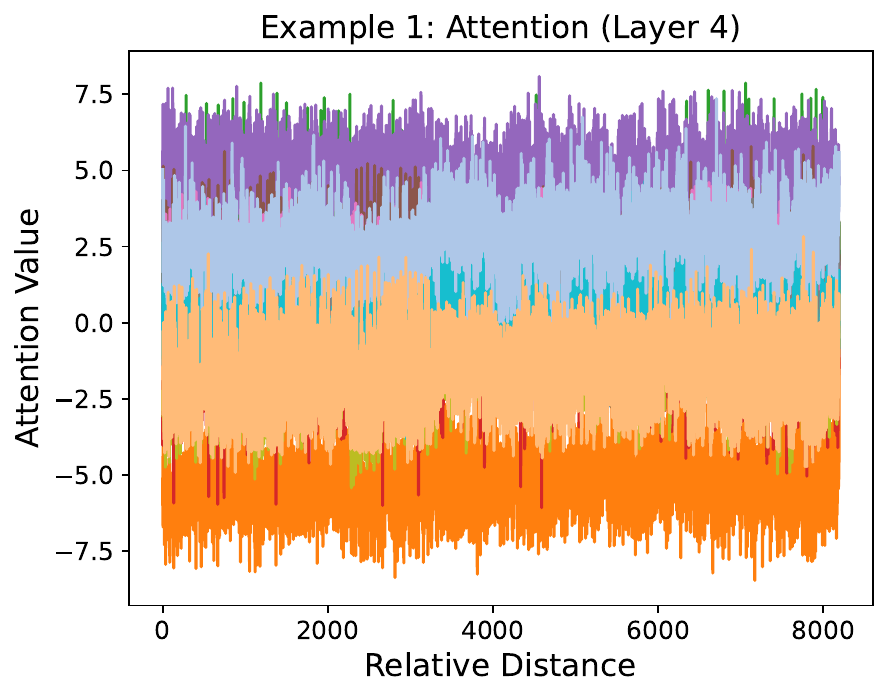}
\hspace{0in}
\includegraphics[width=0.32\textwidth]{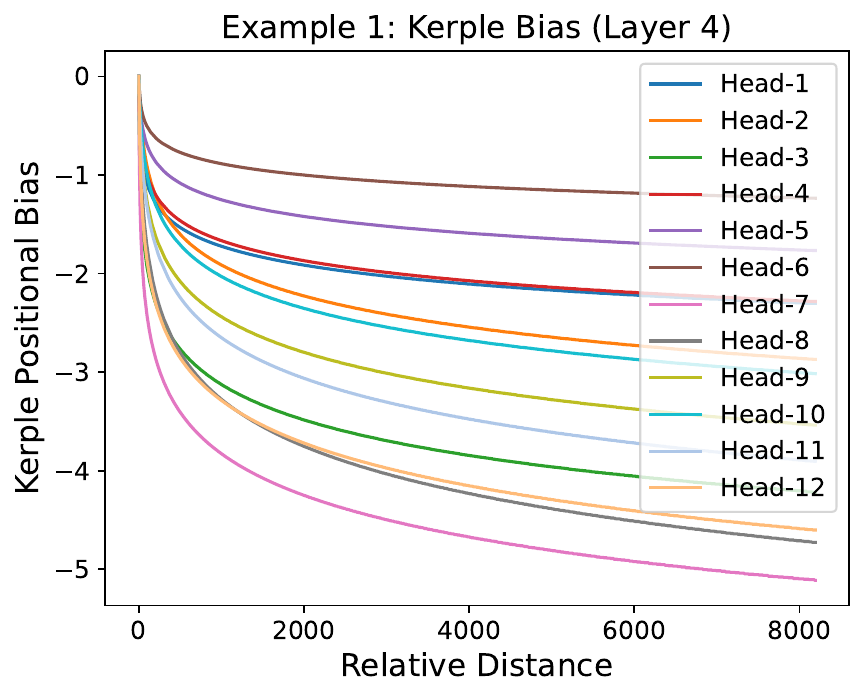}
\hspace{0in}
\includegraphics[width=0.32\textwidth]{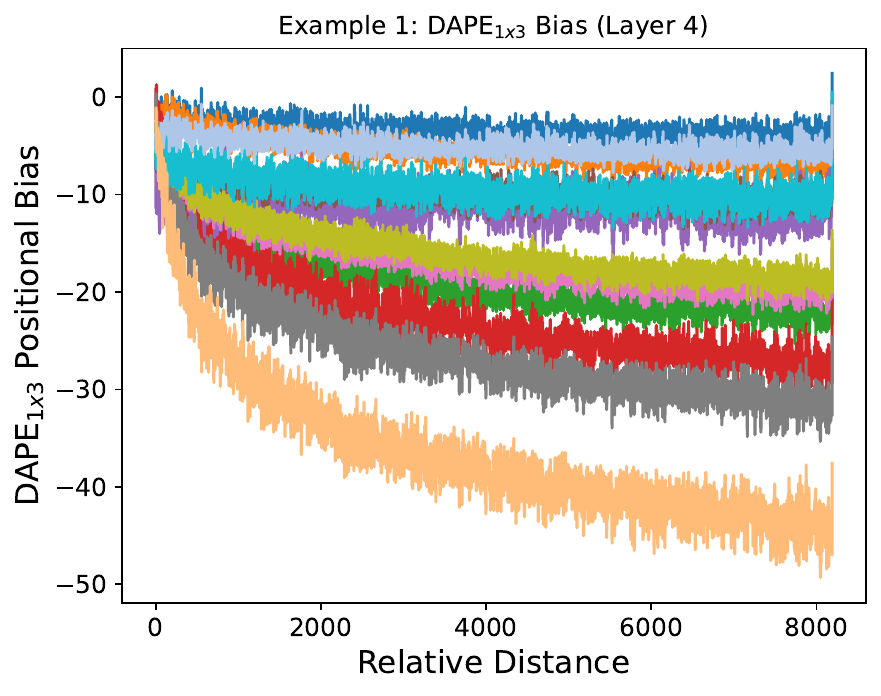}
\hspace{0in}

\includegraphics[width=0.32\textwidth]{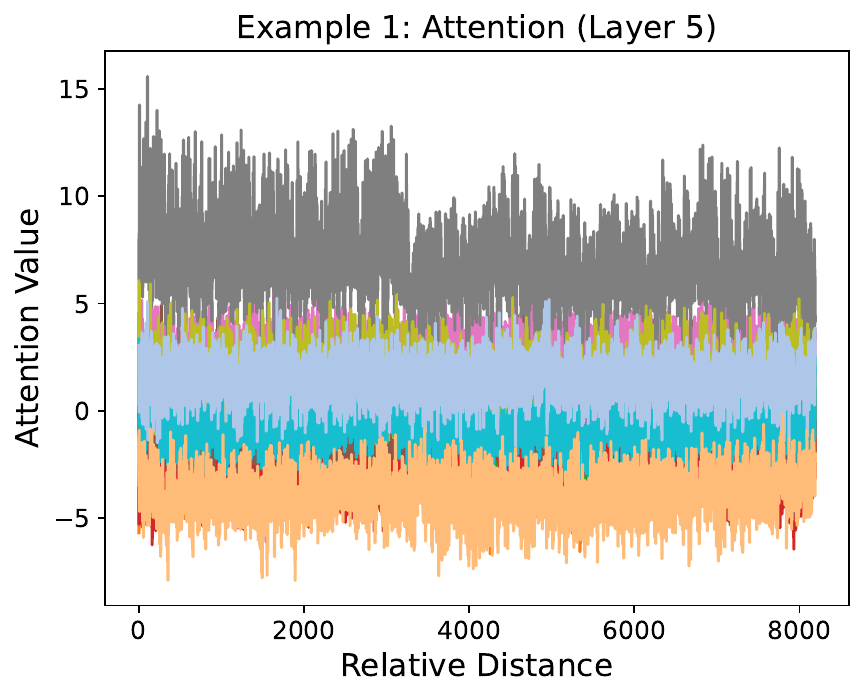}
\hspace{0in}
\includegraphics[width=0.32\textwidth]{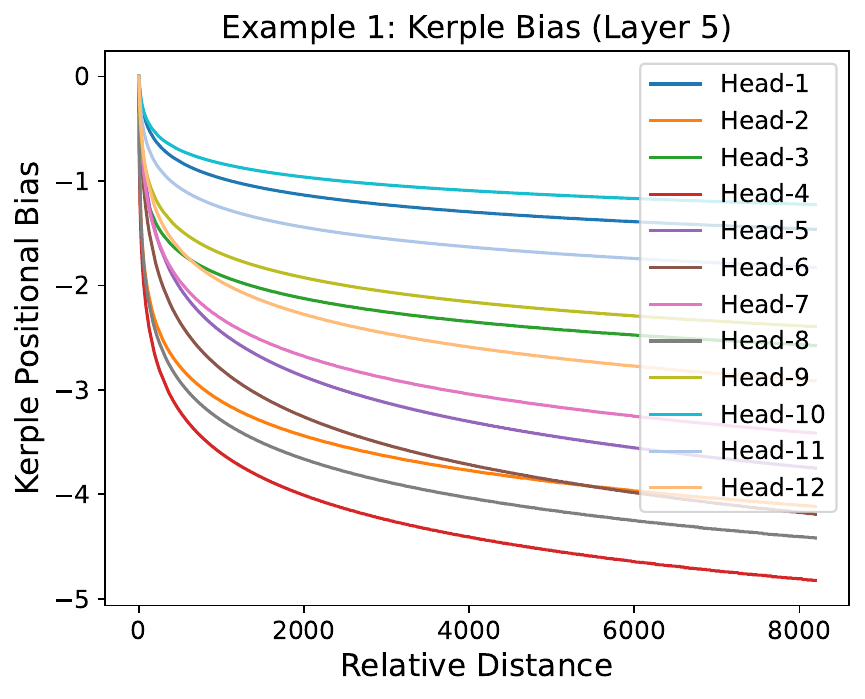}
\hspace{0in}
\includegraphics[width=0.32\textwidth]{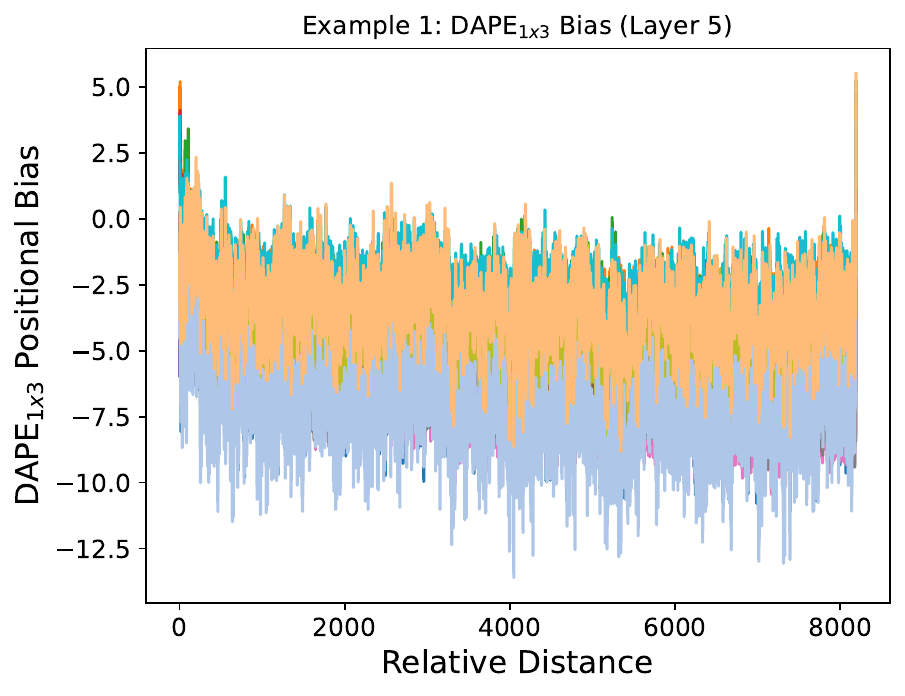}
\hspace{0in}

\hspace{0in}
\caption{
\small
\textbf{Evaluation Length 8192 Example 1: Part 1. From Left to Right: (1) The Attention is $\mX \mW_Q(\mX \mW_K)^{\top}$; (2) The Kerple bias is $\mB$; (3) The \methodShortName (with Kerple) bias is $f( \mX \mW_Q(\mX \mW_K)^{\top},\mB)$.
}
}
\end{figure}

\begin{figure}[htbp]
\setlength{\abovecaptionskip}{0.1cm}
\centering

\includegraphics[width=0.32\textwidth]{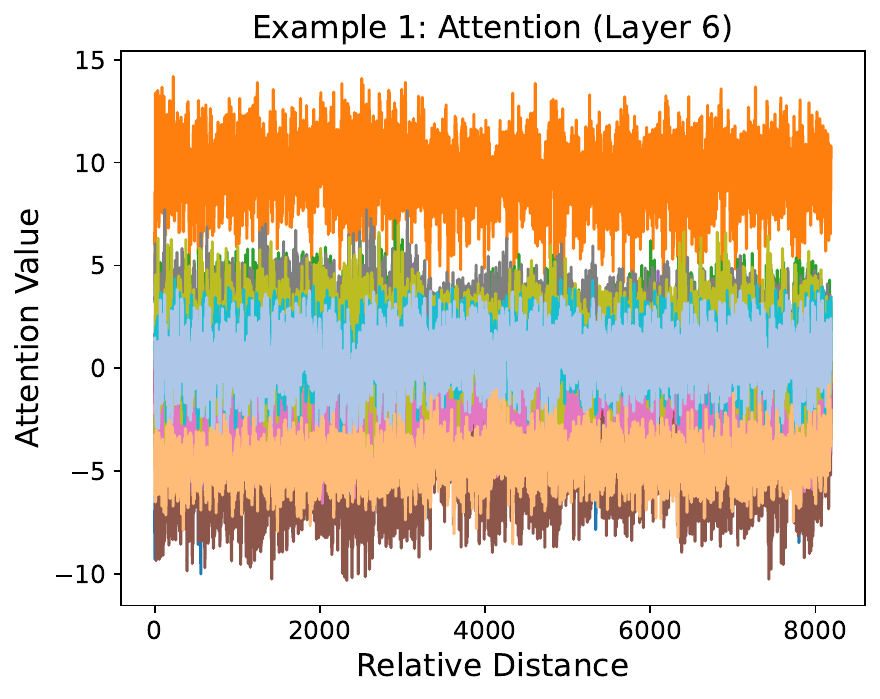}
\hspace{0in}
\includegraphics[width=0.32\textwidth]{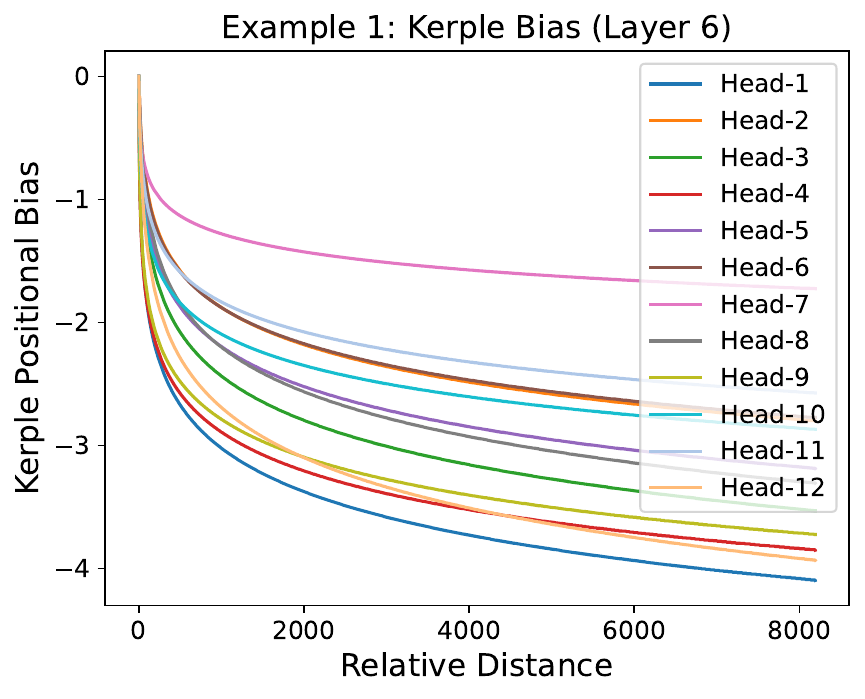}
\hspace{0in}
\includegraphics[width=0.32\textwidth]{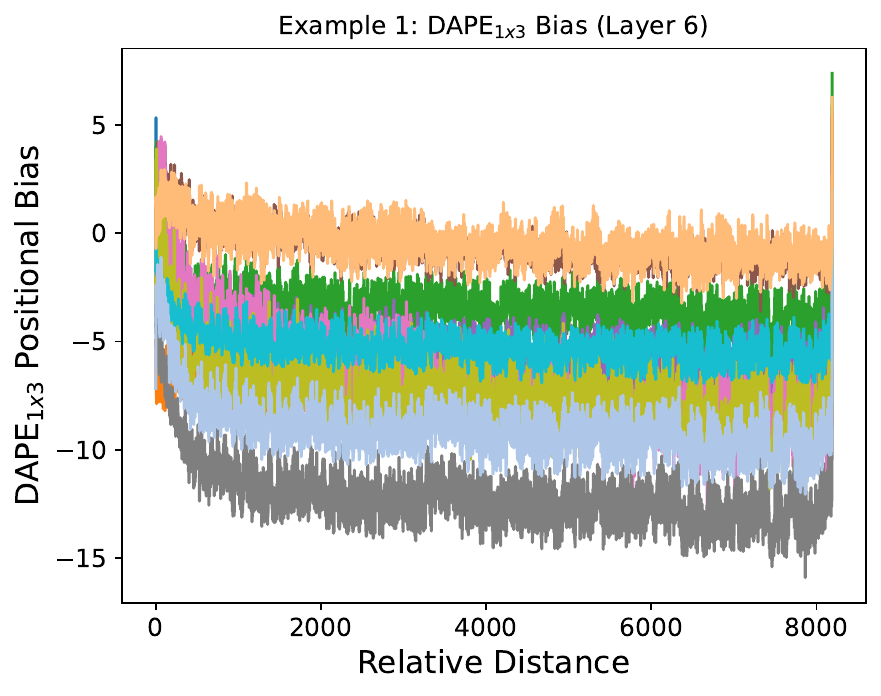}
\hspace{0in}

\includegraphics[width=0.32\textwidth]{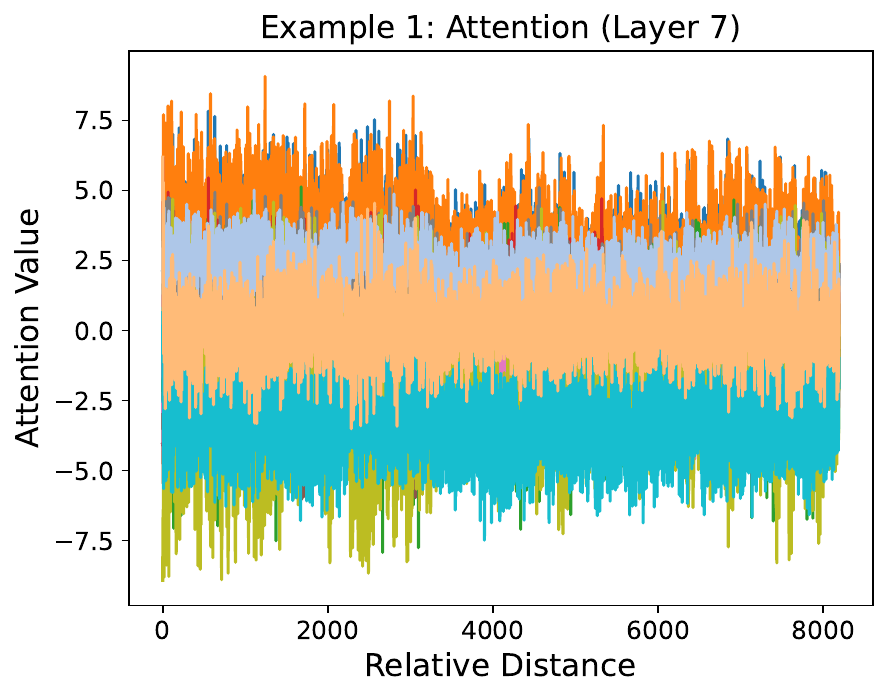}
\hspace{0in}
\includegraphics[width=0.32\textwidth]{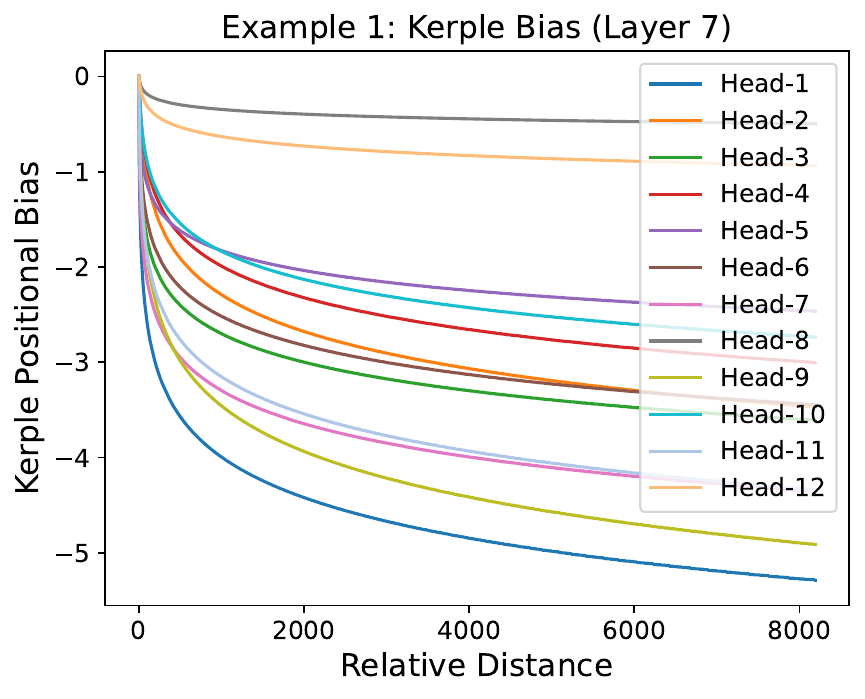}
\hspace{0in}
\includegraphics[width=0.32\textwidth]{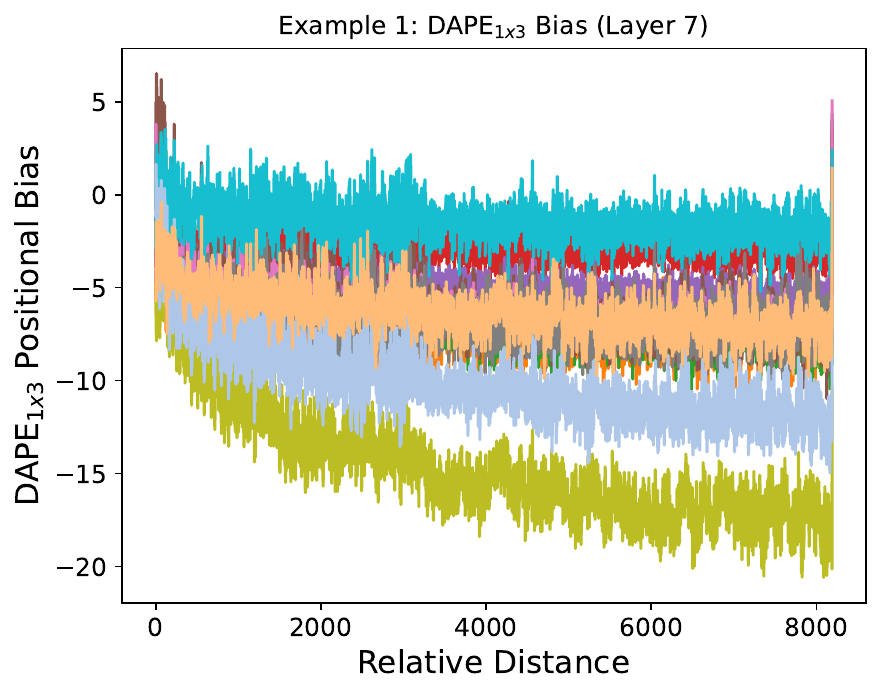}
\hspace{0in}

\includegraphics[width=0.32\textwidth]{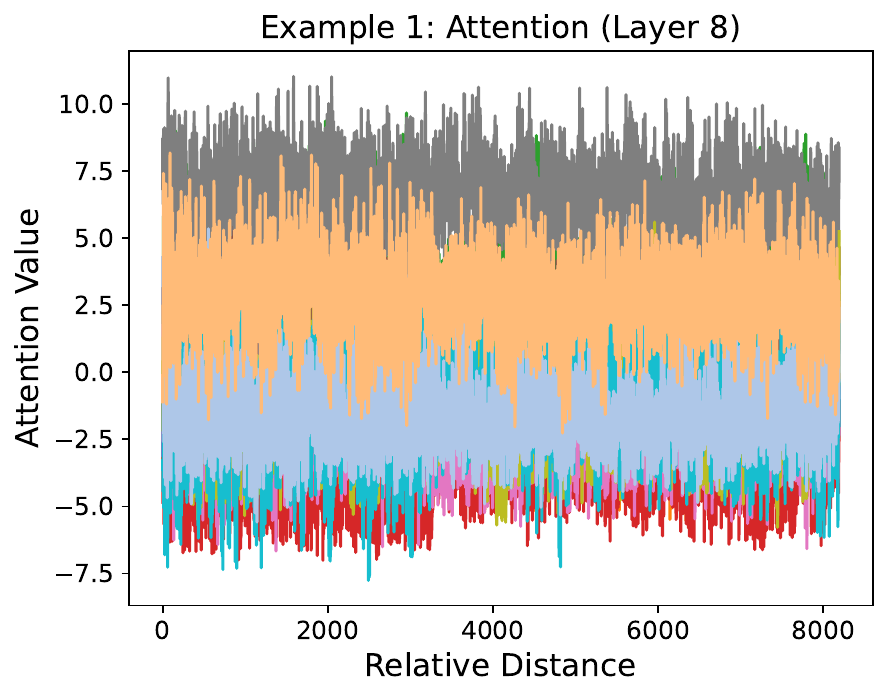}
\hspace{0in}
\includegraphics[width=0.32\textwidth]{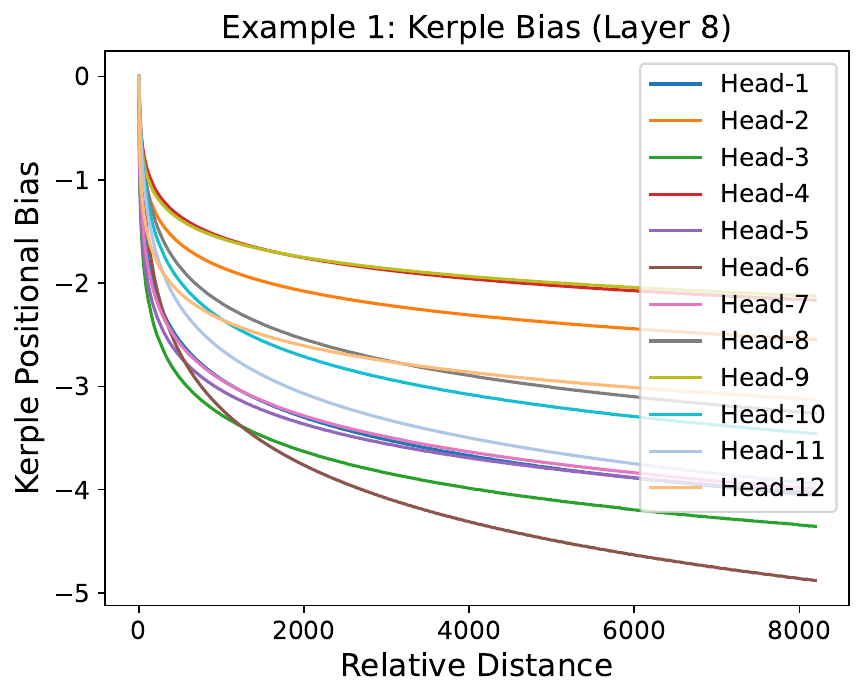}
\hspace{0in}
\includegraphics[width=0.32\textwidth]{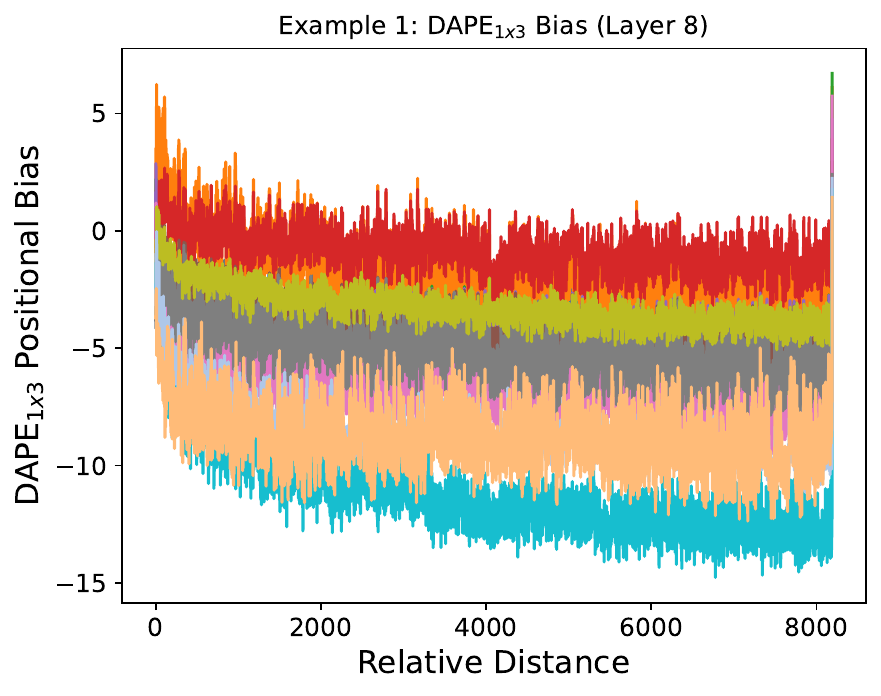}
\hspace{0in}

\includegraphics[width=0.32\textwidth]{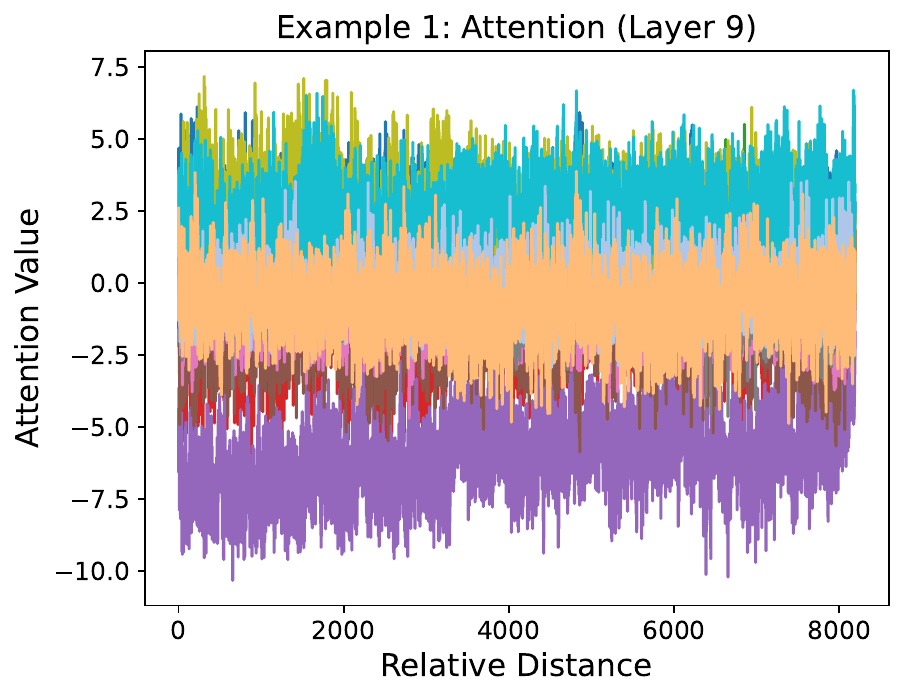}
\hspace{0in}
\includegraphics[width=0.32\textwidth]{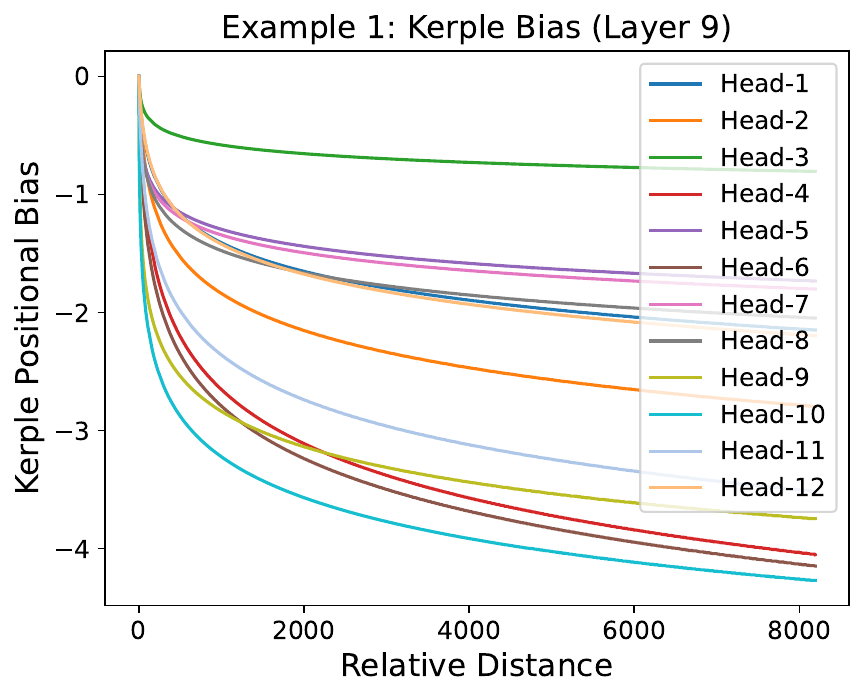}
\hspace{0in}
\includegraphics[width=0.32\textwidth]{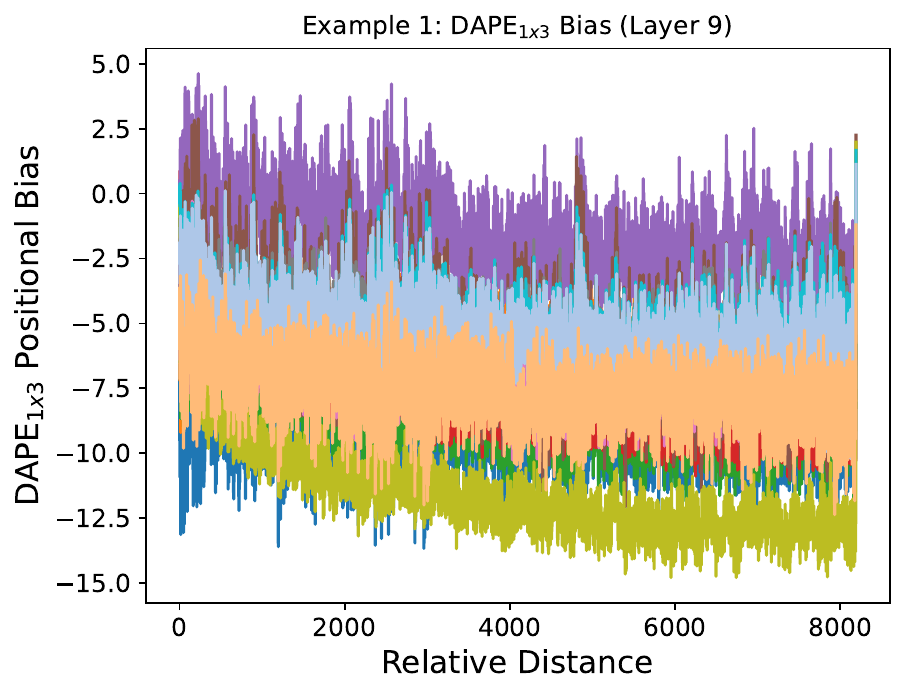}

\includegraphics[width=0.32\textwidth]{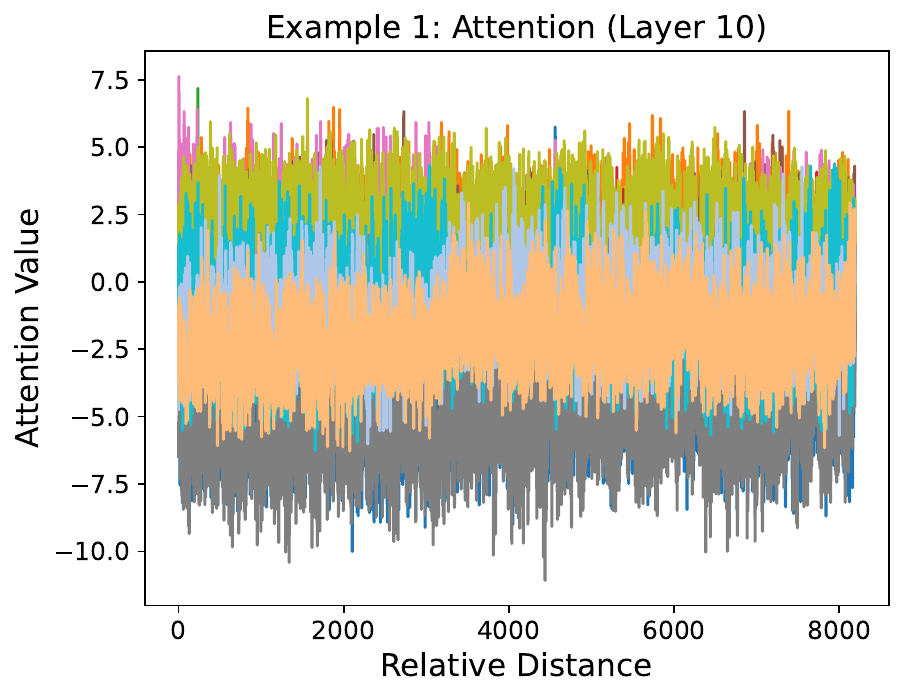}
\hspace{0in}
\includegraphics[width=0.32\textwidth]{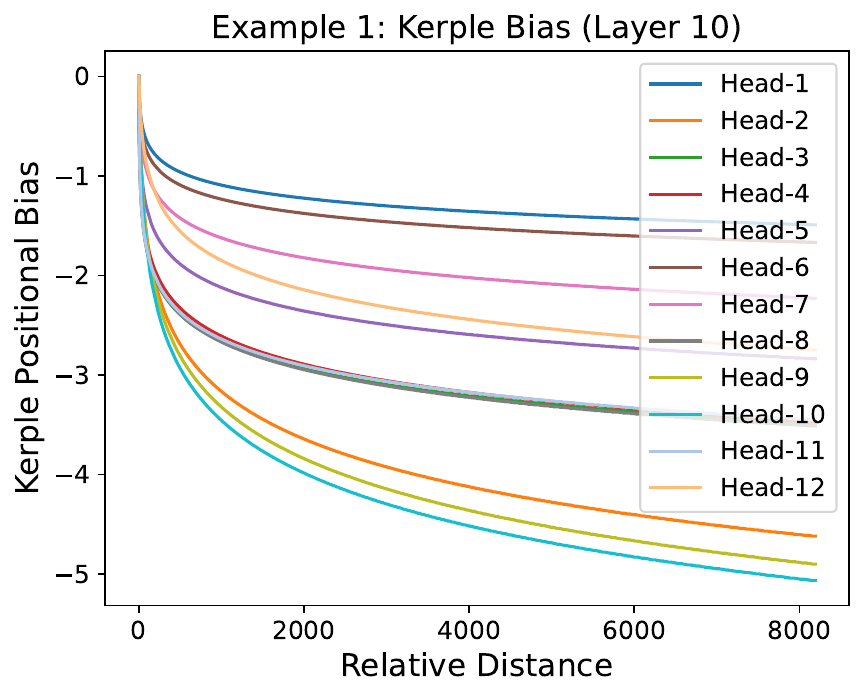}
\hspace{0in}
\includegraphics[width=0.32\textwidth]{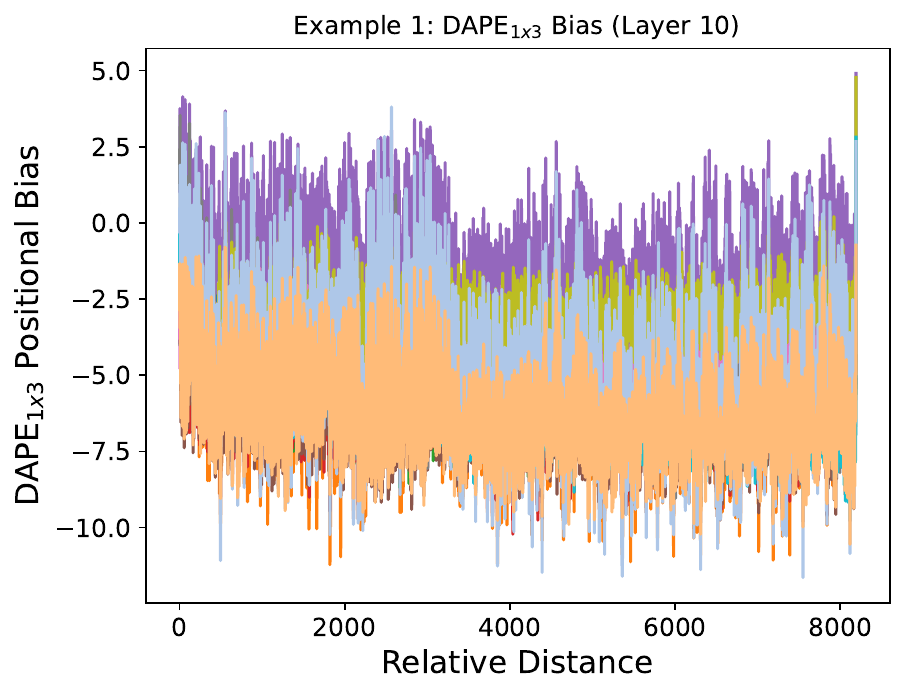}

\hspace{0in}
\caption{
\small
\textbf{Evaluation Length 8192 Example 1: Part 2. From Left to Right: (1) The Attention is $\mX \mW_Q(\mX \mW_K)^{\top}$; (2) The Kerple bias is $\mB$; (3) The \methodShortName (with Kerple) bias is $f( \mX \mW_Q(\mX \mW_K)^{\top},\mB)$.
}
}
\end{figure}

\newpage
\begin{figure}[htbp]
\setlength{\abovecaptionskip}{0.1cm}
\centering

\includegraphics[width=0.32\textwidth]{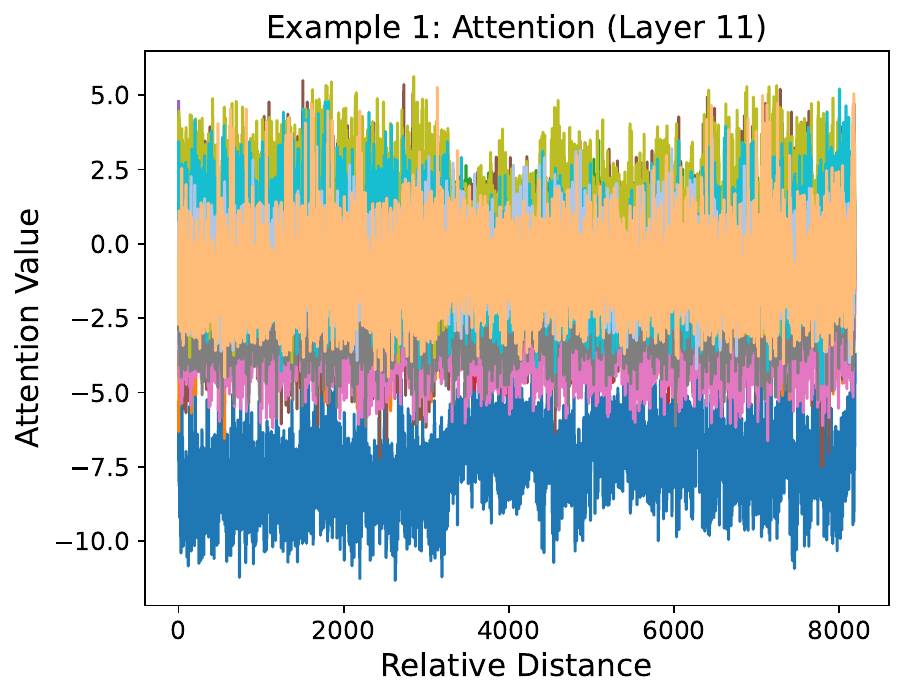}
\hspace{0in}
\includegraphics[width=0.32\textwidth]{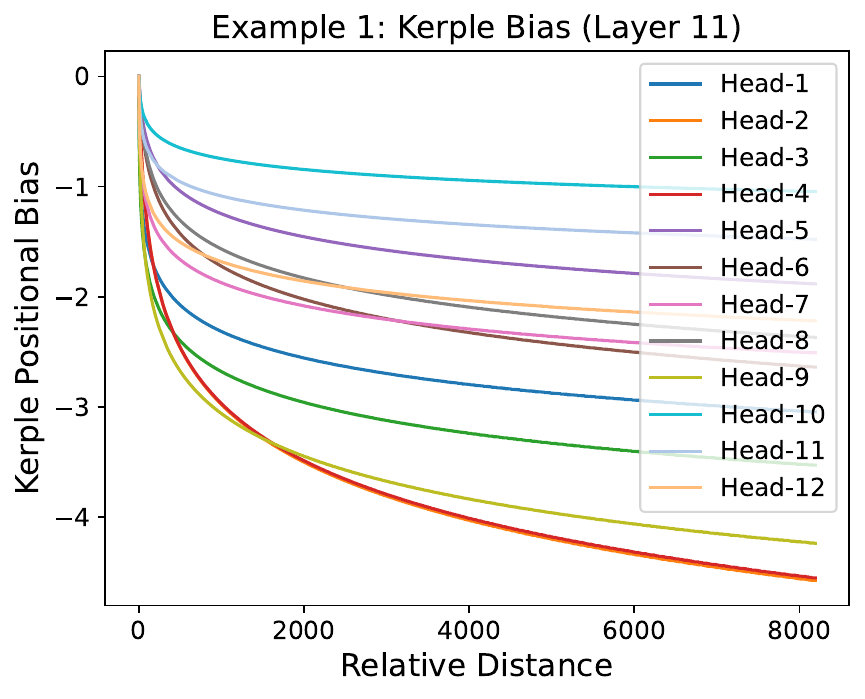}
\hspace{0in}
\includegraphics[width=0.32\textwidth]{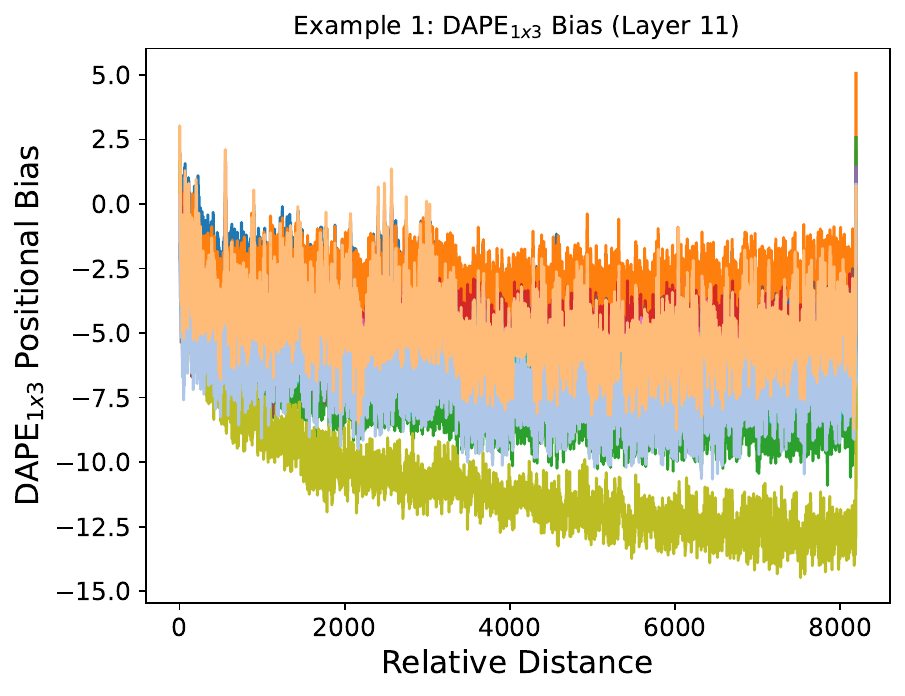}
\hspace{0in}

\includegraphics[width=0.32\textwidth]{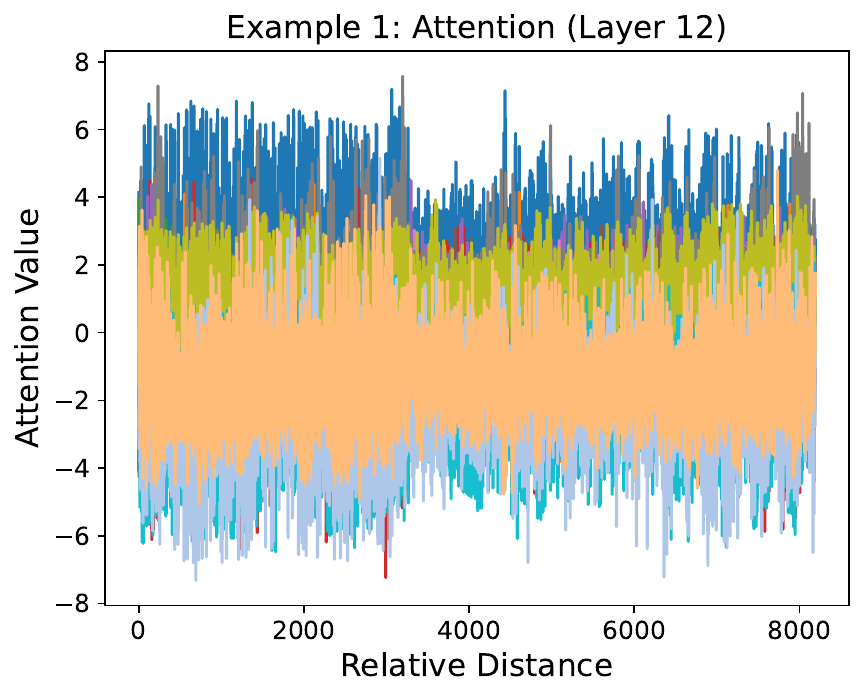}
\hspace{0in}
\includegraphics[width=0.32\textwidth]{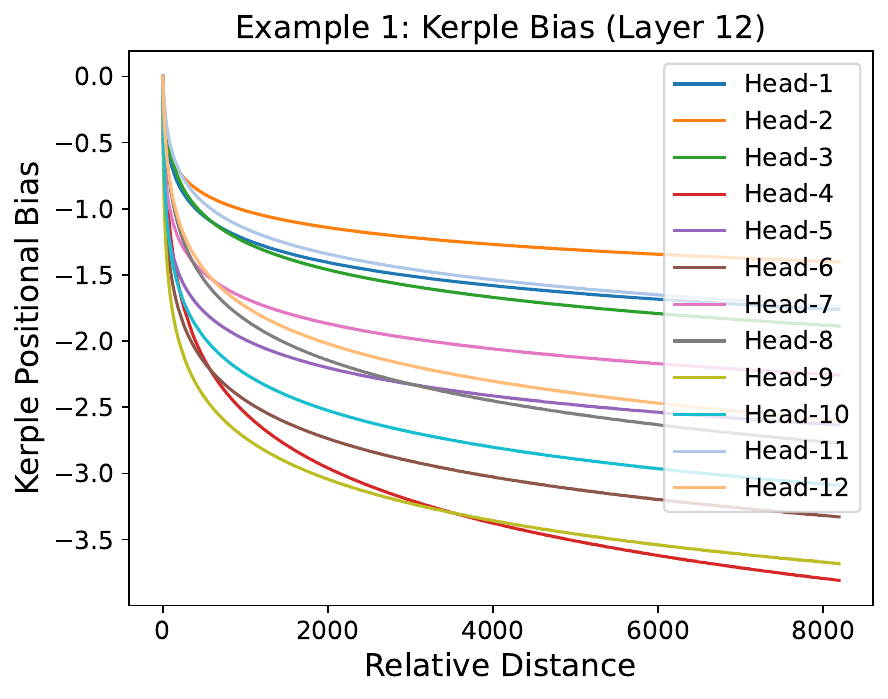}
\hspace{0in}
\includegraphics[width=0.32\textwidth]{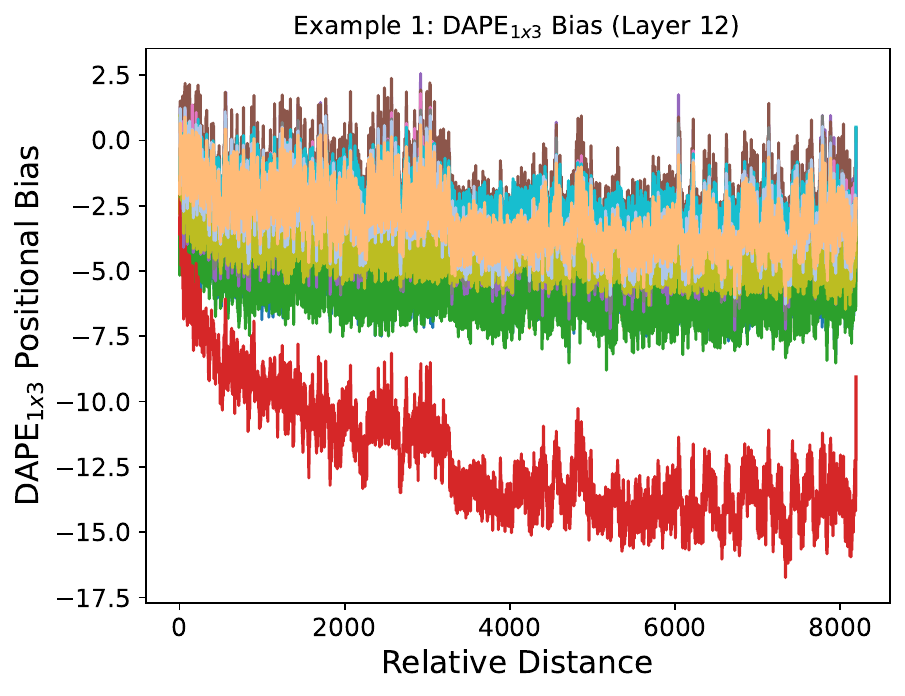}
\caption{
\small
\textbf{Evaluation Length 8192 Example 1: Part 3. From Left to Right: (1) The Attention is $\mX \mW_Q(\mX \mW_K)^{\top}$; (2) The Kerple bias is $\mB$; (3) The \methodShortName (with Kerple) bias is $f( \mX \mW_Q(\mX \mW_K)^{\top},\mB)$.
}
}
\end{figure}

\newpage

\begin{figure}[htbp]
\setlength{\abovecaptionskip}{0.1cm}
\centering
\includegraphics[width=0.32\textwidth]{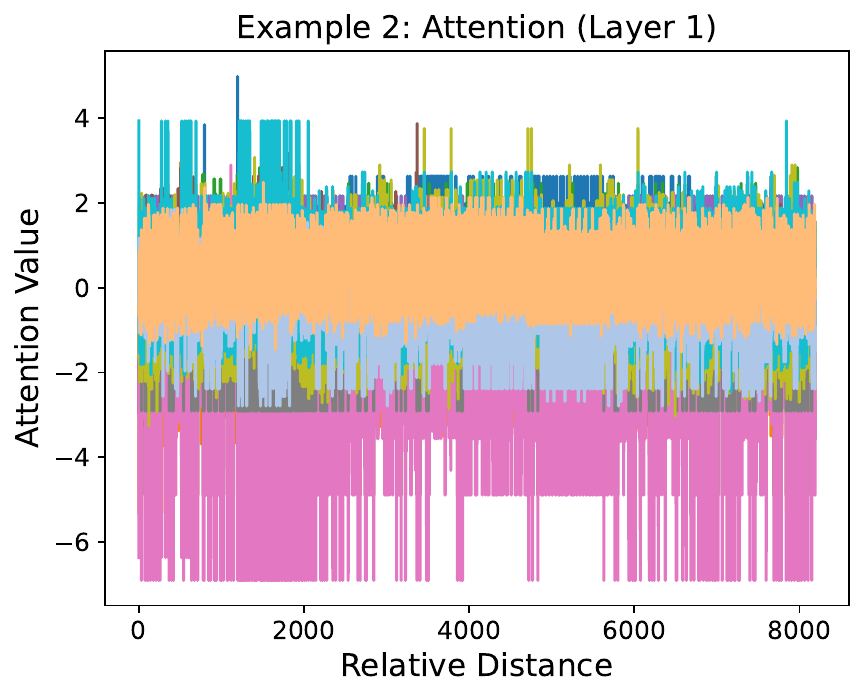}
\hspace{0in}
\includegraphics[width=0.32\textwidth]{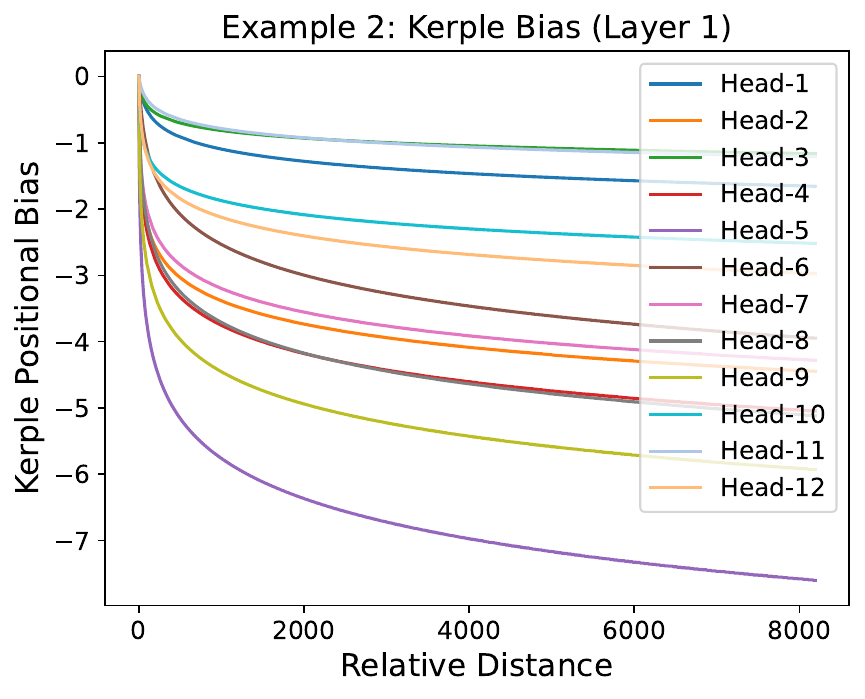}
\hspace{0in}
\includegraphics[width=0.32\textwidth]{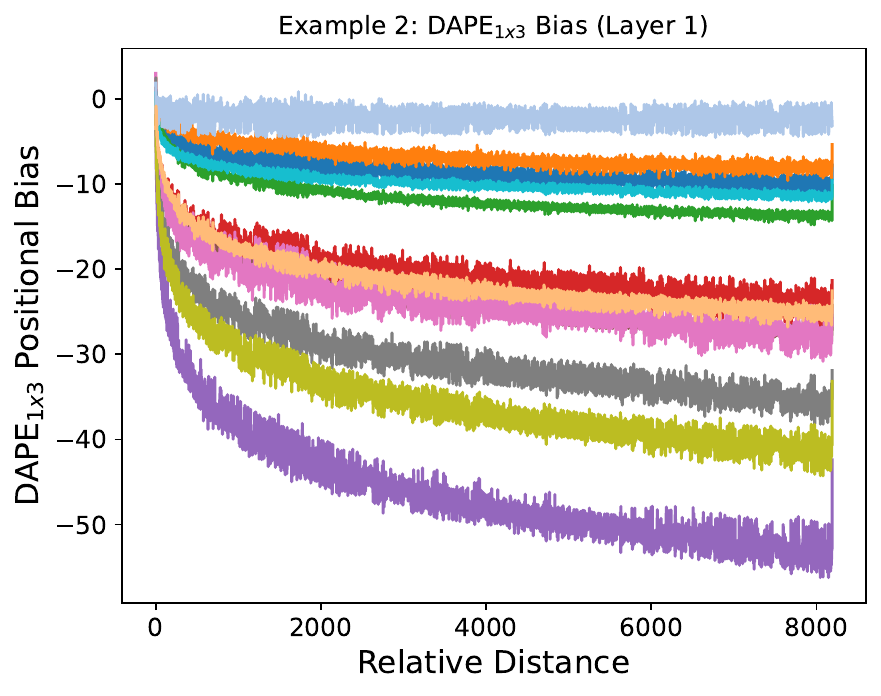}
\hspace{0in}

\includegraphics[width=0.32\textwidth]{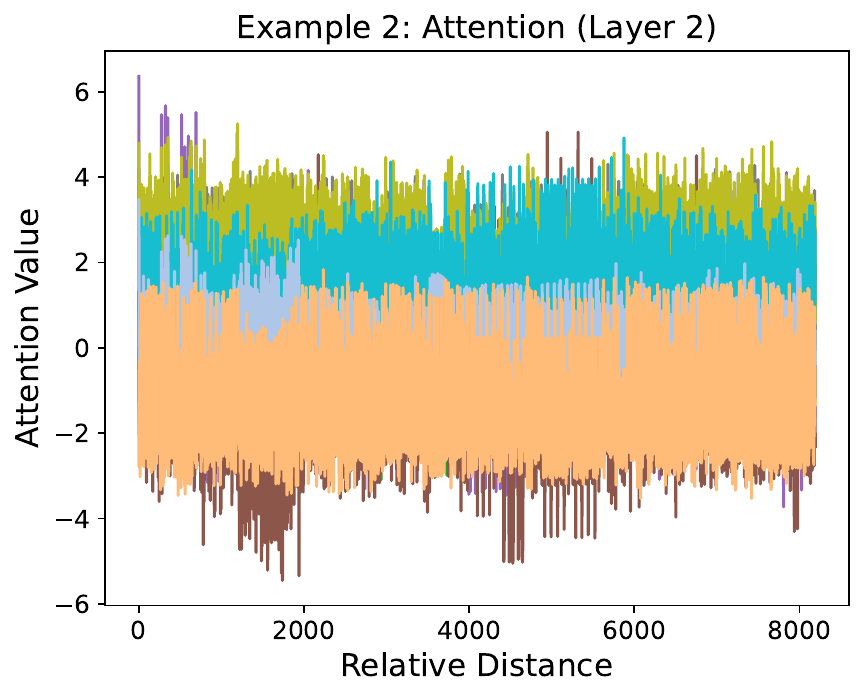}
\hspace{0in}
\includegraphics[width=0.32\textwidth]{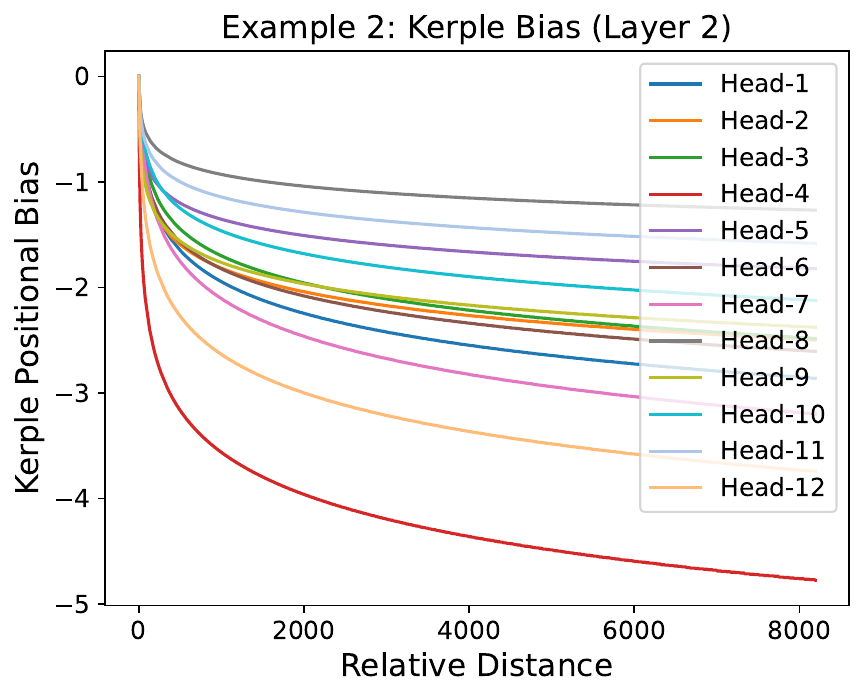}
\hspace{0in}
\includegraphics[width=0.32\textwidth]{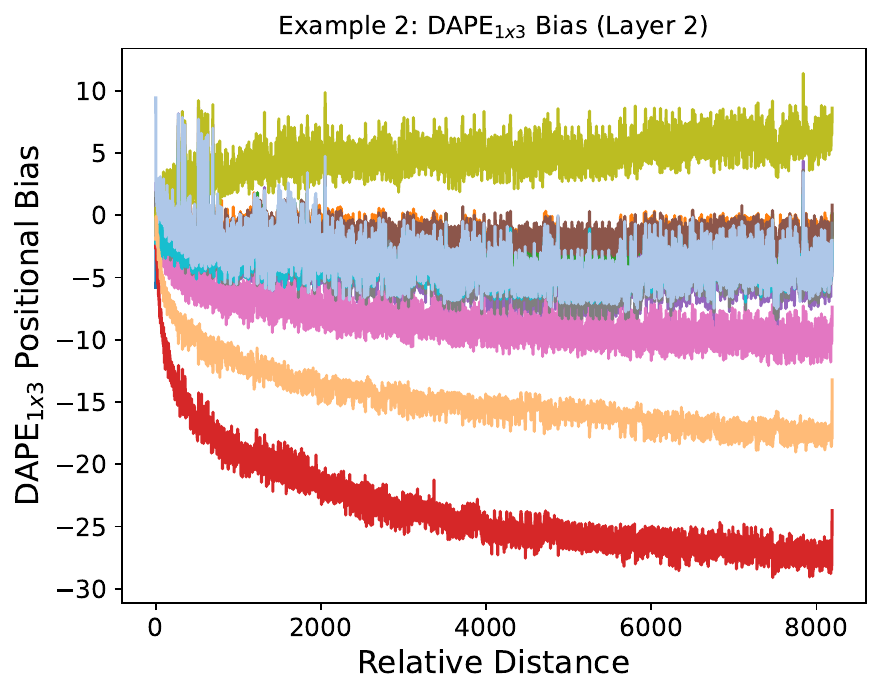}
\hspace{0in}

\includegraphics[width=0.32\textwidth]{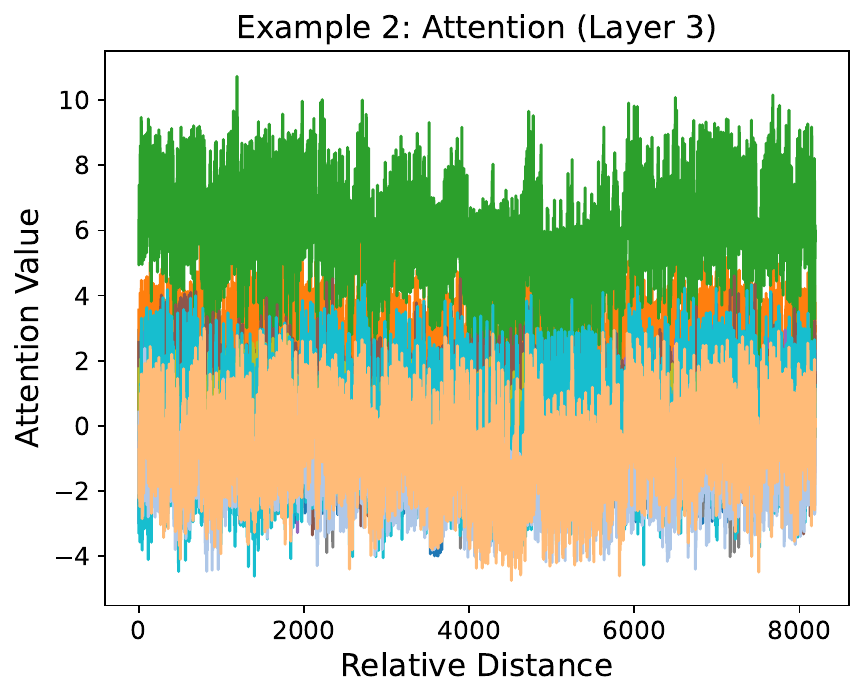}
\hspace{0in}
\includegraphics[width=0.32\textwidth]{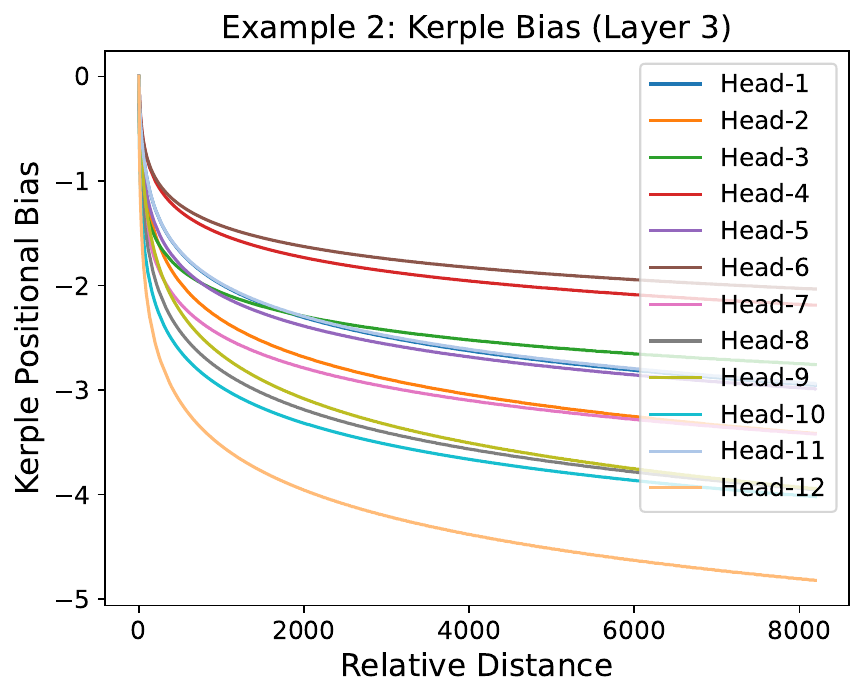}
\hspace{0in}
\includegraphics[width=0.32\textwidth]{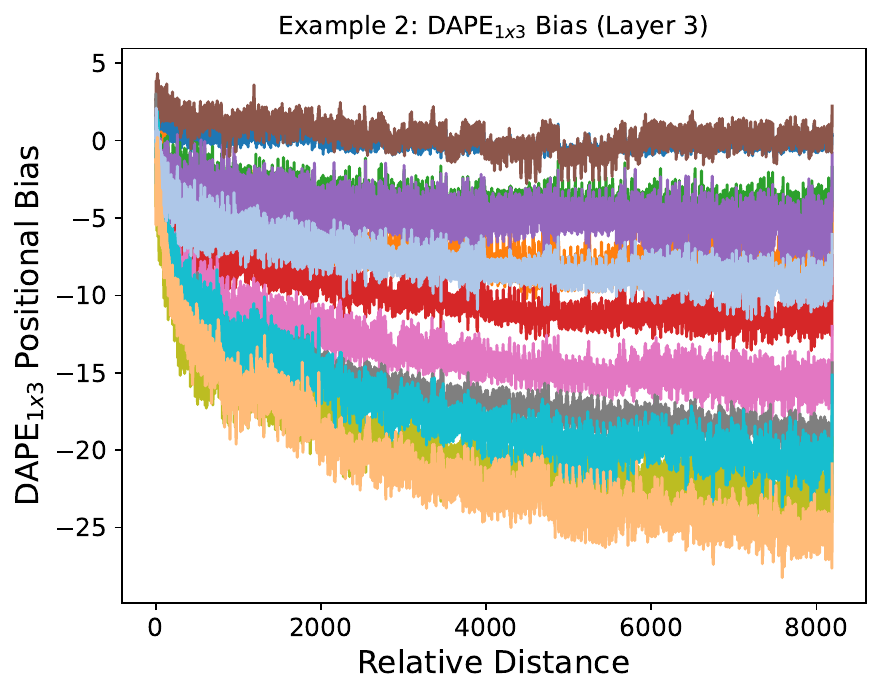}
\hspace{0in}

\includegraphics[width=0.32\textwidth]{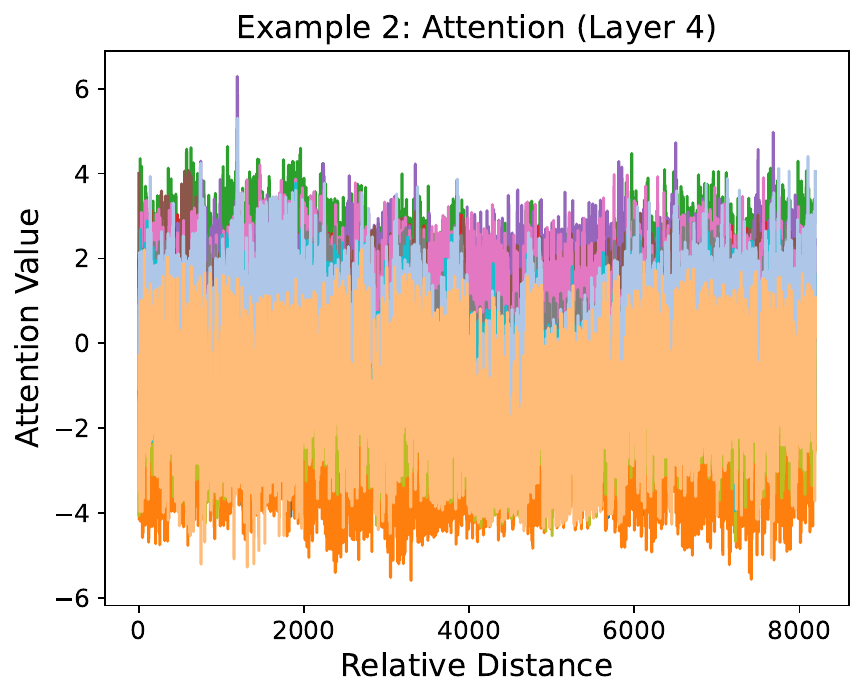}
\hspace{0in}
\includegraphics[width=0.32\textwidth]{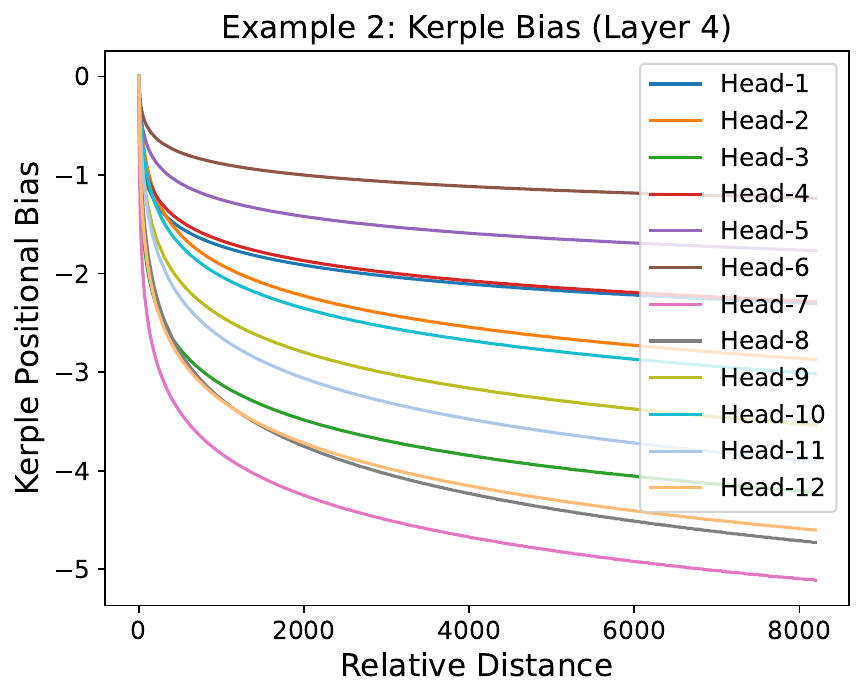}
\hspace{0in}
\includegraphics[width=0.32\textwidth]{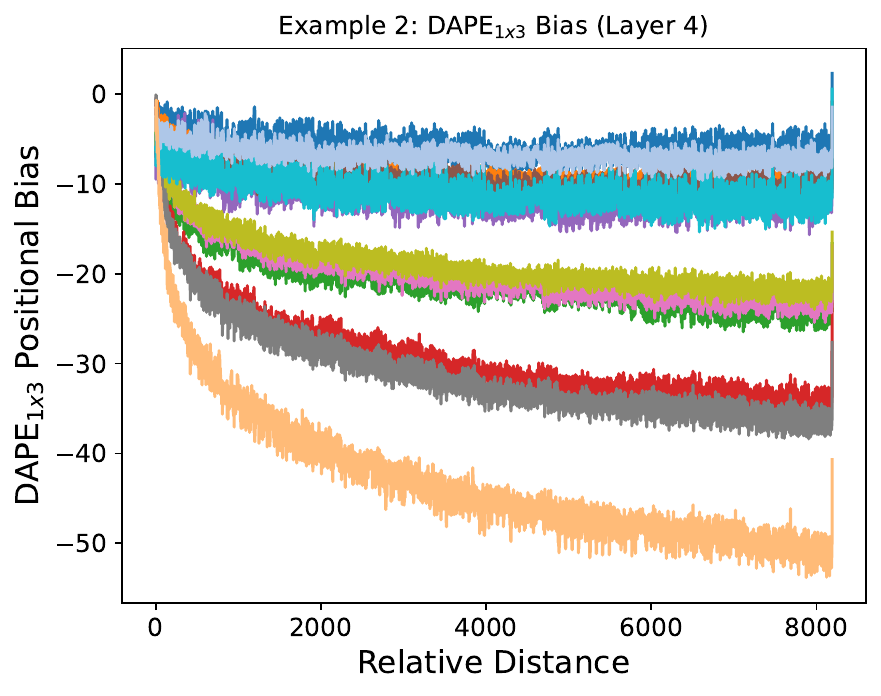}
\hspace{0in}

\includegraphics[width=0.32\textwidth]{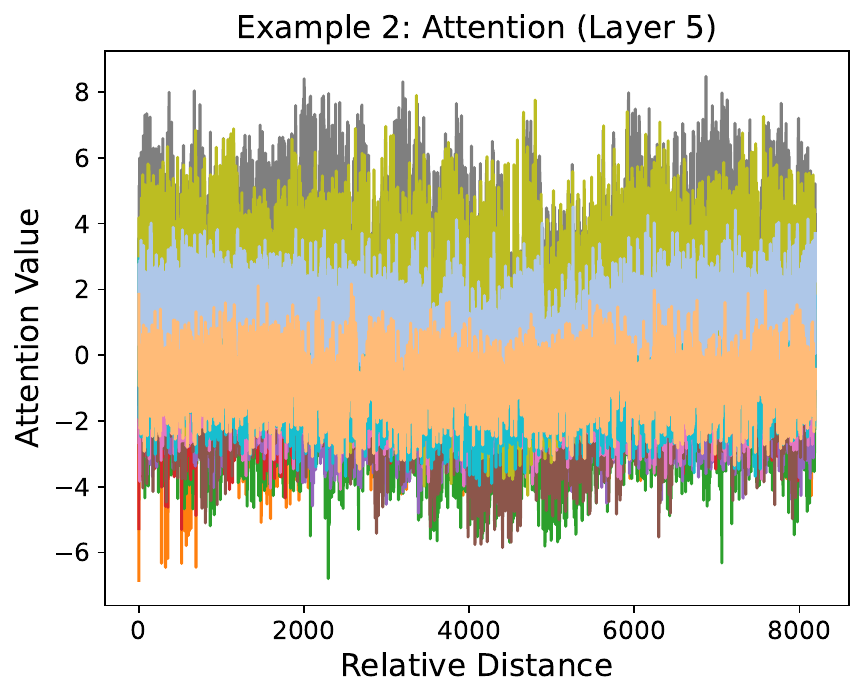}
\hspace{0in}
\includegraphics[width=0.32\textwidth]{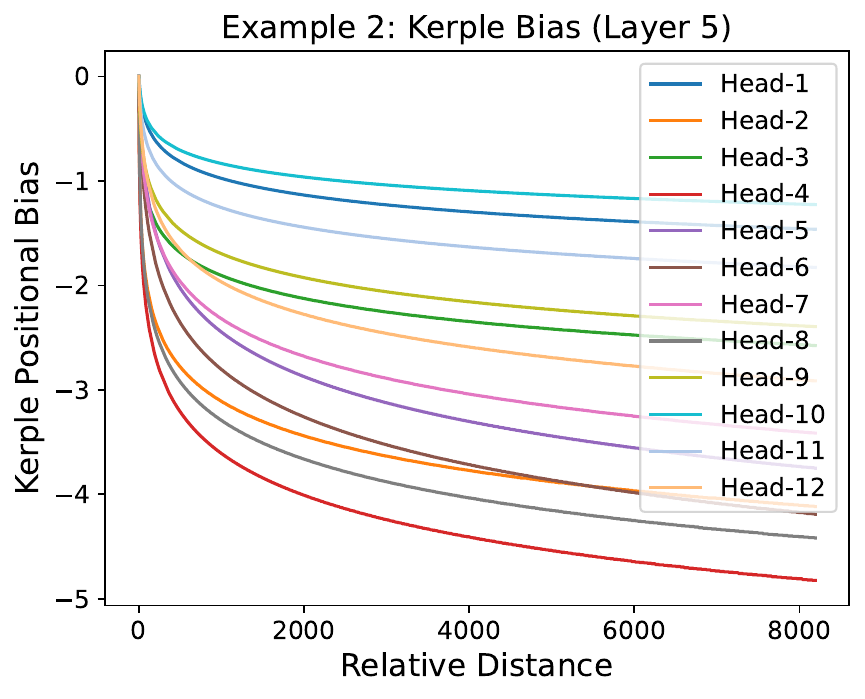}
\hspace{0in}
\includegraphics[width=0.32\textwidth]{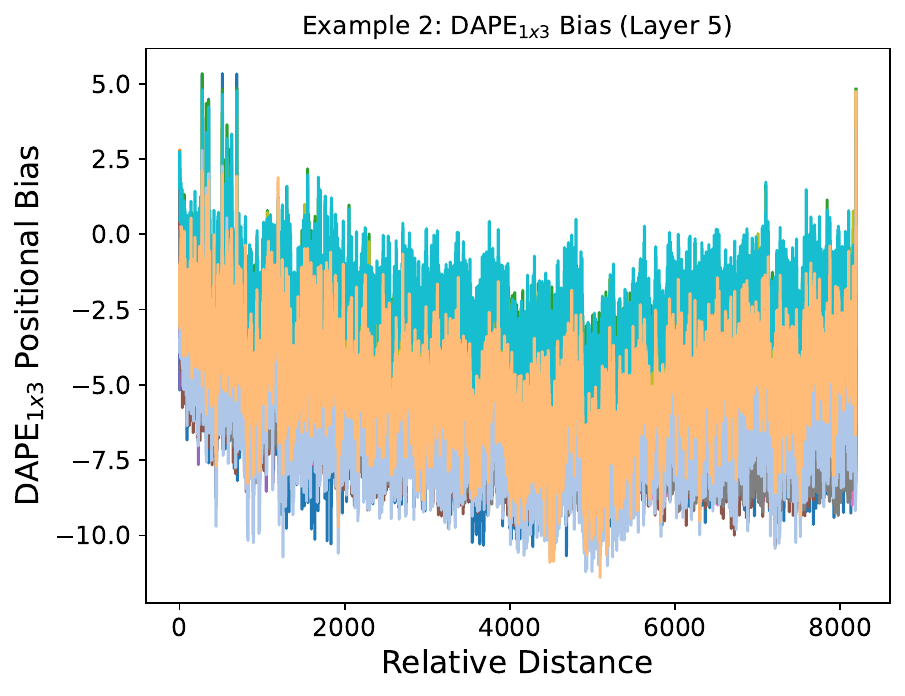}
\hspace{0in}

\includegraphics[width=0.32\textwidth]{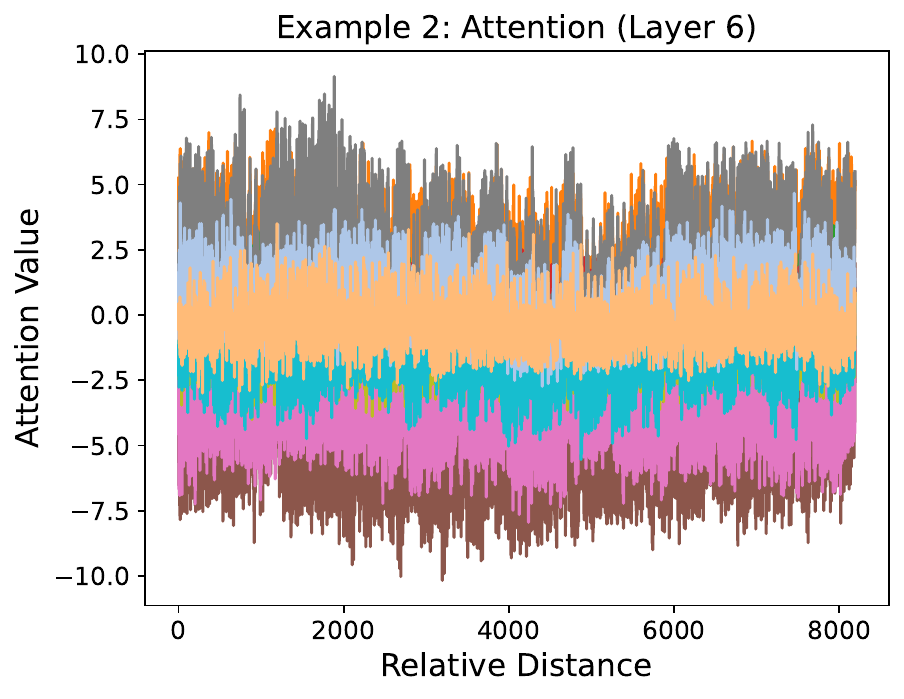}
\hspace{0in}
\includegraphics[width=0.32\textwidth]{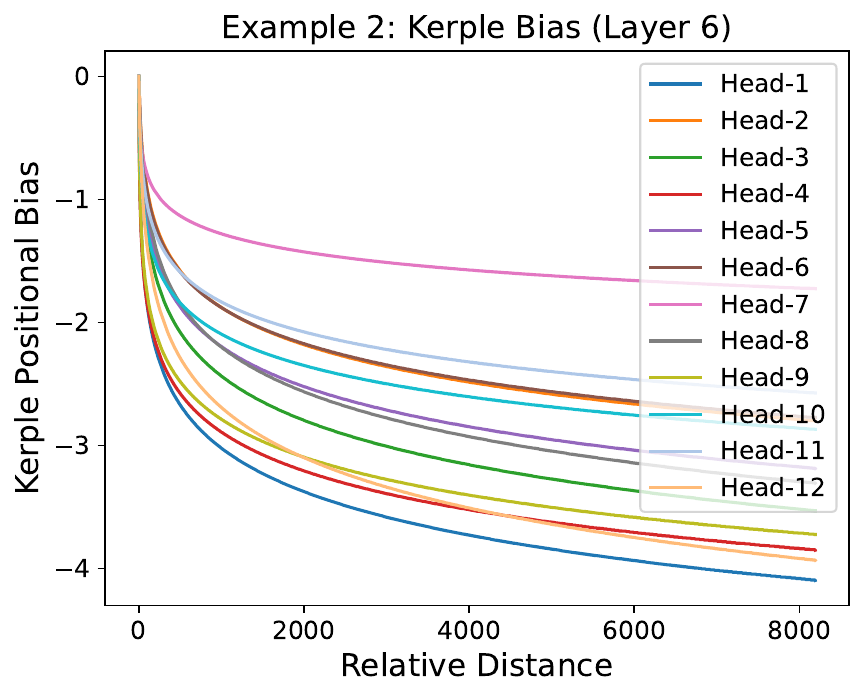}
\hspace{0in}
\includegraphics[width=0.32\textwidth]{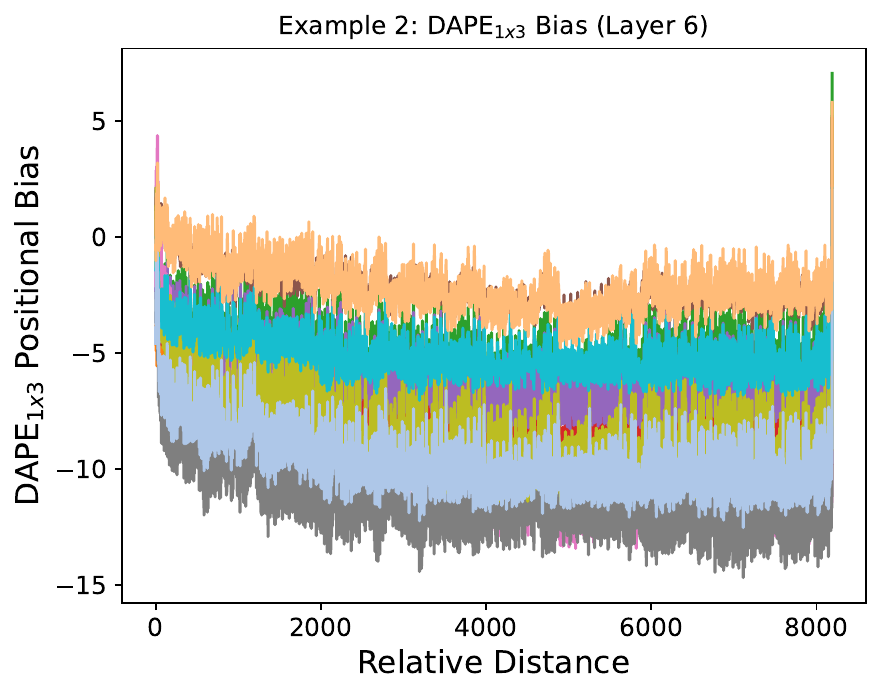}
\hspace{0in}

\hspace{0in}
\caption{
\small
\textbf{Evaluation Length 8192 Example 2: Part 1. From Left to Right: (1) The Attention is $\mX \mW_Q(\mX \mW_K)^{\top}$; (2) The Kerple bias is $\mB$; (3) The \methodShortName (with Kerple) bias is $f( \mX \mW_Q(\mX \mW_K)^{\top},\mB)$.
}
}
\end{figure}

\begin{figure}[htbp]
\setlength{\abovecaptionskip}{0.1cm}
\centering

\includegraphics[width=0.32\textwidth]{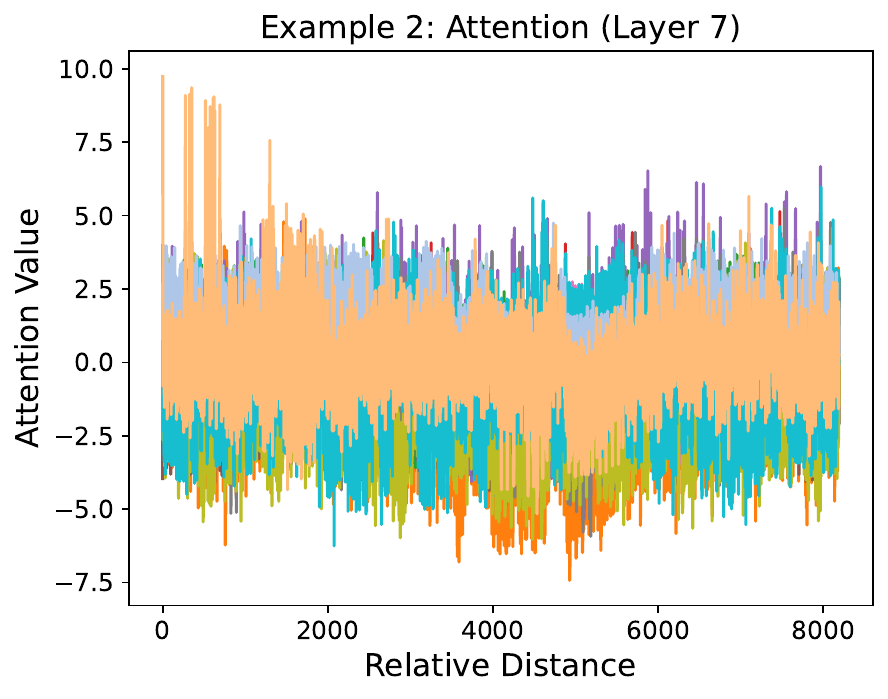}
\hspace{0in}
\includegraphics[width=0.32\textwidth]{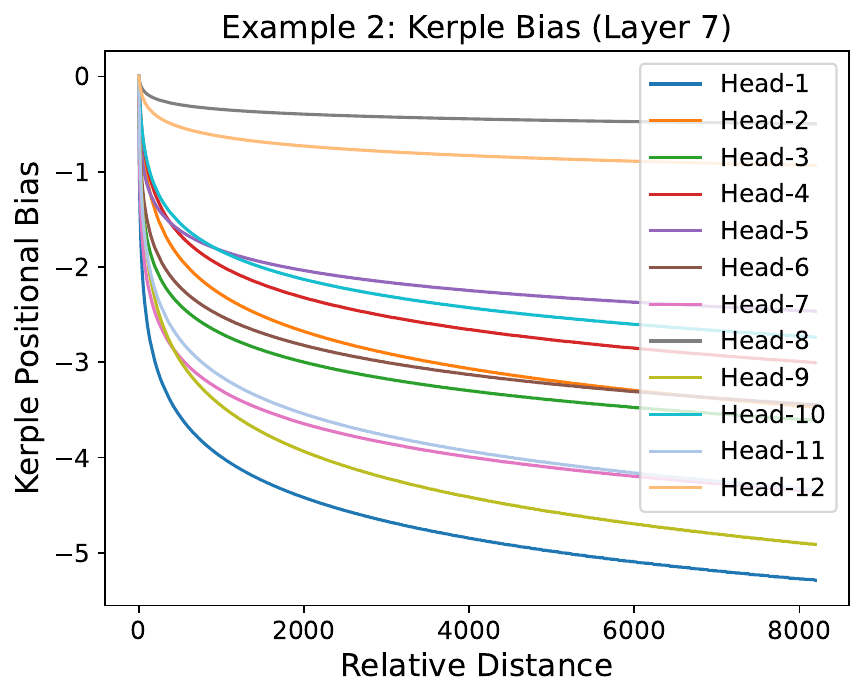}
\hspace{0in}
\includegraphics[width=0.32\textwidth]{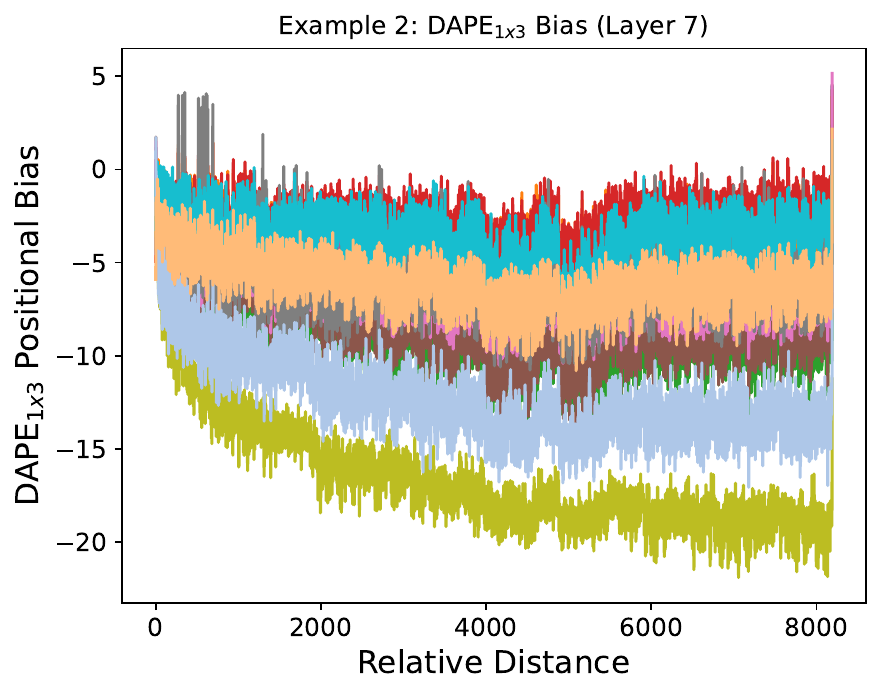}
\hspace{0in}

\includegraphics[width=0.32\textwidth]{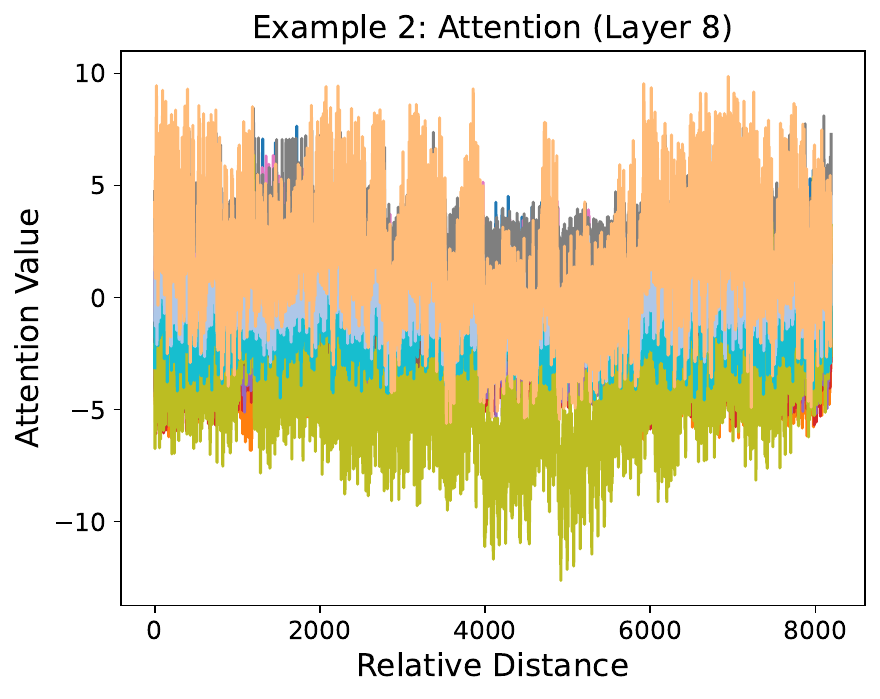}
\hspace{0in}
\includegraphics[width=0.32\textwidth]{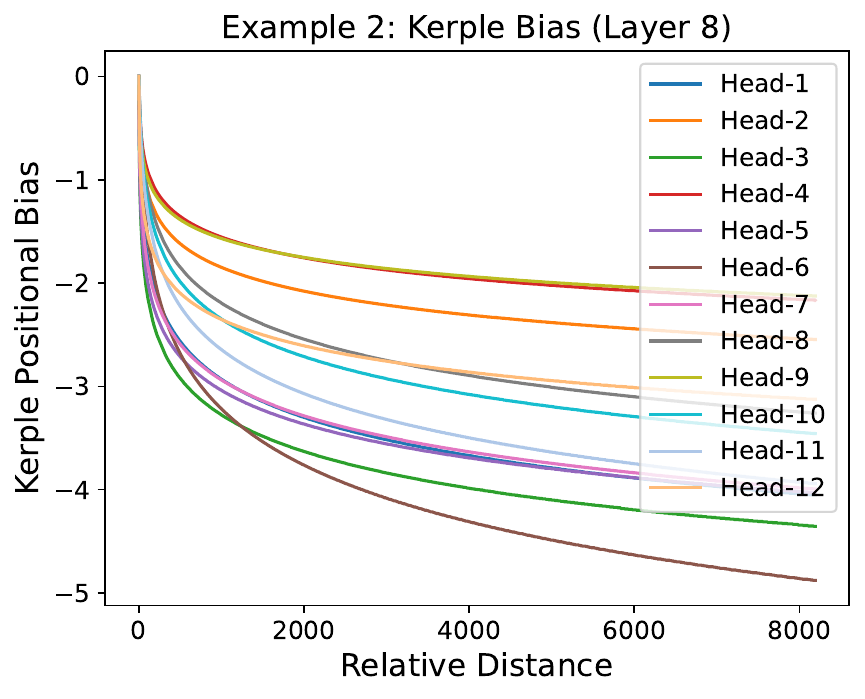}
\hspace{0in}
\includegraphics[width=0.32\textwidth]{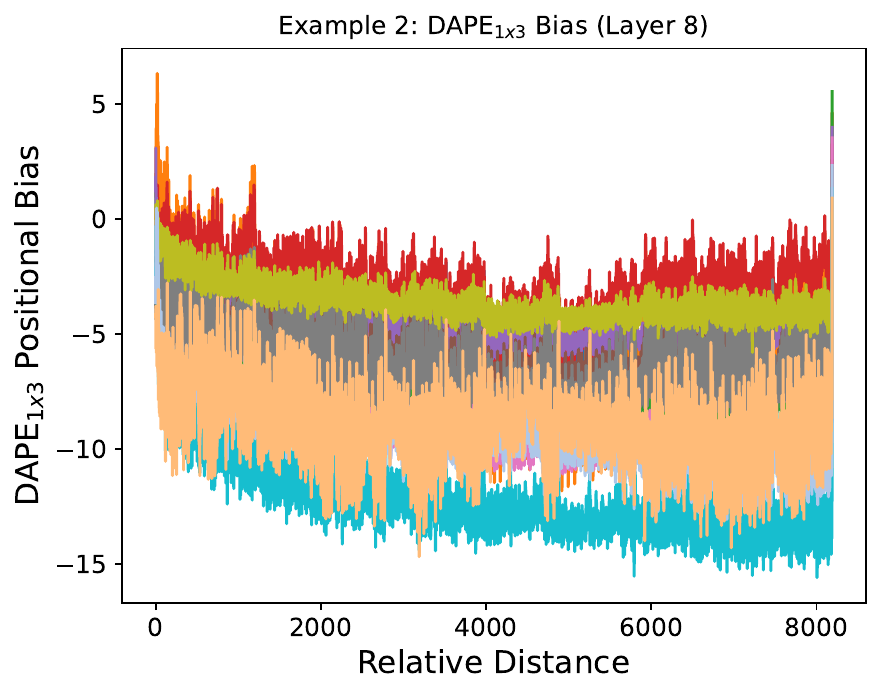}
\hspace{0in}

\includegraphics[width=0.32\textwidth]{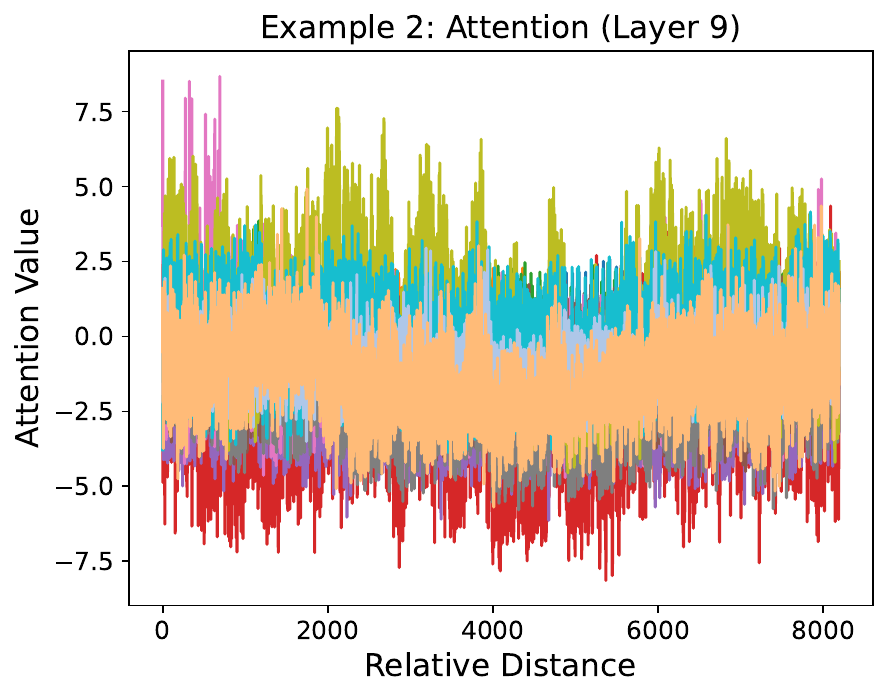}
\hspace{0in}
\includegraphics[width=0.32\textwidth]{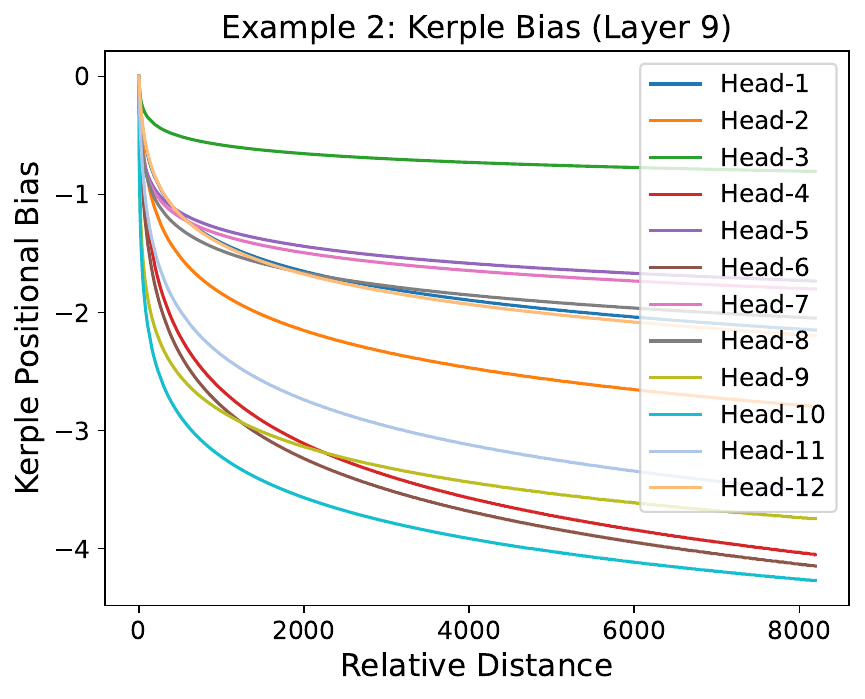}
\hspace{0in}
\includegraphics[width=0.32\textwidth]{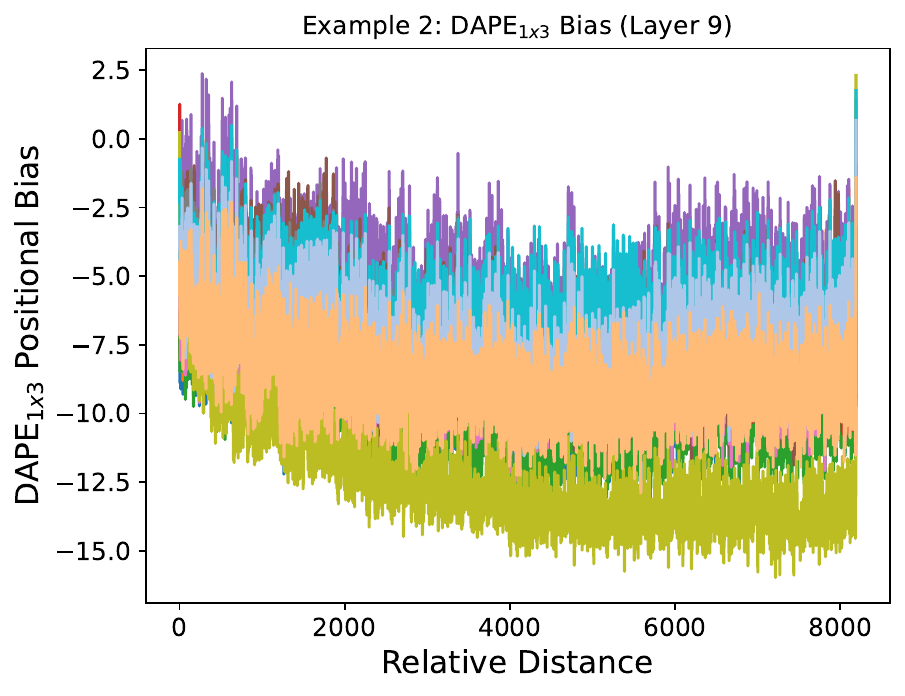}

\includegraphics[width=0.32\textwidth]{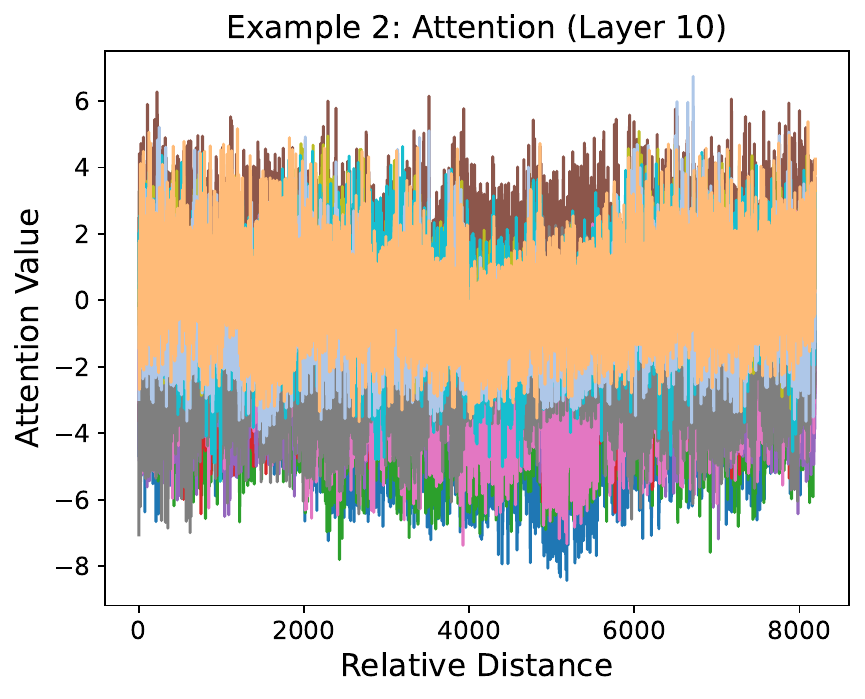}
\hspace{0in}
\includegraphics[width=0.32\textwidth]{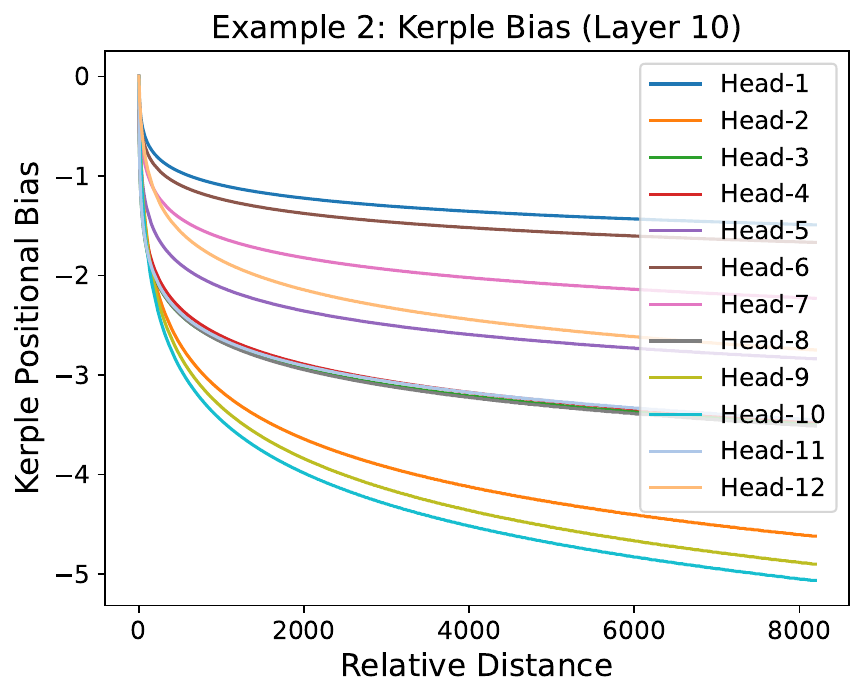}
\hspace{0in}
\includegraphics[width=0.32\textwidth]{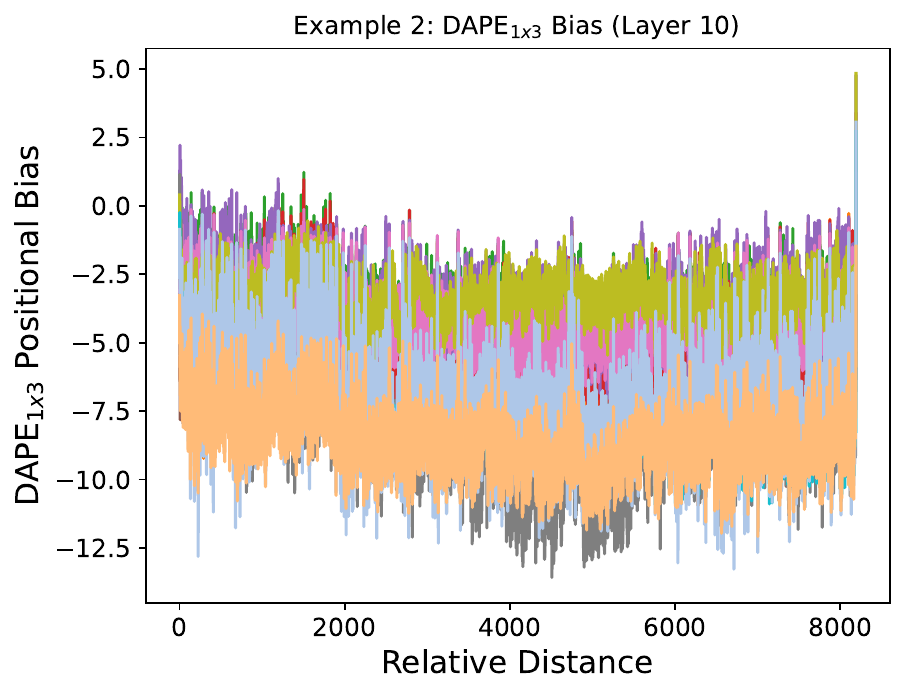}

\includegraphics[width=0.32\textwidth]{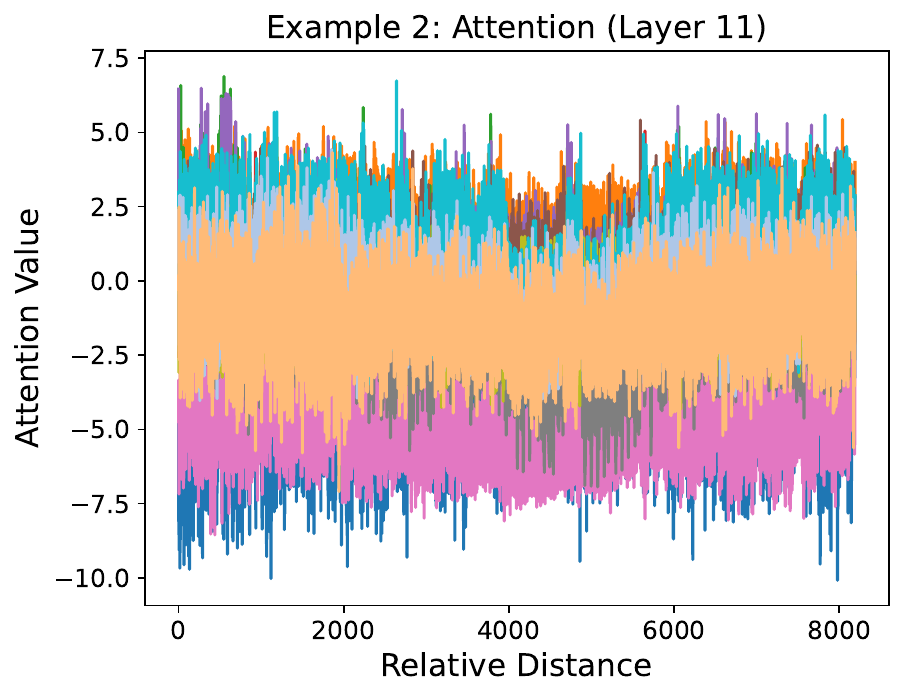}
\hspace{0in}
\includegraphics[width=0.32\textwidth]{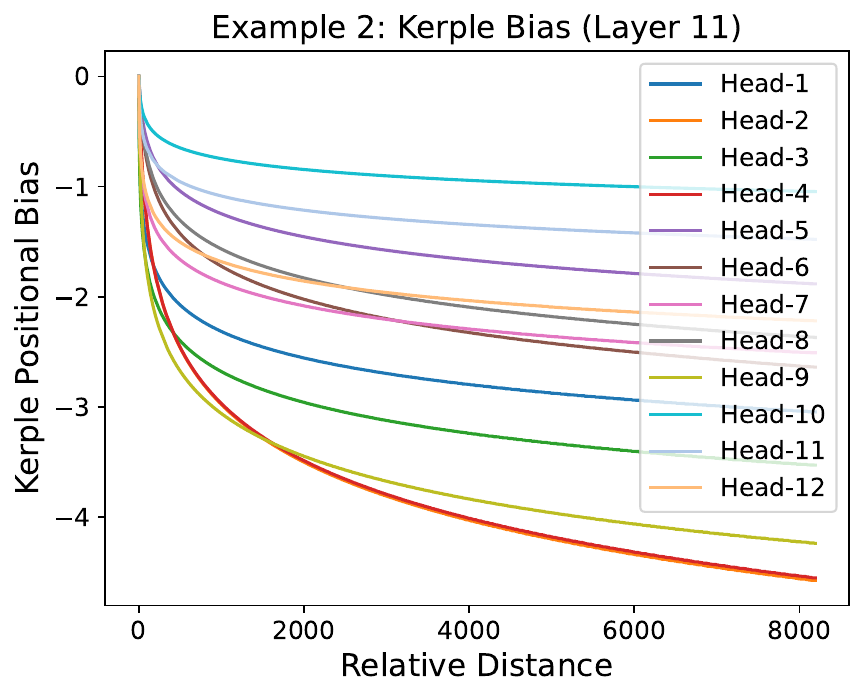}
\hspace{0in}
\includegraphics[width=0.32\textwidth]{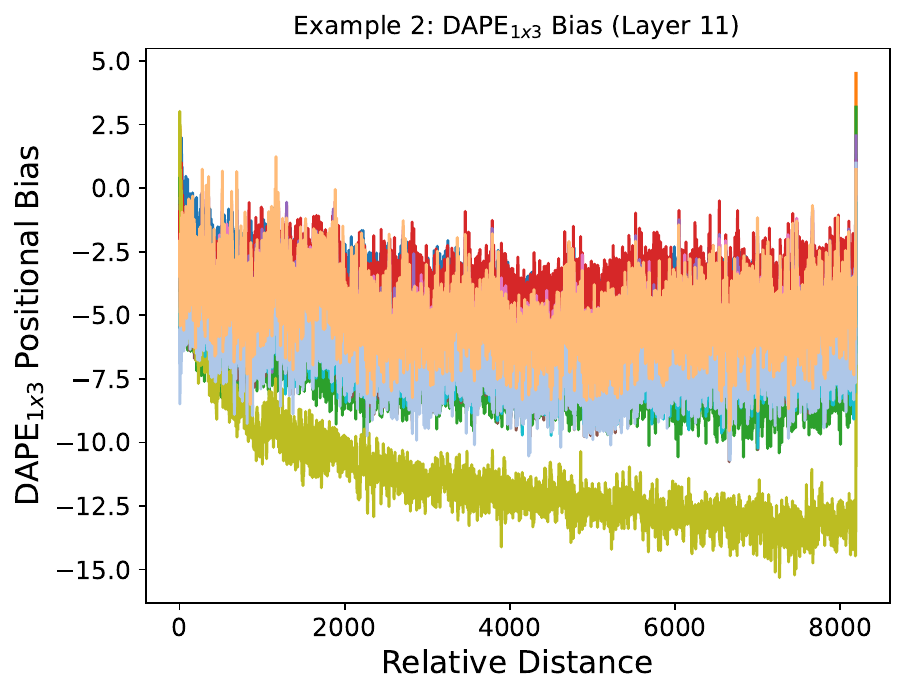}
\hspace{0in}

\includegraphics[width=0.32\textwidth]{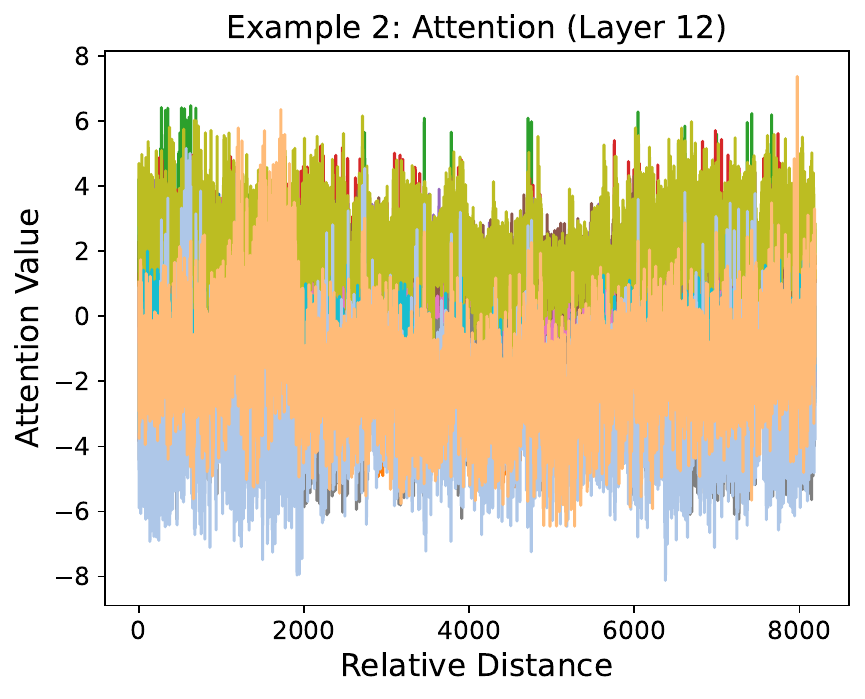}
\hspace{0in}
\includegraphics[width=0.32\textwidth]{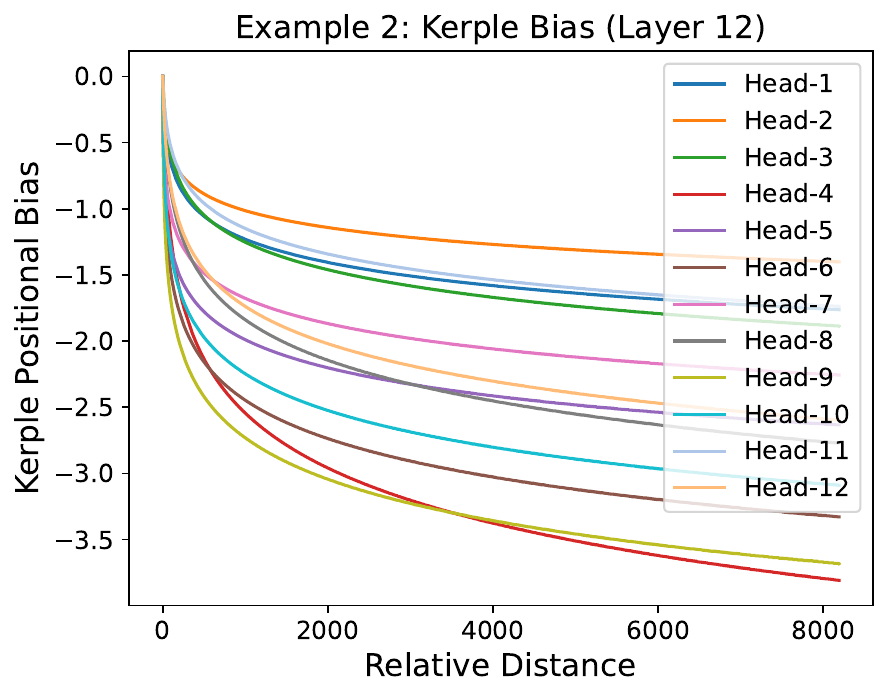}
\hspace{0in}
\includegraphics[width=0.32\textwidth]{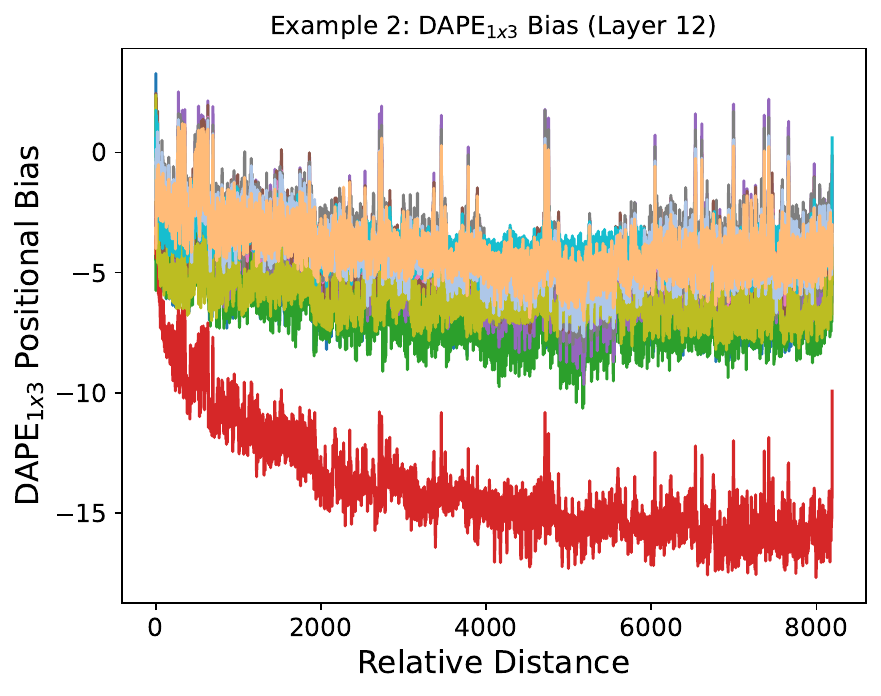}

\hspace{0in}
\caption{
\small
\textbf{Evaluation Length 8192 Example 2: Part 2. From Left to Right: (1) The Attention is $\mX \mW_Q(\mX \mW_K)^{\top}$; (2) The Kerple bias is $\mB$; (3) The \methodShortName (with Kerple) bias is $f( \mX \mW_Q(\mX \mW_K)^{\top},\mB)$.
}
}
\end{figure}

\clearpage
\newpage
\section{Implementation}
\label{appendix: implementation}
In this section, we present the implementation of the proposed \methodShortName module in \texttt{PyTorch} \citep{paszke2019pytorch}.

\definecolor{lightgreen}{rgb}{0,0.8,0}
\definecolor{darkgreen}{rgb}{0,0.8.0.2}
\definecolor{backcolour}{rgb}{0.97,0.97,0.94}
\lstset{language=Python,
basicstyle=\footnotesize,
breaklines=true,
backgroundcolor = \color{backcolour},
keywordstyle=\color{blue}\ttfamily,
stringstyle=\color{lightgreen}\ttfamily,
commentstyle=\color{gray}\ttfamily,
xleftmargin=2.5em,xrightmargin=0.5em, aboveskip=1em,
morecomment=[l][\color{darkgreen}]{\#}}

\begin{lstlisting}
import torch
import torch.nn as nn

class DAPEV2(nn.Module):
  def __init__(self, head_number=12, mlp_width=32,kernel_size=3):
    """
    DAPEV2 attention bias module.

    Args:
      num_heads: number of attention heads.
      mlp_width: Width of MLP.
      kernel_size: convolution kernel size.
    """
    super(DAPEV2, self).__init__()


    self.mlp =  nn.Sequential(
            nn.Conv2d(in_channels=head_number*2, out_channels=mlp_width,kernel_size=(1,kernel_size),stride=(1,1),padding=(0,kernel_size//2),dilation=(1,1)),
            nn.LeakyReLU(),
            nn.Conv2d(in_channels=mlp_width, out_channels=head_number,kernel_size=(1,kernel_size),stride=(1,1),padding=(0,kernel_size//2),dilation=(1,1)))

  def forward(self, attention: torch.Tensor, bias: torch.Tensor):
    """
    Args:
      attention: input sequence, which is q^T * k,
         shape [bsz, num_heads, seq_len, seq_len]
      bias: bias matrix, which can be generated by ALiBi, Kerple 
      FIRE or other additive position encodings
         shape [1,num_heads, seq_len, seq_len]

    Returns:
      attention with DAPEV2,
      shape [bsz, num_heads, seq_len, seq_len]
    """
    bias_tile=torch.tile(fire_bias, (x.shape[0],1,1,1) )
    attention_bias_concat=torch.cat( (attention, bias_tile), dim=1)
    attention_bias_concat=torch.tril(attention_bias_concat)
    attention_bias_concat=self.mlp(attention_bias_concat)
    

    return attention+bias+attention_bias_concat
\end{lstlisting}

\end{document}